%% file: arxiv.tex
\documentclass[10pt,twocolumn,letterpaper]{article}

\usepackage[pagenumbers]{cvpr}

\input{preamble}
\definecolor{cvprblue}{rgb}{0.21,0.49,0.74}
\usepackage[pagebackref,breaklinks,colorlinks,allcolors=cvprblue]{hyperref}

\usepackage{graphicx}
\usepackage{amsmath}
\usepackage{amssymb}
\usepackage{booktabs}
\usepackage{tikz}
\usetikzlibrary{spy}
\usepackage{multirow}
\usepackage{subcaption}
\usepackage{overpic}

\usepackage{colortbl}  %
\usepackage{xcolor}    %
\definecolor{color3}{rgb}{0.95,0.95,0.95}

\newcommand{\myname}[0]{HairGuard} %

\def\paperID{8052} %
\def\confName{CVPR}
\def\confYear{2026}

\title{Guardians of the Hair: \\ Rescuing Soft Boundaries in Depth, Stereo, and Novel Views}

\author{Xiang Zhang$^{1,2}$\quad Yang Zhang$^{2}$\quad Lukas Mehl$^{2}$\quad Markus Gross$^{1,2}$\quad Christopher Schroers$^{2}$\\
$^{1}$ETH Zürich \qquad $^{2}$DisneyResearch$|$Studios
}

\begin{document}

\twocolumn[{%
\renewcommand\twocolumn[1][]{#1}%
\maketitle
\centering
\includegraphics[width=\linewidth]{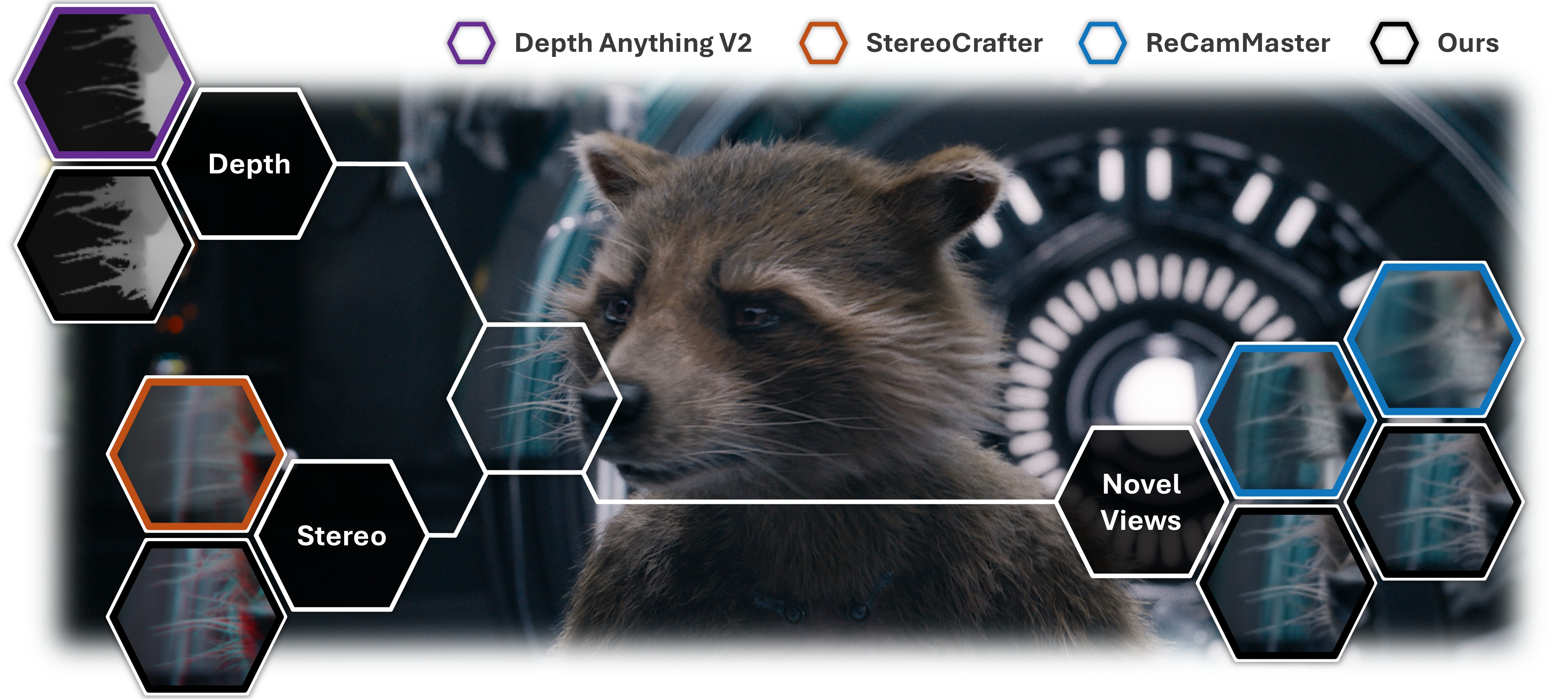}
\vspace{-1em}
\captionof{figure}{\textbf{Our Mission.} The Guardians of the Hair (\textbf{\myname}) aim to rescue soft boundary details, \eg, thin hairs, where foreground and background are mixed in the observed color. Previous state-of-the-art approaches often suffer from missing details (see depth estimation results), degraded texture (see stereo results, displayed in anaglyph), and inconsistent geometry (see novel views) in soft boundaries.
In contrast, \myname\ preserves fine-grained soft boundary details and demonstrates strong performance across diverse tasks.  \vspace{2em}}
\label{fig:teaser}
}]

\input{sec/abstract}
\input{sec/intro}

\input{sec/relatedwork}

\input{sec/approach}

\input{sec/experiments}

\input{sec/conclusion}

\newpage
\appendix

\section*{Supplementary Material}
The supplementary material is organized as follows: We first provide more implementation details in \cref{sec:supp-implement}. Then, the model performance, including robustness, plug-and-play performance, and computational complexity, is analyzed in \cref{sec:supp-analysis}. Following that, we show more experiments on the depth fixer and the color fuser in \cref{sec:supp-fixer} and \cref{sec:supp-fuser}, respectively. Afterward, we analyze the limitations of \myname\ and discuss potential future directions in \cref{sec:supp-limitation}. In the end, more visual comparisons on ablation study, monocular depth estimation, stereo conversion, and novel view synthesis are provided in \cref{sec:supp-visuals}.

\section{More Implementation Details}\label{sec:supp-implement}

\subsection{Marvel-10K Dataset}
The Marvel-10K dataset consists of 501 stereo video sequences from 5 Marvel movies: \textit{Ant-Man and the Wasp: Quantumania} (85 scenes), \textit{Black Panther: Wakanda Forever} (83 scenes), \textit{Doctor Strange in the Multiverse of Madness} (119 scenes), \textit{Guardians of the Galaxy Vol. 3} (101 scenes), and \textit{Thor: Love and Thunder} (113 scenes). Since movie frames are highly correlated within shots, we subsample them to select meaningful frames and exclude the studio intros, credits, and black frames. Each video sequence corresponds to a single shot and consists of 25 stereo pairs. For stereo conversion evaluation, we use left-view images as inputs and the right-view images as ground truth. As shown in \cref{fig:supp-marvel-examples}, the Marvel-10K dataset features computer-generated characters, intense motions, complex lighting, and uncommon cinematic scenes, making it highly challenging and suitable for evaluating algorithms in real-world applications such as film production.

\input{figs/supp/fig-marvel_examples}

\subsection{Color Fuser}
We provide more implementation details about the dual skip module in our color fuser. Given the inpainted image $I_{inpaint}$ and the warped image $I_{warp}$, we first extract multi-scale features $\{F_{inpaint}\}, \{F_{warp}\}$ using the frozen VAE encoder. We also generate multi-scale warped masks $\{M_{warp}\}$ by resizing the original mask to each feature scale via nearest neighbor downsampling. Finally, we use an additional residual block in the VAE decoder to fuse the skipped features and masks at each feature scale, and inject the fused feature into the VAE decoder in a residual fashion,
\begin{equation*}
    F = F_{dec} + \operatorname{ResBlock}(F_{dec},F_{inpaint}, F_{warp},M_{warp}).
\end{equation*}
$\operatorname{ResBlock}(\cdot)$ indicates a residual block. $F_{dec}$ and $F$ correspond to the original decoder feature and the fused feature, respectively. Zero initialization is applied for the additional residual blocks during training.

\section{More Analysis on Model Performance}\label{sec:supp-analysis}

\input{figs/supp/fig-robustness_scene}

\subsection{Robustness and Generalization}
Since our depth fixer is trained on a relatively small synthetic dataset (approximately 20K samples), one concern is its robustness and generalization ability in complex scenes. To this end, we evaluate the performance of the depth fixer on the challenging Marvel-10K dataset, which is not seen during training. As shown in \cref{fig:supp-robust-scene}, the depth fixer can automatically identify soft boundary regions in various scenarios. In scenes without soft boundaries (\eg, the top two rows in \cref{fig:supp-robust-scene}), the depth fixer maintains the depth quality of the base depth model for robust zero-shot estimation. In complex scenes such as bright/dark environments, occlusions, and multiple targets (bottom three rows in \cref{fig:supp-robust-scene}), our depth fixer still exhibits promising performance in extracting and fixing soft boundaries, showcasing its robustness in real-world applications. This is attributed to the decoupling of depth estimation and soft-boundary refinement in our depth fixer.
Thanks to the proposed gated residual mechanism, we can leverage the base depth model for zero-shot transfer and focus solely on refining soft boundaries, thereby achieving strong generalization performance with efficient training.

\input{figs/supp/fig-plug-depth}
\input{tabs/supp/tab-plug-splatdiff}
\subsection{Plug-and-Play Performance}
Benefiting from the gated residual module, our depth fixer can be applied to improve depth predictions from different depth models in a plug-and-play manner. As visualized in \cref{fig:supp-plug-depth}, we apply the depth fixer to two depth models with different characteristics: Depth Pro captures better details but often predicts inaccurate depth values in boundaries~\cite{depthpro}, and UniDepthV2 tends to produce depth results with smoothed boundaries~\cite{unidepthv2}. Despite the different distributions of depth maps, our depth fixer maintains robust performance in predicting soft boundary regions and fixing depth details.

\par 
We further evaluate the plug-and-play capability of the depth fixer on video depth models, \eg, Video Depth Anything~\cite{chen2025videodepthanything}. Although the depth fixer is trained only on image datasets, it still exhibits remarkable performance in improving video depth results, as shown in \cref{fig:supp-plug-video-depth}. Thanks to the gated residual mechanism, our depth fixer only corrects the depth in soft boundary regions while preserving the temporal consistency of video depth results. Besides, the depth fixer shows stable performance in estimating soft boundary regions even under occluded scenes (\eg, see the predicted gate maps in \cref{fig:supp-plug-video-depth}), demonstrating its robustness in complex scenarios.

\par
In addition to enhancing depth estimation methods, our depth fixer can also be integrated with novel view synthesis models for performance improvement. For instance, we combine depth fixer with the previous novel view synthesis approach SplatDiff~\cite{splatdiff}, and evaluate its performance on the Marvel-10K dataset. Since the depth fixer improves the warping results by fixing soft boundary details in depth (detailed in \cref{sec:supp-warping}), its combination with SplatDiff shows a consistent performance gain across all metrics, as reported in \cref{tab:supp-plug-splatdiff}.

\input{tabs/supp/tab-complexity}
\subsection{Computational Complexity}
In \cref{tab:supp-complexity}, we compare the computational complexity, \ie, model size, peak GPU memory, and inference speed, of \myname\ with previous state-of-the-art novel view synthesis methods. We further break down the complexity of each component in \myname, and the results show that the scene painter dominates the computational cost in our framework. Since we apply the depth fixer, scene painter, and color fuser in a sequential way, the peak GPU memory of \myname\ equals that of the scene painter. As our primary contributions lie in the depth fixer and the color fuser, the scene painter can be replaced with a more lightweight variant for better efficiency. In summary, our \myname\ achieves state-of-the-art performance while maintaining competitive computational efficiency, as demonstrated in \cref{tab:supp-complexity}.

\input{tabs/supp/tab-depth-warping}

\section{More Experiments on Depth Fixer}\label{sec:supp-fixer}

\subsection{Warping Performance}\label{sec:supp-warping}
In this section, we analyze the influence of the depth fixer on view synthesis tasks. To focus on the impact of depth maps, we directly assess the quality of the warped images, without applying the scene painter and color fuser. In addition, since the proposed depth fixer only modifies the depth on the predicted soft boundary regions, \ie, regions with gate $G<1$, we compute pixel-level metrics only on these regions.
We apply our depth fixer in a plug-and-play fashion to improve the prediction from three state-of-the-art depth models (Depth Anything V2~\cite{yang2024depthanythingv2}, Depth Pro~\cite{depthpro}, and UniDepthV2~\cite{unidepthv2}), and compare the forward warping performance on the Marvel-10K dataset. As shown in \cref{tab:supp-depth-warping}, our depth fixer helps preserve more soft boundary details during forward warping, leading to consistent and significant improvements across different base depth models.

\subsection{Gated Residual}
Following the same experimental setting in \cref{sec:supp-warping}, we compare the performance of the depth fixer with different output mechanisms. Although the direct prediction and vanilla residual mechanisms help improve the depth on the soft boundary regions, they often cover redundant background regions and fail to capture the fine-grained details, as illustrated in \cref{fig:fixer_visuals} of the main paper. By accurately localizing the soft boundary regions with the estimated gate map, our gated residual facilitates precise depth refinement and achieves the best performance as shown in \cref{tab:supp-fixer-ablation} (\#3 \vs \#1-2).

\input{tabs/supp/tab-supp-fixer-ablation}

\subsection{Loss Function}
We train the depth fixer with the $\ell_1$ loss $\mathcal{L}_1$ and the image matting loss $\mathcal{L}_{\alpha}$~\cite{yao2024vitmatte}. Specifically, the image matting loss $\mathcal{L}_{\alpha}$ is formulated as
\begin{equation}
    \mathcal{L}_{\alpha} = \mathcal{L}_{1} + \mathcal{L}_{lap} + \mathcal{L}_{gp},
\end{equation}
where $\mathcal{L}_{lap}, \mathcal{L}_{gp}$ indicate the Laplacian loss~\cite{lischke2020laploss} and the gradient loss~\cite{dai2022gploss}, respectively. Inspired by the success of such a loss combination in the image matting task~\cite{yao2024vitmatte}, we adopt it to improve the detail extraction performance of our depth fixer. To verify its effectiveness, we train an additional model with $\mathcal{L}_1$ loss only and keep the other training settings unchanged. The results in \cref{tab:supp-fixer-ablation} show the better performance of the proposed loss combination (\#5 \vs \#4).

\subsection{Model Prior}
Identifying soft boundary regions is a challenging task, relying on a comprehensive understanding of semantic context and geometric layout. To this end, we initialize the feature branch of our depth fixer with the pre-trained Depth Anything V2~\cite{yang2024depthanythingv2}, which has been trained on large-scale datasets to acquire robust image and geometry priors. Benefiting from this, our depth fixer achieves strong performance with efficient training (only $\sim$20K training samples are used), as shown in \cref{tab:supp-fixer-ablation} (\#7 \vs \#6).

\subsection{Edge Guidance}
Object boundaries, especially regions with significant depth variations, play a crucial role in 3D tasks like view synthesis, where disocclusions and geometric distortions commonly occur. Thus, we extract edge cues from the input depth to guide the depth fixer. 
The depth gradients provided by the edge guidance enable more accurate localization of soft boundaries, leading to improved warping performance (\#9 \vs \#8 in \cref{tab:supp-fixer-ablation}).

\subsection{Alpha Threshold}
The alpha threshold $\alpha_{th}$ used in generating ground-truth depth is critical to the performance of depth fixer. As shown in \cref{fig:supp-alpha_thresh}, the model trained with a higher $\alpha_{th}$ exhibit finer delineation of depth boundaries with less redundant background regions, but it tends to ignore the areas with low opacity, \eg, very thin hair. In our design, we opt for a lower $\alpha_{th}$ to preserve as many soft boundary details as possible, which shows the best warping performance in \cref{tab:supp-fixer-ablation} (\#12 \vs \#10-11). The scene painter and the color fuser are then employed to fix redundant background regions during view synthesis, as illustrated in \cref{fig:fuser_visual} of the main paper. Nevertheless, a higher $\alpha_{th}$ can be used to train the depth fixer for different tasks, \eg, 3D segmentation or point cloud reconstruction, where precise depth boundaries are preferred.

\input{figs/supp/fig-alpha_thresh}

\input{tabs/supp/tab-fuser-ablation}

\input{figs/supp/fig-limitation}

\section{More Experiments on Color Fuser}\label{sec:supp-fuser}

\subsection{VAE Prior}
We build the color fuser upon a pre-trained VAE to harness its reconstruction prior for better view synthesis performance. To investigate the impact of the VAE prior, we train an additional color fuser from scratch using the same training settings. As shown in \cref{tab:supp-fuser-ablation}, the color fuser with VAE prior significantly outperforms its counterpart in visual quality (\#2 \vs \#1).

\subsection{Dual Skip}
Based on the VAE architecture, we further design a dual skip module to utilize the fine-grained features of the inpainted and warped images. To verify its effectiveness, we train two additional variants: one without skip connections (\#3 in \cref{tab:supp-fuser-ablation}) and one with a single skip connection (\#4). For model \#3, we expand the input channel of the VAE encoder and concatenate the inpainted image, warped image, and warped mask as its input. Regarding model \#4, we add a single skip to utilize the multi-scale features of the warped images. The results in \cref{tab:supp-fuser-ablation} show that model \#4 achieves a significant performance gain over model \#3 by alleviating detail compression in the VAE encoding. By further exploiting the features of inpainted images, our dual skip module yields the best reconstruction performance with high-quality texture details.

\input{figs/supp/fig-ablation}
\section{Limitation and Discussion}\label{sec:supp-limitation}
Despite the remarkable performance achieved by \myname, some limitations remain: 
\begin{itemize}
    \item \textit{Depth errors beyond soft boundaries:} The depth fixer relies on the gated residual mechanism to locate and fix soft boundary details, which benefits precise refinement and plug-and-play deployment. However, it is difficult for the depth fixer to correct depth errors beyond the soft boundary regions, as depicted in \cref{fig:supp-limitation-depth-base}. A possible solution is to train a depth fixer specialized for a given depth model, \eg, by using the model’s predictions instead of the synthesized inputs during training. Thus, the depth fixer could better adapt to the characteristics of the base depth model and achieve better fixing performance.
    \item \textit{Single-layer depth representation:} For view synthesis, we propose a color fuser to utilize the fine-grained texture from the warped images. However, due to the single-layer depth representation, the naive forward warping approach may produce geometric distortions in complex scenes containing multiple depth layers per pixel, \eg, transparent objects as shown in \cref{fig:supp-limitation-single_layer}. We attempted to address this limitation by estimating layered outputs comprising foreground color and depth, background color and depth, and an opacity map for composition. While this layered representation demonstrated advantages in certain cases, our trial experiments showed that it suffers from limited generalization capability, likely due to the increased complexity of the estimation. Thus, a potential solution is to collect large-scale training datasets to gain a strong prior for robust performance. Another possible direction is to employ dense 3D representations, \eg, 3D Gaussians~\cite{kerbl3Dgs}, to handle occlusions and overlapping surfaces.  
\end{itemize}

\section{More Visual Results}\label{sec:supp-visuals}

\subsection{Ablation Study}
\cref{fig:supp-ablation-visual} provides visual results for the ablation study conducted in the main paper (detailed in \cref{tab:stereo-ablation} of the main paper). Since depth quality is critical for forward warping performance, depth estimation errors in Depth Anything V2~\cite{yang2024depthanythingv2} often result in distorted structures in the soft boundary regions like thin hairs. By fixing depth details via the proposed depth fixer, better hair structures are preserved in the warped images, as shown in the green box of \cref{fig:supp-ablation-visual}. The scene painter is then applied to generate realistic contents for the disoccluded regions. However, the inpainted images often exhibit different texture details due to diffusion hallucination and pixel-to-latent compression. To this end, we propose a color fuser that adaptively combines the warped and inpainted images, generating novel views with consistent geometry and high-fidelity textures.

\subsection{Monocular Depth Estimation}
We provide more visual results of monocular depth estimation on the AIM-500 and P3M-10K datasets in \cref{fig:supp-depth-visual-aim} and \cref{fig:supp-depth-visual-p3m}, respectively. Compared with previous methods, our depth fixer shows robust performance in capturing soft boundary details across diverse targets and scenes. In some challenging cases with very thin hair structures, \eg, top few rows of \cref{fig:supp-depth-visual-p3m}, the depth fixer still recovers fine-grained depth details with sharp boundaries.

\subsection{Stereo Conversion}
\cref{fig:supp-stereo-visuals-marvel} compares the stereo conversion performance of \myname\ with the state-of-the-art methods on the Marvel-10K dataset. Due to the generative nature of the underlying models, previous stereo conversion approaches often suffer from texture hallucination and degraded details in the conversion results, \eg, see the top two rows in \cref{fig:supp-stereo-visuals-marvel}. By utilizing the fine-grained details of warped images via the color fuser, our \myname\ achieves high-quality stereo conversion performance with consistent geometry and texture.

\subsection{Novel View Synthesis}
We show more qualitative comparisons of novel view synthesis on the challenging AIM-500 and P3M-10K datasets in \cref{fig:supp-nvs-visual-aim} and \cref{fig:supp-nvs-visual-p3m}. Previous approaches often produce hallucinated textures that are inconsistent with the input image, \eg, see  ViewCrafter~\cite{yu2024viewcrafter} and ReCamMaster~\cite{recammaster} in the top few rows of \cref{fig:supp-nvs-visual-aim}. Although the recent method SplatDiff recovers better details~\cite{splatdiff}, its performance is highly dependent on the quality of the estimated depth maps. Hence, the depth errors in soft boundary regions often lead to artifacts in the synthesized novel views, \eg, top few rows in \cref{fig:supp-nvs-visual-p3m}. In contrast, the proposed \myname\ first fixes depth in the soft boundary regions to ensure geometry consistency, and then utilizes the color fuser to recover high-fidelity texture details, achieving state-of-the-art novel view synthesis performance.

\input{figs/supp/fig-depth_est_visuals_aim}

\input{figs/supp/fig-depth_est_visuals_p3m}

\input{figs/supp/fig-stereo_visuals_marvel}

\input{figs/supp/fig-nvs_visuals_aim}

\input{figs/supp/fig-nvs_visuals_p3m}

{
    \small
    \bibliographystyle{ieeenat_fullname}
    \bibliography{main}
}

\end{document}

%% file: sec/abstract.tex
\begin{abstract}
    Soft boundaries, like thin hairs, are commonly observed in natural and computer-generated imagery, but they remain challenging for 3D vision due to the ambiguous mixing of foreground and background cues. This paper introduces \textbf{Guardians of the Hair (\myname)}, a framework designed to recover fine-grained soft boundary details in 3D vision tasks. Specifically, we first propose a novel data curation pipeline that leverages image matting datasets for training and design a depth fixer network to automatically identify soft boundary regions.
    With a gated residual module, the depth fixer refines depth precisely around soft boundaries while maintaining global depth quality, allowing plug-and-play integration with state-of-the-art depth models. For view synthesis, we perform depth-based forward warping to retain high-fidelity textures, followed by a generative scene painter that fills disoccluded regions and eliminates redundant background artifacts within soft boundaries. Finally, a color fuser adaptively combines warped and inpainted results to produce novel views with consistent geometry and fine-grained details. Extensive experiments demonstrate that \myname\ achieves state-of-the-art performance across monocular depth estimation, stereo image/video conversion, and novel view synthesis, with significant improvements in soft boundary regions.
\end{abstract}

%% file: sec/intro.tex
\section{Introduction}
\label{sec:intro}

Driven by recent advances in foundation models and large-scale visual datasets~\cite{rombach2022stablediffusion,schuhmann2022laion}, significant progress has been witnessed in the field of 3D vision, including depth estimation~\cite{ranftl2020midas,ke2024marigold,yang2024depthanything,zhang2024betterdepth}, stereo conversion~\cite{stereodiffusion,mono2stereo,stereocrafter}, and novel view synthesis~\cite{splatdiff,recammaster,yu2024viewcrafter}.
These techniques play a crucial role in understanding and reconstructing 3D scenes, with broad applications in robotics, autonomous driving, augmented/virtual reality (AR/VR), and film production~\cite{wang2019pseudo,gao2024cat3d,mehl2024stereoconv,bahmani2025lyra,liang2024wonderland}. Although existing methods have shown promising performance in general scenarios~\cite{yang2024depthanything,recammaster}, generating geometrically consistent and visually realistic results in scenes with soft boundaries, \eg, hairs and thin structures, remains highly challenging (\cref{fig:teaser}).

\par 
Soft boundaries are ubiquitous and often arise when pixels receive mixed contributions from both the foreground and background, due to thin/semi-transparent structures or alpha blending in rendering~\cite{li2023mattingsurvey,yao2024vitmatte}. Thus, they are commonly observed in natural images like shots containing animals and humans, as well as computer-generated imagery such as \cref{fig:teaser}. The mixture of foreground and background pixels makes pixel-wise 3D estimation particularly challenging and ill-posed, since such regions exhibit uncertain correspondence and depth ambiguity.

\input{figs/main/fig-problem_stat}

\par 
Existing methods often struggle to capture accurate and fine-grained soft boundaries, as illustrated in \cref{fig:teaser}. For example, the state-of-the-art monocular depth estimation method Depth Anything V2~\cite{yang2024depthanythingv2} fails to extract the fine details of hairs and produces broken depth results (\cref{fig:teaser} and \cref{fig:problem_stat_results}). Although the recent approach Depth Pro~\cite{depthpro} achieves improved detail preservation in depth estimation, the estimated depth around soft boundaries often falls behind the true surface, leading to detached hairs as shown in the point cloud renders of \cref{fig:problem_stat_results} (red box). Since depth estimation is often required by \textit{explicit} 3D vision methods~\cite{yu2024viewcrafter,mono2stereo,splatdiff}, the depth errors tend to propagate to the subsequent stages, resulting in sub-optimal performance. In the field of stereo conversion and novel view synthesis, one emerging trend is to generate new viewpoints in an \textit{implicit} manner without depth~\cite{you2024nvssolver,gao2024cat3d,recammaster}. By utilizing the rich prior knowledge learned in foundation generative models~\cite{rombach2022stablediffusion,wan2025wan}, these implicit approaches can effectively handle complex occlusion and geometry in 3D world. However, due to the generative nature of the underlying foundation models, implicit 3D vision methods often suffer from hallucination issues and thus generate inconsistent texture details in soft boundaries (\eg, see ReCamMaster~\cite{recammaster} in \cref{fig:teaser}). Meanwhile, most foundation generative models are designed in the latent space for computational efficiency~\cite{rombach2022stablediffusion,xing2025dynamicrafter,wan2025wan}. Such a design often results in texture degradation due to pixel-to-latent compression~\cite{splatdiff}, as illustrated by the StereoCrafter~\cite{stereocrafter} results in \cref{fig:teaser}.

\par 
In the realm of 2D vision, image matting provides an explicit formulation for soft boundaries by estimating an opacity map (\ie, alpha matte) to model the pixel mixture along the transition between foreground and background~\cite{li2023mattingsurvey,yao2024vitmatte}.
Inspired by the matting formulation, we leverage image matting datasets to improve soft boundary modeling and propose \textbf{Guardians of the Hair (\myname)} to rescue soft boundary details in 3D tasks. Specifically, \myname\ consists of three teammates: \textit{depth fixer}, \textit{scene painter}, and \textit{color fuser}. 
By utilizing matting datasets in training data curation, our depth fixer learns to identify soft boundary regions and correct depth predictions with a gated residual module. This design not only enables precise depth refinement over soft boundaries, but also supports plug-and-play integration with zero-shot depth models for robust performance. For view synthesis, we first perform forward warping using the fixed depth, followed by a generative scene painter that synthesizes realistic disoccluded regions and corrects geometric errors caused by warping. Finally, to address texture hallucination and detail compression in generative models, we propose a color fuser to preserve fine-grained details and ensure geometrically consistent view synthesis via a dual skip module. As shown in \cref{fig:teaser}, the components of \myname\ work collaboratively to achieve remarkable performance across different 3D vision tasks.

\par 
In a nutshell, our main contributions are three-fold:
\begin{itemize}
    \item We present {\myname} to capture, model, and reconstruct fine-grained soft boundary details in 3D vision tasks. Extensive experiments verify the effectiveness and superiority of \myname\ across monocular depth estimation, stereo image/video conversion, and novel view synthesis.
    \item We design novel data curation strategies to leverage image matting datasets for training, enabling \myname\ to automatically identify and fix soft boundaries without relying on manually crafted cues, \eg, trimaps~\cite{li2023mattingsurvey,yao2024vitmatte}.
    \item We propose a depth fixer with a gated residual module, which enables precise depth refinement in soft boundary regions for plug-and-play enhancement. 
    Additionally, we design a dual skip architecture in the color fuser to ensure geometrically consistent and high-quality view synthesis.
\end{itemize}

%% file: figs/main/fig-problem_stat.tex
\def\imgWidth{0.32\linewidth} %
\def\scc{(-1.9,-1.4)}

\def\rebigone{(-0.5, -0.5)} %
\def\rebigtwo{(-0.5, 0.5)} %

\def\rebigthree{(2.8, 0.7)} %
\def\rebigfour{(2.8, -0.7)} %

\def\zoomone{(-0.2,-0.3)} %
\def\zoomtwo{(0.33,0.4)} %
\def\zoomthree{(0.9,0.7)} %
\def\zoomfour{(-0.15,-0.3)} %
\def\zoomfive{(0,0.32)} %
\def\zoomsix{(-0.1,0.32)} %

\def\ssizz{0.8cm} %
\def\ssmag{3}

\def\ssizzimg{1.2cm} %

\begin{figure}[t]
\centering
\tikzstyle{img} = [rectangle, minimum width=\imgWidth, draw=black]
    \begin{subfigure}{\linewidth}
      \begin{subfigure}{\imgWidth}
        \begin{tikzpicture}[spy using outlines={magnification=\ssmag,size=\ssizz},inner sep=0]
            \node [align=center, img] {\includegraphics[width=\textwidth]{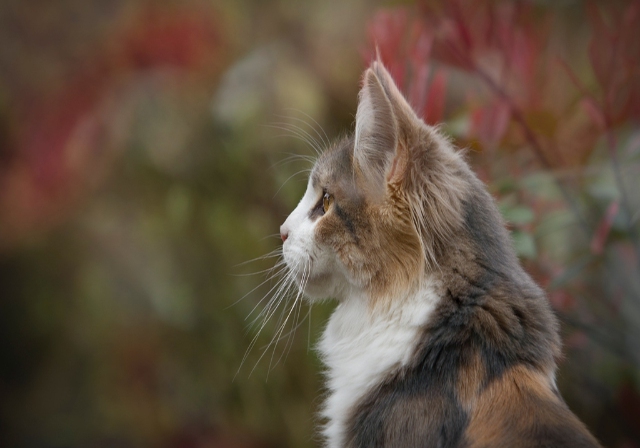}};
            \spy [draw=green] on \zoomone in node [left] at \rebigone;
            \spy [draw=red] on \zoomsix in node [left] at \rebigtwo;
    	\end{tikzpicture}
     \caption*{Input Image}
    \end{subfigure}
    \begin{subfigure}{\imgWidth}
		\begin{tikzpicture}[spy using outlines={magnification=\ssmag,size=\ssizz},inner sep=0]
            \node [align=center, img] {\includegraphics[width=\textwidth]{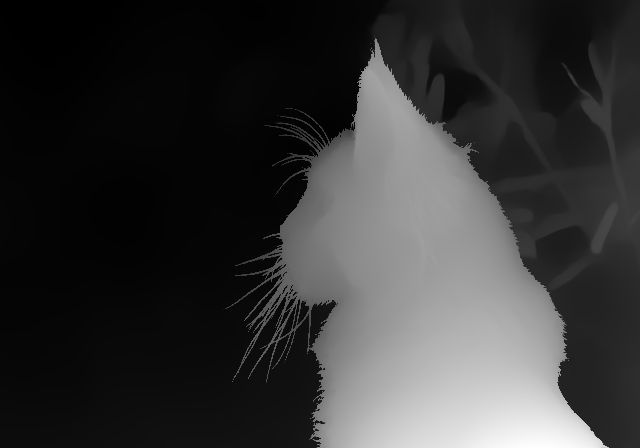}};
            \spy [draw=green] on \zoomone in node [left] at \rebigone;
            \spy [draw=red] on \zoomsix in node [left] at \rebigtwo;
    	\end{tikzpicture}
     \caption*{Depth ({Ours})}
    \end{subfigure}
    \begin{subfigure}{\imgWidth}
        \begin{tikzpicture}[spy using outlines={magnification=\ssmag,size=\ssizz},inner sep=0]
            \node [align=center, img] {\includegraphics[width=\textwidth]{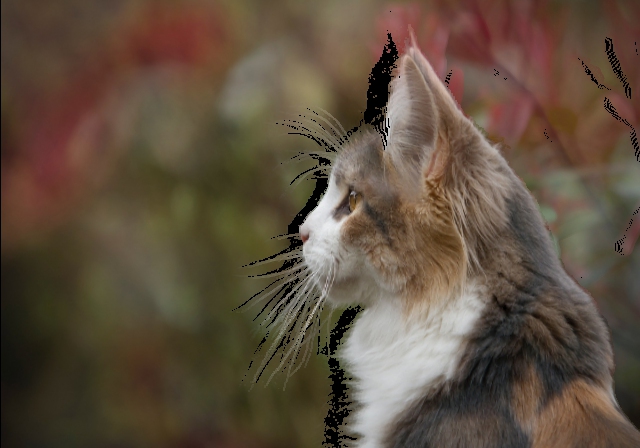}};
            \spy [draw=green] on \zoomfour in node [left] at \rebigone;
            \spy [draw=red] on \zoomfive in node [left] at \rebigtwo;
    	\end{tikzpicture}
      \caption*{Point Render ({Ours})}
      \end{subfigure}
    \end{subfigure}
      \\ %
      \begin{subfigure}{\linewidth}
      \begin{subfigure}{\imgWidth}
        \begin{tikzpicture}[spy using outlines={magnification=\ssmag,size=\ssizz},inner sep=0]
            \node [align=center, img] {\includegraphics[width=\textwidth]{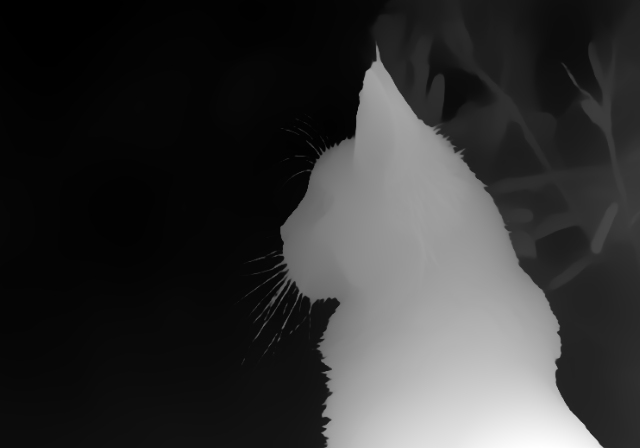}};
            \spy [draw=green] on \zoomone in node [left] at \rebigone;
            \spy [draw=red] on \zoomsix in node [left] at \rebigtwo;
    	\end{tikzpicture}
     \caption*{Depth (DAv2)}
    \end{subfigure}
    \begin{subfigure}{\imgWidth}
		\begin{tikzpicture}[spy using outlines={magnification=\ssmag,size=\ssizz},inner sep=0]
            \node [align=center, img] {\includegraphics[width=\textwidth]{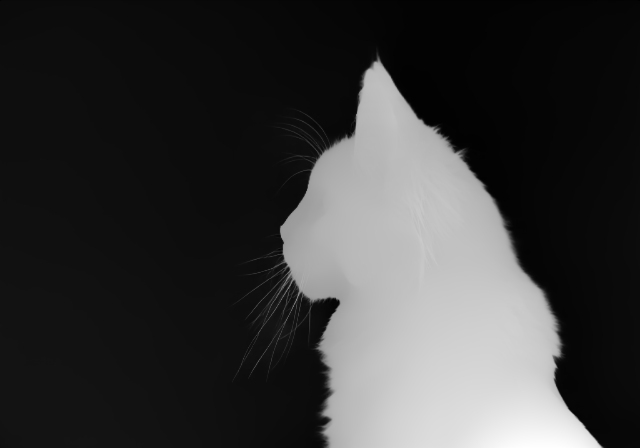}};
            \spy [draw=green] on \zoomone in node [left] at \rebigone;
            \spy [draw=red] on \zoomsix in node [left] at \rebigtwo;
    	\end{tikzpicture}
     \caption*{Depth (DPro)}
    \end{subfigure}
    \begin{subfigure}{\imgWidth}
        \begin{tikzpicture}[spy using outlines={magnification=\ssmag,size=\ssizz},inner sep=0]
            \node [align=center, img] {\includegraphics[width=\textwidth]{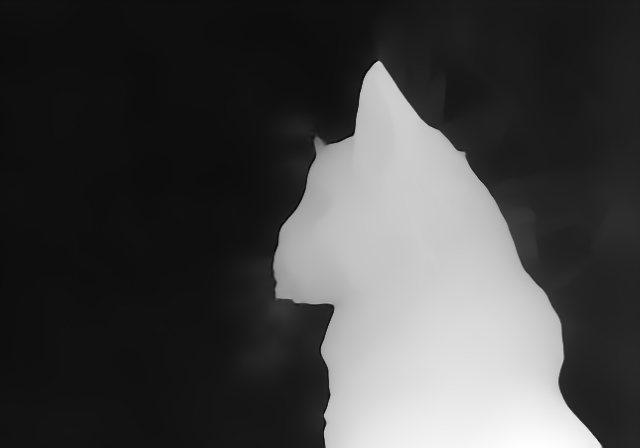}};
            \spy [draw=green] on \zoomone in node [left] at \rebigone;
            \spy [draw=red] on \zoomsix in node [left] at \rebigtwo;
    	\end{tikzpicture}
      \caption*{Depth (UDv2)}
      \end{subfigure}
      \\ %
       \begin{subfigure}{\imgWidth}
        \begin{tikzpicture}[spy using outlines={magnification=\ssmag,size=\ssizz},inner sep=0]
            \node [align=center, img] {\includegraphics[width=\textwidth]{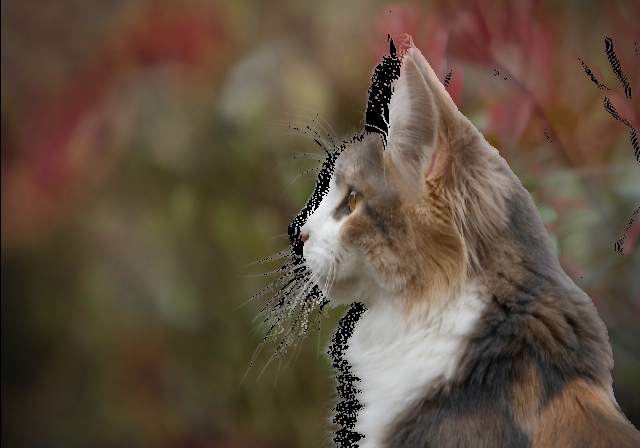}};
            \spy [draw=green] on \zoomfour in node [left] at \rebigone;
            \spy [draw=red] on \zoomfive in node [left] at \rebigtwo;
    	\end{tikzpicture}
     \caption*{Point Render (DAv2)}
    \end{subfigure}
    \begin{subfigure}{\imgWidth}
		\begin{tikzpicture}[spy using outlines={magnification=\ssmag,size=\ssizz},inner sep=0]
            \node [align=center, img] {\includegraphics[width=\textwidth]{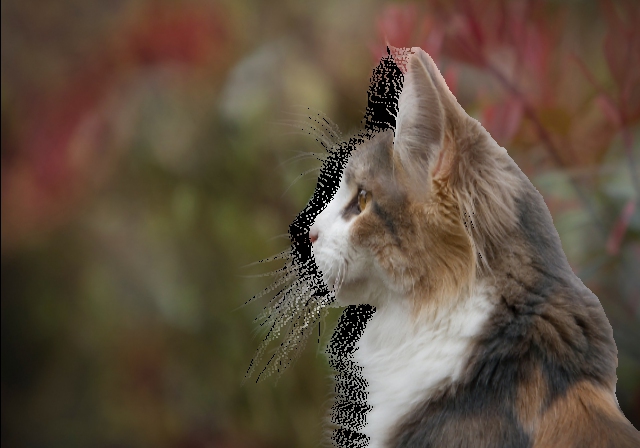}};
            \spy [draw=green] on \zoomfour in node [left] at \rebigone;
            \spy [draw=red] on \zoomfive in node [left] at \rebigtwo;
    	\end{tikzpicture}
     \caption*{Point Render (DPro)}
    \end{subfigure}
    \begin{subfigure}{\imgWidth}
        \begin{tikzpicture}[spy using outlines={magnification=\ssmag,size=\ssizz},inner sep=0]
            \node [align=center, img] {\includegraphics[width=\textwidth]{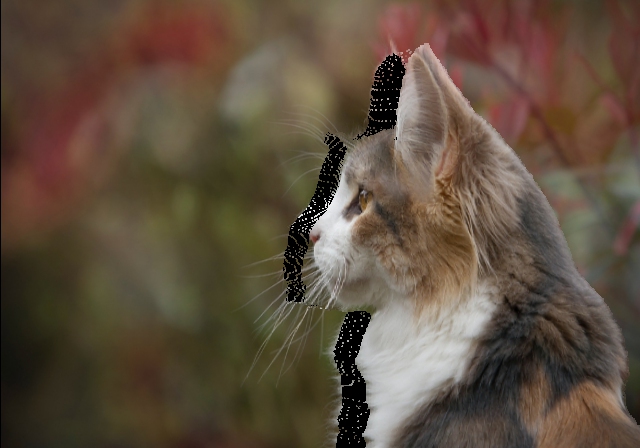}};
            \spy [draw=green] on \zoomfour in node [left] at \rebigone;
            \spy [draw=red] on \zoomfive in node [left] at \rebigtwo;
        \end{tikzpicture}
      \caption*{Point Render (UDv2)}
      \end{subfigure}
      \end{subfigure}
      \vspace{-1.5em}
    \caption{\textbf{Soft boundaries.} Existing depth estimation methods often struggle to capture accurate depth in soft boundaries, resulting in discontinuous depth ({\color{red}red} box) and broken boundaries ({\color{green}green} box). DAv2, DPro, and UDv2 represent Depth Anything V2~\cite{yang2024depthanythingv2}, Depth Pro~\cite{depthpro}, and UniDepthV2~\cite{unidepthv2}, respectively.}
    \vspace{-1em}
    \label{fig:problem_stat_results}
\end{figure}

%% file: sec/relatedwork.tex
\section{Related Work}
\label{sec:relatedwork}

\subsection{Monocular Depth Estimation}\label{subsec:relatedwork-depthestimation}
Monocular depth estimation aims to infer scene geometry from a single image~\cite{chen2016single,chen2020oasis,yin2021leres,zhang2022hdn,ranftl2021dpt,yin2023metric3d,hu2024metric3dv2,piccinelli2024unidepth}, a fundamentally ill-posed problem due to the loss of depth cues during projection. To achieve zero-shot depth estimation, early attempts employ mixed training datasets to obtain a strong geometric prior of the scene~\cite{ranftl2020midas,ranftl2021dpt,yang2024depthanything}. Marigold also proposes to utilize the rich prior knowledge in generative foundation models, \eg, Stable Diffusion~\cite{rombach2022stablediffusion}, to efficiently approach zero-shot estimation~\cite{ke2024marigold}. Recently, several approaches have been proposed to improve the details of depth maps~\cite{zhang2024betterdepth,depthpro,yang2024depthanythingv2}. For example, Depth Anything V2 exploits high-quality depth supervision in synthetic datasets and learns to extract fine details from input images~\cite{yang2024depthanythingv2}. Meanwhile, Depth Pro designs a training protocol to combine real and synthetic datasets for metric depth estimation and fine boundary preservation. The recent approach UniDepthV2 also proposes an edge-guided loss to improve the sharpness of edges in the depth output~\cite{unidepthv2}. Despite these advances, existing methods still struggle in soft boundary regions, often producing missing or discontinuous depth estimates. In contrast, our \myname\ precisely localizes soft boundaries and reconstructs fine-grained depth details, as shown in \cref{fig:problem_stat_results}.

\subsection{Stereo Conversion}\label{subsec:relatedwork-stereoconversion}
The goal of stereo conversion is to generate right-view images/videos from left-view inputs~\cite{xie2016deep3d,mehl2024stereoconv,mono2stereo,dai2024svg,wang2023learning}, which has gained increasing attention due to its practical application in 3D video production and immersive media. With the rapid progress of generative foundation models~\cite{rombach2022stablediffusion,blattmann2023stablevideodiffuion,wan2025wan}, an emerging trend is to utilize learned generative and geometry priors for stereo conversion~\cite{mono2stereo,stereocrafter,geyer2025eye2eye,shvetsova2025m2svid}. For image-based conversion, StereoDiffusion introduces a training-free latent modification strategy using Stable Diffusion~\cite{stereodiffusion}, and Mono2Stereo designs dual conditioning and edge-consistency losses to enhance stereo quality~\cite{mono2stereo}. Recently, an increasing number of works have focused on leveraging video generative models for stereo video conversion~\cite{stereocrafter,geyer2025eye2eye,shvetsova2025m2svid}. 
For instance, StereoCrafter designs a tiled diffusion strategy to generate stereoscopic videos from high-resolution and long video inputs~\cite{stereocrafter}. Based on Stable Video Diffusion~\cite{blattmann2023stablevideodiffuion}, M2SVid devises a spatio-temporal aggregation mechanism to leverage information from neighboring frames and achieves high-quality inpainting performance~\cite{shvetsova2025m2svid}. Eye2Eye~\cite{geyer2025eye2eye} further synthesizes stereo videos without explicit depth projection, effectively handling specular and transparent surfaces using video diffusion priors. 
However, due to the generative nature of diffusion models, current stereo conversion approaches often suffer from texture hallucination and loss of fine details, as shown in \cref{fig:teaser}. To overcome these limitations, a color fuser network is designed in \myname\ to recover high-fidelity texture details.

\subsection{Novel View Synthesis}\label{subsec:relatedwork-novelviewsynthesis}
Novel view synthesis has attracted considerable interest in computer vision community for its ability to render photo-realistic images from novel viewpoints~\cite{wiles2020synsin,wimbauer2023bts,jampani2021slide,szymanowicz2024splatterimage,mildenhall2020nerf,kerbl3Dgs,yu2024viewcrafter,you2024nvssolver}. A popular trend is to perform 3D scene reconstruction from input images for novel view synthesis, such as Multi-Plane Image (MPI)~\cite{li2021mine,tucker2020svmpi,adampi}, Neural Radiance Field (NeRF)~\cite{mildenhall2020nerf,yu2021pixelnerf}, and 3D Gaussian Splatting (3DGS)~\cite{kerbl3Dgs,szymanowicz2024flash3d,xu2024depthsplat}. More recently, diffusion-based approaches have emerged as a powerful alternative, leveraging generative priors to produce high-fidelity novel views without requiring explicit 3D reconstruction~\cite{chan2023gennvs,liu2023zero123,zheng2024free3d,sargent2024zeronvs,seo2024genwarp,gu2025diffusionasshader}. For example, ReCamMaster introduces frame-dimension conditioning to enhance view consistency in video diffusion models~\cite{recammaster}, but its results often suffer from texture inconsistency due to diffusion hallucination (\eg, see \cref{fig:teaser}). Another recent work, SplatDiff, integrates depth-guided pixel splatting with diffusion models to achieve high-fidelity view synthesis~\cite{splatdiff}. However, its performance is highly dependent on the quality of depth, which often contains errors around soft boundaries (\cref{fig:problem_stat_results}). By comparison, our \myname\ combines a depth fixer and a color fuser to jointly correct depth inaccuracies and restore fine-grained texture details, achieving geometrically consistent and photo-realistic novel views (\cref{fig:teaser}).

%% file: sec/approach.tex
\section{\myname}
\label{sec:approach}

\input{figs/main/fig-pipeline}

Following the formulation in image matting~\cite{li2023mattingsurvey,yao2024vitmatte}, the observed image $I$ can be expressed as an alpha composition between the foreground $I_{FG}$ and the background $I_{BG}$, \ie,
\begin{equation}\label{eq:alpha_composition}
    I = \alpha \cdot I_{FG} + (1-\alpha) \cdot I_{BG},
\end{equation}
where $\alpha\in[0,1]$ denotes the opacity map (alpha matte). Soft boundaries can be defined as regions with mixed foreground and background pixels, \ie, $\alpha\in(0,1)$, posing ambiguity in depth and color correspondence. To handle these challenging areas in depth estimation (\cref{subsec:approach-depth}), we design a depth fixer to automatically localize soft boundaries and refine depth predictions, as shown in \cref{fig:pipeline}. For view synthesis tasks (\cref{subsec:approach-viewsyn}), we first perform forward warping based on the fixed depth, and then apply the generative scene painter to fill the unknown regions like disoccluded areas. Finally, our color fuser combines warped and inpainted results for high-quality view synthesis.

\subsection{Depth Estimation}\label{subsec:approach-depth}

\input{figs/main/fig-depth_fixer}

Given an image and its depth map (\eg, estimation results from Depth Anything V2~\cite{yang2024depthanythingv2}), our depth fixer aims to automatically identify soft boundary regions and perform precise depth correction. However, several challenges exist: 
\begin{itemize}
    \item \textit{High-quality annotation.} Most existing depth datasets focus on scenes with hard boundaries, lacking fine-grained depth annotations around soft boundary regions.
    \item \textit{Automatic localization.} Estimation in soft boundaries often relies on hand-crafted cues like trimaps~\cite{yao2024vitmatte}, hindering generalization and applicability to complex scenes.
    \item \textit{Precise refinement.} Achieving precise depth correction in soft boundary regions without compromising the global depth quality remains an open challenge.
\end{itemize}

\noindent \textbf{Dataset Curation.} Collecting large-scale datasets with high-quality depth annotations in soft boundary regions could be time-consuming and impractical. Thus, we address the \textit{high-quality annotation} issue by utilizing the existing image matting datasets, which contain diverse targets with soft boundaries and the corresponding opacity maps (\ie, alpha mattes). As shown in \cref{fig:fixer_data}, we use matting datasets as foreground datasets $\mathcal{I}_{FG}=\{(\alpha,I_{FG})\}$ and image datasets as background datasets $\mathcal{I}_{BG}=\{(I_{BG})\}$. Since alpha mattes usually exhibit smooth transitions in soft boundaries, which are not aligned with depth characteristics, we first obtain alpha masks $M_{\alpha}$ by thresholding $\alpha$ with $\alpha_{th}$, \ie, $M_{\alpha}=\{p\mid\alpha_{th} < \alpha(p)\}$. Then, we generate foreground depth $d_{FG}$ by
\begin{equation}
    d_{FG} = M_{\alpha} \odot \operatorname{Depth}(I_{FG}),
\end{equation}
where $\operatorname{Depth}(\cdot)$ represents depth estimation methods, and a green background is added to $I_{FG}$ to enhance the contrast in depth estimation. Afterward, we obtain the background depth as $d_{BG}=\operatorname{Depth}(I_{BG})$ and randomly sample two depth values from $[ d_{min}, d_{max}]$ to rescale $d_{FG}$ for data augmentation, where $d_{min}=\max_{p\in M_{\alpha}}(d_{BG}(p))$ to ensure correct depth ordering and $d_{max}$ is a predefined constant. Finally, we blend $d_{FG}$ and $d_{BG}$ by depth composition:
\begin{equation}\label{eq:depth_composition}
    d = d_{FG} \odot M_{\alpha} + d_{BG} \odot (1-M_{\alpha}).
\end{equation}
Using \cref{eq:depth_composition}, one can create depth training pairs $\{(d_{in}, d_{GT})\}$ for the depth fixer by varying the threshold $\alpha_{th}$. A lower $\alpha_{th}$ is used to generate depth labels $d_{GT}$ with fine details in soft boundaries, and a higher $\alpha_{th}$ is used to simulate depth inputs $d_{in}$ with broken or missing depth in these regions. Additionally, we apply a random Gaussian blur to $M_{\alpha}$ when generating $d_{in}$, but use the unblurred mask in \cref{eq:depth_composition} to produce $d_{GT}$ with sharp boundaries.

\par
\noindent \textbf{Network Design.} 
As illustrated in \cref{fig:fixer_network}, our depth fixer has two main branches: a feature branch built upon DINOv2~\cite{oquab2023dinov2} and DPT~\cite{ranftl2021dpt} to extract deep features and image semantics, and a pixel branch based on U-Net~\cite{ronneberger2015unet} to capture local structures and boundary details. 
To address the \textit{automatic localization} problem, we propose to infer soft boundaries directly from images and depth maps. In particular, we first generate explicit edge guidance $e$ by applying the Sobel operator to the input depth, \ie, $e=\operatorname{Sobel}(d_{in})$, and then concatenate $e$ with image $I_{in}$ and depth $d_{in}$ as inputs to the pixel branch. With this design, our depth fixer is able to focus on the regions with high depth gradients and learn to automatically identify soft boundaries with image semantics and geometric layouts.

\par 
For \textit{precise refinement}, we propose a gated residual mechanism to refine depth only in soft boundary regions while preserving global depth quality. Specifically, we first model the soft boundary regions by predicting a gate map $G\in[0,1]$, where $G<1$ indicates soft boundary regions. Then, the gated residual is performed to obtain the refined depth $\hat{d}$, 
\begin{equation}
    \hat{d} = d_{in}\cdot G + d_{res} \cdot (1-G),
\end{equation}
where $d_{res}$ is the estimated depth residual. Compared with the direct prediction of refined depth or vanilla residual approach, our gated residual better preserves sharp and fine-grained details in soft boundaries, as shown in \cref{fig:fixer_visuals}. Furthermore, the gating mechanism decouples depth estimation and soft-boundary fixing, and thus our depth fixer can be seamlessly integrated with state-of-the-art depth models to achieve robust and detail-preserving performance.

\par
\noindent \textbf{Model Training.}
Directly training the depth fixer using standard depth losses~\cite{ranftl2020midas} tends to yield a trivial solution where the gate collapses to $G=\mathbf{1}$. Hence, we propose a two-stage strategy to learn depth refinement in a local-to-global manner. We first generate a soft boundary mask $M_{soft}$ by thresholding the ground-truth alpha matte, \ie, $M_{soft}= \{p \mid \alpha_{min} < \alpha(p) < \alpha_{max}\}$, with constants $\alpha_{min},\alpha_{max}$ determining the soft boundary areas. The learning objective for the first stage $\mathcal{L}^{stage1}_{depth}$ is defined as
\begin{equation}
    \mathcal{L}^{stage1}_{depth} = \mathcal{L}_1(\hat{d},d_{GT}) + \mathcal{L}_{\alpha}(\hat{d}\odot M_{soft},d_{GT}\odot M_{soft}),
\end{equation}
where $\mathcal{L}_1$ denotes the $\ell_1$ loss, and $\mathcal{L}_{\alpha}$ is an image matting loss from ViTMatte~\cite{yao2024vitmatte} to facilitate detail extraction. Although $\mathcal{L}^{stage1}_{depth}$ prevents the trivial solution of $G=\mathbf{1}$ by imposing stronger penalties on soft boundaries, it often introduces halo artifacts around these regions, as illustrated in \cref{fig:fixer_visuals}. Thus, in the second stage, we apply the constraint $\mathcal{L}_{\alpha}$ globally to improve the overall depth quality, \ie,
\begin{equation}
    \mathcal{L}^{stage2}_{depth} = \mathcal{L}_{\alpha}(\hat{d},d_{GT}).
\end{equation}
\cref{fig:fixer_visuals} shows that our two-stage training achieves fine-grained details while preserving global depth quality.

\input{figs/main/fig-color_fuser}

\input{tabs/main/tab-depth-edge}

\subsection{View Synthesis}\label{subsec:approach-viewsyn}

For view synthesis, we first perform forward warping based on the fixed depth to preserve fine details and soft boundaries. However, the mixture of foreground and background in soft boundaries often introduces redundant background colors in the warped results (\eg, see the green box in \cref{fig:fuser_visual}). Since existing multi-view datasets mainly contain hard boundaries, we propose to model such characteristics by leveraging matting datasets in data curation (\cref{fig:fuser_data}).

\noindent \textbf{Dataset Curation.}
Given a sequence of background images $\{I_{BG}\}$ from multi-view datasets, we first predict the optical flow $\{f_{BG}\}$ between all pairs of images via an off-the-shelf optical flow estimator~\cite{xu2023unimatch}. To synthesize the warped results of soft boundaries, we sample a foreground image $I_{FG}$ from the matting dataset, and generate the foreground flow $f_{FG}$ with a random displacement vector $(u,v)$ for all pixels, ensuring purely translational motion within the image plane. Then, we perform flow composition using alpha mask $M_{\alpha}$,
\begin{equation}
    f = f_{FG} \odot M_{\alpha} + f_{BG} \odot (1-M_{\alpha}).
\end{equation}
Since the foreground only moves within the image plane, ground-truth views $\{I_{GT}\}$ can be easily synthesized by applying \cref{eq:alpha_composition} to the foreground $I_{FG}$ and background images $\{I_{BG}\}$. Although the foreground motions are relatively simple, the background regions preserve realistic viewpoint changes and complex camera motions for robust training. Finally, we perform forward warping with flows $\{f\}$ to generate the warped images and masks, and fine-tune our scene painter to fit the characteristics of soft boundaries while inpainting disoccluded regions. The aligned synthesis strategy in SplatDiff~\cite{splatdiff} is also applied to the background regions for precise viewpoint control.

\par
\noindent \textbf{Color Fuser.}
Although the scene painter is able to eliminate redundant background in soft boundaries, its generative nature tends to hallucinate inconsistent texture details (\eg, see the red box in \cref{fig:fuser_visual}).
To this end, we propose the color fuser to adaptively combine warped and inpainted images. As shown in \cref{fig:fuser_network}, we build the color fuser upon a pre-trained Variational Auto-Encoder (VAE)~\cite{podell2023sdxl} to harness its reconstruction prior. Since VAE models often suffer from detail compression~\cite{splatdiff}, a dual skip module is designed to propagate fine-grained features for fusion. Specifically, we first extract multi-scale features of the inpainted and warped images via a frozen VAE encoder. These features are then concatenated with the warped masks and fed into the VAE decoder to compensate for texture details. Based on our curated view synthesis dataset, we further synthesize input inpainted images with hallucinated textures by applying the scene painter to ground-truth images $\{I_{GT}\}$. Finally, we fine-tune the VAE decoder using the following objective:
\begin{equation}
    \mathcal{L}_{color} = \mathcal{L}_1(\hat{I},I_{GT}) + \lambda \cdot \mathcal{L}_{lpips}(\hat{I},I_{GT}),
\end{equation}
where the balancing parameter $\lambda=0.1$, $\hat{I}$ denotes the outputs of the color fuser, and $\mathcal{L}_{lpips}$ indicates perceptual loss~\cite{zhang2018lpips}. Compared with the warped and inpainted results, our color fuser produces the best novel views with high-quality texture and geometry (\cref{fig:fuser_visual}).

%% file: figs/main/fig-pipeline.tex
\begin{figure}[t]
  \centering
  \includegraphics[width=\linewidth]{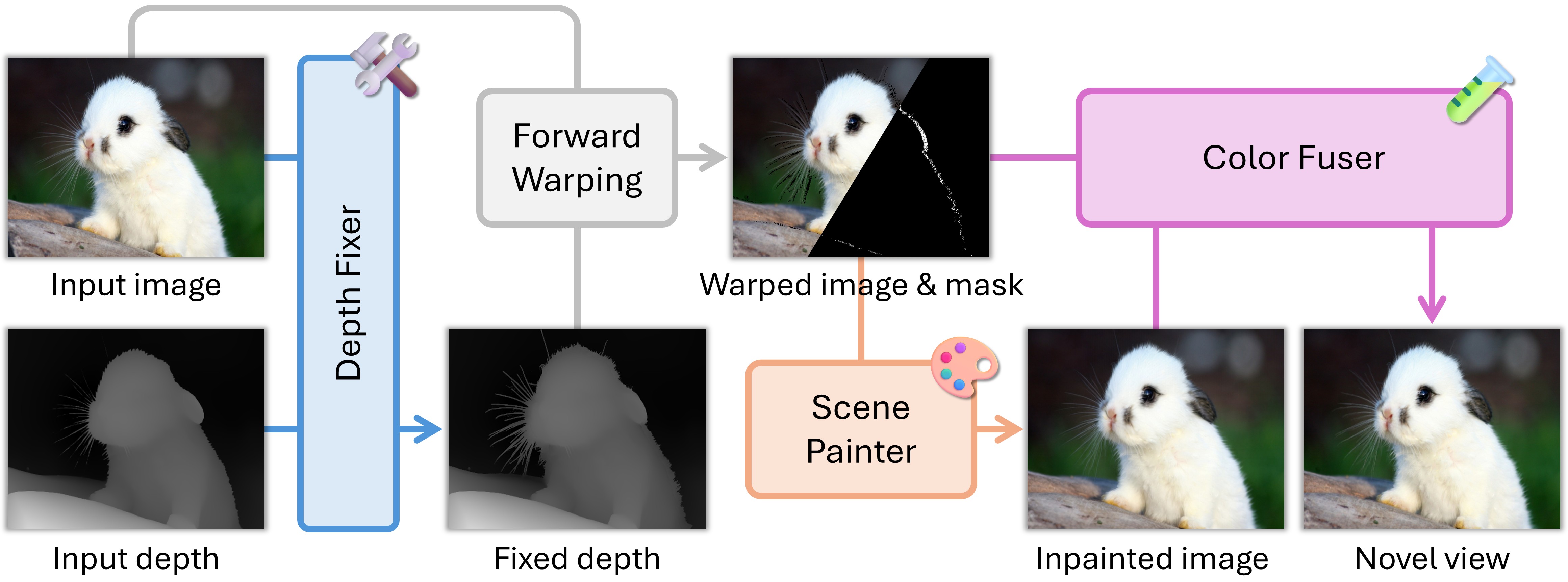}
  \vspace{-1.5em}
  \caption{\textbf{\myname\ pipeline.} Given an input image and its estimated depth, we first design a depth fixer to refine depth predictions around soft boundary regions. The fixed depth is then used for forward warping to generate preliminary novel views, which are fed into the scene painter for disocclusion inpainting. Finally, our color fuser adaptively combines the warped and inpainted results to produce geometrically and visually consistent novel views.}
  \vspace{-1em}
  \label{fig:pipeline}
\end{figure}

%% file: figs/main/fig-depth_fixer.tex
\def\imgWidth{0.32\linewidth} %
\def\scc{(-1.9,-1.4)}

\def\rebigone{(-0.5, 0.5)} %
\def\rebigtwo{(-0.5, 0.5)} %

\def\zoomone{(-0.33,0.4)} %

\def\ssizz{0.8cm} %
\def\ssmag{3}

\begin{figure}[t]
  \centering
  \tikzstyle{img} = [rectangle, minimum width=\imgWidth, draw=black]
  \begin{subfigure}[b]{\linewidth}
  \includegraphics[width=\linewidth]{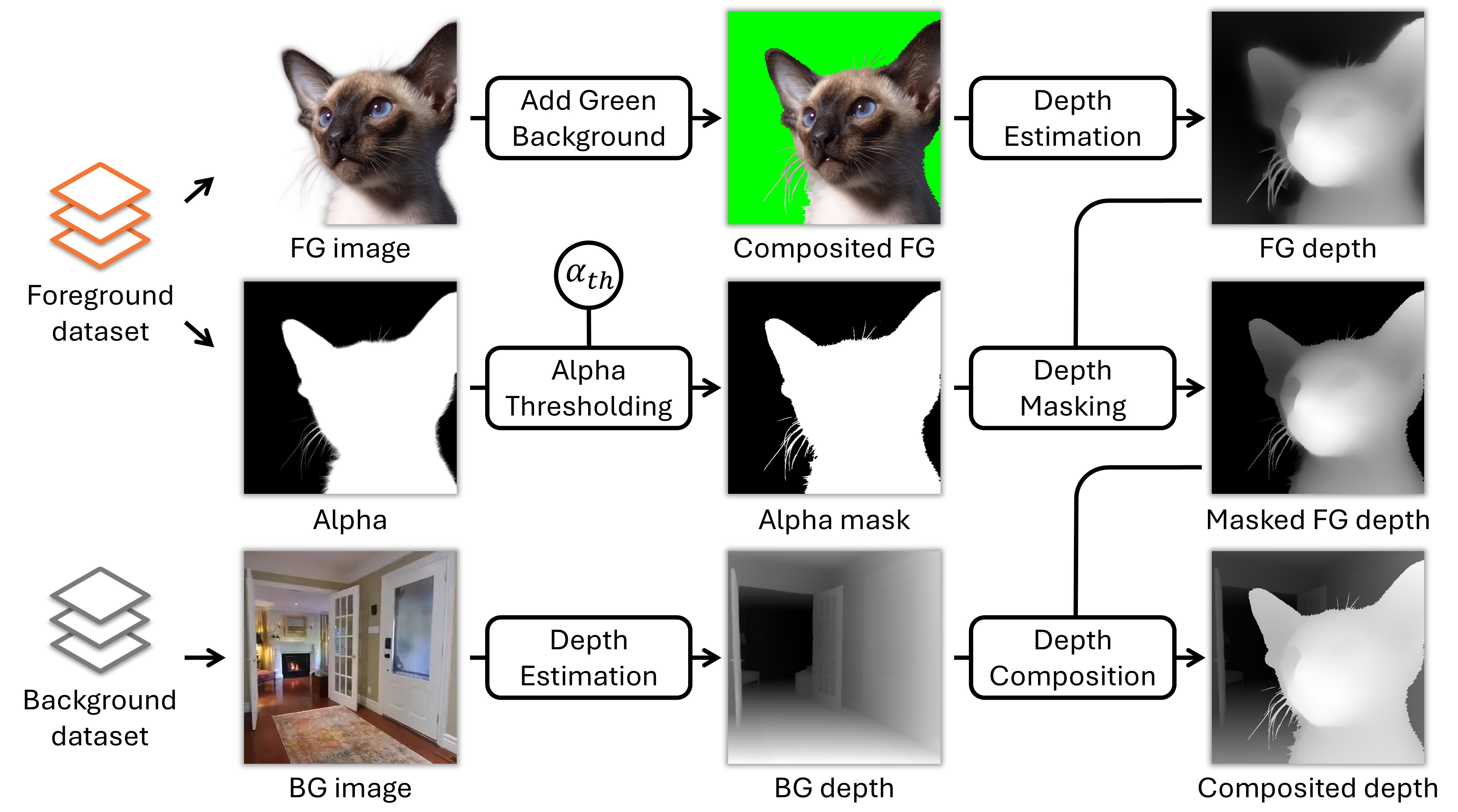}
  \caption{Training data curation with matting datasets}
  \label{fig:fixer_data}
  \end{subfigure}
  \begin{subfigure}[b]{\linewidth}
   \includegraphics[width=\linewidth]{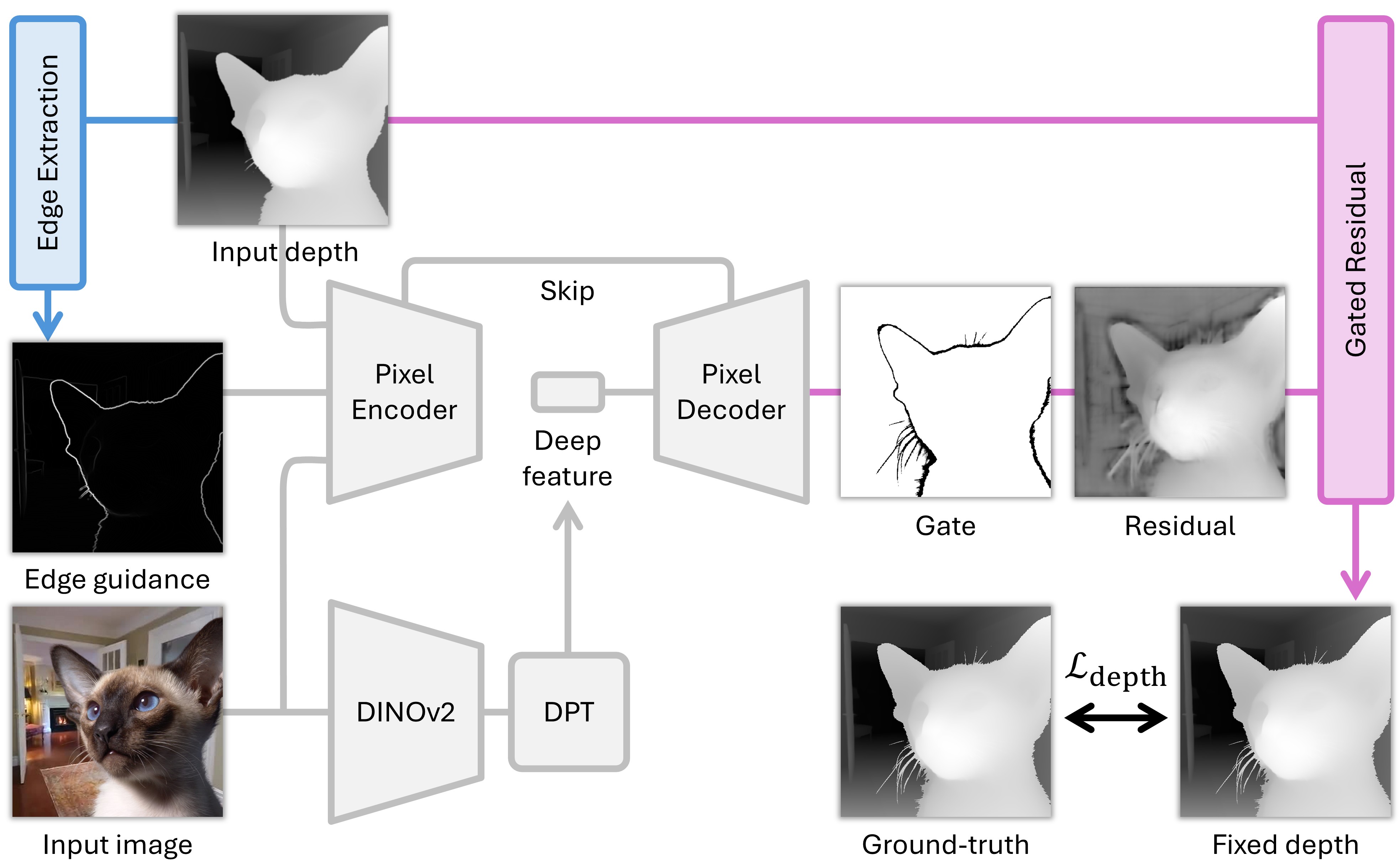}
  \caption{Network architecture}
  \label{fig:fixer_network}
  \end{subfigure}
  \begin{subfigure}{\linewidth}
  \centering   
      \begin{subfigure}{\imgWidth}
        \begin{tikzpicture}[spy using outlines={magnification=\ssmag,size=\ssizz},inner sep=0]
            \node [align=center, img] {\includegraphics[width=\textwidth]{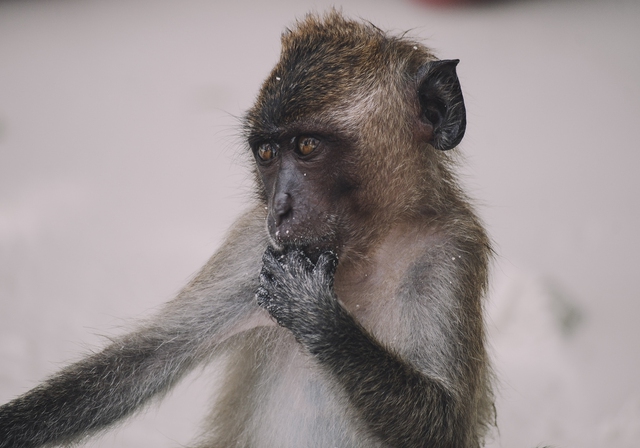}};
            \spy [draw=green] on \zoomone in node [left] at \rebigone;
    	\end{tikzpicture}
     \caption*{Input Image}
    \end{subfigure}
    \begin{subfigure}{\imgWidth}
		\begin{tikzpicture}[spy using outlines={magnification=\ssmag,size=\ssizz},inner sep=0]
            \node [align=center, img] {\includegraphics[width=\textwidth]{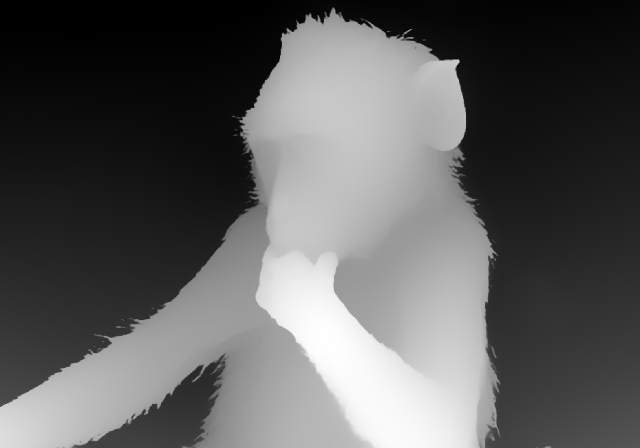}};
            \spy [draw=green] on \zoomone in node [left] at \rebigone;
    	\end{tikzpicture}
     \caption*{Input Depth}
    \end{subfigure}
    \begin{subfigure}{\imgWidth}
        \begin{tikzpicture}[spy using outlines={magnification=\ssmag,size=\ssizz},inner sep=0]
            \node [align=center, img] {\includegraphics[width=\textwidth]{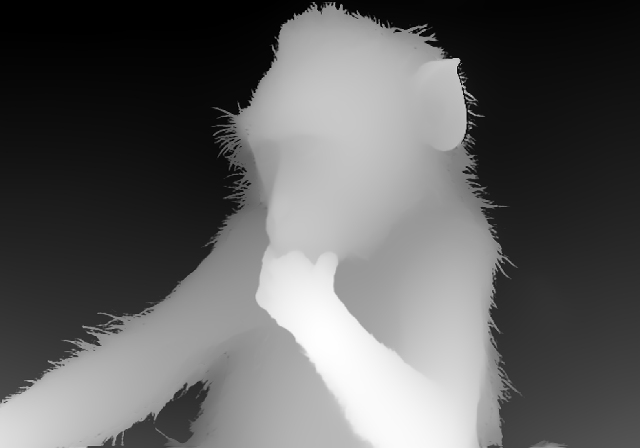}};
            \spy [draw=green] on \zoomone in node [left] at \rebigone;
    	\end{tikzpicture}
      \caption*{Ours}
      \end{subfigure}
      \\ %
       \begin{subfigure}{\imgWidth}
        \begin{tikzpicture}[spy using outlines={magnification=\ssmag,size=\ssizz},inner sep=0]
            \node [align=center, img] {\includegraphics[width=\textwidth]{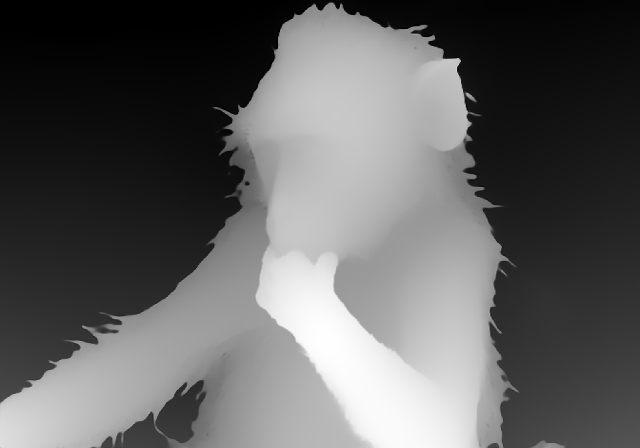}};
            \spy [draw=green] on \zoomone in node [left] at \rebigone;
    	\end{tikzpicture}
     \caption*{Direct Prediction }
    \end{subfigure}
    \begin{subfigure}{\imgWidth}
		\begin{tikzpicture}[spy using outlines={magnification=\ssmag,size=\ssizz},inner sep=0]
            \node [align=center, img] {\includegraphics[width=\textwidth]{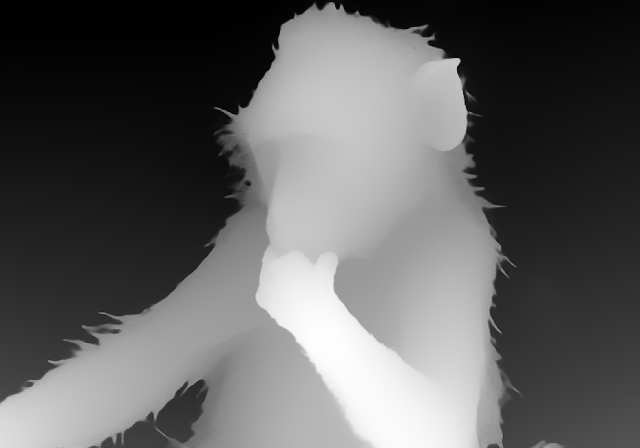}};
            \spy [draw=green] on \zoomone in node [left] at \rebigone;
    	\end{tikzpicture}
     \caption*{Vanilla Residual}
    \end{subfigure}
    \begin{subfigure}{\imgWidth}
        \begin{tikzpicture}[spy using outlines={magnification=\ssmag,size=\ssizz},inner sep=0]
            \node [align=center, img] {\includegraphics[width=\textwidth]{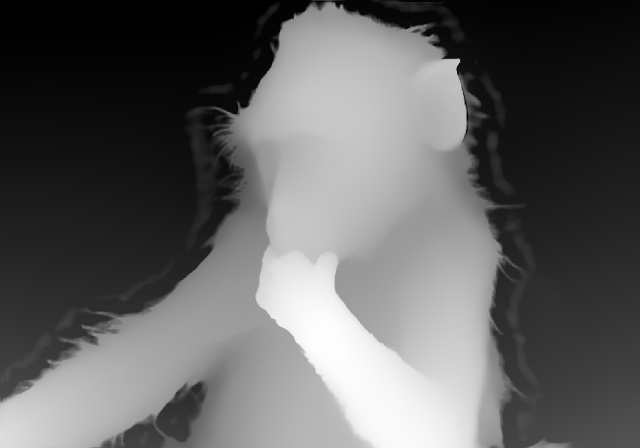}};
            \spy [draw=green] on \zoomone in node [left] at \rebigone;
        \end{tikzpicture}
      \caption*{One-Stage Training}
      \end{subfigure}
      \caption{Comparisons of output mechanisms and training strategies}
      \label{fig:fixer_visuals}
      \end{subfigure}
      \vspace{-1.5em}
  \caption{\textbf{Depth fixer.} (a) We utilize image matting datasets to synthesize training data with fine-grained depth labels in soft boundaries. (b) Instead of relying on manually crafted cues like trimaps~\cite{yao2024vitmatte}, we leverage depth maps and image semantics to automatically identify soft boundary regions. The gated residual module enables precise depth correction in soft boundary areas and thus benefits plug-and-play refinement. (c) Compared with direct prediction and vanilla residual, our gated residual combined with two-stage training achieves the best depth results. }
  \vspace{-1em}
  \label{fig:depth_fixer}
\end{figure}

%% file: figs/main/fig-color_fuser.tex
\def\imgWidth{0.32\linewidth} %
\def\scc{(-1.9,-1.4)}

\def\rebigone{(0.55, -0.55)} %
\def\rebigtwo{(1.3, -0.55)} %

\def\zoomone{(-0.9,-0.75)} %
\def\zoomtwo{(1.05,0.4)} %

\def\ssizz{0.7cm} %
\def\ssmag{3}

\begin{figure}[t]
  \centering
  \tikzstyle{img} = [rectangle, minimum width=\imgWidth, draw=black]
  \begin{subfigure}[b]{\linewidth}
  \includegraphics[width=\linewidth]{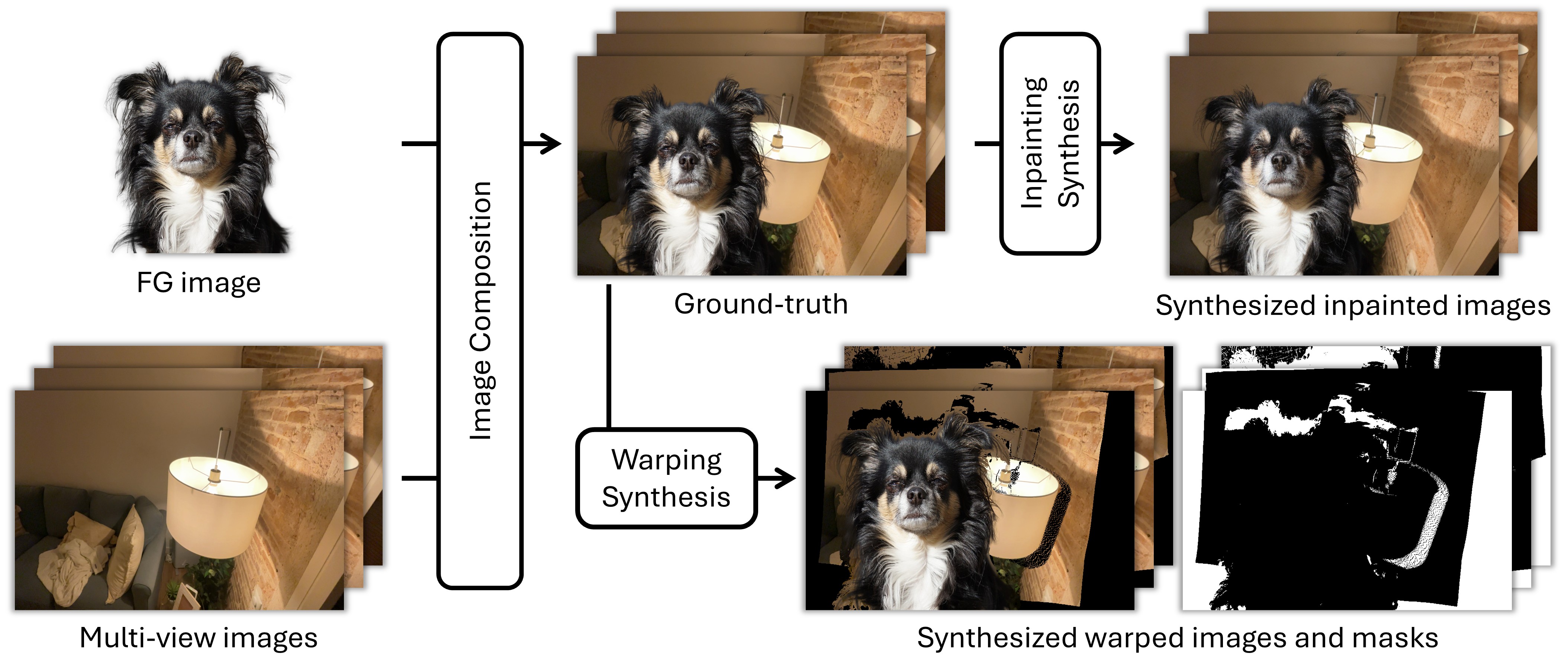}
  \caption{Training data curation with matting datasets}
  \label{fig:fuser_data}
  \end{subfigure}
  \begin{subfigure}[b]{\linewidth}
   \includegraphics[width=\linewidth]{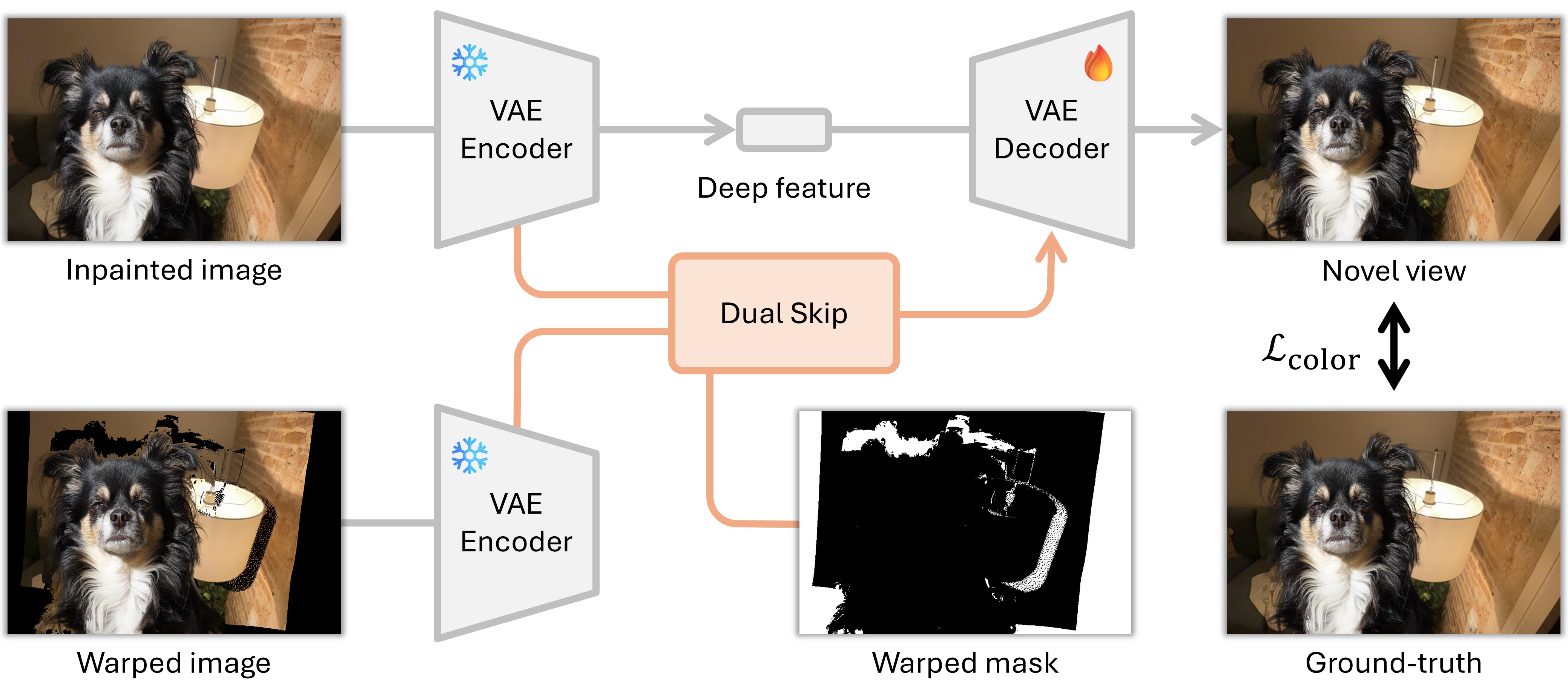}
  \caption{Network architecture}
  \label{fig:fuser_network}
  \end{subfigure}
  \begin{subfigure}{\linewidth}
  \centering   
      \begin{subfigure}{\imgWidth}
        \begin{tikzpicture}[spy using outlines={magnification=\ssmag,size=\ssizz},inner sep=0]
            \node [align=center, img] {\includegraphics[width=\textwidth]{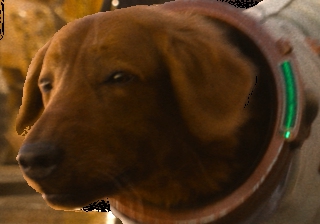}};
            \spy [draw=green] on \zoomone in node [left] at \rebigone;
             \spy [draw=red] on \zoomtwo in node [left] at \rebigtwo;
    	\end{tikzpicture}
     \caption*{Warped Image}
    \end{subfigure}
    \begin{subfigure}{\imgWidth}
		\begin{tikzpicture}[spy using outlines={magnification=\ssmag,size=\ssizz},inner sep=0]
            \node [align=center, img] {\includegraphics[width=\textwidth]{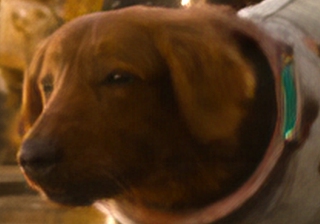}};
            \spy [draw=green] on \zoomone in node [left] at \rebigone;
             \spy [draw=red] on \zoomtwo in node [left] at \rebigtwo;
    	\end{tikzpicture}
     \caption*{Inpainted Image}
    \end{subfigure}
    \begin{subfigure}{\imgWidth}
        \begin{tikzpicture}[spy using outlines={magnification=\ssmag,size=\ssizz},inner sep=0]
            \node [align=center, img] {\includegraphics[width=\textwidth]{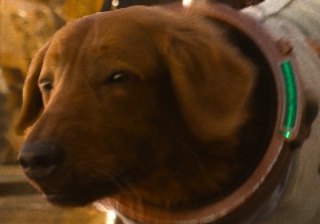}};
            \spy [draw=green] on \zoomone in node [left] at \rebigone;
             \spy [draw=red] on \zoomtwo in node [left] at \rebigtwo;
    	\end{tikzpicture}
      \caption*{Fused Image}
      \end{subfigure}
      \caption{Comparisons of warped, inpainted, and fused images}
      \label{fig:fuser_visual}
    \end{subfigure}
    \vspace{-1.5em}
  \caption{\textbf{Color fuser.} (a) We employ image matting datasets and multi-view datasets to synthesize warped and inpainted results for training. (b) Built upon a pre-trained VAE, we design a dual skip module to leverage the merits of inapinted and warped images. (c) Our color fuser eliminates redundant background colors (\textcolor{green}{green} box in the warped image) and hallucinated textures (\textcolor{red}{red} box in the inpainted image) for high-quality view synthesis.
  }
  \vspace{-1em}
  \label{fig:color_fuser}
\end{figure}

%% file: tabs/main/tab-depth-edge.tex
\begin{table*}[th]
\centering
\caption{\textbf{Zero-shot depth boundary accuracy} on the natural image matting datasets. Depth fixer can be integrated with different depth models in a plug-and-play manner for soft boundary refinement. \textcolor{red}{Best} results are marked. Please see the supplementary for more evaluation. }
\vspace{-0.8em}
\label{tab:depth-boundary}
\footnotesize
\setlength\tabcolsep{3pt}
\begin{tabular}{lccccccccc}
\toprule
\rowcolor{color3}                                       & \multicolumn{4}{c}{\textbf{AIM-500}}                                                                                             &  & \multicolumn{4}{c}{\textbf{P3M-10K}}                                                                                             \\ \cline{2-5} \cline{7-10}
\rowcolor{color3} \multirow{-2}{*}{\textbf{Method}}      & DBE\_comp $\downarrow$                 & DBE\_acc $\downarrow$                  & EP (\%) $\uparrow$                    & ER (\%) $\uparrow$                    &  & DBE\_comp $\downarrow$                 & DBE\_acc $\downarrow$                  & EP (\%) $\uparrow$                    & ER (\%) $\uparrow$                    \\ \midrule
Depth Anything V2~\cite{yang2024depthanythingv2}                                   & 7.93                        & 3.29                        & 19.90                        & 6.50                         &  & 7.53                        & 2.60                        & 26.53                        & 9.37                         \\
\rowcolor{color3} \textbf{Depth Anything V2+Depth Fixer (Ours)}       & {\color{red} 7.19} & {\color{red} 2.10} & {\color{red} 34.56} & {\color{red} 13.08} &  & {\color{red} 7.21} & {\color{red} 1.93} & {\color{red} 36.91} & {\color{red} 13.39} \\ \midrule
Depth Pro~\cite{depthpro}                              & 7.75                        & 3.80                        & 15.92                        & 6.12                         &  & 7.25                        & 3.25                        & 18.36                        & 9.21                         \\
\rowcolor{color3} \textbf{Depth Pro+Depth Fixer (Ours)}  & {\color{red} 6.70} & {\color{red} 2.30} & {\color{red} 35.01} & {\color{red} 17.33} &  & {\color{red} 6.44} & {\color{red} 1.78} & {\color{red} 37.90} & {\color{red} 18.91} \\ \midrule
UniDepthV2~\cite{unidepthv2}                             & 8.34                        & 3.87                        & 19.52                        & 5.14                         &  & 7.73                        & 3.48                        & 20.82                        & 8.12                         \\
\rowcolor{color3} \textbf{UniDepthV2+Depth Fixer (Ours)} & {\color{red} 7.49} & {\color{red} 2.71} & {\color{red} 33.06} & {\color{red} 10.98} &  & {\color{red} 6.89} & {\color{red} 2.05} & {\color{red} 37.71} & {\color{red} 15.12} \\ \bottomrule
\end{tabular}
\vspace{-1em}
\end{table*}

%% file: sec/experiments.tex
\section{Experiments and Analysis}
\label{sec:experiments}

\input{figs/main/fig-exp_depth_est_visual}
\input{tabs/main/tab-depth-zeroshot}

\subsection{Experimental Settings}\label{subsec:experiments-experimentalsettings}
\noindent \textbf{Implementation Details.}
For depth fixer, we curate our training dataset with $\alpha_{min}=0.02$ and $\alpha_{max}=0.98$. We use $\alpha_{th}=\alpha_{min}$ when generating ground-truth depth, and randomly sample $\alpha_{th}\sim \mathcal{U}(\alpha_{min}, \alpha_{max})$ when synthesizing input depth. We implement the depth fixer with Depth Anything V2~\cite{yang2024depthanythingv2} weight initialization for the feature branch. Depth fixer is trained with AdamW optimizer~\cite{loshchilov2018adamw} under $448\times448$ patches, batch size 32,  and $1\times10^{-5}$ learning rate for 35K iterations for both stages.
For scene painter, we employ the pretrained VACE model~\cite{jiang2025vace} based on Wan2.1-1.3B~\cite{wan2025wan}, and fine-tune it under $480\times832$ resolution, batch size 4, and $1\times10^{-5}$ learning rate for 10K iterations.
Regarding the color fuser, we add extra residual blocks in VAE decoder to blend the features from dual skip module, and fine-tune under $448\times448$ patches, batch size 16,  and $1\times10^{-5}$ learning rate for 35K iterations. The total training takes 4 days on 4 NVIDIA RTX A6000 GPUs.

\par 
\noindent \textbf{Datasets.}
For training, we employ two multi-view datasets as background datasets: \textit{RealEstate10K}~\cite{re10k} and \textit{DL3DV-10K}~\cite{ling2024dl3dv}, and three image matting datasets as foreground datasets: \textit{AM-2K}~\cite{li2020am2k}, \textit{Distinctions-646}~\cite{Qiao_2020Distinctions_646}, and \textit{Composition-1K}~\cite{xu2017composition_1k}. For evaluation, we created a \textit{Marvel-10K} dataset composed of 501 stereo videos from Marvel movies, with a total of 12,525 stereo pairs. We also use 5 public depth estimation benchmarks for zero-shot evaluation: \textit{NYUv2}~\cite{SilbermanECCV12nyu}, \textit{KITTI}~\cite{Geiger2012kitti}, \textit{ETH3D}~\cite{schops2017eth3d}, \textit{ScanNet}~\cite{dai2017scannet}, and \textit{DIODE}~\cite{diode_dataset}. In addition, two natural image matting datasets \textit{AIM-500}~\cite{aim_500} and \textit{P3M-10K}~\cite{li2021p3m_10k} are employed to evaluate the real-world performance of \myname.

\subsection{Depth Estimation}\label{subsec:experiments-depthestimation}
We apply depth fixer to improve 3 state-of-the-art depth estimation models: Depth Anything V2~\cite{yang2024depthanythingv2}, Depth Pro~\cite{depthpro}, and UniDepthV2~\cite{unidepthv2} in a plug-and-play manner, and evaluate their performance in terms of depth boundary accuracy and zero-shot depth estimation.

\noindent \textbf{Boundary Accuracy.}
Following Depth Pro~\cite{depthpro}, we employ image matting datasets to evaluate depth accuracy in soft boundaries. Edge-based metrics, \ie, the completeness and accuracy of depth boundaries (DBE\_comp and DBE\_acc)~\cite{koch2018evaluation} and the edge precision and recall (EP and ER)~\cite{hu2019revisiting}, are used to evaluate depth results in soft boundary regions. Our depth fixer better captures fine-grained depth details in soft boundaries (see depth and point cloud results in \cref{fig:exp_quali_res}), and consistently yields significant improvements when integrated with different depth models (\cref{tab:depth-boundary}).

\noindent \textbf{Zero-Shot Performance.} We further test the robustness of the depth fixer on 5 unseen public datasets. Although these datasets rarely contain soft boundaries, \cref{tab:depth-zeroshot} shows that our depth fixer still achieves comparable or slightly better performance under in-the-wild settings. Thanks to the proposed gated residual module, our depth fixer can adaptively fix depth in soft boundaries while maintaining the zero-shot performance of the base depth model, allowing seamless plug-and-play integration with current and future state-of-the-art depth models.

\subsection{Stereo Conversion}\label{subsec:experiments-stereoconversion}

\input{tabs/main/tab-stereo-marvel}

Since stereo conversion is widely applied in film production,
we compare \myname\ with state-of-the-art stereo conversion and novel view synthesis approaches on the Marvel-10K dataset, which features challenging cinematic scenes and talking heads with complex hair structures. Pixel-level metrics (PSNR, SSIM, and RMSE), feature-level metrics (LPIPS~\cite{zhang2018lpips} and DISTS~\cite{dists}), and the stereo metric SIoU~\cite{mono2stereo} are used for evaluation.

\noindent \textbf{Benchmarking on Marvel-10K.} We compare the performance of stereo image conversion (1 frame per sequence) and stereo video conversion (all sequence frames) in \cref{tab:stereo-marvel}. Although our \myname\ focuses mainly on improving soft boundaries, which usually occupy small regions in images, it consistently outperforms existing approaches in all metrics. In addition, \cref{fig:stereo_video_consistency} verifies the superior performance and temporal consistency of \myname\ compared with previous video-based methods.

\noindent \textbf{Ablation Study.} \cref{tab:stereo-ablation} shows the contribution of each component in our \myname. 
We first estimate depth with Depth Anything V2~\cite{yang2024depthanythingv2} and use its warped results as the baseline (\#1).
By fixing depth in soft boundaries, depth fixer achieves better warping performance with higher SIoU (\#2 \vs \#1). Scene painter largely improves perceptual quality by filling disoccluded regions, but suffers from detail compression and texture hallucination (better LPIPS and worse PSNR in \#3). By adaptively combining warped and inpainted images via the color fuser, \myname\ achieves the best results with high-quality textures and stereo effects.

\input{figs/main/fig-stereo_video_consistency}

\input{tabs/main/tab-stereo-ablation}

\input{tabs/main/tab-nvs-metrics}

\subsection{Novel View Synthesis}\label{subsec:experiments-novelviewsynthesis}

\noindent \textbf{Benchmarking on Matting Datasets.} We employ 2 natural image matting datasets to evaluate the novel view synthesis performance on scenes with soft boundaries. Since ground-truth views are not available, we adopt FID~\cite{heusel2017fid} and the average CLIP similarity of
adjacent frames (CLIP-F)~\cite{radford2021clip} for quantitative evaluation. \cref{fig:exp_quali_res} and \cref{tab:nvs-metrics} verify the state-of-the-art performance of \myname\ in real-world scenarios.

\par 
\noindent \textbf{User Study.}
We conducted a user study with 27 participants on the full evaluation sets of AIM-500 and P3M-10K datasets (1000 natural images in total, no hand-picked samples). The participants will see side-by-side novel view video results, vote for their preferred one, and indicate if it is a strong preference. 1332 votes are collected in total, and the results in \cref{fig:nvs_user_study} verify the superiority of \myname.

\input{figs/main/fig-user_study}

%% file: figs/main/fig-exp_depth_est_visual.tex
\def\imgWidth{0.32\linewidth} %
\def\depthWidth{0.19\linewidth} %
\def\pointWidth{0.24\linewidth} %
\def\scc{(-1.9,-1.4)}

\def\rebigone{(-0.5, -0.55)} %
\def\rebigtwo{(-0.9, -0.73)} %
\def\rebigthree{(-0.93, -0.85)} %

\def\zoomone{(-0.6,0.95)} %
\def\zoomtwo{(-0.7,1.1)} %

\def\zoomthree{(-0.85,1.25)} %
\def\zoomfour{(-1,1.25)} %
\def\zoomfive{(-0.93,1.25)} %

\def\ssizz{1.1cm} %
\def\ssmag{3}

\begin{figure*}[t]
\centering
\tikzstyle{img} = [rectangle, minimum width=\imgWidth]
    \centering
    \begin{subfigure}{\linewidth}
    \centering
    \begin{subfigure}{\depthWidth}
        \begin{tikzpicture}[spy using outlines={green,magnification=\ssmag,size=\ssizz},inner sep=0]
            \node [align=center, img] {\includegraphics[width=\textwidth]{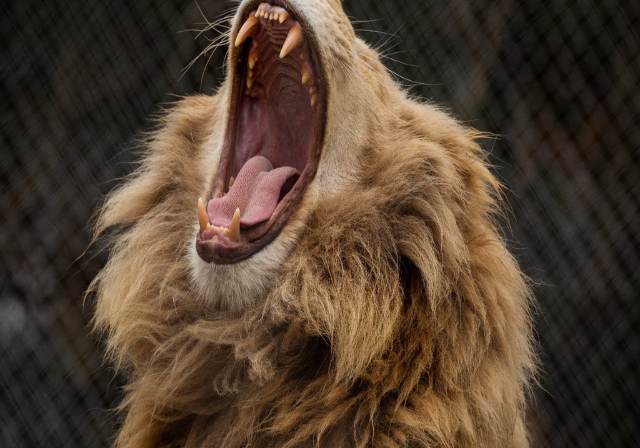}};
            \spy on \zoomone in node [left] at \rebigone;
    	\end{tikzpicture}
        \caption*{Input image}
    \end{subfigure}
    \begin{subfigure}{\depthWidth}
		\begin{tikzpicture}[spy using outlines={green,magnification=\ssmag,size=\ssizz},inner sep=0]
            \node [align=center, img] {\includegraphics[width=\textwidth]{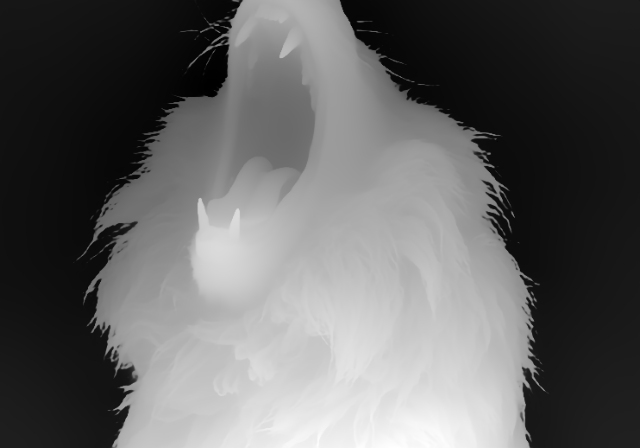}};
            \spy on \zoomone in node [left] at \rebigone;
    	\end{tikzpicture}
        \caption*{Depth Anything V2}
    \end{subfigure}
    \begin{subfigure}{\depthWidth}
        \begin{tikzpicture}[spy using outlines={green,magnification=\ssmag,size=\ssizz},inner sep=0]
            \node [align=center, img] {\includegraphics[width=\textwidth]{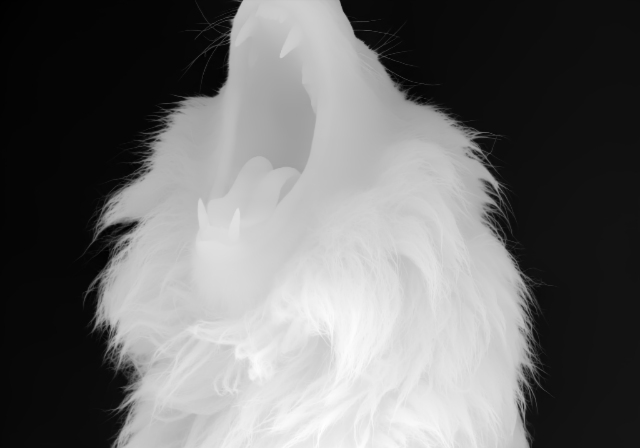}};
            \spy on \zoomone in node [left] at \rebigone;
    	\end{tikzpicture}
        \caption*{Depth Pro}
      \end{subfigure}
    \begin{subfigure}{\depthWidth}
        \begin{tikzpicture}[spy using outlines={green,magnification=\ssmag,size=\ssizz},inner sep=0]
            \node [align=center, img] {\includegraphics[width=\textwidth]{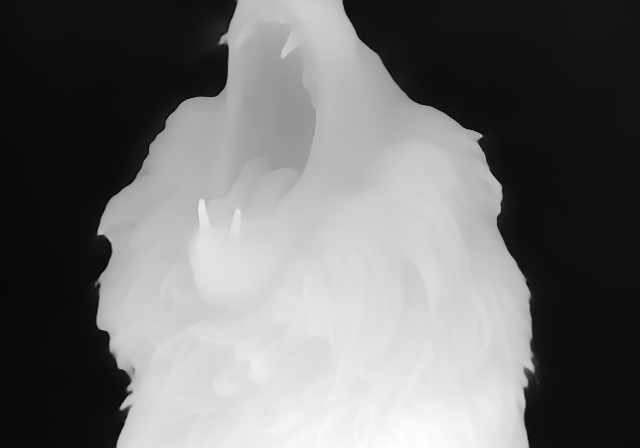}};
            \spy on \zoomone in node [left] at \rebigone;
    	\end{tikzpicture}
        \caption*{UniDepthV2}
      \end{subfigure}
    \begin{subfigure}{\depthWidth}
        \begin{tikzpicture}[spy using outlines={green,magnification=\ssmag,size=\ssizz},inner sep=0]
            \node [align=center, img] {\includegraphics[width=\textwidth]{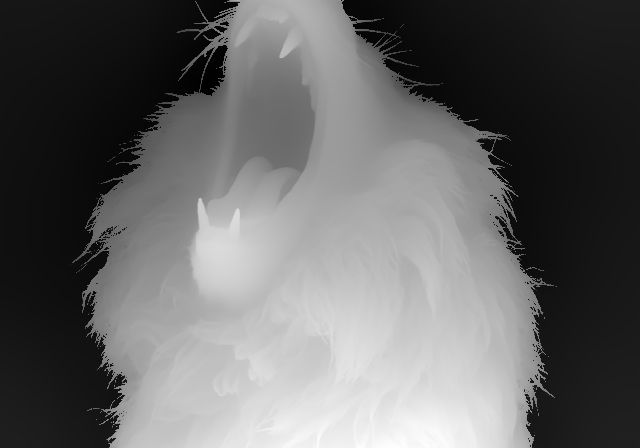}};
            \spy on \zoomone in node [left] at \rebigone;
    	\end{tikzpicture}
        \caption*{Ours}
      \end{subfigure}
    \end{subfigure}
    \begin{subfigure}{\linewidth}
    \centering
         \begin{subfigure}{\pointWidth}
        \begin{tikzpicture}[spy using outlines={green,magnification=\ssmag,size=\ssizz},inner sep=0]
            \node [align=center, img] {\includegraphics[width=\textwidth,trim={35 160 15 130}, clip]{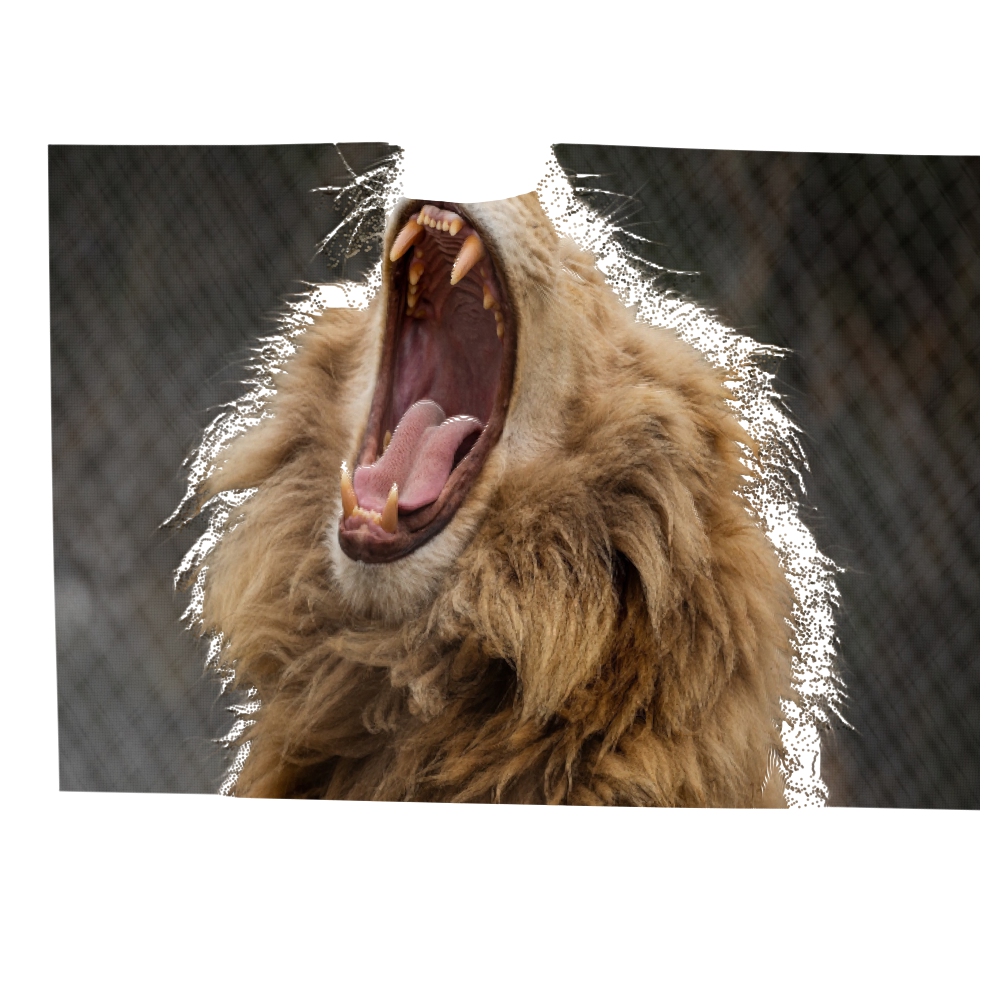}};
            \spy on \zoomtwo in node [left] at \rebigtwo;
    	\end{tikzpicture}
        \caption*{Depth Anything V2}
      \end{subfigure}
      \begin{subfigure}{\pointWidth}
        \begin{tikzpicture}[spy using outlines={green,magnification=\ssmag,size=\ssizz},inner sep=0]
            \node [align=center, img] {\includegraphics[width=\textwidth,trim={35 160 15 130}, clip]{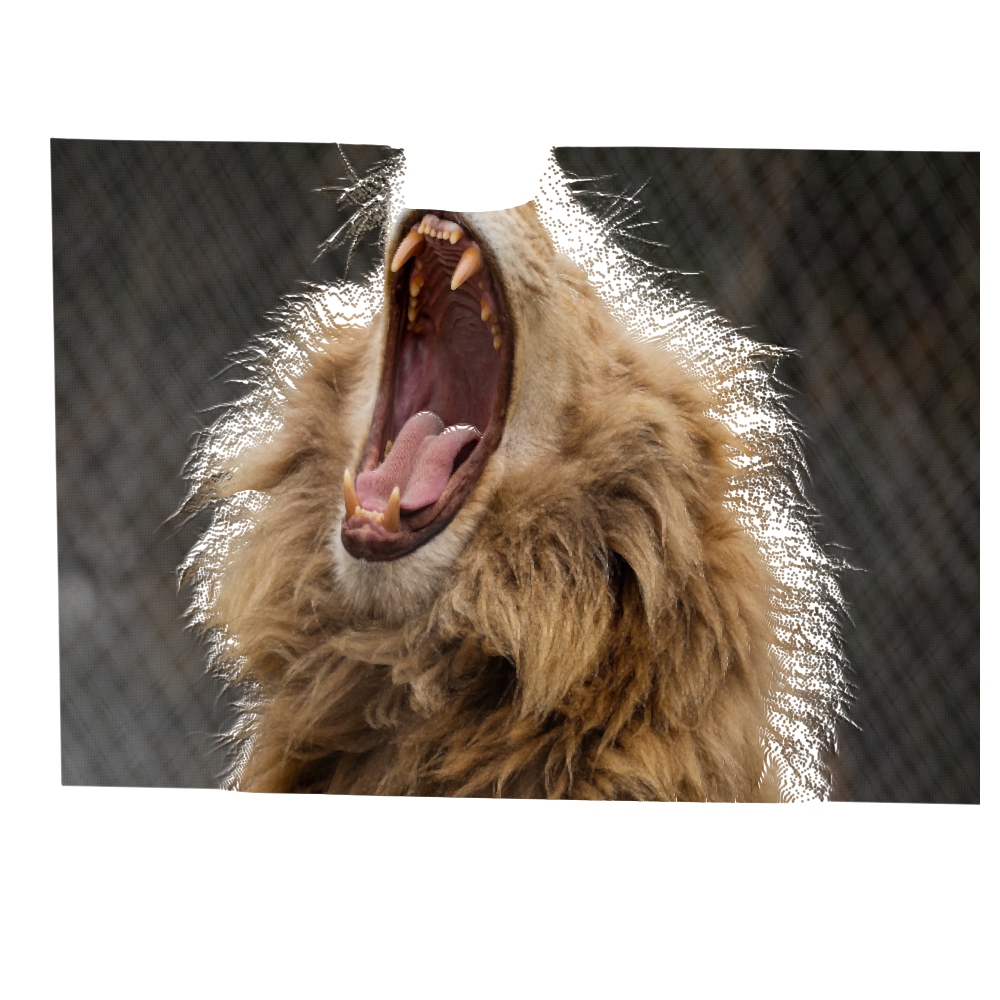}};
            \spy on \zoomtwo in node [left] at \rebigtwo;
    	\end{tikzpicture}
        \caption*{Depth Pro}
      \end{subfigure}
      \begin{subfigure}{\pointWidth}
        \begin{tikzpicture}[spy using outlines={green,magnification=\ssmag,size=\ssizz},inner sep=0]
            \node [align=center, img] {\includegraphics[width=\textwidth,trim={35 160 15 130}, clip]{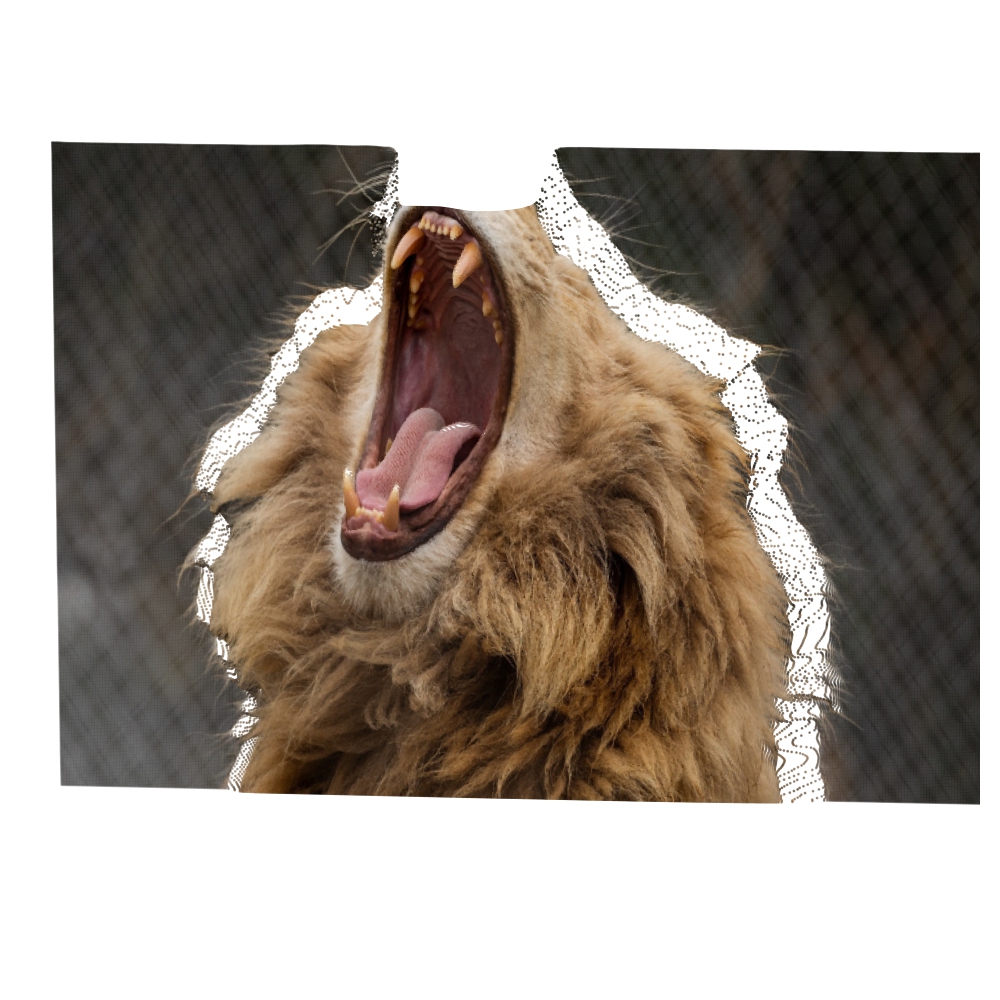}};
            \spy on \zoomtwo in node [left] at \rebigtwo;
    	\end{tikzpicture}
        \caption*{UniDepthV2}
      \end{subfigure}
      \begin{subfigure}{\pointWidth}
        \begin{tikzpicture}[spy using outlines={green,magnification=\ssmag,size=\ssizz},inner sep=0]
            \node [align=center, img] {\includegraphics[width=\textwidth,trim={35 160 15 130}, clip]{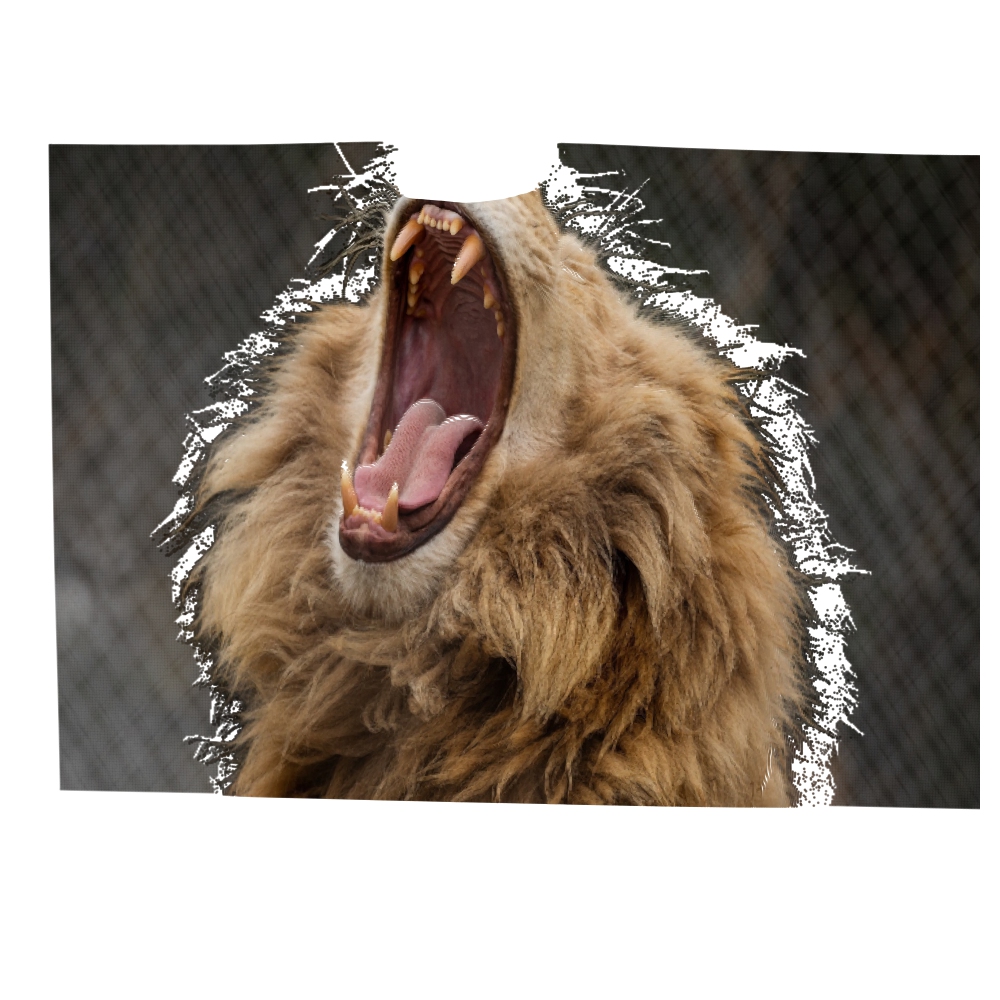}};
            \spy on \zoomtwo in node [left] at \rebigtwo;
    	\end{tikzpicture}
        \caption*{Ours}
      \end{subfigure}
    \end{subfigure}
    \begin{subfigure}{\linewidth}
    \centering
         \begin{subfigure}{\pointWidth}
        \begin{tikzpicture}[spy using outlines={green,magnification=\ssmag,size=\ssizz},inner sep=0]
            \node [align=center, img] {\includegraphics[width=\textwidth,trim={0 0 0 0}, clip]{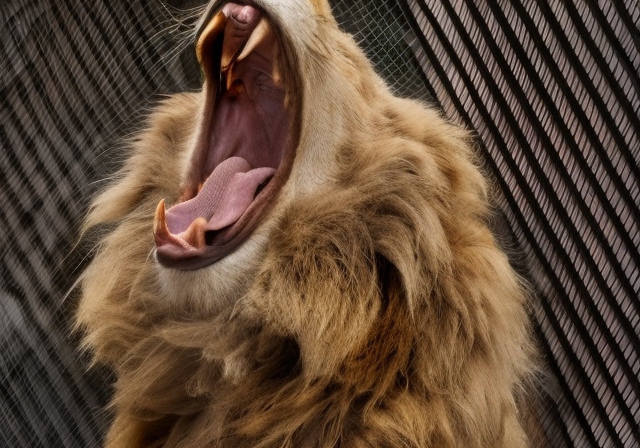}};
            \spy on \zoomfive in node [left] at \rebigthree;
    	\end{tikzpicture}
        \caption*{ViewCrafter}
      \end{subfigure}
      \begin{subfigure}{\pointWidth}
        \begin{tikzpicture}[spy using outlines={green,magnification=\ssmag,size=\ssizz},inner sep=0]
            \node [align=center, img] {\includegraphics[width=\textwidth,trim={0 0 0 0}, clip]{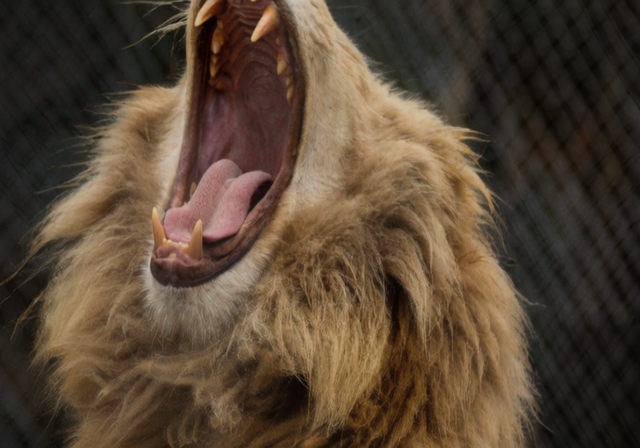}};
            \spy on \zoomfour in node [left] at \rebigthree;
    	\end{tikzpicture}
        \caption*{ReCamMaster}
      \end{subfigure}
      \begin{subfigure}{\pointWidth}
        \begin{tikzpicture}[spy using outlines={green,magnification=\ssmag,size=\ssizz},inner sep=0]
            \node [align=center, img] {\includegraphics[width=\textwidth,trim={0 0 0 0}, clip]{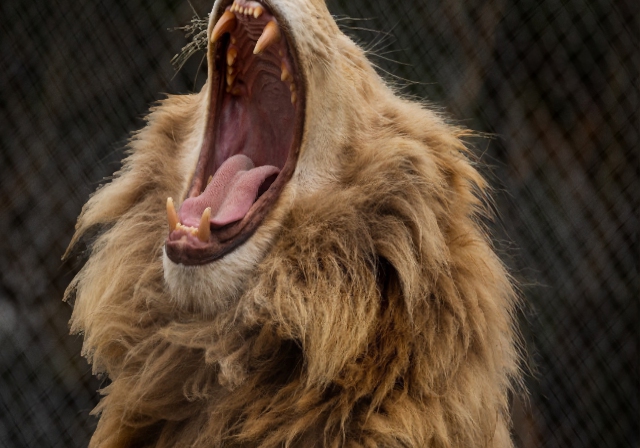}};
            \spy on \zoomthree in node [left] at \rebigthree;
    	\end{tikzpicture}
        \caption*{SplatDiff}
      \end{subfigure}
      \begin{subfigure}{\pointWidth}
        \begin{tikzpicture}[spy using outlines={green,magnification=\ssmag,size=\ssizz},inner sep=0]
            \node [align=center, img] {\includegraphics[width=\textwidth,trim={0 0 0 0}, clip]{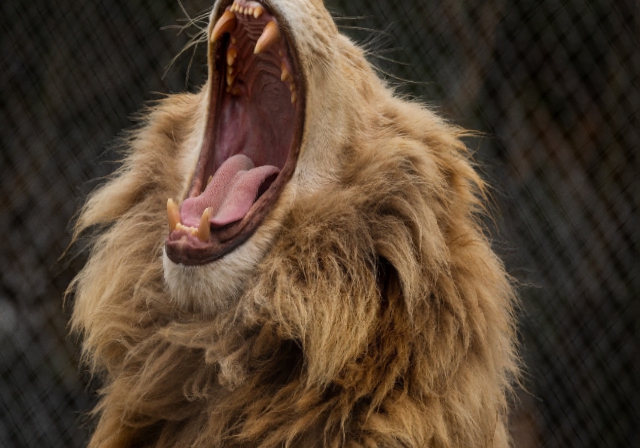}};
            \spy on \zoomthree in node [left] at \rebigthree;
    	\end{tikzpicture}
        \caption*{Ours}
      \end{subfigure}
    \end{subfigure}
    \vspace{-2em}
    \caption{\textbf{Qualitative comparisons} of depth estimation (top), point clouds (middle), and novel view synthesis (bottom). Our \myname\ better preserves soft boundary details in depth results and novel views, without artifacts like broken, detached, or hallucinated hairs. }
    \vspace{-0.5em}
    \label{fig:exp_quali_res}
\end{figure*}

%% file: tabs/main/tab-depth-zeroshot.tex
\begin{table*}[th]
\centering
\caption{\textbf{Zero-shot depth estimation performance.} The depth fixer preserves the zero-shot capability of its base depth model in diverse scenarios. Metrics are shown in percentage. \textcolor{red}{Better} results are marked. Please see the supplementary for more robustness evaluations. }
\vspace{-0.8em}
\label{tab:depth-zeroshot}
\footnotesize
\setlength\tabcolsep{3pt}
\begin{tabular}{lcccccccccccccc}
\toprule
\rowcolor{color3}                                        & \multicolumn{2}{c}{\textbf{NYUv2}} &  & \multicolumn{2}{c}{\textbf{KITTI}} &  & \multicolumn{2}{c}{\textbf{ETH3D}}              &  & \multicolumn{2}{c}{\textbf{ScanNet}} &  & \multicolumn{2}{c}{\textbf{DIODE}}                                   \\ \cline{2-3} \cline{5-6} \cline{8-9} \cline{11-12} \cline{14-15}
\rowcolor{color3}  \multirow{-2}{*}{\textbf{Method}}      & AbsRel $\downarrow$    & $\delta$1 $\uparrow$    &  & AbsRel $\downarrow$    & $\delta$1 $\uparrow$    &  & AbsRel $\downarrow$                    & $\delta$1 $\uparrow$ &  & AbsRel $\downarrow$     & $\delta$1 $\uparrow$     &  & AbsRel $\downarrow$                     & $\delta$1 $\uparrow$                     \\ \midrule
Depth Anything V2~\cite{yang2024depthanythingv2}                                   & 4.27        & 97.86       &  & 7.97        & 94.38       &  & 5.25                        & 98.27    &  & 4.15         & 97.94        &  & 26.24                        & 75.49                        \\
\rowcolor{color3} \textbf{Depth Anything V2+Depth Fixer (Ours)}       & 4.27        & 97.86       &  & 7.97        & 94.38       &  & 5.25                        & 98.27    &  & 4.15         & 97.94        &  & 26.24                        & 75.49                        \\ \midrule
Depth Pro~\cite{depthpro}                              & 4.29        & 97.90       &  & 5.98        & 96.25       &  & 5.23                        & 96.89    &  & 4.11         & 97.98        &  & 22.20                        & 76.28                        \\
\rowcolor{color3} \textbf{Depth Pro+Depth Fixer (Ours)}  & 4.29        & 97.90       &  & 5.98        & 96.25       &  & 5.23                        & 96.89    &  & 4.11         & 97.98        &  & {\color{red} 22.17} & 76.28                        \\ \midrule
UniDepthV2~\cite{unidepthv2}                             & 3.40        & 98.33       &  & 4.67        & 97.42       &  & 3.31                        & 99.16    &  & 3.08         & 98.32        &  & 23.94                        & 75.67                        \\
\rowcolor{color3} \textbf{UniDepthV2+Depth Fixer (Ours)} & 3.40        & 98.33       &  & 4.67        & 97.42       &  & {\color{red} 3.27} & 99.16    &  & 3.08         & 98.32        &  & {\color{red} 23.87} & {\color{red} 75.71} \\ \bottomrule
\end{tabular}
\vspace{-1.5em}
\end{table*}

%% file: tabs/main/tab-stereo-marvel.tex
\begin{table*}[ht]
\centering
\caption{\textbf{Stereo image/video conversion performance} on the Marvel-10K dataset. The \textcolor{red}{best} and \textcolor{blue}{second-best} results are marked. }
\vspace{-0.8em}
\label{tab:stereo-marvel}
\footnotesize
\setlength\tabcolsep{2.9pt}
\begin{tabular}{lccccccccccccc}
\toprule
\rowcolor{color3} & \multicolumn{6}{c}{\textbf{Stereo Image Conversion}}                          &  & \multicolumn{6}{c}{\textbf{Stereo Video Conversion}}                          \\
\cline{2-7} \cline{9-14} \rowcolor{color3}                               \multirow{-2}{*}{\textbf{Method}}      & PSNR $\uparrow$ & SSIM $\uparrow$ & RMSE $\downarrow$ & LPIPS $\downarrow$ & DISTS $\downarrow$ & SIoU $\uparrow$ &  & PSNR $\uparrow$ & SSIM $\uparrow$ & RMSE $\downarrow$ & LPIPS $\downarrow$ & DISTS $\downarrow$ & SIoU $\uparrow$ \\ \midrule
StereoDiffusion~\cite{stereodiffusion}                  & 32.70     & 0.7654    & 6.05        & 0.2177       & 0.0698       & 0.2638    &  & 32.71     & 0.7656    & 6.04        & 0.2172       & 0.0693       & 0.2655    \\
Mono2Stereo~\cite{mono2stereo}                      & 33.65     & 0.8143    & 5.45        & 0.1973       & 0.0690       & 0.2556    &  & 33.63     & 0.8134    & 5.47        & 0.1980       & 0.0691       & 0.2552    \\
StereoCrafter~\cite{stereocrafter}                    & 32.52     & 0.8148    & 6.13        & 0.2330       & 0.1208       & 0.2664    &  & 32.35     & 0.8125    & 6.25        & 0.2381       & 0.1246       & 0.2645    \\
ViewCrafter~\cite{yu2024viewcrafter}                      & 30.69     & 0.6705    & 7.61        & 0.3258       & 0.1330       & 0.2085    &  & 30.73     & 0.6739    & 7.59        & 0.3221       & 0.1312       & 0.2101    \\
NVS-Solver~\cite{you2024nvssolver}                       & 31.18     & 0.7108    & 7.16        & 0.3323       & 0.1793       & 0.2143    &  & 31.44     & 0.7220    & 6.98        & 0.3256       & 0.1743       & 0.2161    \\
ReCamMaster~\cite{recammaster}                      & 30.44     & 0.6118    & 7.82        & 0.4082       & 0.1391       & 0.1798    &  & 30.41     & 0.6107    & 7.84        & 0.4106       & 0.1412       & 0.1797    \\
SplatDiff~\cite{splatdiff} & {\color{blue} 36.23}     & {\color{blue} 0.8857}    & {\color{blue} 4.06}        & {\color{blue} 0.1116}       & {\color{blue} 0.0435}       & {\color{blue} 0.3259}    &  & {\color{blue} 36.24}     & {\color{blue} 0.8858}    & {\color{blue} 4.06}        & {\color{blue} 0.1114}       & {\color{blue} 0.0437}       & {\color{blue}0.3280}    \\
\rowcolor{color3} \textbf{\myname\ (Ours)}                             &  {\color{red} 36.59}         & {\color{red}0.8953}          & {\color{red}3.91}            & {\color{red}0.0909}             &  {\color{red}0.0331}            &  {\color{red}0.3337}         &  &   {\color{red}36.58}        &     {\color{red}0.8953}      &  {\color{red}3.92}           &   {\color{red}0.0911}           &  {\color{red}0.0334}            &   {\color{red}0.3355}        \\ \bottomrule
\end{tabular}
\vspace{-1em}
\end{table*}

%% file: figs/main/fig-stereo_video_consistency.tex
\begin{figure}[th]
  \centering
  \includegraphics[width=.49\linewidth]{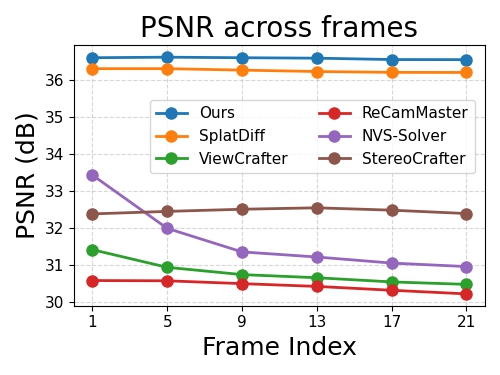}
  \includegraphics[width=.49\linewidth]{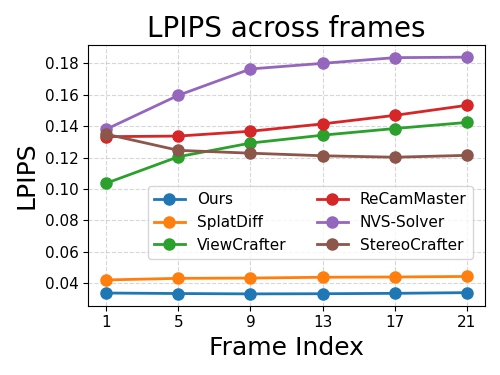}
  \vspace{-1em}
  \caption{\textbf{Stereo video conversion performance} on Marvel-10K.}
  \vspace{-1.5em}\label{fig:stereo_video_consistency}
\end{figure}

%% file: tabs/main/tab-stereo-ablation.tex
\begin{table}[th]
\centering
\caption{\textbf{Ablation on stereo image conversion} on the Marvel-10K dataset. The \textcolor{red}{best} and \textcolor{blue}{second-best} results are marked. Please see the supplementary for more detailed ablation studies. }
\vspace{-0.8em}
\label{tab:stereo-ablation}
\footnotesize
\setlength\tabcolsep{4pt}
\begin{tabular}{ccccccccc}
\toprule
\rowcolor{color3} &  & Depth     & Scene     & Color     &  & \multicolumn{3}{c}{\textbf{Marvel-10K}} \\ \cline{7-9} \rowcolor{color3} \multirow{-2}{*}{\textbf{Exp}}      
                    &  & Fixer     & Painter   & Fuser     &  & PSNR $\uparrow$         & LPIPS $\downarrow$      & SIoU $\uparrow$      \\ \midrule
\#1                 &  &           &           &           &  & 36.26      & 0.1490       & 0.3097     \\
\#2                 &  & $\checkmark$ &           &           &  & {\color{blue} 36.28}      & 0.1458      & {\color{blue} 0.3118 }     \\
\#3                 &  & $\checkmark$ & $\checkmark$ &           &  &  35.82            &  {\color{blue} 0.1246 }          & 0.3015           \\
\#4                 &  & $\checkmark$ & $\checkmark$ & $\checkmark$ &  &  {\color{red} 36.59}            &  {\color{red} 0.0909 }          & {\color{red} 0.3337}           \\ \bottomrule
\end{tabular}
\vspace{-1em}
\end{table}

%% file: tabs/main/tab-nvs-metrics.tex
\begin{table}[th]
\centering
\caption{\textbf{Novel view synthesis performance} on the natural image matting datasets. The \textcolor{red}{best} and \textcolor{blue}{second-best} results are marked.}
\vspace{-0.8em}
\label{tab:nvs-metrics}
\footnotesize
\setlength\tabcolsep{3pt}
\begin{tabular}{llccccc}
\toprule
\rowcolor{color3} &  & \multicolumn{2}{c}{\textbf{AIM-500}} &  & \multicolumn{2}{c}{\textbf{P3M-10K}} \\ \cline{3-4} \cline{6-7} \rowcolor{color3} \multirow{-2}{*}{\textbf{Method}}
                                 &  & FID $\downarrow$         & CLIP-F $\uparrow$     &  & FID $\downarrow$        & CLIP-F $\uparrow$    \\ \midrule
NVS-Solver~\cite{you2024nvssolver}                       &  & 51.71        & 97.24      &  & 55.12           & 96.66      \\
ViewCrafter~\cite{yu2024viewcrafter}                      &  & 33.43         & 99.03      &  &  35.40                &  98.73          \\
ReCamMaster~\cite{recammaster}                      &  & 57.19           & 97.95      &  &  62.80               & 97.09           \\
SplatDiff~\cite{splatdiff}                        &  & {\color{blue} 19.26}         & {\color{blue} 99.36}      &  &    {\color{blue} 21.61}             &  {\color{blue} 99.09}          \\
\rowcolor{color3} \textbf{\myname\ (Ours)}                             &  & {\color{red} 18.82}         & {\color{red}99.38}      &  &    {\color{red}21.38}          &    {\color{red}99.11}        \\ \bottomrule
\end{tabular}
\vspace{-1em}
\end{table}

%% file: figs/main/fig-user_study.tex
\begin{figure}[th]
  \centering
  \includegraphics[width=\linewidth,trim={0 30 0 0}, clip]{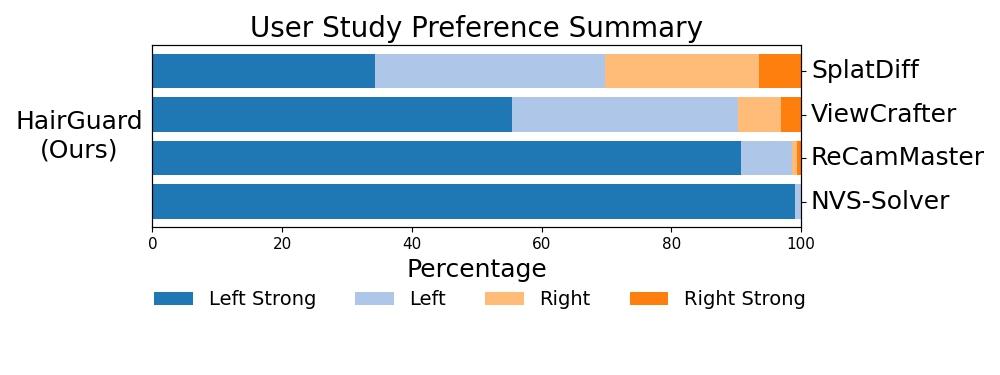}
  \vspace{-2em}
  \caption{\textbf{User study on novel view synthesis.} A survey of 27 participants (1332 votes in total) shows the superiority of our \myname\ compared with previous state-of-the-art approaches.}
  \label{fig:nvs_user_study}
\end{figure}

%% file: sec/conclusion.tex
\section{Conclusion}
\label{sec:conclusion}

This paper presents \myname\ to address the challenges of soft boundaries in 3D vision tasks. By utilizing image matting datasets, we train a depth fixer to automatically identify soft boundary regions and correct depth results in a plug-and-play manner. For view synthesis tasks, the scene painter and color fuser are employed to fix geometric errors in novel views while preserving high-quality texture details. Extensive experiments on monocular depth estimation, stereo image/video conversion, and novel view synthesis verify the state-of-the-art performance of \myname.

%% file: figs/supp/fig-marvel_examples.tex
\begin{figure}[th]
    \centering
    \includegraphics[width=0.32\linewidth]{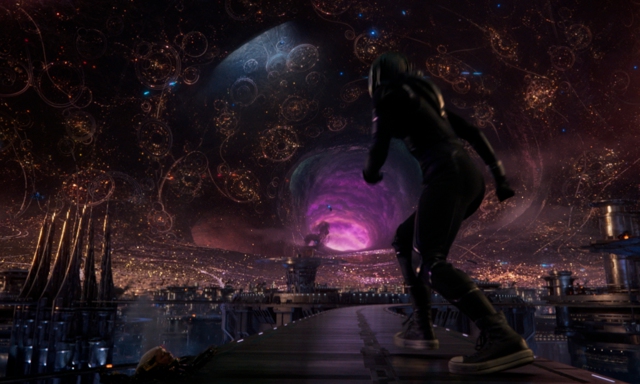}
    \includegraphics[width=0.32\linewidth]{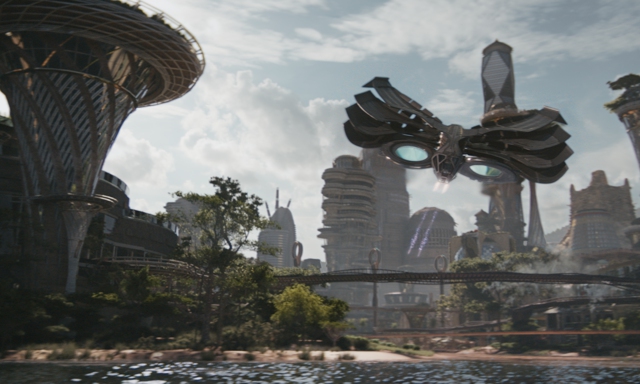}
    \includegraphics[width=0.32\linewidth]{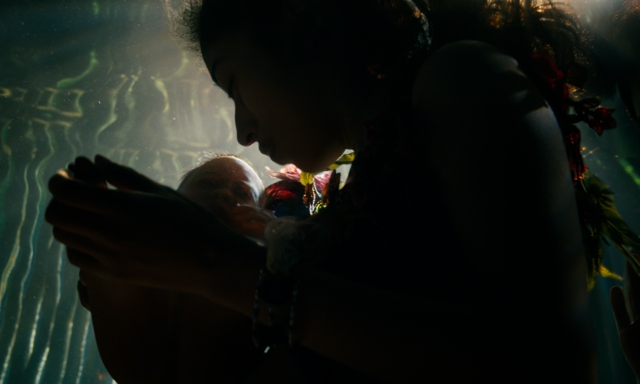}
    \\
    \includegraphics[width=0.32\linewidth]{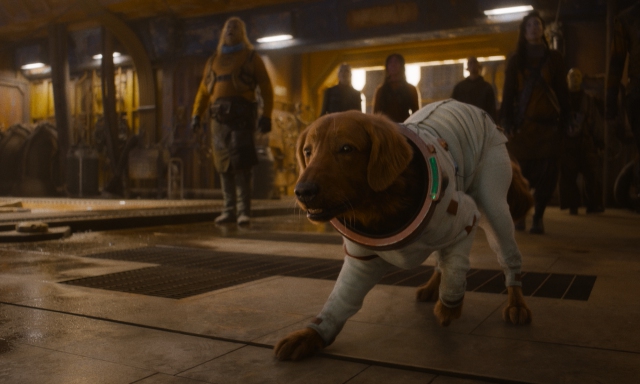}
    \includegraphics[width=0.32\linewidth]{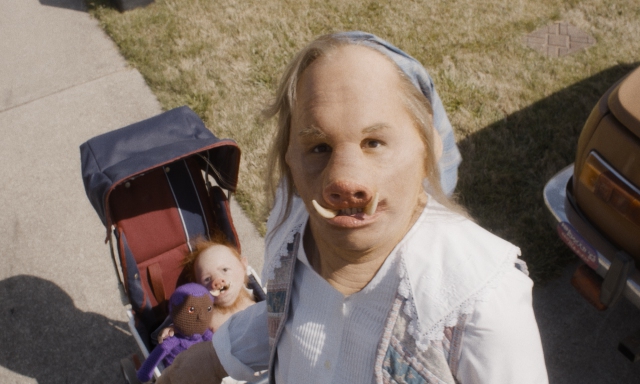}
    \includegraphics[width=0.32\linewidth]{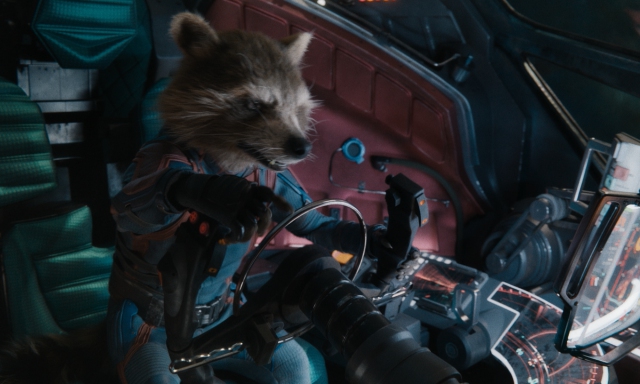}
    \\
    \includegraphics[width=0.32\linewidth]{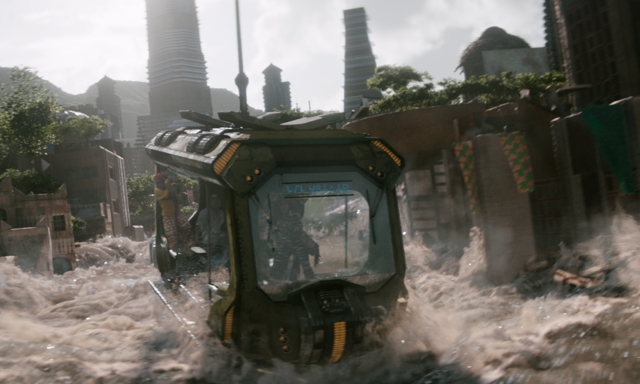}
    \includegraphics[width=0.32\linewidth]{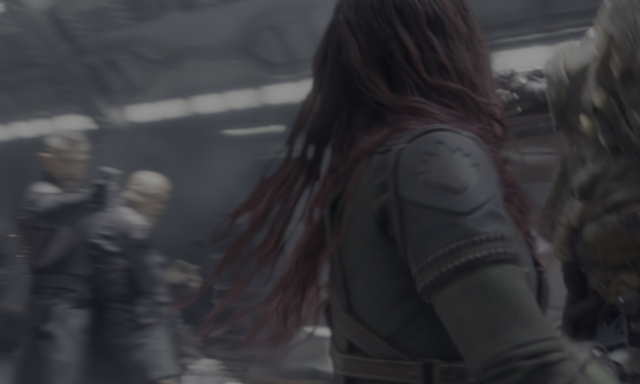}
    \includegraphics[width=0.32\linewidth]{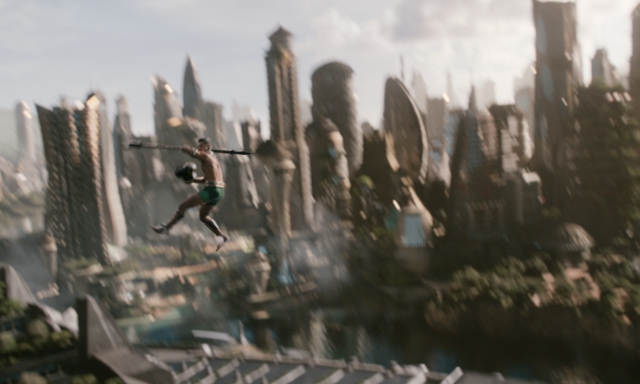}
    \vspace{-0.5em}
    \caption{\textbf{Example images in the Marvel-10K dataset.}}
    \label{fig:supp-marvel-examples}
\end{figure}

%% file: figs/supp/fig-robustness_scene.tex
\def\imgWidth{0.32\linewidth} %
\def\depthWidth{0.24\linewidth} %
\def\pointWidth{0.24\linewidth} %
\def\scc{(-1.9,-1.4)}

\def\rebigone{(-0.9, -0.6)} %
\def\rebigtwo{(2.0, -0.6)} %
\def\rebigthree{(-0.9, 0.6)} %

\def\zoomone{(-0.5,-0)} %
\def\zoomtwo{(-1.2,0.8)} %

\def\zoomthree{(-0.68,0.33)} %

\def\ssizz{1.1cm} %
\def\ssmag{3}

\begin{figure*}[t]
\centering
\tikzstyle{img} = [rectangle, minimum width=\imgWidth]
    \centering
    \begin{subfigure}{\linewidth}
    \centering
    \begin{subfigure}{\depthWidth}
		\begin{tikzpicture}[spy using outlines={green,magnification=\ssmag,size=\ssizz},inner sep=0]
            \node [align=center, img] {\includegraphics[width=\textwidth]{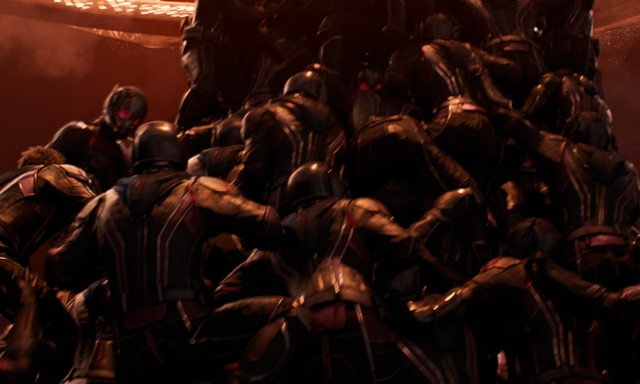}};
    	\end{tikzpicture}
    \end{subfigure}
    \begin{subfigure}{\depthWidth}
        \begin{tikzpicture}[spy using outlines={green,magnification=\ssmag,size=\ssizz},inner sep=0]
            \node [align=center, img] {\includegraphics[width=\textwidth]{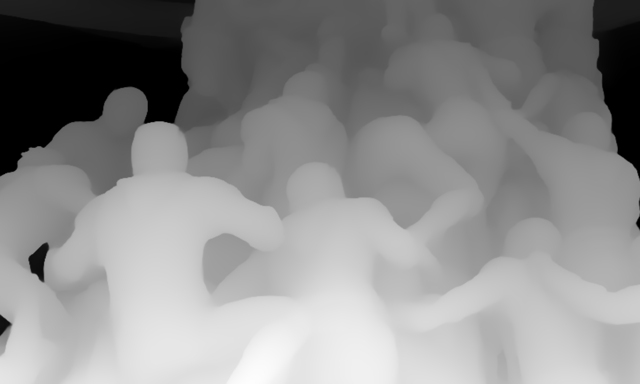}};
    	\end{tikzpicture}
      \end{subfigure}
    \begin{subfigure}{\depthWidth}
        \begin{tikzpicture}[spy using outlines={green,magnification=\ssmag,size=\ssizz},inner sep=0]
            \node [align=center, img] {\includegraphics[width=\textwidth]{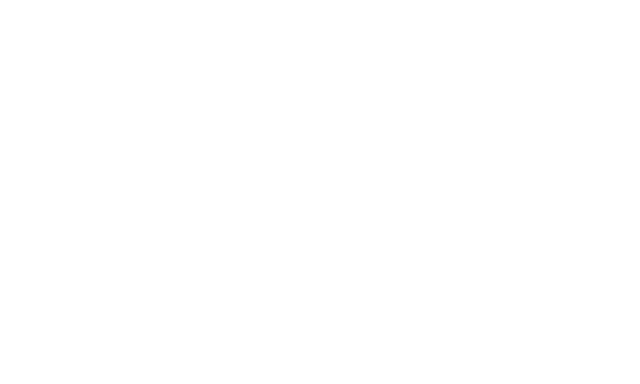}};
    	\end{tikzpicture}
      \end{subfigure}
    \begin{subfigure}{\depthWidth}
        \begin{tikzpicture}[spy using outlines={green,magnification=\ssmag,size=\ssizz},inner sep=0]
            \node [align=center, img] {\includegraphics[width=\textwidth]{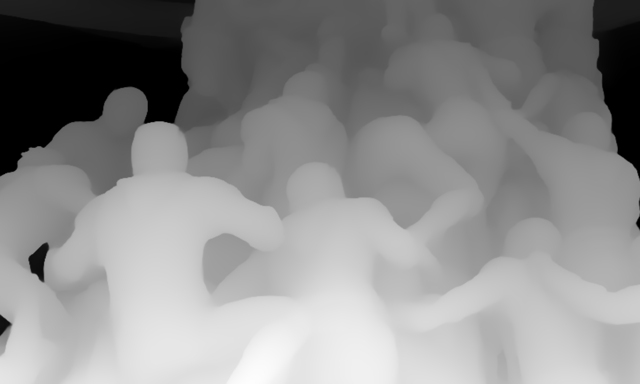}};
    	\end{tikzpicture}
      \end{subfigure}
    \end{subfigure}
    \begin{subfigure}{\linewidth}
    \centering
    \begin{subfigure}{\depthWidth}
		\begin{tikzpicture}[spy using outlines={green,magnification=\ssmag,size=\ssizz},inner sep=0]
            \node [align=center, img] {\includegraphics[width=\textwidth]{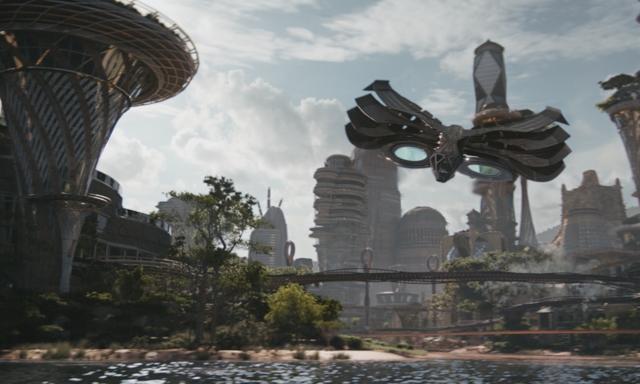}};
    	\end{tikzpicture}
    \end{subfigure}
    \begin{subfigure}{\depthWidth}
        \begin{tikzpicture}[spy using outlines={green,magnification=\ssmag,size=\ssizz},inner sep=0]
            \node [align=center, img] {\includegraphics[width=\textwidth]{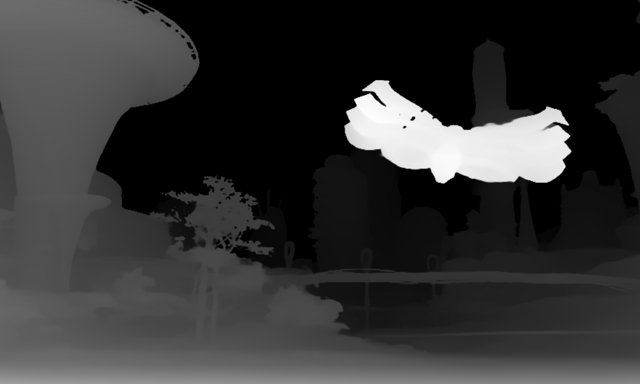}};
    	\end{tikzpicture}
      \end{subfigure}
    \begin{subfigure}{\depthWidth}
        \begin{tikzpicture}[spy using outlines={green,magnification=\ssmag,size=\ssizz},inner sep=0]
            \node [align=center, img] {\includegraphics[width=\textwidth]{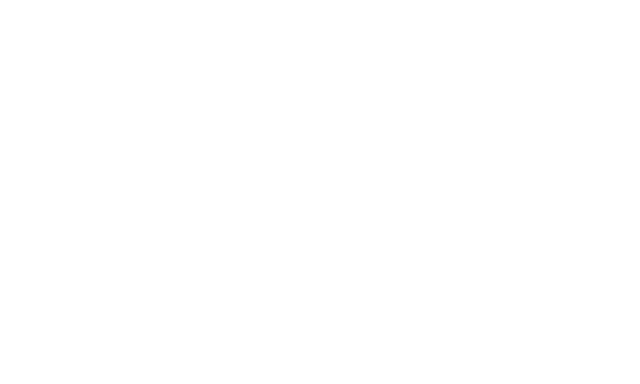}};
    	\end{tikzpicture}
      \end{subfigure}
    \begin{subfigure}{\depthWidth}
        \begin{tikzpicture}[spy using outlines={green,magnification=\ssmag,size=\ssizz},inner sep=0]
            \node [align=center, img] {\includegraphics[width=\textwidth]{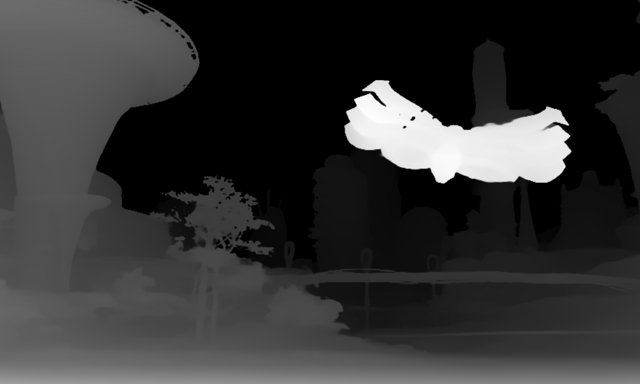}};
    	\end{tikzpicture}
      \end{subfigure}
    \end{subfigure}
    \begin{subfigure}{\linewidth}
    \centering
    \begin{subfigure}{\depthWidth}
		\begin{tikzpicture}[spy using outlines={green,magnification=\ssmag,size=\ssizz},inner sep=0]
            \node [align=center, img] {\includegraphics[width=\textwidth]{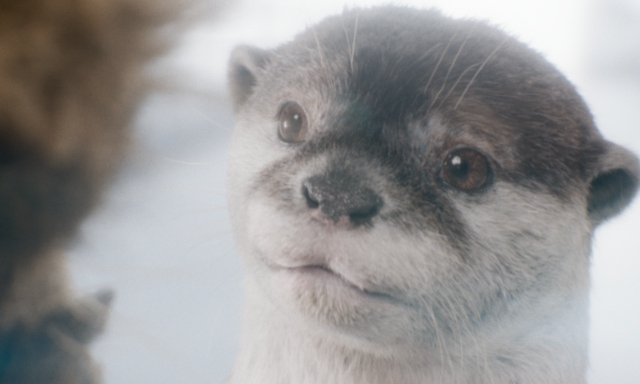}};
            \spy on \zoomthree in node [left] at \rebigthree;
    	\end{tikzpicture}
    \end{subfigure}
    \begin{subfigure}{\depthWidth}
        \begin{tikzpicture}[spy using outlines={green,magnification=\ssmag,size=\ssizz},inner sep=0]
            \node [align=center, img] {\includegraphics[width=\textwidth]{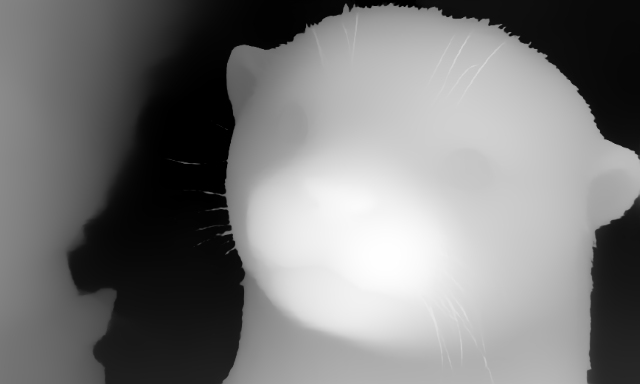}};
            \spy on \zoomthree in node [left] at \rebigthree;
    	\end{tikzpicture}
      \end{subfigure}
    \begin{subfigure}{\depthWidth}
        \begin{tikzpicture}[spy using outlines={green,magnification=\ssmag,size=\ssizz},inner sep=0]
            \node [align=center, img] {\includegraphics[width=\textwidth]{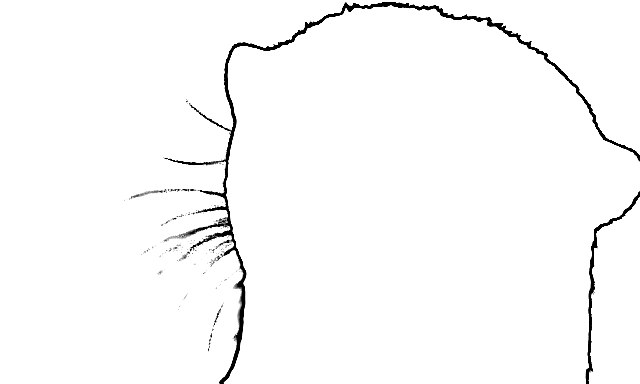}};
            \spy on \zoomthree in node [left] at \rebigthree;
    	\end{tikzpicture}
      \end{subfigure}
    \begin{subfigure}{\depthWidth}
        \begin{tikzpicture}[spy using outlines={green,magnification=\ssmag,size=\ssizz},inner sep=0]
            \node [align=center, img] {\includegraphics[width=\textwidth]{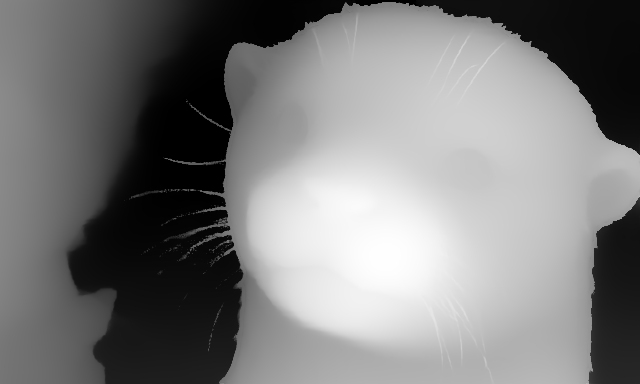}};
            \spy on \zoomthree in node [left] at \rebigthree;
    	\end{tikzpicture}
      \end{subfigure}
    \end{subfigure}
    \begin{subfigure}{\linewidth}
    \centering
    \begin{subfigure}{\depthWidth}
		\begin{tikzpicture}[spy using outlines={green,magnification=\ssmag,size=\ssizz},inner sep=0]
            \node [align=center, img] {\includegraphics[width=\textwidth]{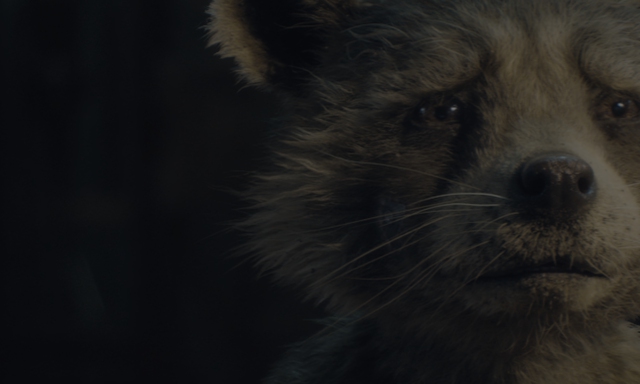}};
            \spy on \zoomone in node [left] at \rebigone;
    	\end{tikzpicture}
    \end{subfigure}
    \begin{subfigure}{\depthWidth}
        \begin{tikzpicture}[spy using outlines={green,magnification=\ssmag,size=\ssizz},inner sep=0]
            \node [align=center, img] {\includegraphics[width=\textwidth]{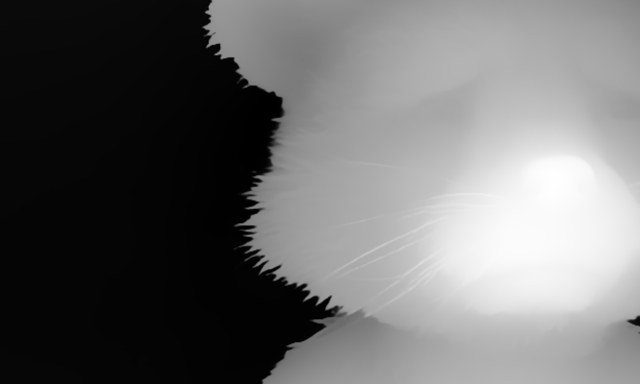}};
            \spy on \zoomone in node [left] at \rebigone;
    	\end{tikzpicture}
      \end{subfigure}
    \begin{subfigure}{\depthWidth}
        \begin{tikzpicture}[spy using outlines={green,magnification=\ssmag,size=\ssizz},inner sep=0]
            \node [align=center, img] {\includegraphics[width=\textwidth]{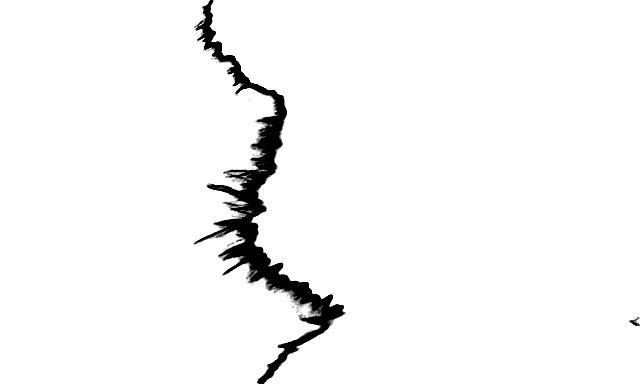}};
            \spy on \zoomone in node [left] at \rebigone;
    	\end{tikzpicture}
      \end{subfigure}
    \begin{subfigure}{\depthWidth}
        \begin{tikzpicture}[spy using outlines={green,magnification=\ssmag,size=\ssizz},inner sep=0]
            \node [align=center, img] {\includegraphics[width=\textwidth]{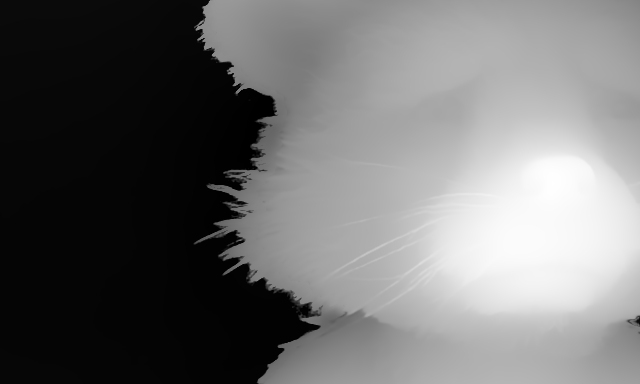}};
            \spy on \zoomone in node [left] at \rebigone;
    	\end{tikzpicture}
      \end{subfigure}
    \end{subfigure}
    \begin{subfigure}{\linewidth}
    \centering
    \begin{subfigure}{\depthWidth}
		\begin{tikzpicture}[spy using outlines={green,magnification=\ssmag,size=\ssizz},inner sep=0]
            \node [align=center, img] {\includegraphics[width=\textwidth]{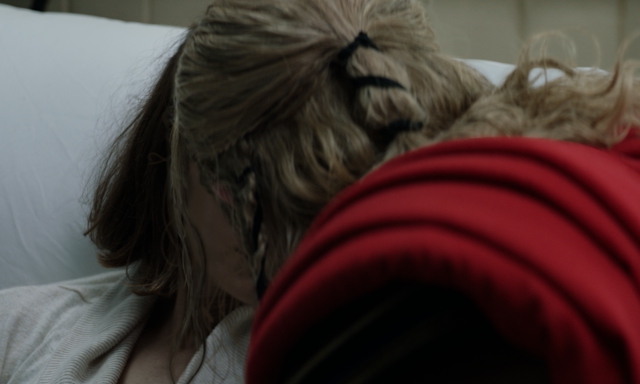}};
            \spy on \zoomtwo in node [left] at \rebigtwo;
    	\end{tikzpicture}
        \caption*{Input Image}
    \end{subfigure}
    \begin{subfigure}{\depthWidth}
        \begin{tikzpicture}[spy using outlines={green,magnification=\ssmag,size=\ssizz},inner sep=0]
            \node [align=center, img] {\includegraphics[width=\textwidth]{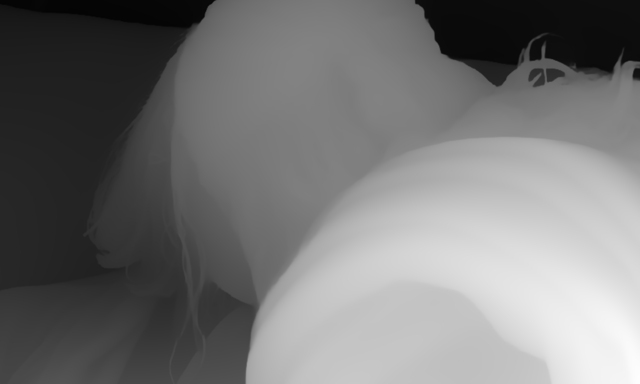}};
            \spy on \zoomtwo in node [left] at \rebigtwo;
    	\end{tikzpicture}
        \caption*{Depth Anything V2}
      \end{subfigure}
    \begin{subfigure}{\depthWidth}
        \begin{tikzpicture}[spy using outlines={green,magnification=\ssmag,size=\ssizz},inner sep=0]
            \node [align=center, img] {\includegraphics[width=\textwidth]{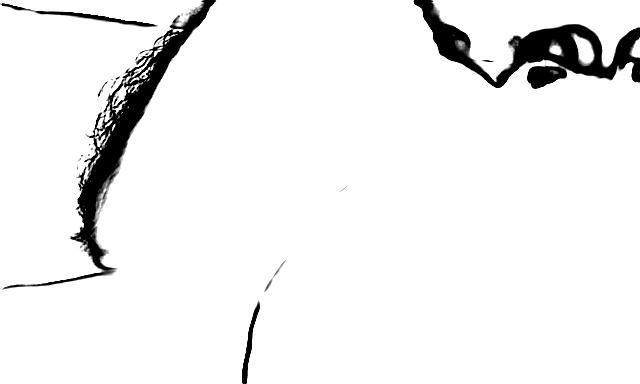}};
            \spy on \zoomtwo in node [left] at \rebigtwo;
    	\end{tikzpicture}
        \caption*{Predicted Gate}
      \end{subfigure}
    \begin{subfigure}{\depthWidth}
        \begin{tikzpicture}[spy using outlines={green,magnification=\ssmag,size=\ssizz},inner sep=0]
            \node [align=center, img] {\includegraphics[width=\textwidth]{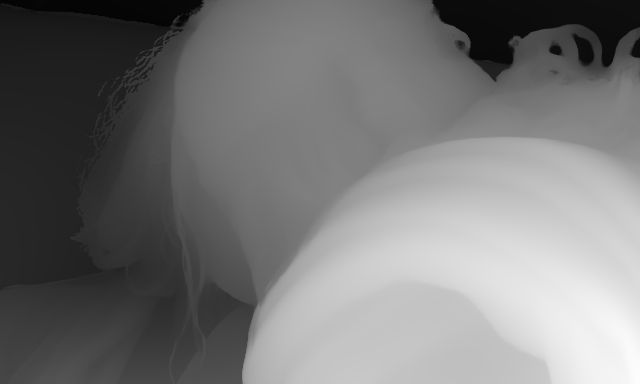}};
            \spy on \zoomtwo in node [left] at \rebigtwo;
    	\end{tikzpicture}
        \caption*{Fixed Depth (Ours)}
      \end{subfigure}
    \end{subfigure}
    \vspace{-1.5em}
    \caption{\textbf{Performance of depth fixer under challenging scenarios.} The regions with the predicted gate $G<1$ indicate the estimated soft boundary regions. Even under complex environments, \eg, heavy occlusion, extreme lighting conditions, and multiple targets, our depth fixer can automatically identify soft boundary regions and perform precise fixing. }
    \label{fig:supp-robust-scene}
\end{figure*}

%% file: figs/supp/fig-plug-depth.tex
\def\imgWidth{0.32\linewidth} %
\def\depthWidth{0.24\linewidth} %
\def\pointWidth{0.24\linewidth} %
\def\scc{(-1.9,-1.4)}

\def\rebigone{(-0.9, 0.83)} %
\def\rebigtwo{(2, -0.83)} %
\def\rebigthree{(-0.93, -0.55)} %

\def\zoomone{(-1.1,-1)} %
\def\zoomtwo{(0.8,0.05)} %

\def\zoomthree{(-0.2,-0.22)} %
\def\zoomfour{(-0.42,-0.25)} %
\def\zoomfive{(-0.42,-0.28)} %
\def\zoomsix{(-0.1,-0.15)} %

\def\ssizz{1.1cm} %
\def\ssmag{3}

\begin{figure*}[t]
\centering
\tikzstyle{img} = [rectangle, minimum width=\imgWidth]
    \centering
    \begin{subfigure}{\linewidth}
    \centering
    \begin{subfigure}{\depthWidth}
		\begin{tikzpicture}[spy using outlines={green,magnification=\ssmag,size=\ssizz},inner sep=0]
            \node [align=center, img] {\includegraphics[width=\textwidth]{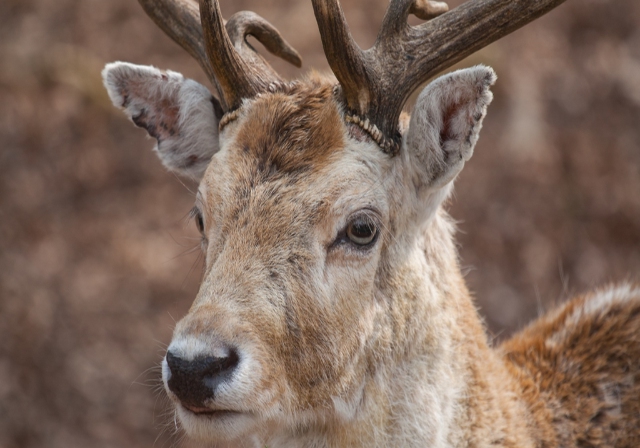}};
            \spy on \zoomone in node [left] at \rebigone;
    	\end{tikzpicture}
        \caption*{Input Image}
    \end{subfigure}
    \begin{subfigure}{\depthWidth}
        \begin{tikzpicture}[spy using outlines={green,magnification=\ssmag,size=\ssizz},inner sep=0]
            \node [align=center, img] {\includegraphics[width=\textwidth]{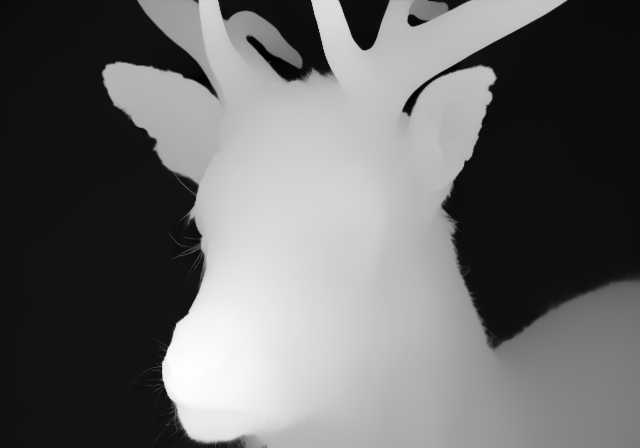}};
            \spy on \zoomone in node [left] at \rebigone;
    	\end{tikzpicture}
        \caption*{Depth Pro}
      \end{subfigure}
    \begin{subfigure}{\depthWidth}
        \begin{tikzpicture}[spy using outlines={green,magnification=\ssmag,size=\ssizz},inner sep=0]
            \node [align=center, img] {\includegraphics[width=\textwidth]{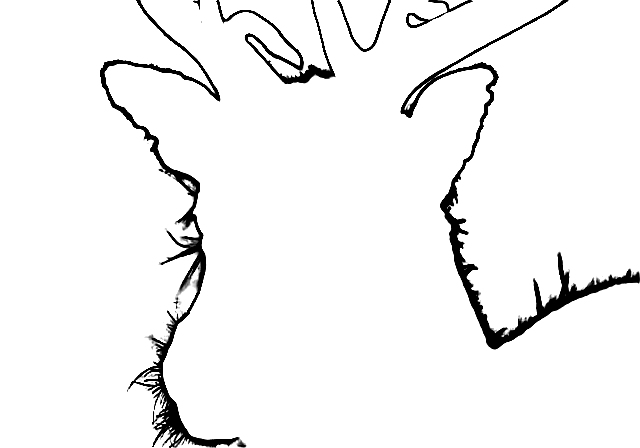}};
            \spy on \zoomone in node [left] at \rebigone;
    	\end{tikzpicture}
        \caption*{Predicted Gate}
      \end{subfigure}
    \begin{subfigure}{\depthWidth}
        \begin{tikzpicture}[spy using outlines={green,magnification=\ssmag,size=\ssizz},inner sep=0]
            \node [align=center, img] {\includegraphics[width=\textwidth]{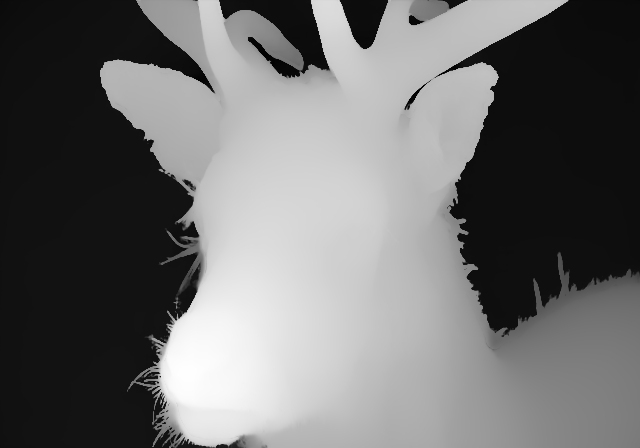}};
            \spy on \zoomone in node [left] at \rebigone;
    	\end{tikzpicture}
        \caption*{Fixed Depth (Ours)}
      \end{subfigure}
    \end{subfigure}
    \begin{subfigure}{\linewidth}
    \centering
    \begin{subfigure}{\depthWidth}
		\begin{tikzpicture}[spy using outlines={green,magnification=\ssmag,size=\ssizz},inner sep=0]
            \node [align=center, img] {\includegraphics[width=\textwidth]{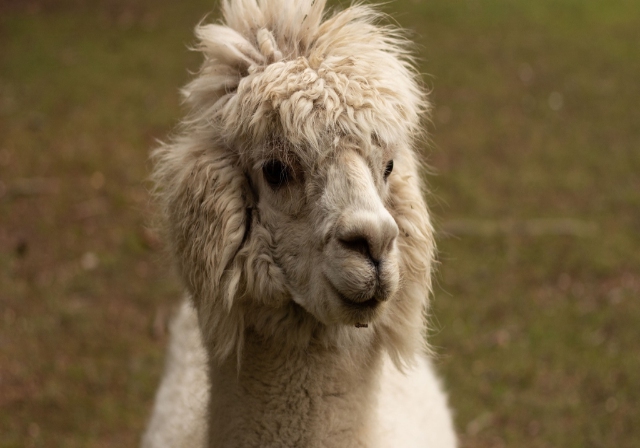}};
            \spy on \zoomtwo in node [left] at \rebigtwo;
    	\end{tikzpicture}
        \caption*{Input Image}
    \end{subfigure}
    \begin{subfigure}{\depthWidth}
        \begin{tikzpicture}[spy using outlines={green,magnification=\ssmag,size=\ssizz},inner sep=0]
            \node [align=center, img] {\includegraphics[width=\textwidth]{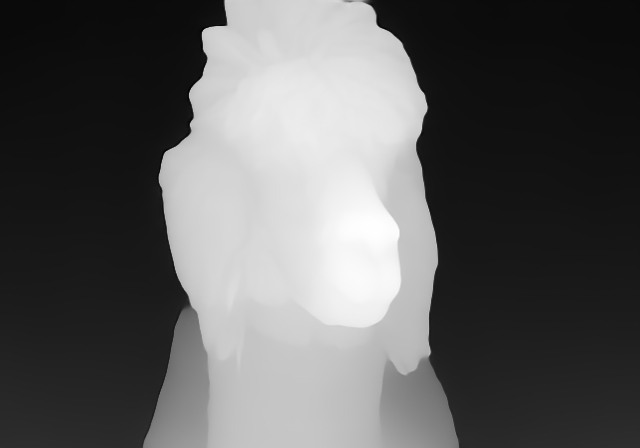}};
            \spy on \zoomtwo in node [left] at \rebigtwo;
    	\end{tikzpicture}
        \caption*{UniDepthV2}
      \end{subfigure}
    \begin{subfigure}{\depthWidth}
        \begin{tikzpicture}[spy using outlines={green,magnification=\ssmag,size=\ssizz},inner sep=0]
            \node [align=center, img] {\includegraphics[width=\textwidth]{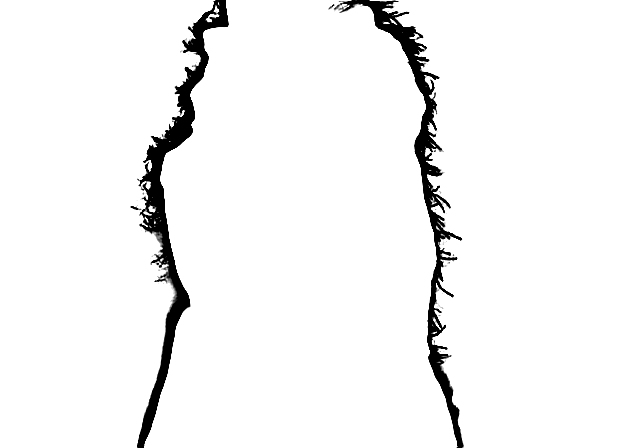}};
            \spy on \zoomtwo in node [left] at \rebigtwo;
    	\end{tikzpicture}
        \caption*{Predicted Gate}
      \end{subfigure}
    \begin{subfigure}{\depthWidth}
        \begin{tikzpicture}[spy using outlines={green,magnification=\ssmag,size=\ssizz},inner sep=0]
            \node [align=center, img] {\includegraphics[width=\textwidth]{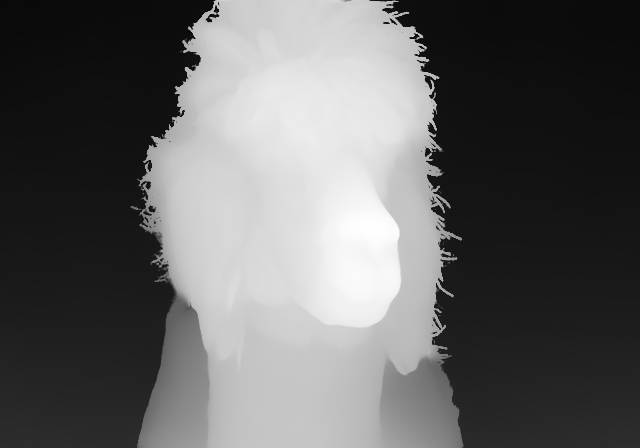}};
            \spy on \zoomtwo in node [left] at \rebigtwo;
    	\end{tikzpicture}
        \caption*{Fixed Depth (Ours)}
      \end{subfigure}
      \caption{Plug-and-play refinement on image-based depth models}
      \label{fig:supp-plug-depth}
    \end{subfigure}
    \\ %
    \begin{subfigure}{\linewidth}
    \centering
    \begin{subfigure}{\depthWidth}
		\begin{tikzpicture}[spy using outlines={green,magnification=\ssmag,size=\ssizz},inner sep=0]
            \node [align=center, img] {\includegraphics[width=\textwidth]{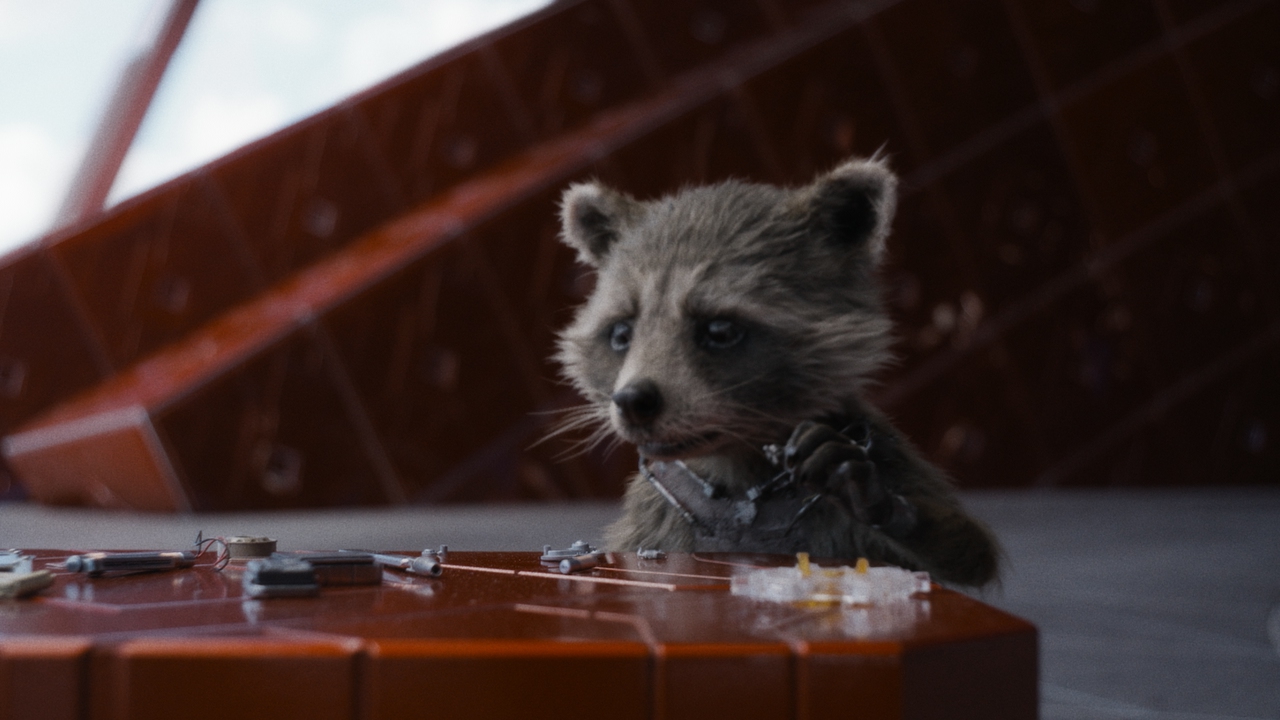}};
            \spy on \zoomthree in node [left] at \rebigthree;
    	\end{tikzpicture}
    \end{subfigure}
    \begin{subfigure}{\depthWidth}
        \begin{tikzpicture}[spy using outlines={green,magnification=\ssmag,size=\ssizz},inner sep=0]
            \node [align=center, img] {\includegraphics[width=\textwidth]{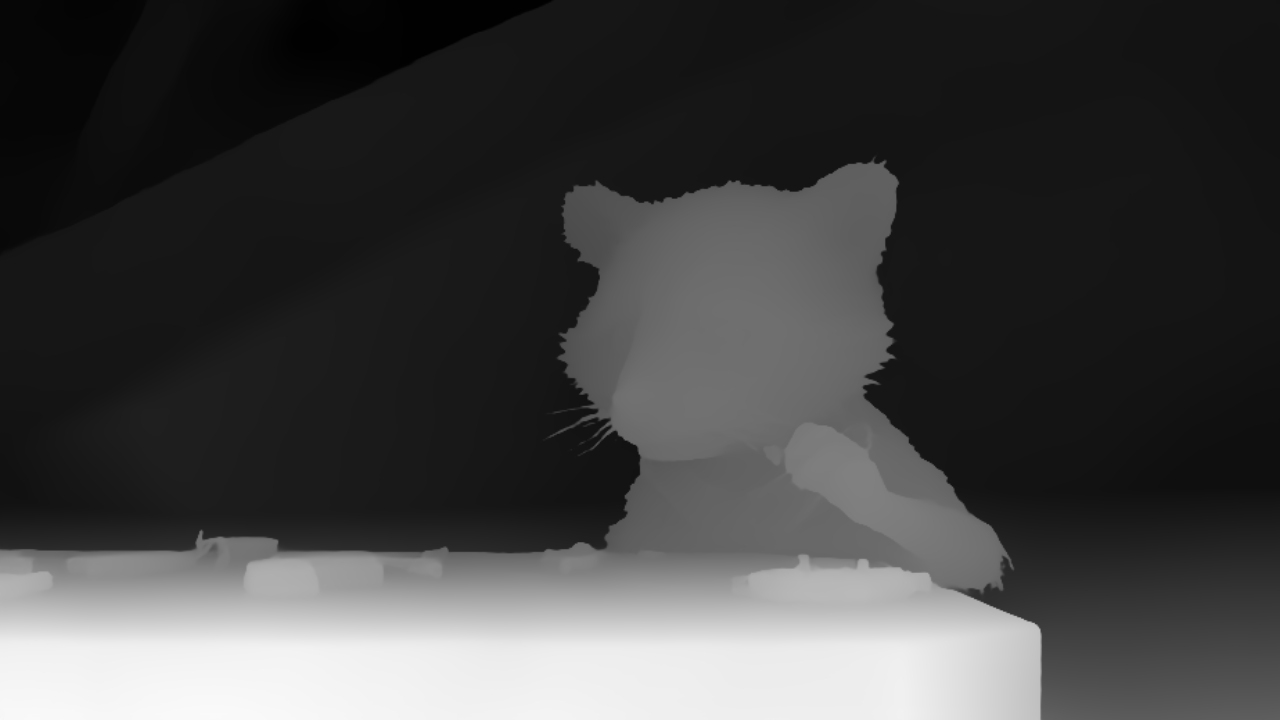}};
            \spy on \zoomthree in node [left] at \rebigthree;
    	\end{tikzpicture}
      \end{subfigure}
    \begin{subfigure}{\depthWidth}
        \begin{tikzpicture}[spy using outlines={green,magnification=\ssmag,size=\ssizz},inner sep=0]
            \node [align=center, img] {\includegraphics[width=\textwidth]{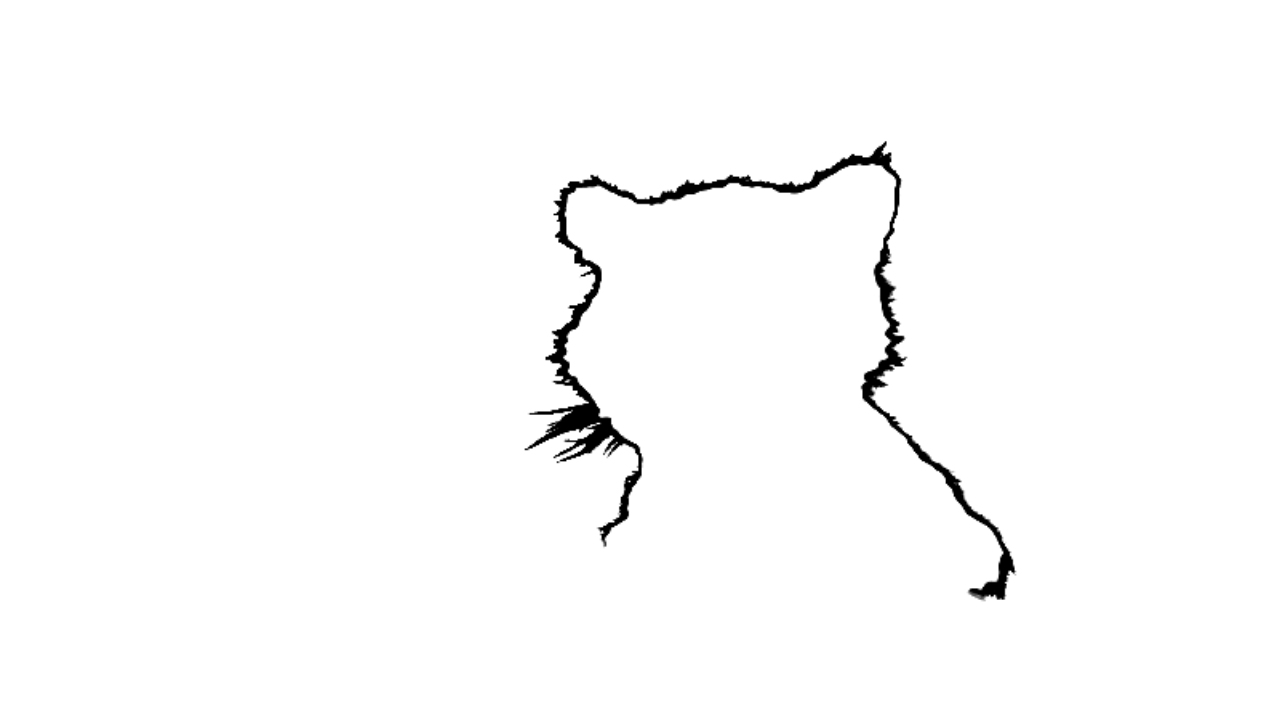}};
            \spy on \zoomthree in node [left] at \rebigthree;
    	\end{tikzpicture}
      \end{subfigure}
    \begin{subfigure}{\depthWidth}
        \begin{tikzpicture}[spy using outlines={green,magnification=\ssmag,size=\ssizz},inner sep=0]
            \node [align=center, img] {\includegraphics[width=\textwidth]{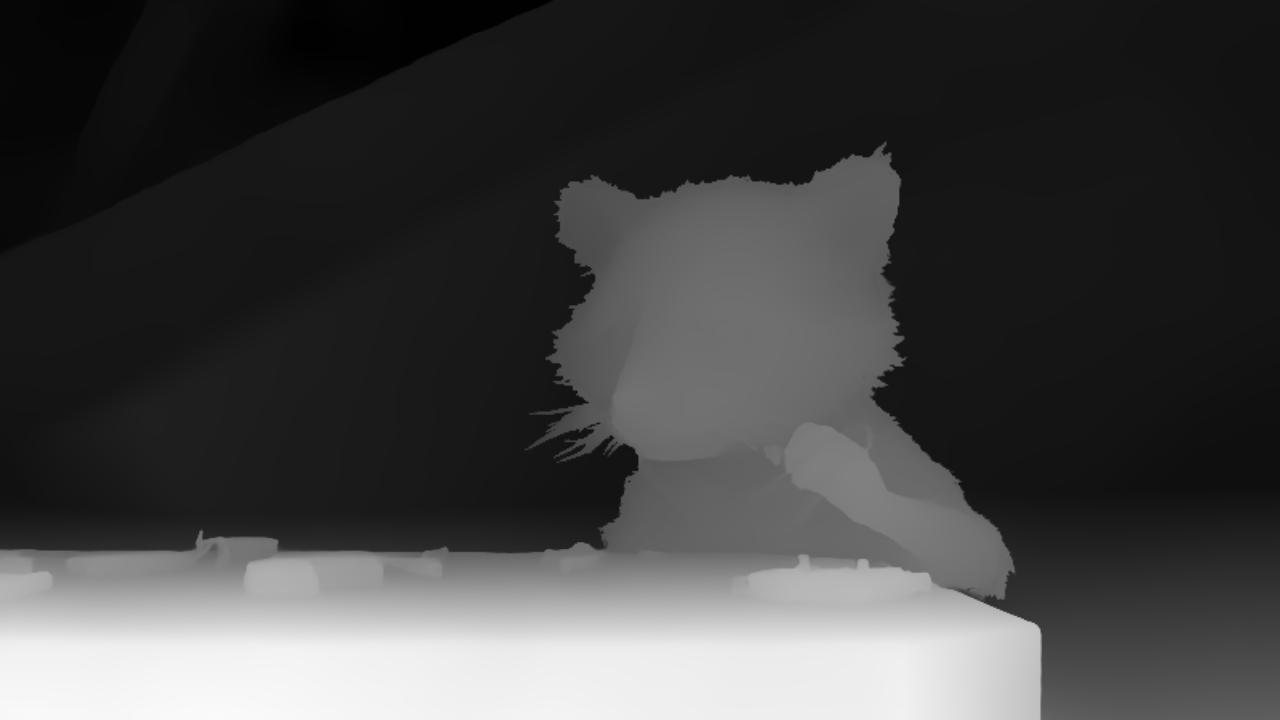}};
            \spy on \zoomthree in node [left] at \rebigthree;
    	\end{tikzpicture}
      \end{subfigure}
      \\
          \begin{subfigure}{\depthWidth}
		\begin{tikzpicture}[spy using outlines={green,magnification=\ssmag,size=\ssizz},inner sep=0]
            \node [align=center, img] {\includegraphics[width=\textwidth]{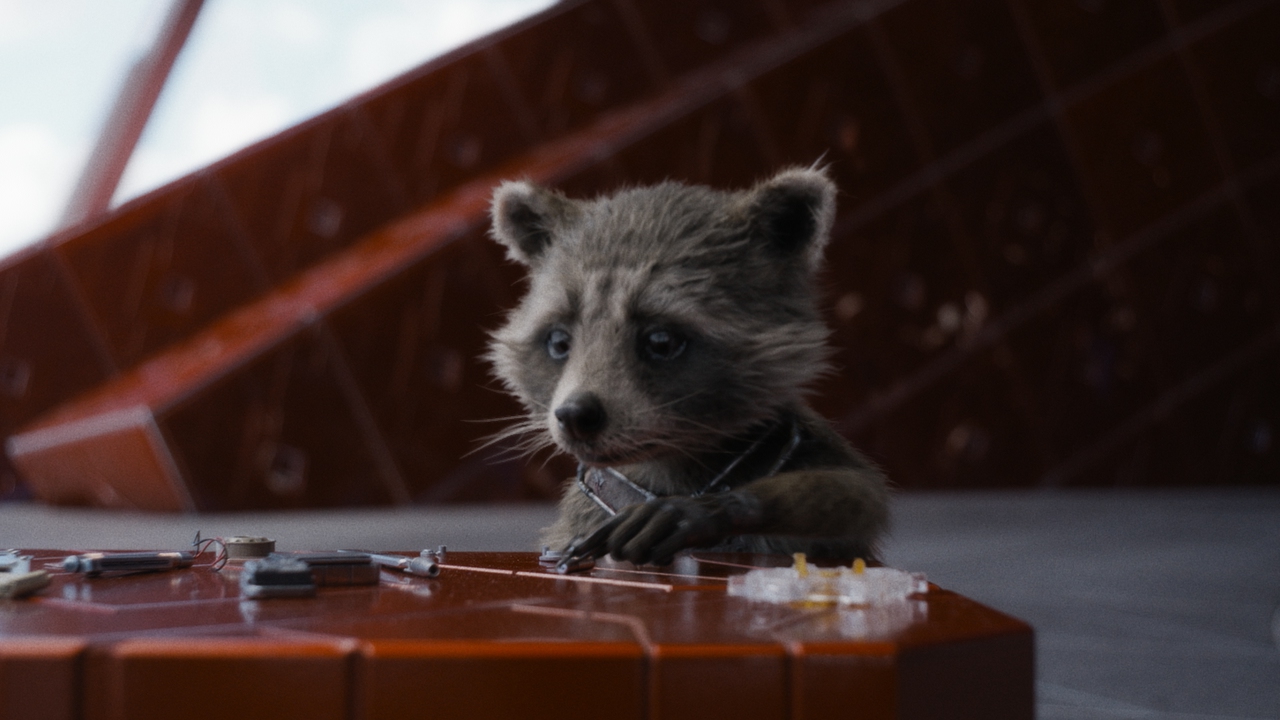}};
            \spy on \zoomfour in node [left] at \rebigthree;
    	\end{tikzpicture}
    \end{subfigure}
    \begin{subfigure}{\depthWidth}
        \begin{tikzpicture}[spy using outlines={green,magnification=\ssmag,size=\ssizz},inner sep=0]
            \node [align=center, img] {\includegraphics[width=\textwidth]{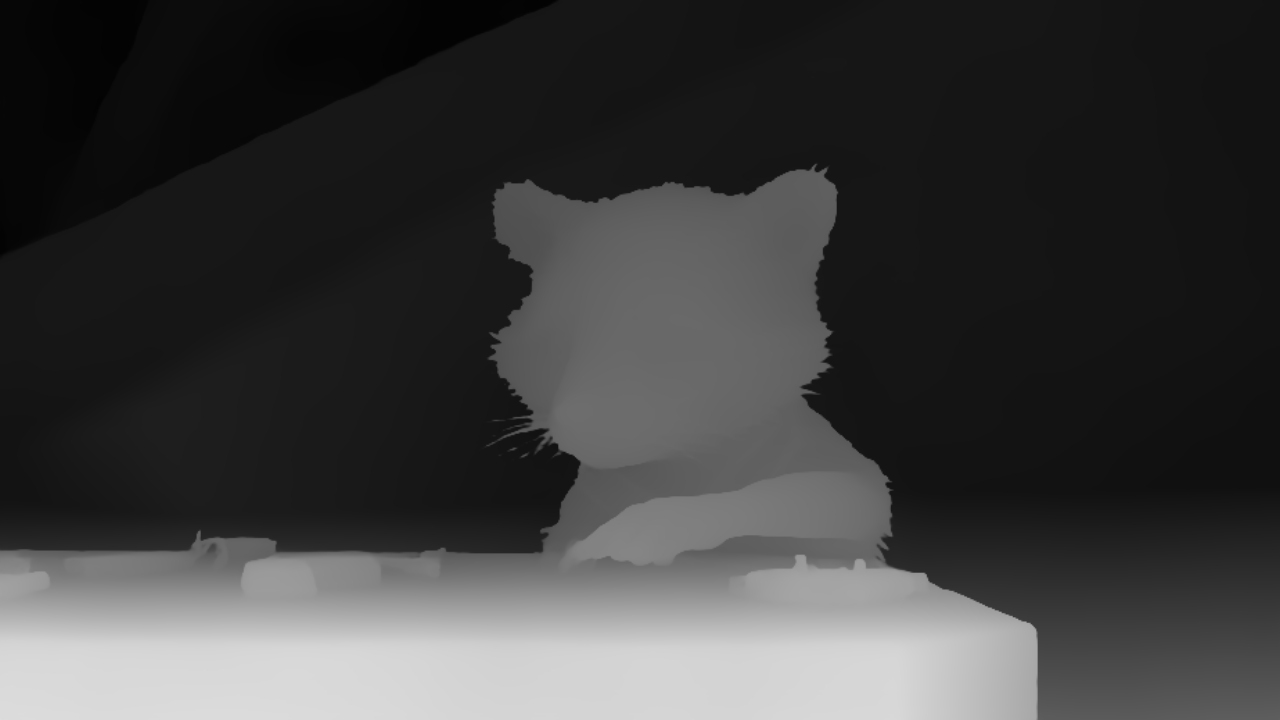}};
            \spy on \zoomfour in node [left] at \rebigthree;
    	\end{tikzpicture}
      \end{subfigure}
    \begin{subfigure}{\depthWidth}
        \begin{tikzpicture}[spy using outlines={green,magnification=\ssmag,size=\ssizz},inner sep=0]
            \node [align=center, img] {\includegraphics[width=\textwidth]{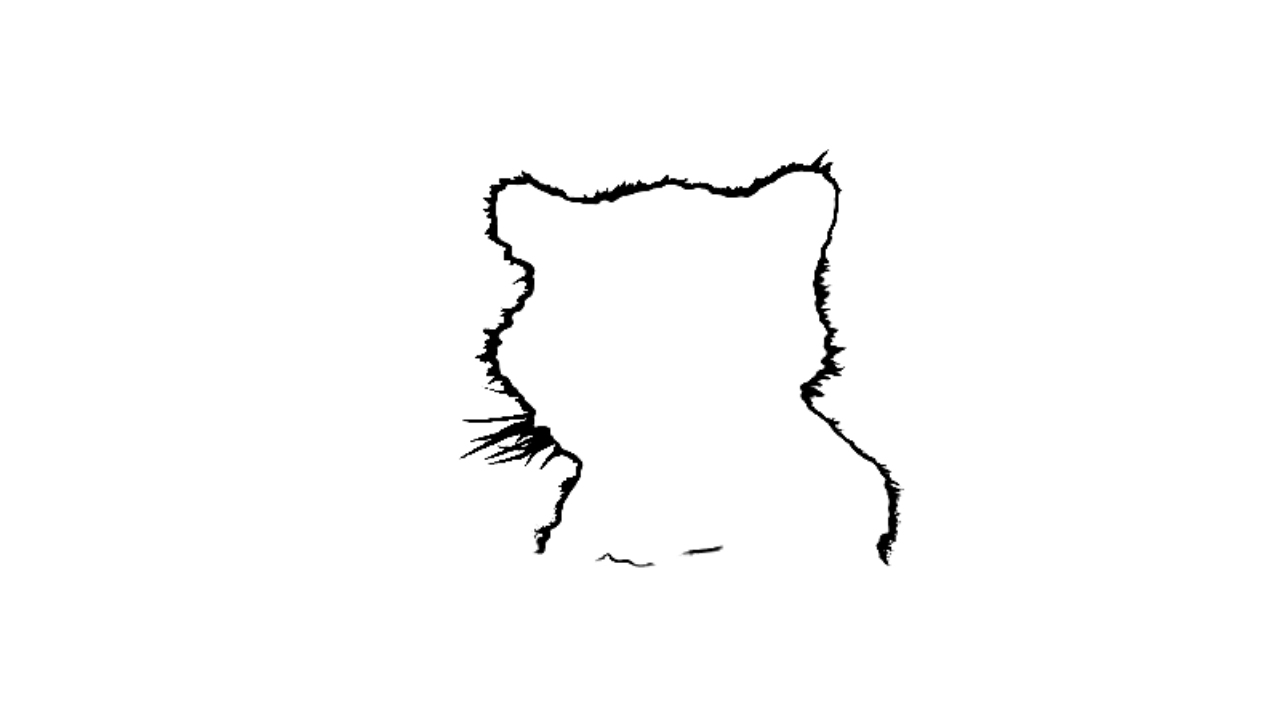}};
            \spy on \zoomfour in node [left] at \rebigthree;
    	\end{tikzpicture}
      \end{subfigure}
    \begin{subfigure}{\depthWidth}
        \begin{tikzpicture}[spy using outlines={green,magnification=\ssmag,size=\ssizz},inner sep=0]
            \node [align=center, img] {\includegraphics[width=\textwidth]{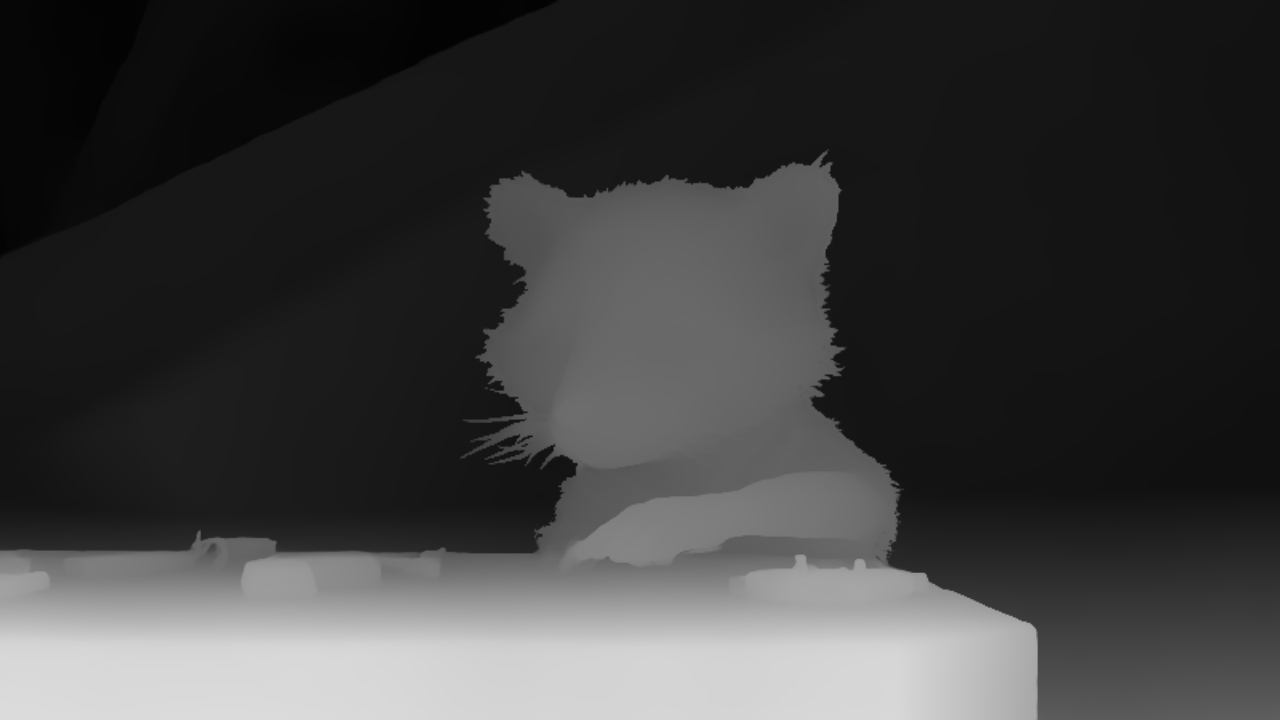}};
            \spy on \zoomfour in node [left] at \rebigthree;
    	\end{tikzpicture}
      \end{subfigure}
      \\
          \begin{subfigure}{\depthWidth}
		\begin{tikzpicture}[spy using outlines={green,magnification=\ssmag,size=\ssizz},inner sep=0]
            \node [align=center, img] {\includegraphics[width=\textwidth]{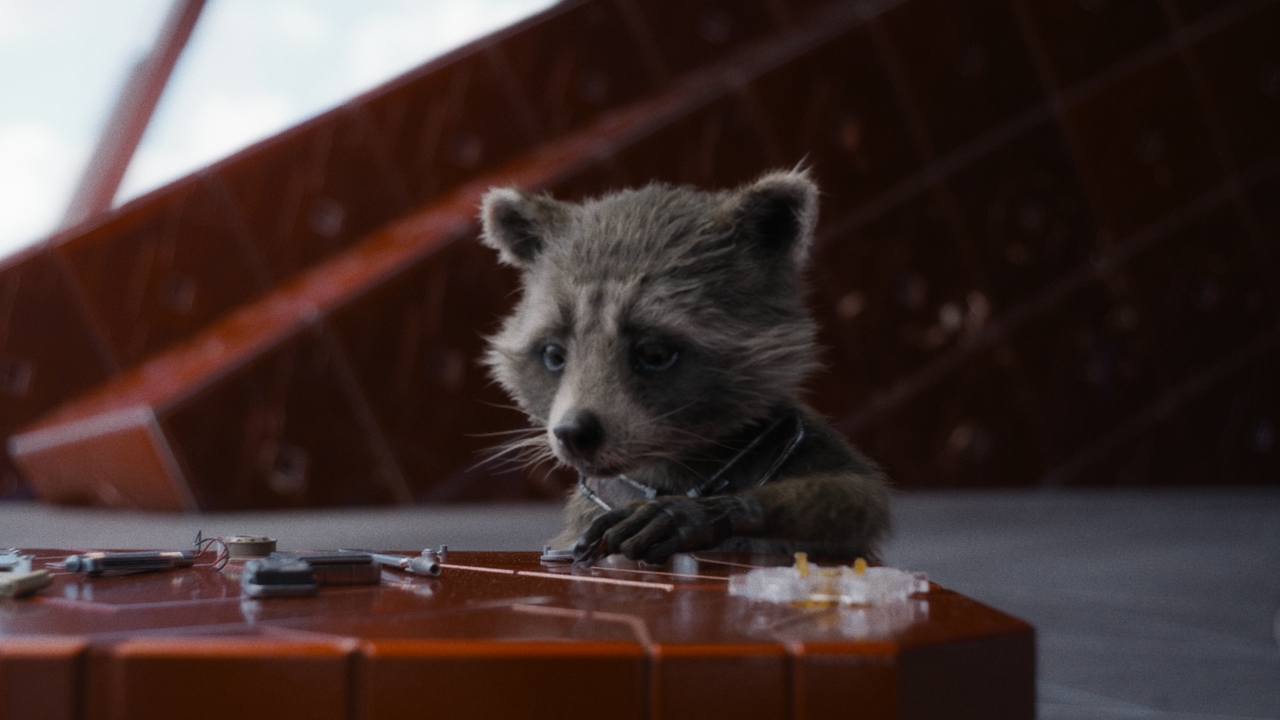}};
            \spy on \zoomfive in node [left] at \rebigthree;
    	\end{tikzpicture}
    \end{subfigure}
    \begin{subfigure}{\depthWidth}
        \begin{tikzpicture}[spy using outlines={green,magnification=\ssmag,size=\ssizz},inner sep=0]
            \node [align=center, img] {\includegraphics[width=\textwidth]{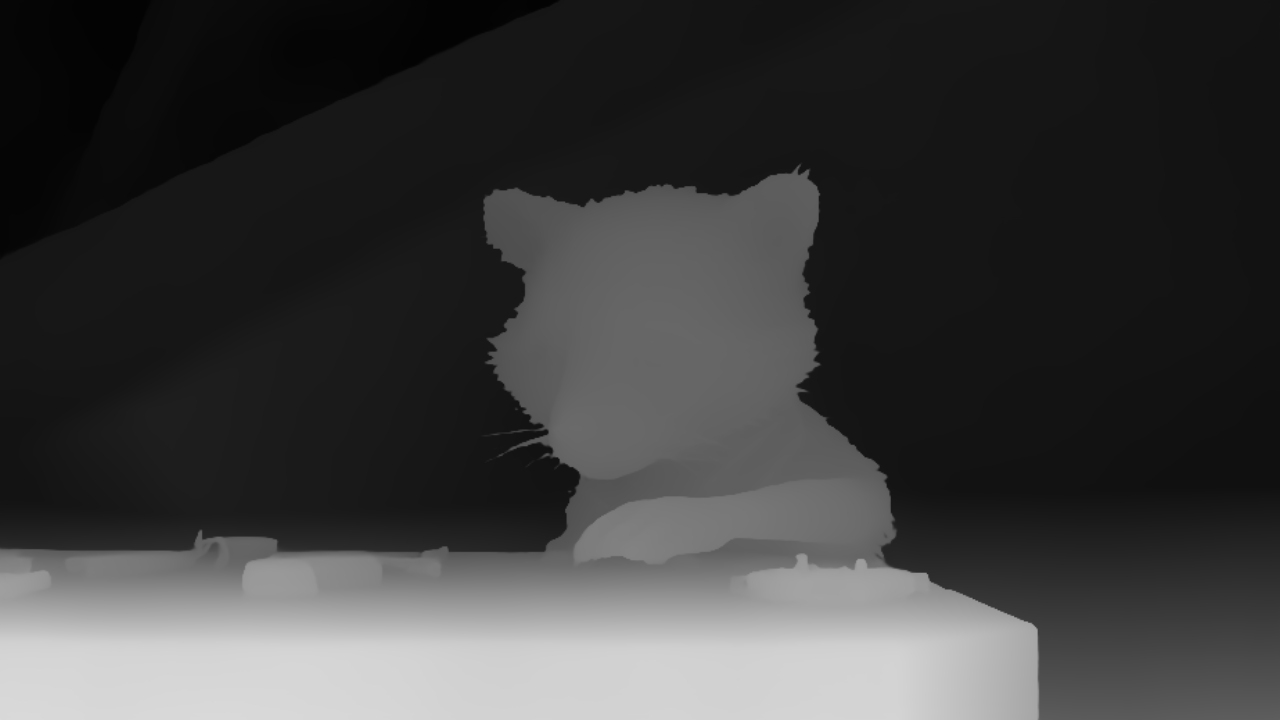}};
            \spy on \zoomfive in node [left] at \rebigthree;
    	\end{tikzpicture}
      \end{subfigure}
    \begin{subfigure}{\depthWidth}
        \begin{tikzpicture}[spy using outlines={green,magnification=\ssmag,size=\ssizz},inner sep=0]
            \node [align=center, img] {\includegraphics[width=\textwidth]{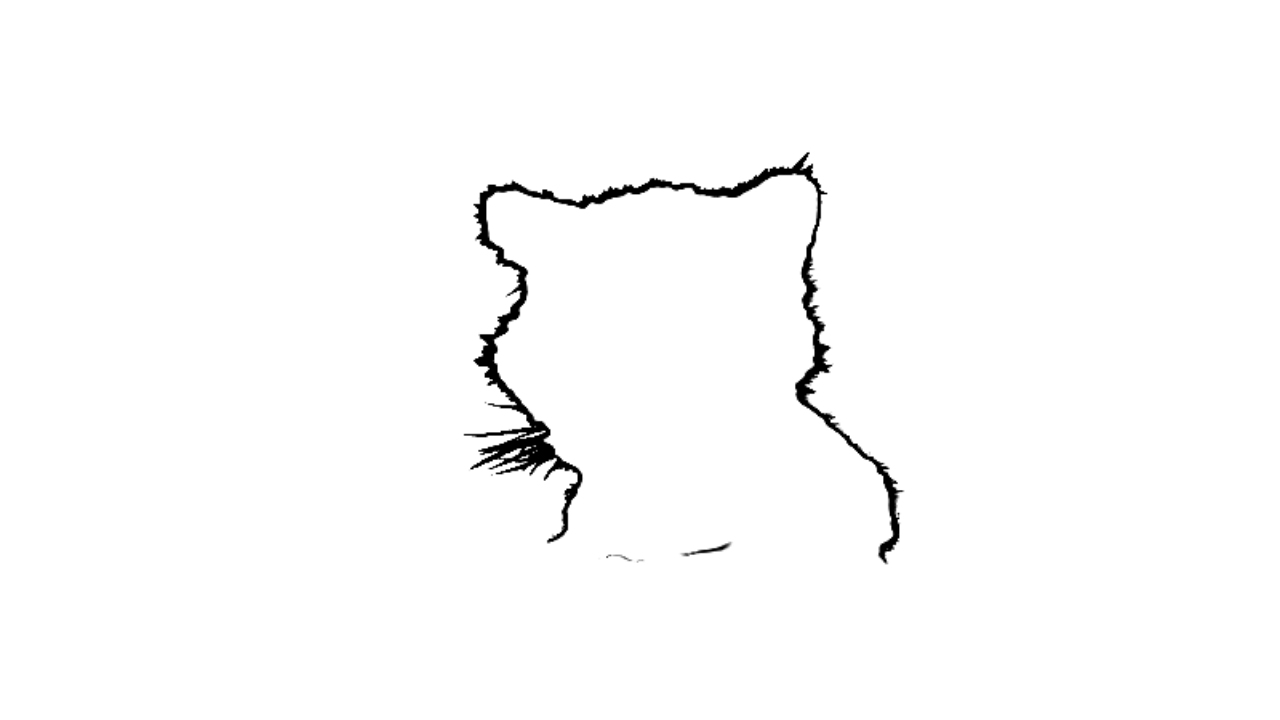}};
            \spy on \zoomfive in node [left] at \rebigthree;
    	\end{tikzpicture}
      \end{subfigure}
    \begin{subfigure}{\depthWidth}
        \begin{tikzpicture}[spy using outlines={green,magnification=\ssmag,size=\ssizz},inner sep=0]
            \node [align=center, img] {\includegraphics[width=\textwidth]{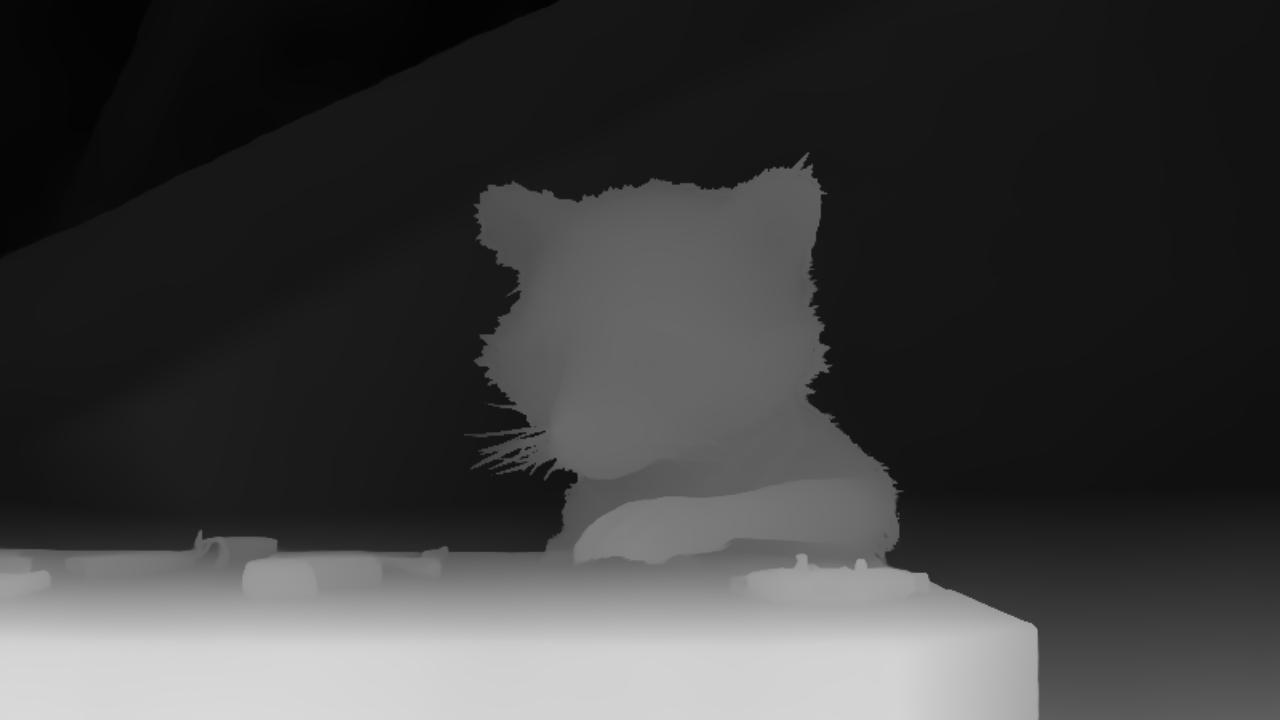}};
            \spy on \zoomfive in node [left] at \rebigthree;
    	\end{tikzpicture}
      \end{subfigure}
          \\
          \begin{subfigure}{\depthWidth}
		\begin{tikzpicture}[spy using outlines={green,magnification=\ssmag,size=\ssizz},inner sep=0]
            \node [align=center, img] {\includegraphics[width=\textwidth]{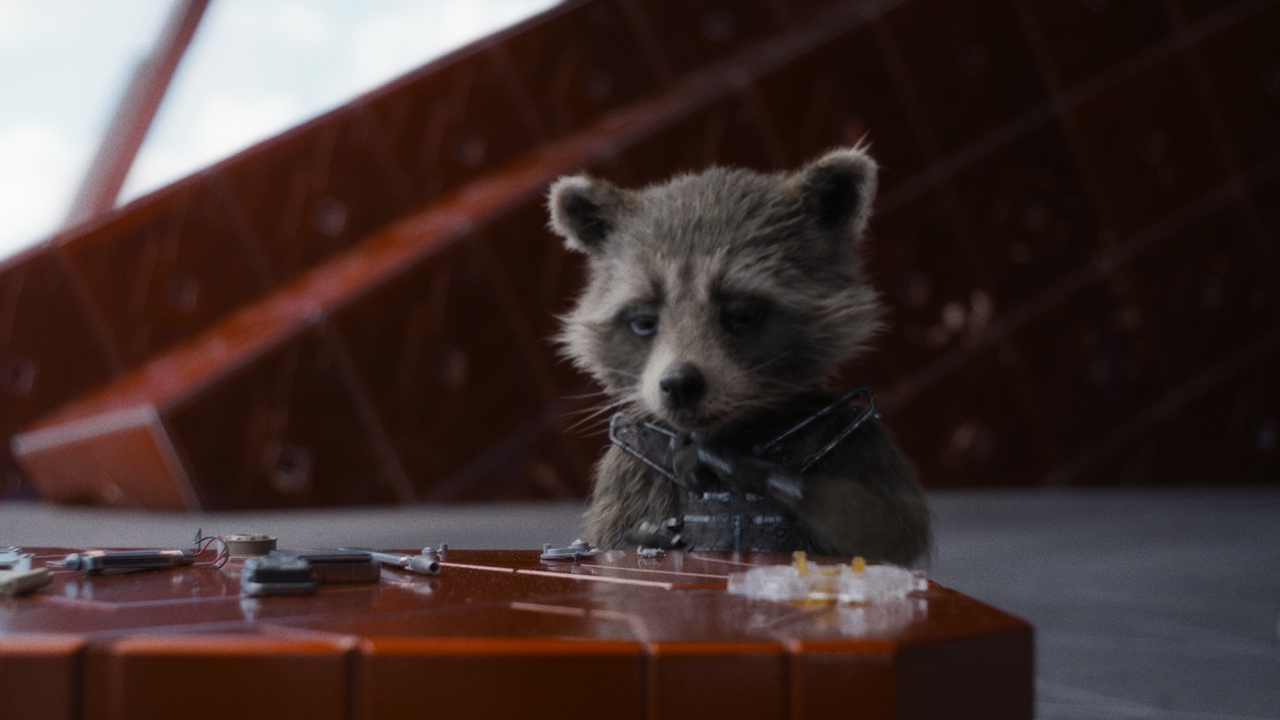}};
            \spy on \zoomsix in node [left] at \rebigthree;
    	\end{tikzpicture}
        \caption*{Input Video Frames}
    \end{subfigure}
    \begin{subfigure}{\depthWidth}
        \begin{tikzpicture}[spy using outlines={green,magnification=\ssmag,size=\ssizz},inner sep=0]
            \node [align=center, img] {\includegraphics[width=\textwidth]{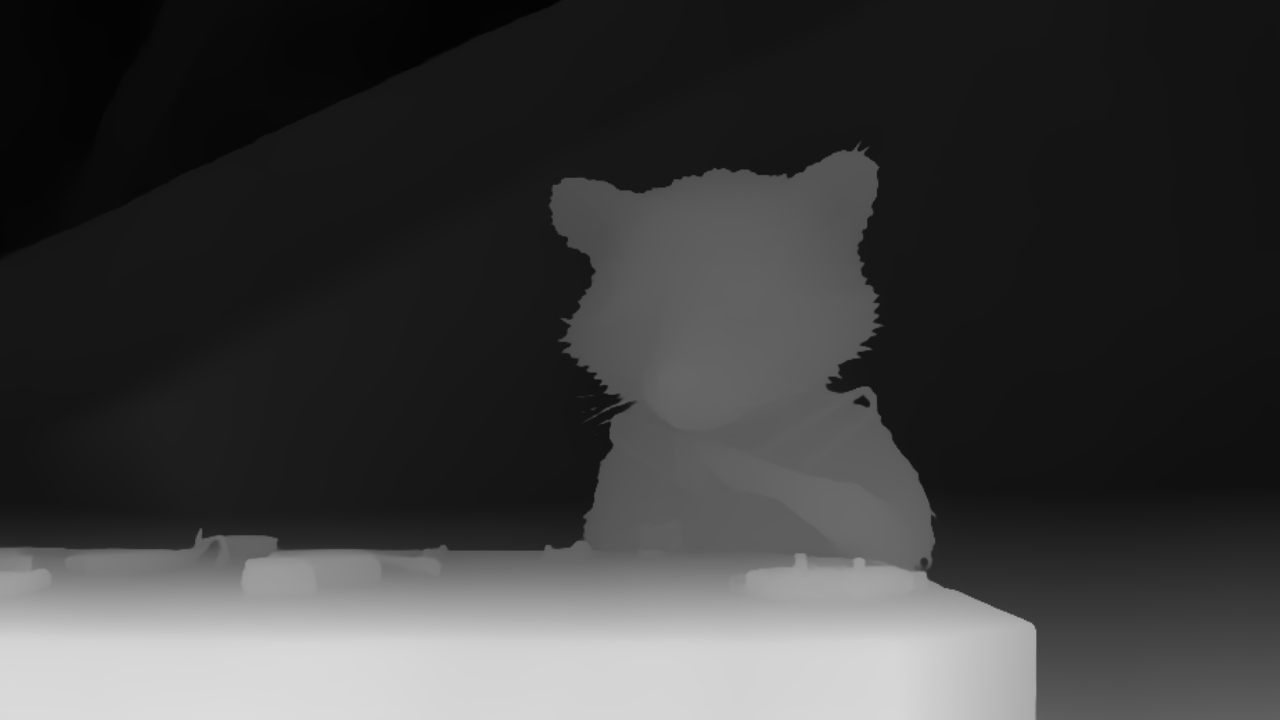}};
            \spy on \zoomsix in node [left] at \rebigthree;
    	\end{tikzpicture}
        \caption*{Video Depth Anything}
      \end{subfigure}
    \begin{subfigure}{\depthWidth}
        \begin{tikzpicture}[spy using outlines={green,magnification=\ssmag,size=\ssizz},inner sep=0]
            \node [align=center, img] {\includegraphics[width=\textwidth]{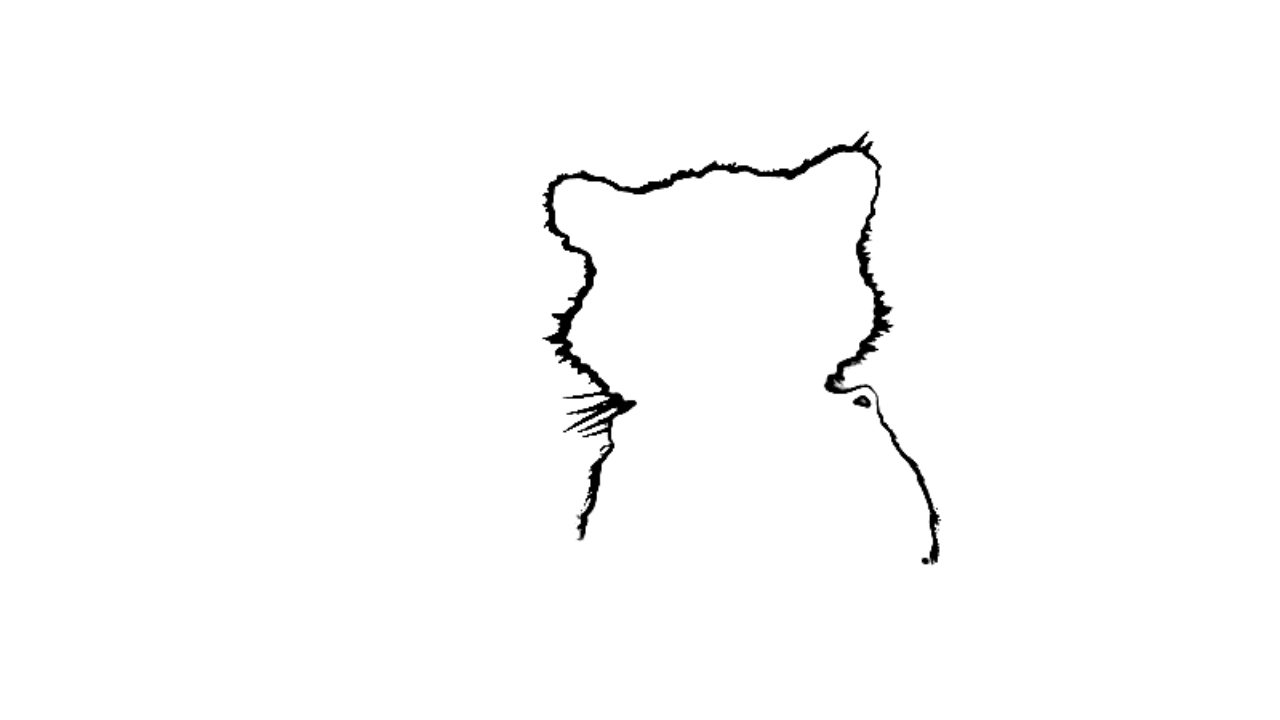}};
            \spy on \zoomsix in node [left] at \rebigthree;
    	\end{tikzpicture}
        \caption*{Predicted Gate}
      \end{subfigure}
    \begin{subfigure}{\depthWidth}
        \begin{tikzpicture}[spy using outlines={green,magnification=\ssmag,size=\ssizz},inner sep=0]
            \node [align=center, img] {\includegraphics[width=\textwidth]{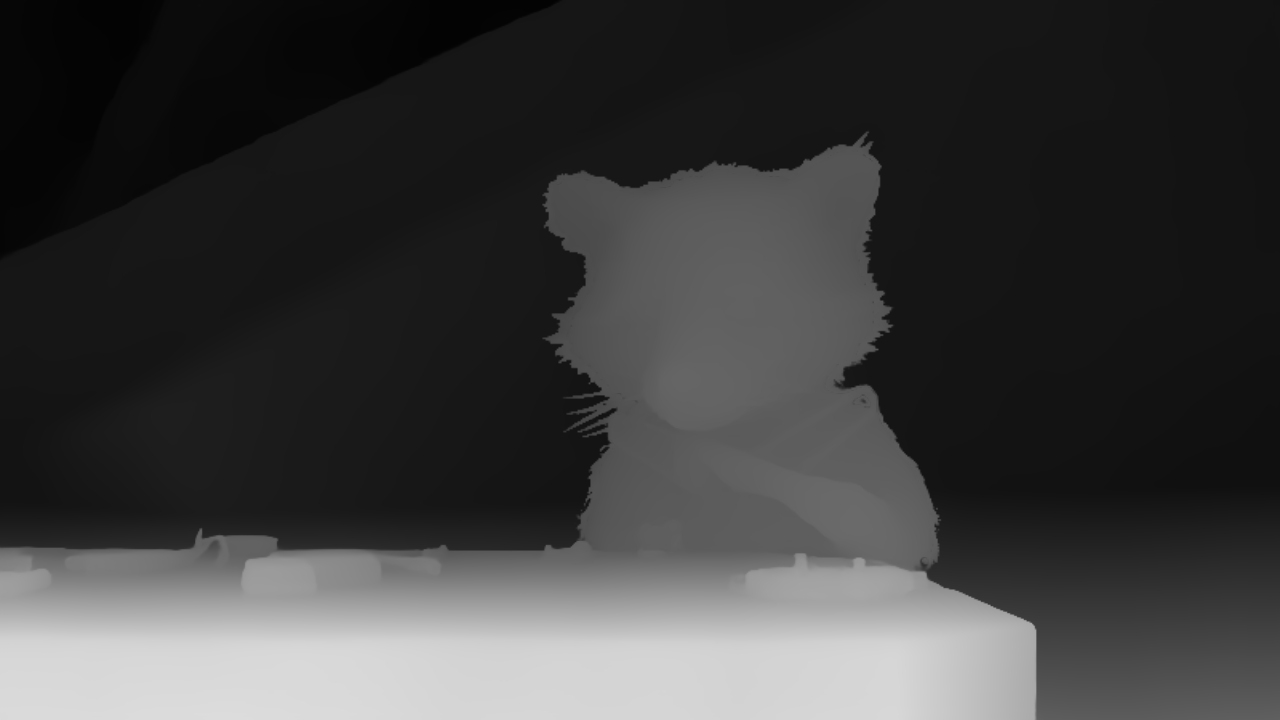}};
            \spy on \zoomsix in node [left] at \rebigthree;
    	\end{tikzpicture}
        \caption*{Fixed Depth (Ours)}
      \end{subfigure}
      \caption{Plug-and-play refinement on video-based depth models}
      \label{fig:supp-plug-video-depth}
    \end{subfigure}
    \vspace{-1.5em}
    \caption{\textbf{Plug-and-play performance of depth fixer.} The depth fixer can be integrated with different depth models, \eg, image-based models in (a) and video-based models in (b),  in a plug-and-play fashion for soft boundary refinement. Although the depth fixer is trained only on image datasets, it can be directly applied to improve video-based models such as Video Depth Anything~\cite{chen2025videodepthanything}, without additional re-training. Leveraging the gated residual mechanism, the depth fixer preserves the temporal consistency of the video depth model while achieving stable performance in identifying soft boundaries and recovering fine-grained details, even in complex scenes with occlusions. }
    \label{fig:supp-plug-and-play}
\end{figure*}

%% file: tabs/supp/tab-plug-splatdiff.tex
\begin{table*}[ht]
\centering
\caption{\textbf{Plug-and-play stereo image/video conversion performance} on the Marvel-10K dataset. The \textcolor{red}{best} results are marked. }
\vspace{-0.8em}
\label{tab:supp-plug-splatdiff}
\footnotesize
\setlength\tabcolsep{2.5pt}
\begin{tabular}{lccccccccccccc}
\toprule
\rowcolor{color3} & \multicolumn{6}{c}{\textbf{Stereo Image Conversion}}                          &  & \multicolumn{6}{c}{\textbf{Stereo Video Conversion}}                          \\
\cline{2-7} \cline{9-14} \rowcolor{color3}                               \multirow{-2}{*}{\textbf{Method}}      & PSNR $\uparrow$ & SSIM $\uparrow$ & RMSE $\downarrow$ & LPIPS $\downarrow$ & DISTS $\downarrow$ & SIoU $\uparrow$ &  & PSNR $\uparrow$ & SSIM $\uparrow$ & RMSE $\downarrow$ & LPIPS $\downarrow$ & DISTS $\downarrow$ & SIoU $\uparrow$ \\ \midrule
SplatDiff~\cite{splatdiff} & {\color{black} 36.23}     & {\color{black} 0.8857}    & {\color{black} 4.06}        & {\color{black} 0.1116}       & {\color{black} 0.0435}       & {\color{black} 0.3259}    &  & {\color{black} 36.24}     & {\color{black} 0.8858}    & {\color{black} 4.06}        & {\color{black} 0.1114}       & {\color{black} 0.0437}       & {\color{black}0.3280}    \\
\rowcolor{color3} \textbf{SplatDiff+Depth Fixer (Ours)}                             &  {\color{red} 36.38}         & {\color{red}0.8915}          & {\color{red}4.00}            & {\color{red}0.0974}             &  {\color{red}0.0348}            &  {\color{red}0.3309}         &  &   {\color{red}36.39}        &     {\color{red}0.8917}      &  {\color{red}3.99}           &   {\color{red}0.0972}           &  {\color{red}0.0351}            &   {\color{red}0.3326}        \\\bottomrule
\end{tabular}
\vspace{-1em}
\end{table*}

%% file: tabs/supp/tab-complexity.tex
\begin{table}[th]
\centering
\caption{\textbf{Complexity comparison} with previous state-of-the-art novel view synthesis methods on the Marvel-10K dataset at a resolution of $384\times640$, evaluated using an NVIDIA GeForce RTX 4090 GPU. For diffusion-based methods, we only take into account the model sizes of the latent diffusion model and the VAE model. * means that the method runs out of memory, and thus we perform inference at a lower resolution $256\times448$ for reference. The complexity of each component in \myname\, \ie, depth fixer, scene painter, and color fuser, is also reported. Since the three components are applied sequentially, the peak GPU memory of \myname\ equals that of the scene painter. The \textcolor{red}{best} and \textcolor{blue}{second-best} results are marked. }
\vspace{-0.8em}
\label{tab:supp-complexity}
\footnotesize
\setlength\tabcolsep{3pt}
\begin{tabular}{lccc}
\toprule
\rowcolor{color3} \textbf{Method}        & \textbf{Model Size} & \textbf{Peak GPU Mem.} & \textbf{Infer. Speed}    \\ \midrule
ViewCrafter~\cite{yu2024viewcrafter}   & 2.22 B     & \textcolor{blue}{14.91 G}       & \textcolor{red}{47.83 s}  \\
NVS-Solver*~\cite{you2024nvssolver}    & 2.25 B     & 21.60 G       & 100.00 s \\
ReCamMaster~\cite{recammaster}   & \textcolor{red}{1.51 B}     & 17.11 G       & 684.44 s \\
SplatDiff~\cite{splatdiff}     & 2.28 B     & 22.88 G       & \textcolor{blue}{52.28 s}  \\
\rowcolor{color3} \textbf{\myname\ (Ours)}          & \textcolor{blue}{1.86 B}     & \textcolor{red}{10.65 G}       & 95.23 s  \\ \midrule
\rowcolor{color3} \textbf{Depth Fixer (Ours)}   & 0.32 B     & 1.84 G        & 0.03 s   \\
\rowcolor{color3} \textbf{Scene Painter (Ours)} & 1.44 B     & 10.65 G       & 95.19 s  \\
\rowcolor{color3} \textbf{Color Fuser (Ours)}   & 0.10 B     & 3.15 G        & 0.01 s   \\ \bottomrule
\end{tabular}
\end{table}

%% file: tabs/supp/tab-depth-warping.tex
\begin{table}[t]
\centering
\caption{\textbf{Warping performance} using different depth maps on the Marvel-10K dataset. Metrics are computed only on the soft boundary regions. \textcolor{red}{Best} results are marked. }
\vspace{-0.8em}
\label{tab:supp-depth-warping}
\footnotesize
\setlength\tabcolsep{2.5pt}
\begin{tabular}{lccc}
\toprule
\rowcolor{color3} \textbf{Method}                                 & PSNR $\uparrow$  & SSIM $\uparrow$  & RMSE $\downarrow$  \\ \midrule
Depth Anything V2~\cite{yang2024depthanythingv2}                                   & 30.18 & 0.4495 & 8.08 \\
\rowcolor{color3} \textbf{Depth Anything V2+Depth Fixer (Ours)}       & \textcolor{red}{31.07} & \textcolor{red}{0.5140} & \textcolor{red}{7.36}  \\ \midrule
Depth Pro~\cite{depthpro}                              & 31.12 & 0.5591 & 7.28  \\
\rowcolor{color3} \textbf{Depth Pro+Depth Fixer (Ours)}  & \textcolor{red}{31.77} & \textcolor{red}{0.6144} & \textcolor{red}{6.79} \\ \midrule
UniDepthV2~\cite{unidepthv2}                             & 31.31 & 0.5637 & 7.12 \\
\rowcolor{color3} \textbf{UniDepthV2+Depth Fixer (Ours)} & \textcolor{red}{32.23} & \textcolor{red}{0.6261} & \textcolor{red}{6.46}  \\ \bottomrule
\end{tabular}
\end{table}

%% file: tabs/supp/tab-supp-fixer-ablation.tex
\begin{table}[t]
\centering
\caption{\textbf{Ablation study of depth fixer} on the Marvel-10K dataset. The ablations about gated residual, loss function, model prior, edge guidance, and alpha threshold correspond to experiments \#1-3, \#4-5, \#6-7, \#8-9, and \#10-12, respectively. We use Depth Anything V2 as the base depth model~\cite{yang2024depthanythingv2}. Metrics are computed only on the soft boundary regions. \textcolor{red}{Best} results are marked.}
\vspace{-0.8em}
\label{tab:supp-fixer-ablation}
\footnotesize
\setlength\tabcolsep{4pt}
\begin{tabular}{ccccccc}
\toprule
\rowcolor{color3}                     &  &                              &  & \multicolumn{3}{c}{\textbf{Marvel-10K}}                                                             \\ \cline{5-7} 
\rowcolor{color3} \multirow{-2}{*}{\textbf{Exp}} &  & \multirow{-2}{*}{\textbf{Strategies}} &  & PSNR $\uparrow$                         & SSIM $\uparrow$                         & RMSE $\downarrow$                       \\ \midrule
\#1                  &  & Direct Prediction            &  & 30.66                        & 0.5124                        & 7.67                        \\
\#2                  &  & Vanilla Residual             &  & 30.37                        & 0.5009                        & 7.88                        \\ 
\rowcolor{color3} \#3                  &  & {Gated Residual (Ours)}      &  & \textcolor{red}{31.07}                        & \textcolor{red}{0.5140}                        & \textcolor{red}{7.36}                        \\ \midrule
\#4                  &  & $\mathcal{L}_1$ Only                      &  & 30.58                        & 0.5057                        & 7.74                        \\ 
\rowcolor{color3} \#5                  &  & {$\mathcal{L}_1$ + $\mathcal{L}_{\alpha}$ (Ours)}        &  & {\color{red} 31.07} & {\color{red} 0.5140} & {\color{red} 7.36} \\ \midrule
\#6                  &  & w/o Model Prior              &  & 30.26                        & 0.4668                        & 8.00                        \\
\rowcolor{color3} \#7                  &  & {w/ Model Prior (Ours)}      &  & {\color{red} 31.07} & {\color{red} 0.5140} & {\color{red} 7.36} \\ \midrule
\#8                  &  & w/o Edge Guidance            &  & 30.66                        & 0.5057                        & 7.67                        \\ 
\rowcolor{color3} \#9                  &  & {w/ Edge Guidance (Ours)}    &  & {\color{red} 31.07} & {\color{red} 0.5140} & {\color{red} 7.36} \\ \midrule
\#10                 &  & $\alpha_{th}=0.1$                &  & 30.43                        & 0.4774                        & 7.87                        \\ 
\#11                 &  & $\alpha_{th}=0.05$               &  & 30.55                        & 0.4917                        & 7.76                        \\ 
\rowcolor{color3} \#12                 &  & $\alpha_{th}=0.02$ (Ours)      &  & {\color{red} 31.07} & {\color{red} 0.5140} & {\color{red} 7.36} \\
\bottomrule
\end{tabular}
\end{table}

%% file: figs/supp/fig-alpha_thresh.tex
\def\imgWidth{0.49\linewidth} %
\def\scc{(-1.9,-1.4)}

\def\rebigone{(-0.8, -0.78)} %
\def\rebigtwo{(-0.8, 0.78)} %

\def\zoomone{(-0.32,-0.05)} %
\def\zoomtwo{(0.5,1.2)} %

\def\ssizz{1.2cm} %
\def\ssmag{3}

\begin{figure}[t]
\centering
\tikzstyle{img} = [rectangle, minimum width=\imgWidth, draw=black]
    \centering
    \begin{subfigure}{\imgWidth}
        \begin{tikzpicture}[spy using outlines={green,magnification=\ssmag,size=\ssizz},inner sep=0]
            \node [align=center, img] {\includegraphics[width=\textwidth]{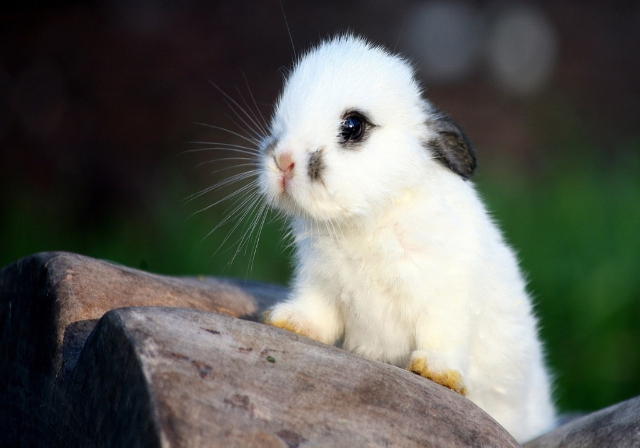}};
            \spy on \zoomone in node [left] at \rebigone;
            \spy [draw=red] on \zoomtwo in node [left] at \rebigtwo;
    	\end{tikzpicture}
        \caption*{Input Image}
    \end{subfigure}
    \begin{subfigure}{\imgWidth}
		\begin{tikzpicture}[spy using outlines={green,magnification=\ssmag,size=\ssizz},inner sep=0]
            \node [align=center, img] {\includegraphics[width=\textwidth]{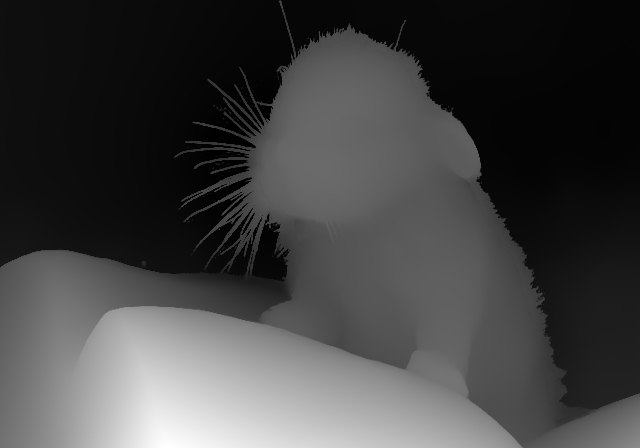}};
            \spy on \zoomone in node [left] at \rebigone;
            \spy [draw=red] on \zoomtwo in node [left] at \rebigtwo;
    	\end{tikzpicture}
        \caption*{$\alpha_{th}=0.02$}
    \end{subfigure}
    \begin{subfigure}{\imgWidth}
        \begin{tikzpicture}[spy using outlines={green,magnification=\ssmag,size=\ssizz},inner sep=0]
            \node [align=center, img] {\includegraphics[width=\textwidth]{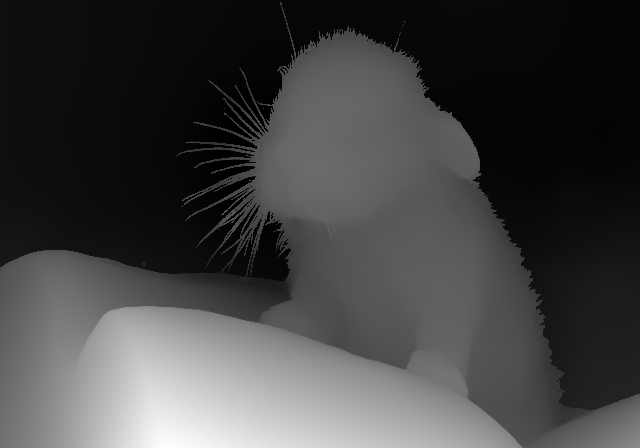}};
            \spy on \zoomone in node [left] at \rebigone;
            \spy [draw=red] on \zoomtwo in node [left] at \rebigtwo;
    	\end{tikzpicture}
        \caption*{$\alpha_{th}=0.05$}
      \end{subfigure}
    \begin{subfigure}{\imgWidth}
        \begin{tikzpicture}[spy using outlines={green,magnification=\ssmag,size=\ssizz},inner sep=0]
            \node [align=center, img] {\includegraphics[width=\textwidth]{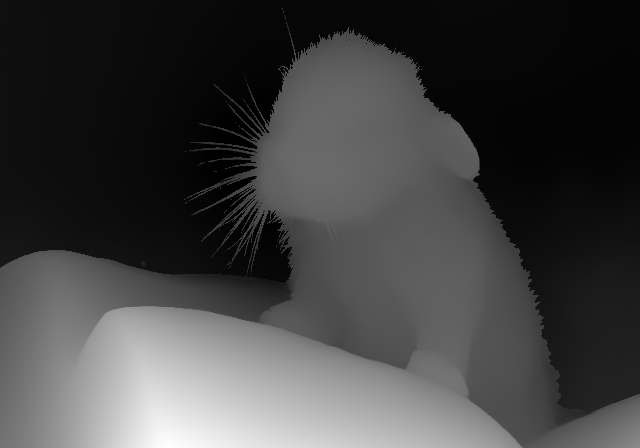}};
            \spy on \zoomone in node [left] at \rebigone;
            \spy [draw=red] on \zoomtwo in node [left] at \rebigtwo;
    	\end{tikzpicture}
        \caption*{$\alpha_{th}=0.1$}
      \end{subfigure}
      \vspace{-0.5em}
    \caption{\textbf{Performance of the depth fixer trained with different alpha thresholds.} A higher threshold $\alpha_{th}$ leads to less redundant background in the depth map (\eg, \textcolor{green}{green} box), while a lower threshold $\alpha_{th}$ improves the coverage of fine-grained details, \eg, the very thin hair in the \textcolor{red}{red} box. }
    \label{fig:supp-alpha_thresh}
\end{figure}

%% file: tabs/supp/tab-fuser-ablation.tex
\begin{table}[th]
\centering
\caption{\textbf{Ablation study of color fuser} on the Marvel-10K dataset. The ablations about VAE prior and skip mechanisms correspond to experiments \#1-2 and \#3-5, respectively. \textcolor{red}{Best} results are marked.}
\vspace{-0.8em}
\label{tab:supp-fuser-ablation}
\footnotesize
\setlength\tabcolsep{3.5pt}
\begin{tabular}{cccccccc}
\toprule
\rowcolor{color3} &  & &  & \multicolumn{4}{c}{\textbf{Marvel-10K}}  \\  \cline{5-8} \rowcolor{color3} \multirow{-2}{*}{\textbf{Exp}} &  & \multirow{-2}{*}{\textbf{Strategies}}
                           &  & PSNR $\uparrow$  & LPIPS $\downarrow$  & DISTS $\downarrow$  & FID $\downarrow$  \\ \midrule
\#1                  &  & w/o VAE Prior               &  & \textcolor{red}{36.61} & 0.0965 & 0.0366 & 7.99  \\
\rowcolor{color3} \#2                  &  & w/ VAE Prior (Ours)         &  & 36.59 & \textcolor{red}{0.0909} & \textcolor{red}{0.0331} & \textcolor{red}{7.19}  \\ \midrule
\#3                  &  & w/o Skip                    &  & 34.96 & 0.1664 & 0.0807 & 18.11 \\
\#4                  &  & Sinlge Skip                 &  & 36.46 & 0.0919 & 0.0337 & 7.34  \\
\rowcolor{color3} \#5                  &  & Dual Skip (Ours)            &  & \textcolor{red}{36.59} & \textcolor{red}{0.0909} & \textcolor{red}{0.0331} & \textcolor{red}{7.19}  \\ \bottomrule
\end{tabular}
\end{table}

%% file: figs/supp/fig-limitation.tex
\def\imgWidth{0.238\textwidth} %
\def\scc{(-1.9,-1.4)}

\def\rebigtwo{(-0.95, -0.85)} %

\def\rezero{(-0.8,0.15)} %

\def\reone{(0.21,0.4)} %
\def\retwo{(0.4,0.48)} %
\def\rethree{(-0.1,0.5)} %

\def\ssizz{1cm} %
\def\ssmag{3}

\begin{figure*}[!t] 
\centering
\tikzstyle{img} = [rectangle, minimum width=\imgWidth, draw=black]
\centering
\begin{subfigure}{\textwidth}
\centering
    \begin{subfigure}{\imgWidth}
        \begin{tikzpicture}[spy using outlines={green,magnification=\ssmag,size=\ssizz},inner sep=0]
            \node [align=center, img] {\includegraphics[width=\textwidth]{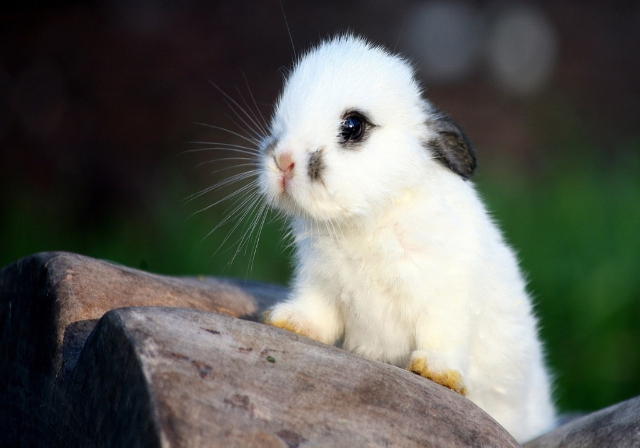}};
            \spy on \rezero in node [left] at \rebigtwo;
    	\end{tikzpicture}
     \caption*{Input Image}
    \end{subfigure}
    \begin{subfigure}{\imgWidth}
        \begin{tikzpicture}[spy using outlines={green,magnification=\ssmag,size=\ssizz},inner sep=0]
            \node [align=center, img] {\includegraphics[width=\textwidth]{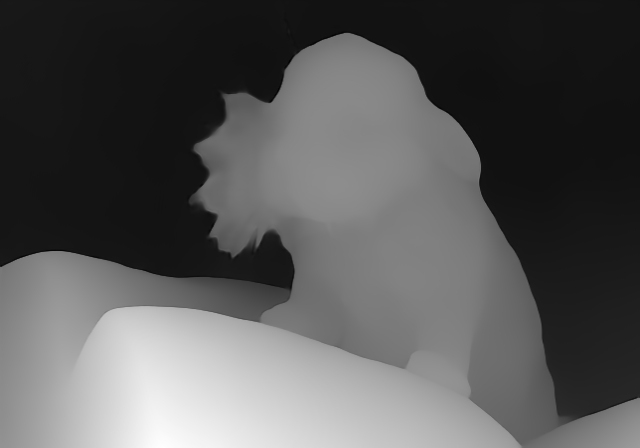}};
            \spy on \rezero in node [left] at \rebigtwo;
    	\end{tikzpicture}
     \caption*{UniDepthV2}
    \end{subfigure}
    \begin{subfigure}{\imgWidth}
		\begin{tikzpicture}[spy using outlines={green,magnification=\ssmag,size=\ssizz},inner sep=0]
            \node [align=center, img] {\includegraphics[width=\textwidth]{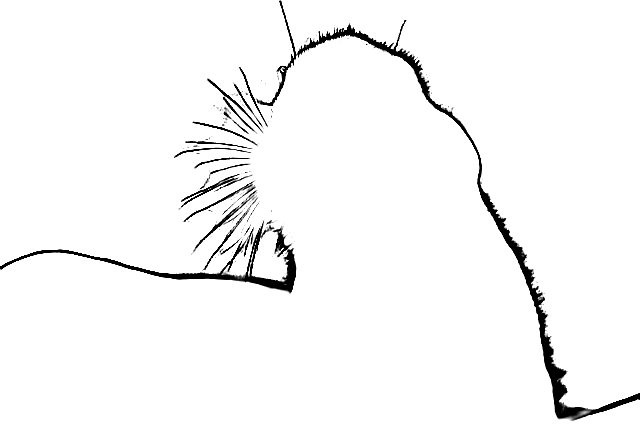}};
            \spy on \rezero in node [left] at \rebigtwo;
    	\end{tikzpicture}
     \caption*{Predicted Gate}
    \end{subfigure}
        \begin{subfigure}{\imgWidth}
        \begin{tikzpicture}[spy using outlines={green,magnification=\ssmag,size=\ssizz},inner sep=0]
            \node [align=center, img] {\includegraphics[width=\textwidth]{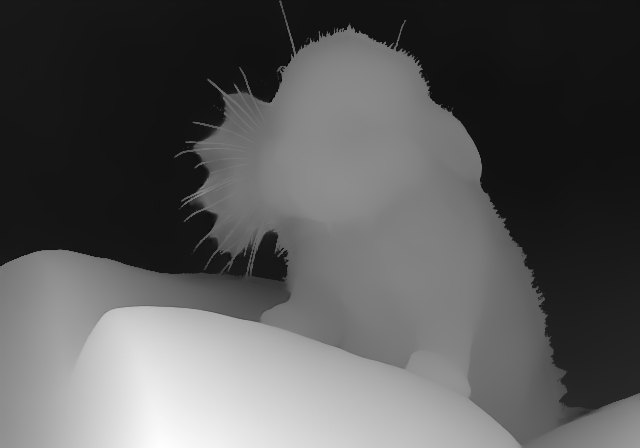}};
            \spy on \rezero in node [left] at \rebigtwo;
    	\end{tikzpicture}
      \caption*{UniDepthV2+Depth Fixer}
    \end{subfigure}
    \caption{Depth errors beyond soft boundaries}
    \label{fig:supp-limitation-depth-base}
      \end{subfigure}
\begin{subfigure}{\textwidth}
\centering
    \begin{subfigure}{\imgWidth}
        \begin{tikzpicture}[spy using outlines={green,magnification=\ssmag,size=\ssizz},inner sep=0]
            \node [align=center, img] {\includegraphics[width=\textwidth]{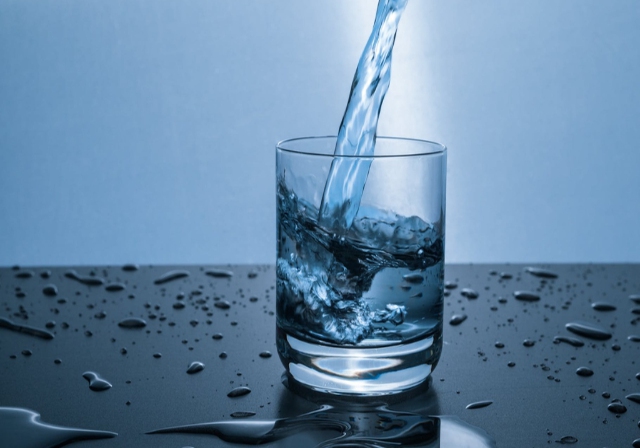}};
            \spy on \reone in node [left] at \rebigtwo;
    	\end{tikzpicture}
     \caption*{Input Image}
    \end{subfigure}
    \begin{subfigure}{\imgWidth}
        \begin{tikzpicture}[spy using outlines={green,magnification=\ssmag,size=\ssizz},inner sep=0]
            \node [align=center, img] {\includegraphics[width=\textwidth]{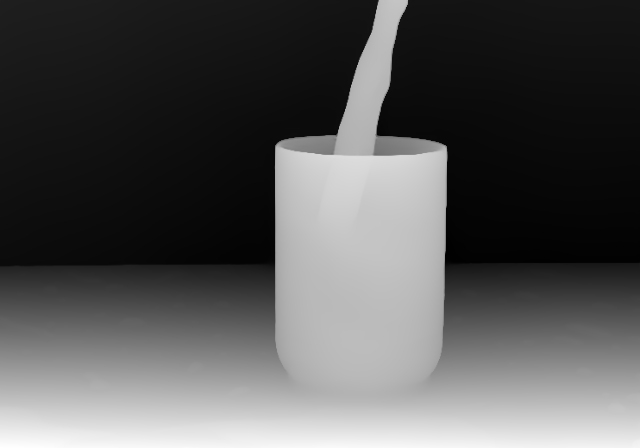}};
            \spy on \reone in node [left] at \rebigtwo;
    	\end{tikzpicture}
     \caption*{Depth}
    \end{subfigure}
    \begin{subfigure}{\imgWidth}
		\begin{tikzpicture}[spy using outlines={green,magnification=\ssmag,size=\ssizz},inner sep=0]
            \node [align=center, img] {\includegraphics[width=\textwidth]{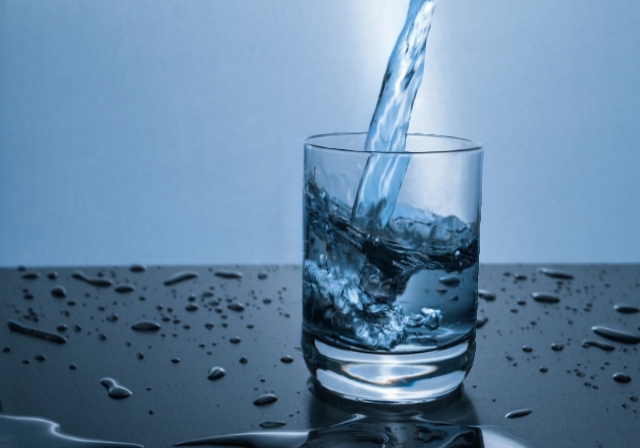}};
            \spy on \retwo in node [left] at \rebigtwo;
    	\end{tikzpicture}
     \caption*{1st Novel View}
    \end{subfigure}
        \begin{subfigure}{\imgWidth}
        \begin{tikzpicture}[spy using outlines={green,magnification=\ssmag,size=\ssizz},inner sep=0]
            \node [align=center, img] {\includegraphics[width=\textwidth]{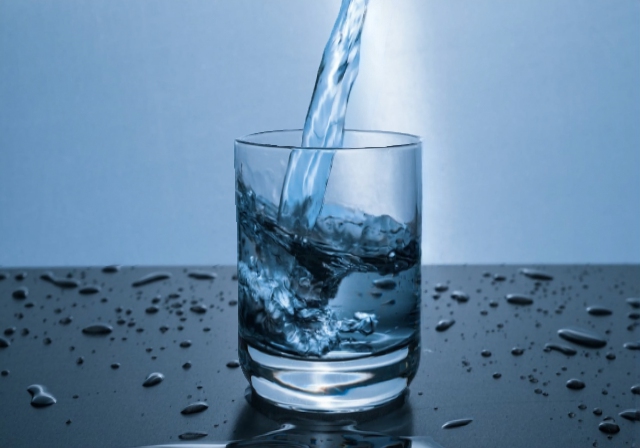}};
            \spy on \rethree in node [left] at \rebigtwo;
    	\end{tikzpicture}
      \caption*{2nd Novel View}
    \end{subfigure}
    \caption{Single-layer depth representation}
    \label{fig:supp-limitation-single_layer}
    \end{subfigure}
    \vspace{-1.5em}
    \caption{\textbf{Failure cases.} (a) Since the depth fixer only fixes the depth in the soft boundaries (represented by the regions with gate $G<1$), it is difficult to correct depth errors beyond soft boundaries in the prediction of the base depth model. (b) Due to the limitation of single-layer depth representation, the synthesized novel view might fail to correct the geometric errors caused by forward warping.}
 \label{fig:supp-limitation}
\end{figure*}

%% file: figs/supp/fig-ablation.tex
\def\imgWidth{0.32\linewidth} %
\def\depthWidth{0.19\linewidth} %
\def\pointWidth{0.24\linewidth} %
\def\scc{(-1.9,-1.4)}

\def\rebigone{(-0.6, -0.38)} %
\def\rebigtwo{(1.6, -0.38)} %

\def\rebigthree{(-0.6, 0.6)} %
\def\rebigfour{(-0.6, 0.6)} %

\def\zoomfour{(-0.25,-0.3)} %
\def\zoomfive{(-0.35,-0.3)} %

\def\zoomsix{(-0.45,-0.6)} %
\def\zoomsixori{(-0.55,-0.6)} %
\def\zoomseven{(0.95,0.05)} %
\def\zoomsevenori{(0.85,0.05)} %

\def\zoomnine{(0.15,0.05)} %
\def\zoomten{(0,0.05)} %

\def\ssizz{1cm} %
\def\ssmag{3}

\begin{figure*}[t]
\centering
\tikzstyle{img} = [rectangle, minimum width=\imgWidth]
    \centering
        \begin{subfigure}{\linewidth}
    \begin{subfigure}{\linewidth}
    \centering
    \begin{subfigure}{\depthWidth}
        \begin{tikzpicture}[spy using outlines={green,magnification=\ssmag,size=\ssizz},inner sep=0]
            \node [align=center, img] {\includegraphics[width=\textwidth]{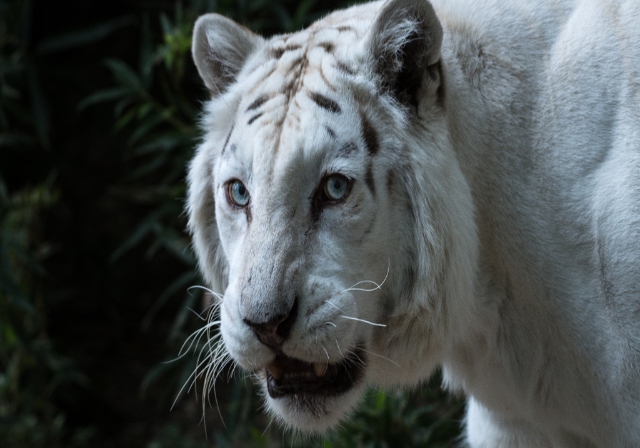}};
            \spy on \zoomsixori in node [left] at \rebigthree;
    	\end{tikzpicture}
    \end{subfigure}
    \begin{subfigure}{\depthWidth}
		\begin{tikzpicture}[spy using outlines={green,magnification=\ssmag,size=\ssizz},inner sep=0]
            \node [align=center, img] {\includegraphics[width=\textwidth]{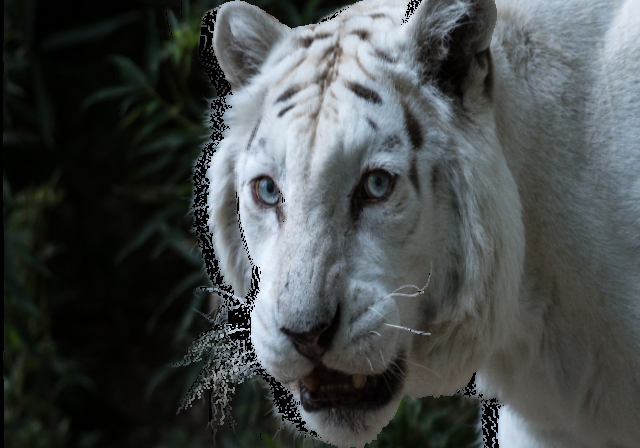}};
            \spy on \zoomsix in node [left] at \rebigthree;
    	\end{tikzpicture}
    \end{subfigure}
    \begin{subfigure}{\depthWidth}
        \begin{tikzpicture}[spy using outlines={green,magnification=\ssmag,size=\ssizz},inner sep=0]
            \node [align=center, img] {\includegraphics[width=\textwidth]{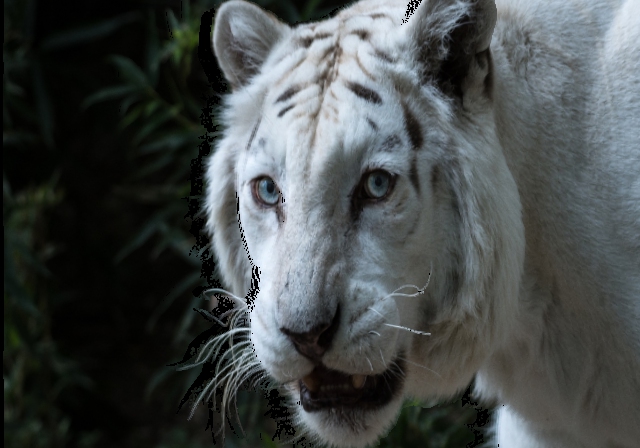}};
            \spy on \zoomsix in node [left] at \rebigthree;
    	\end{tikzpicture}
      \end{subfigure}
    \begin{subfigure}{\depthWidth}
        \begin{tikzpicture}[spy using outlines={green,magnification=\ssmag,size=\ssizz},inner sep=0]
            \node [align=center, img] {\includegraphics[width=\textwidth]{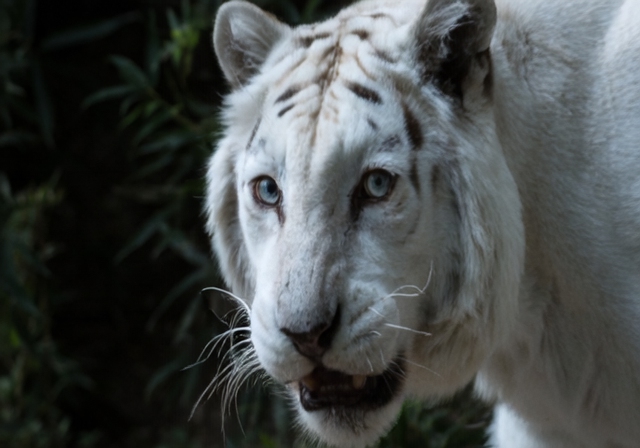}};
            \spy on \zoomsix in node [left] at \rebigthree;
    	\end{tikzpicture}
      \end{subfigure}
    \begin{subfigure}{\depthWidth}
        \begin{tikzpicture}[spy using outlines={green,magnification=\ssmag,size=\ssizz},inner sep=0]
            \node [align=center, img] {\includegraphics[width=\textwidth]{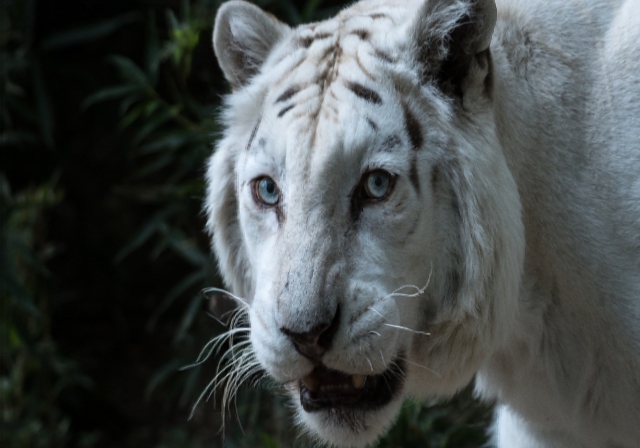}};
            \spy on \zoomsix in node [left] at \rebigthree;
    	\end{tikzpicture}
      \end{subfigure}
    \end{subfigure}
    \begin{subfigure}{\linewidth}
    \centering
    \begin{subfigure}{\depthWidth}
        \begin{tikzpicture}[spy using outlines={green,magnification=\ssmag,size=\ssizz},inner sep=0]
            \node [align=center, img] {\includegraphics[width=\textwidth]{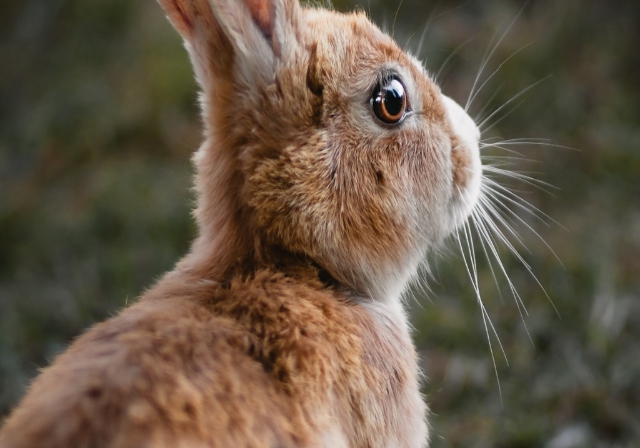}};
            \spy on \zoomsevenori in node [left] at \rebigfour;
    	\end{tikzpicture}
        \caption*{\centering Input \\ Image}
    \end{subfigure}
    \begin{subfigure}{\depthWidth}
		\begin{tikzpicture}[spy using outlines={green,magnification=\ssmag,size=\ssizz},inner sep=0]
            \node [align=center, img] {\includegraphics[width=\textwidth]{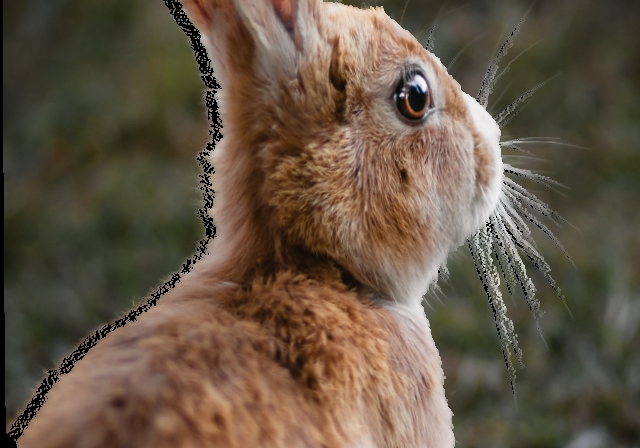}};
            \spy on \zoomseven in node [left] at \rebigfour;
    	\end{tikzpicture}
        \caption*{\centering Original Warped Image \\ (using Depth Anyting V2)}
    \end{subfigure}
    \begin{subfigure}{\depthWidth}
        \begin{tikzpicture}[spy using outlines={green,magnification=\ssmag,size=\ssizz},inner sep=0]
            \node [align=center, img] {\includegraphics[width=\textwidth]{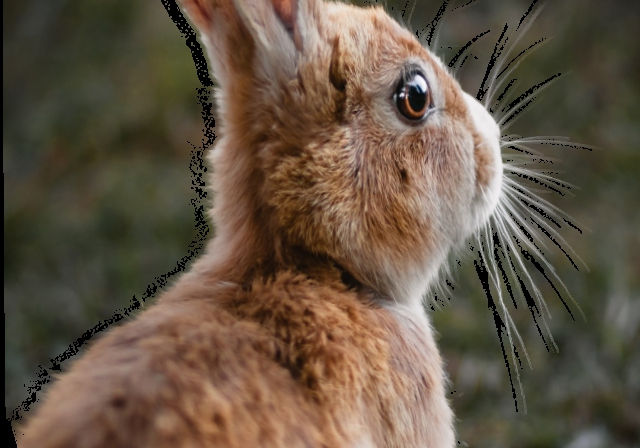}};
            \spy on \zoomseven in node [left] at \rebigfour;
    	\end{tikzpicture}
        \caption*{\centering Fixed Warped Image \\ (applying Depth Fixer)}
      \end{subfigure}
    \begin{subfigure}{\depthWidth}
        \begin{tikzpicture}[spy using outlines={green,magnification=\ssmag,size=\ssizz},inner sep=0]
            \node [align=center, img] {\includegraphics[width=\textwidth]{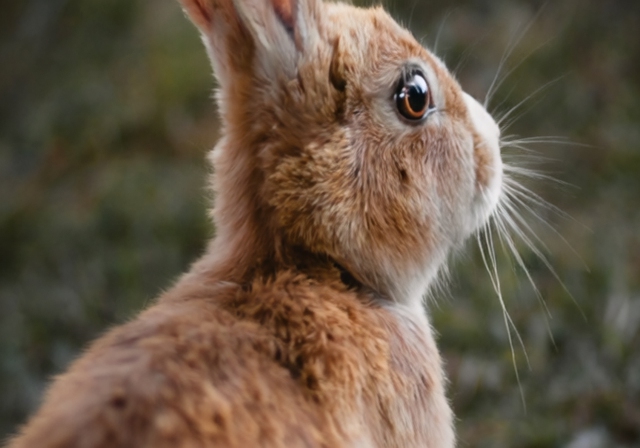}};
            \spy on \zoomseven in node [left] at \rebigfour;
    	\end{tikzpicture}
        \caption*{\centering Inpainted Image \\ (applying Scene Painter)}
      \end{subfigure}
    \begin{subfigure}{\depthWidth}
        \begin{tikzpicture}[spy using outlines={green,magnification=\ssmag,size=\ssizz},inner sep=0]
            \node [align=center, img] {\includegraphics[width=\textwidth]{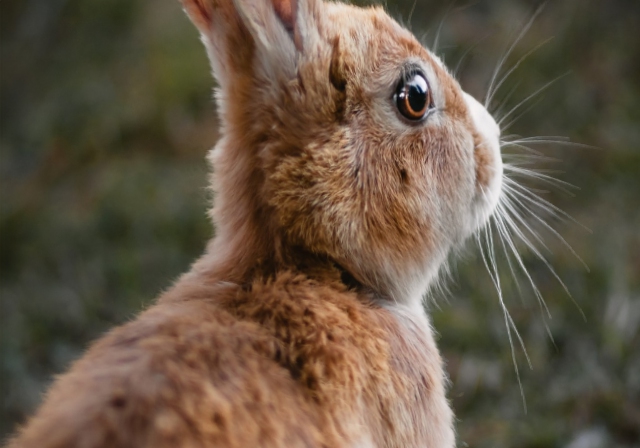}};
            \spy on \zoomseven in node [left] at \rebigfour;
    	\end{tikzpicture}
        \caption*{\centering Fused Image \\ (applying Color Fuser)}
      \end{subfigure}
    \end{subfigure}
    \caption{Visual results on the AIM-500 dataset}
    \end{subfigure}
    \begin{subfigure}{\linewidth}
    \begin{subfigure}{\linewidth}
    \centering
    \begin{subfigure}{\depthWidth}
        \begin{tikzpicture}[spy using outlines={green,magnification=\ssmag,size=\ssizz},inner sep=0]
            \node [align=center, img] {\includegraphics[width=\textwidth]{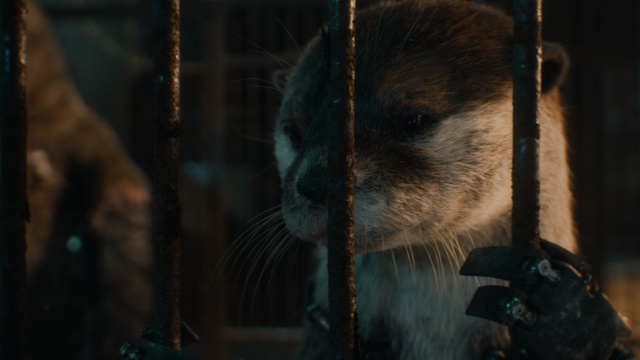}};
            \spy on \zoomfour in node [left] at \rebigone;
    	\end{tikzpicture}
    \end{subfigure}
    \begin{subfigure}{\depthWidth}
		\begin{tikzpicture}[spy using outlines={green,magnification=\ssmag,size=\ssizz},inner sep=0]
            \node [align=center, img] {\includegraphics[width=\textwidth]{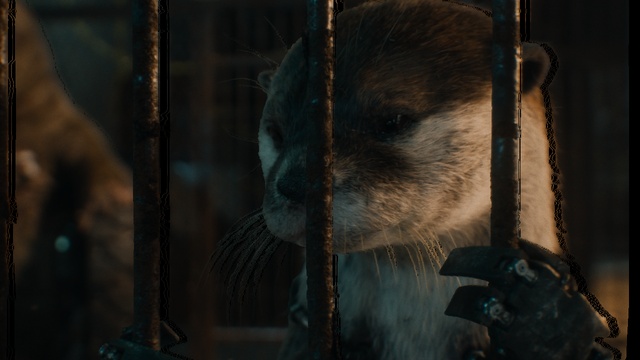}};
            \spy on \zoomfive in node [left] at \rebigone;
    	\end{tikzpicture}
    \end{subfigure}
    \begin{subfigure}{\depthWidth}
        \begin{tikzpicture}[spy using outlines={green,magnification=\ssmag,size=\ssizz},inner sep=0]
            \node [align=center, img] {\includegraphics[width=\textwidth]{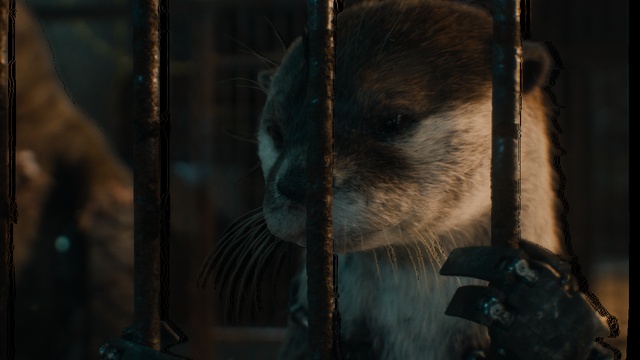}};
            \spy on \zoomfive in node [left] at \rebigone;
    	\end{tikzpicture}
      \end{subfigure}
    \begin{subfigure}{\depthWidth}
        \begin{tikzpicture}[spy using outlines={green,magnification=\ssmag,size=\ssizz},inner sep=0]
            \node [align=center, img] {\includegraphics[width=\textwidth]{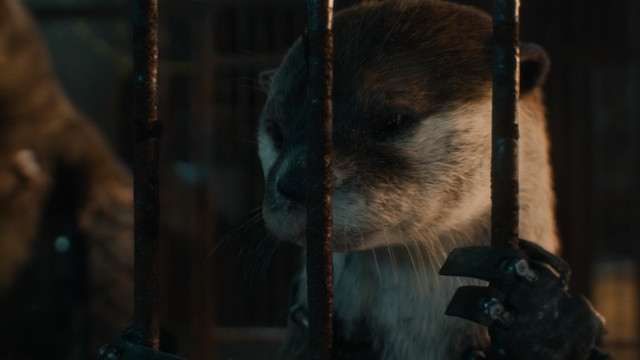}};
            \spy on \zoomfive in node [left] at \rebigone;
    	\end{tikzpicture}
      \end{subfigure}
    \begin{subfigure}{\depthWidth}
        \begin{tikzpicture}[spy using outlines={green,magnification=\ssmag,size=\ssizz},inner sep=0]
            \node [align=center, img] {\includegraphics[width=\textwidth]{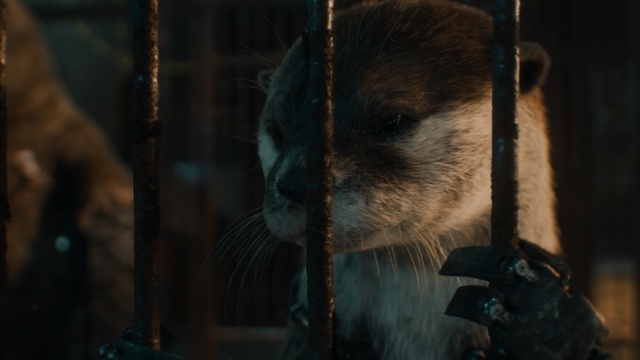}};
            \spy on \zoomfive in node [left] at \rebigone;
    	\end{tikzpicture}
      \end{subfigure}
    \end{subfigure}
    \begin{subfigure}{\linewidth}
    \centering
    \begin{subfigure}{\depthWidth}
        \begin{tikzpicture}[spy using outlines={green,magnification=\ssmag,size=\ssizz},inner sep=0]
            \node [align=center, img] {\includegraphics[width=\textwidth]{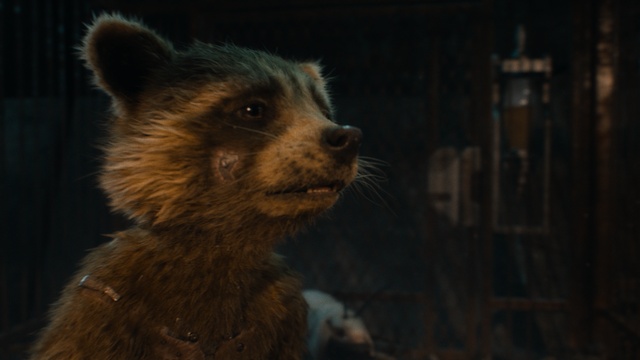}};
            \spy on \zoomnine in node [left] at \rebigtwo;
    	\end{tikzpicture}
        \caption*{\centering Input \\ Image}
    \end{subfigure}
    \begin{subfigure}{\depthWidth}
		\begin{tikzpicture}[spy using outlines={green,magnification=\ssmag,size=\ssizz},inner sep=0]
            \node [align=center, img] {\includegraphics[width=\textwidth]{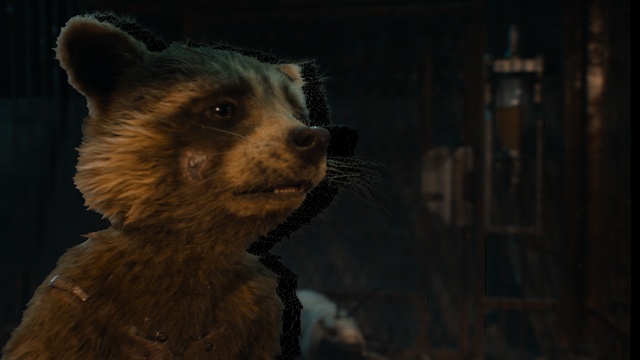}};
            \spy on \zoomten in node [left] at \rebigtwo;
    	\end{tikzpicture}
        \caption*{\centering Original Warped Image \\ (using Depth Anyting V2)}
    \end{subfigure}
    \begin{subfigure}{\depthWidth}
        \begin{tikzpicture}[spy using outlines={green,magnification=\ssmag,size=\ssizz},inner sep=0]
            \node [align=center, img] {\includegraphics[width=\textwidth]{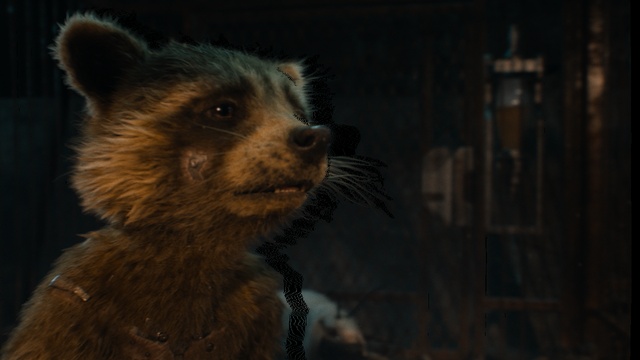}};
            \spy on \zoomten in node [left] at \rebigtwo;
    	\end{tikzpicture}
        \caption*{\centering Fixed Warped Image \\ (applying Depth Fixer)}
      \end{subfigure}
    \begin{subfigure}{\depthWidth}
        \begin{tikzpicture}[spy using outlines={green,magnification=\ssmag,size=\ssizz},inner sep=0]
            \node [align=center, img] {\includegraphics[width=\textwidth]{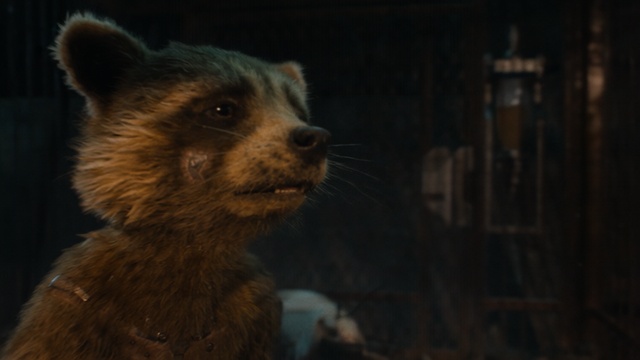}};
            \spy on \zoomten in node [left] at \rebigtwo;
    	\end{tikzpicture}
        \caption*{\centering Inpainted Image \\ (applying Scene Painter)}
      \end{subfigure}
    \begin{subfigure}{\depthWidth}
        \begin{tikzpicture}[spy using outlines={green,magnification=\ssmag,size=\ssizz},inner sep=0]
            \node [align=center, img] {\includegraphics[width=\textwidth]{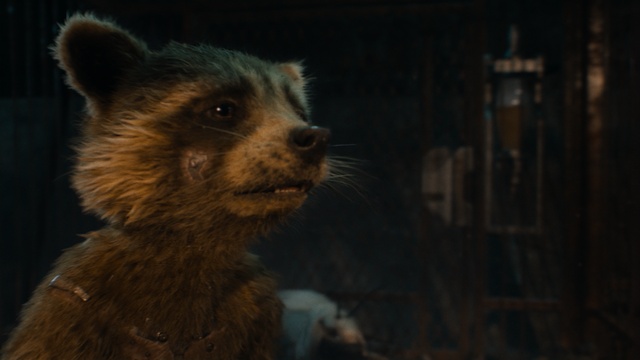}};
            \spy on \zoomten in node [left] at \rebigtwo;
    	\end{tikzpicture}
        \caption*{\centering Fused Image \\ (applying Color Fuser)}
      \end{subfigure}
    \end{subfigure}
    \caption{Visual results on the Marvel-10K dataset}
    \end{subfigure}
    \vspace{-1.5em}
    \caption{\textbf{Visual results of ablation study} on the AIM-500 and Marvel-10K datasets. Due to depth estimation errors, the original warped images often contain broken or distorted structures in thin hairs. Our depth fixer improves the warping performance by fixing the soft boundary regions in the depth. The scene painter is employed to fill disoccluded regions in the warped images, but the inpainted results often suffer from hallucinated details that are inconsistent with the input image (\eg, see hairs in the \textcolor{green}{green} box, particularly in the Marvel-10K examples). By adaptively combining the warped and inpainted images, the color fuser produces high-quality results with consistent texture and geometry. }
    \label{fig:supp-ablation-visual}
\end{figure*}

%% file: figs/supp/fig-depth_est_visuals_aim.tex
\def\imgWidth{0.32\linewidth} %
\def\depthWidth{0.19\linewidth} %
\def\pointWidth{0.24\linewidth} %
\def\scc{(-1.9,-1.4)}

\def\rebigone{(-0.5, -0.55)} %
\def\rebigtwo{(1.6, -0.55)} %

\def\zoomone{(0.05,0.15)} %
\def\zoomtwo{(-0.83,0.15)} %
\def\zoomthree{(0,-0.9)} %
\def\zoomfour{(0.6,0.45)} %
\def\zoomfive{(-0.2,-0.9)} %

\def\zoomsix{(-0.7,0.4)} %
\def\zoomseven{(-0.65,0.98)} %
\def\zoomeight{(0,0.5)} %
\def\zoomnine{(0.2,0.6)} %

\def\ssizz{1.1cm} %
\def\ssmag{3}

\begin{figure*}[t]
\centering
\tikzstyle{img} = [rectangle, minimum width=\imgWidth]
    \centering
    \begin{subfigure}{\linewidth}
    \centering
    \begin{subfigure}{\depthWidth}
        \begin{tikzpicture}[spy using outlines={green,magnification=\ssmag,size=\ssizz},inner sep=0]
            \node [align=center, img] {\includegraphics[width=\textwidth]{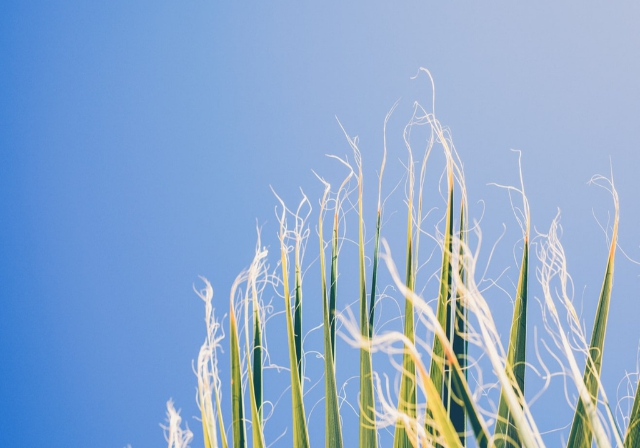}};
            \spy on \zoomone in node [left] at \rebigone;
    	\end{tikzpicture}
    \end{subfigure}
    \begin{subfigure}{\depthWidth}
		\begin{tikzpicture}[spy using outlines={green,magnification=\ssmag,size=\ssizz},inner sep=0]
            \node [align=center, img] {\includegraphics[width=\textwidth]{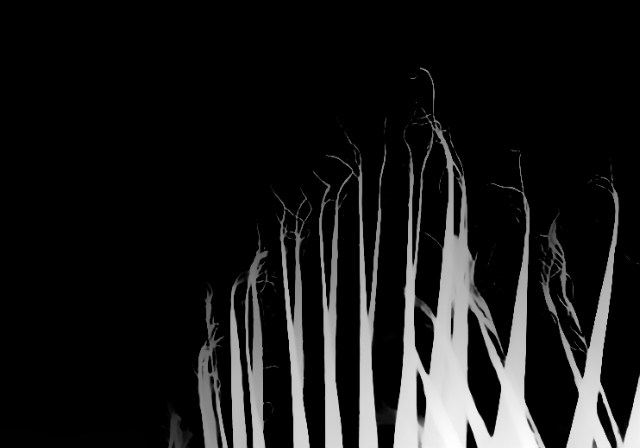}};
            \spy on \zoomone in node [left] at \rebigone;
    	\end{tikzpicture}
    \end{subfigure}
    \begin{subfigure}{\depthWidth}
        \begin{tikzpicture}[spy using outlines={green,magnification=\ssmag,size=\ssizz},inner sep=0]
            \node [align=center, img] {\includegraphics[width=\textwidth]{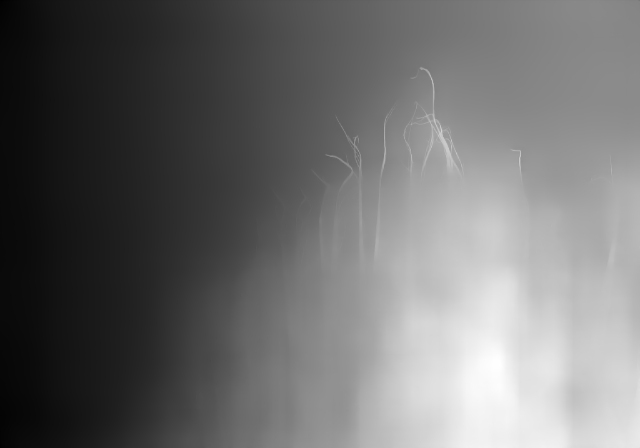}};
            \spy on \zoomone in node [left] at \rebigone;
    	\end{tikzpicture}
      \end{subfigure}
    \begin{subfigure}{\depthWidth}
        \begin{tikzpicture}[spy using outlines={green,magnification=\ssmag,size=\ssizz},inner sep=0]
            \node [align=center, img] {\includegraphics[width=\textwidth]{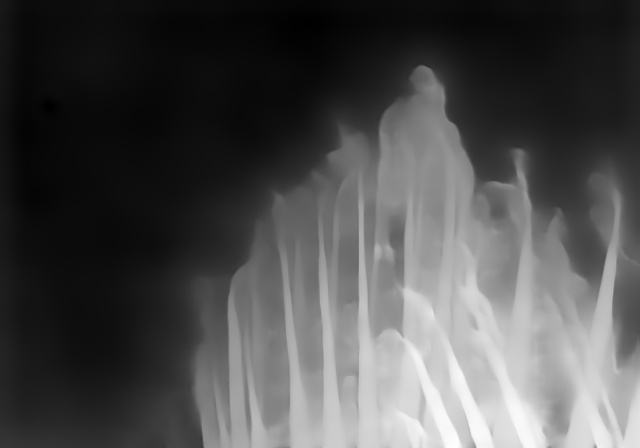}};
            \spy on \zoomone in node [left] at \rebigone;
    	\end{tikzpicture}
      \end{subfigure}
    \begin{subfigure}{\depthWidth}
        \begin{tikzpicture}[spy using outlines={green,magnification=\ssmag,size=\ssizz},inner sep=0]
            \node [align=center, img] {\includegraphics[width=\textwidth]{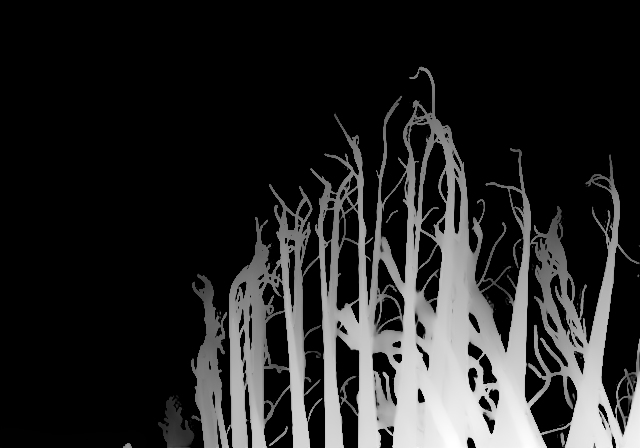}};
            \spy on \zoomone in node [left] at \rebigone;
    	\end{tikzpicture}
      \end{subfigure}
    \end{subfigure}
    \begin{subfigure}{\linewidth}
    \centering
    \begin{subfigure}{\depthWidth}
        \begin{tikzpicture}[spy using outlines={green,magnification=\ssmag,size=\ssizz},inner sep=0]
            \node [align=center, img] {\includegraphics[width=\textwidth]{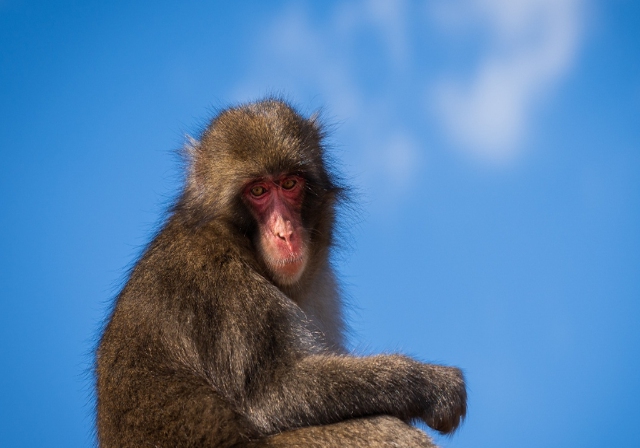}};
            \spy on \zoomtwo in node [left] at \rebigtwo;
    	\end{tikzpicture}
    \end{subfigure}
    \begin{subfigure}{\depthWidth}
		\begin{tikzpicture}[spy using outlines={green,magnification=\ssmag,size=\ssizz},inner sep=0]
            \node [align=center, img] {\includegraphics[width=\textwidth]{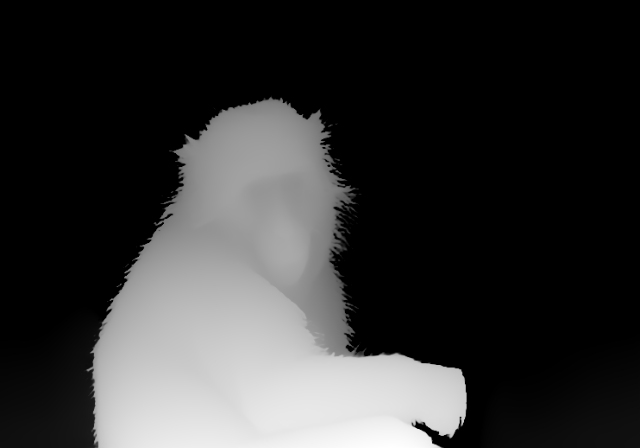}};
            \spy on \zoomtwo in node [left] at \rebigtwo;
    	\end{tikzpicture}
    \end{subfigure}
    \begin{subfigure}{\depthWidth}
        \begin{tikzpicture}[spy using outlines={green,magnification=\ssmag,size=\ssizz},inner sep=0]
            \node [align=center, img] {\includegraphics[width=\textwidth]{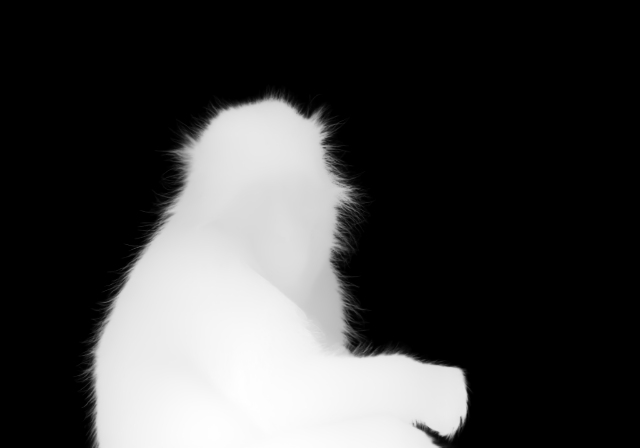}};
            \spy on \zoomtwo in node [left] at \rebigtwo;
    	\end{tikzpicture}
      \end{subfigure}
    \begin{subfigure}{\depthWidth}
        \begin{tikzpicture}[spy using outlines={green,magnification=\ssmag,size=\ssizz},inner sep=0]
            \node [align=center, img] {\includegraphics[width=\textwidth]{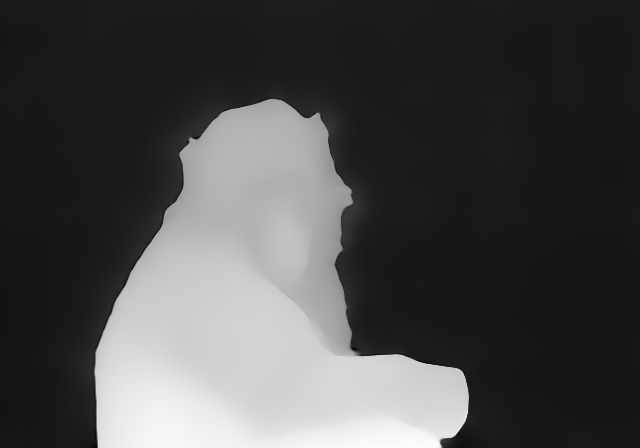}};
            \spy on \zoomtwo in node [left] at \rebigtwo;
    	\end{tikzpicture}
      \end{subfigure}
    \begin{subfigure}{\depthWidth}
        \begin{tikzpicture}[spy using outlines={green,magnification=\ssmag,size=\ssizz},inner sep=0]
            \node [align=center, img] {\includegraphics[width=\textwidth]{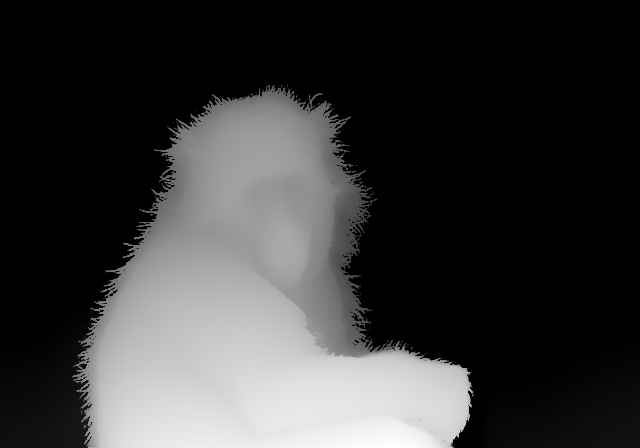}};
            \spy on \zoomtwo in node [left] at \rebigtwo;
    	\end{tikzpicture}
      \end{subfigure}
    \end{subfigure}
    \begin{subfigure}{\linewidth}
    \centering
    \begin{subfigure}{\depthWidth}
        \begin{tikzpicture}[spy using outlines={green,magnification=\ssmag,size=\ssizz},inner sep=0]
            \node [align=center, img] {\includegraphics[width=\textwidth]{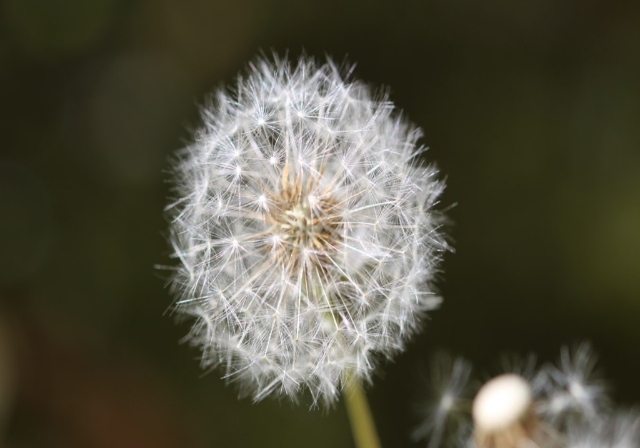}};
            \spy on \zoomthree in node [left] at \rebigone;
    	\end{tikzpicture}
    \end{subfigure}
    \begin{subfigure}{\depthWidth}
		\begin{tikzpicture}[spy using outlines={green,magnification=\ssmag,size=\ssizz},inner sep=0]
            \node [align=center, img] {\includegraphics[width=\textwidth]{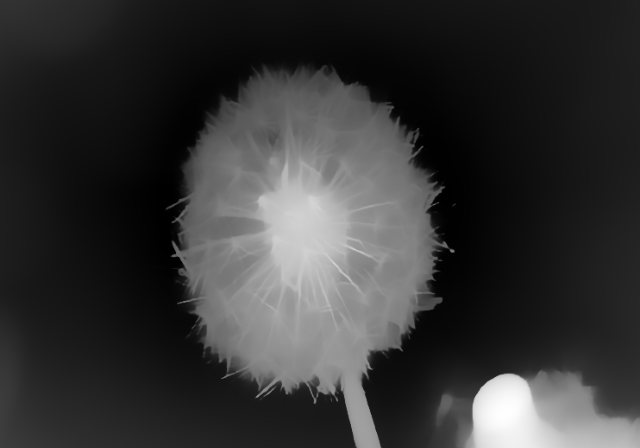}};
            \spy on \zoomthree in node [left] at \rebigone;
    	\end{tikzpicture}
    \end{subfigure}
    \begin{subfigure}{\depthWidth}
        \begin{tikzpicture}[spy using outlines={green,magnification=\ssmag,size=\ssizz},inner sep=0]
            \node [align=center, img] {\includegraphics[width=\textwidth]{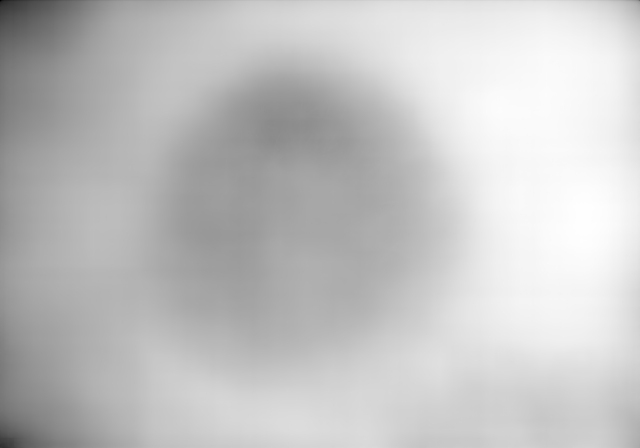}};
            \spy on \zoomthree in node [left] at \rebigone;
    	\end{tikzpicture}
      \end{subfigure}
    \begin{subfigure}{\depthWidth}
        \begin{tikzpicture}[spy using outlines={green,magnification=\ssmag,size=\ssizz},inner sep=0]
            \node [align=center, img] {\includegraphics[width=\textwidth]{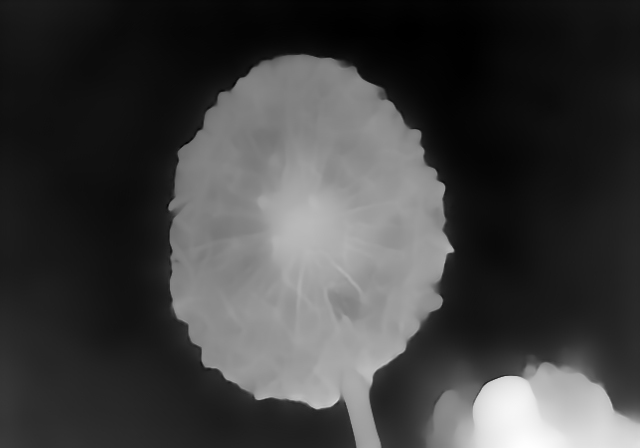}};
            \spy on \zoomthree in node [left] at \rebigone;
    	\end{tikzpicture}
      \end{subfigure}
    \begin{subfigure}{\depthWidth}
        \begin{tikzpicture}[spy using outlines={green,magnification=\ssmag,size=\ssizz},inner sep=0]
            \node [align=center, img] {\includegraphics[width=\textwidth]{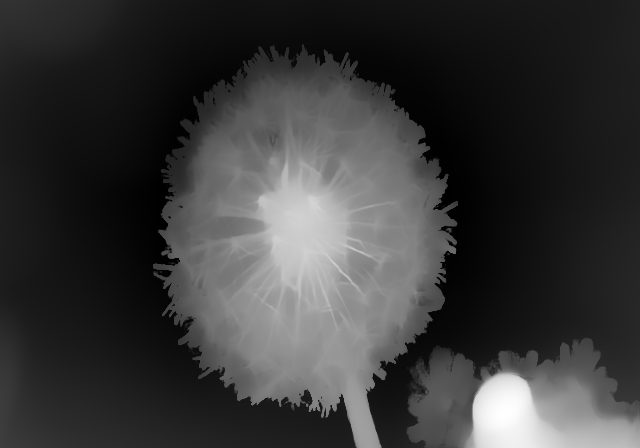}};
            \spy on \zoomthree in node [left] at \rebigone;
    	\end{tikzpicture}
      \end{subfigure}
    \end{subfigure}
    \begin{subfigure}{\linewidth}
    \centering
    \begin{subfigure}{\depthWidth}
        \begin{tikzpicture}[spy using outlines={green,magnification=\ssmag,size=\ssizz},inner sep=0]
            \node [align=center, img] {\includegraphics[width=\textwidth]{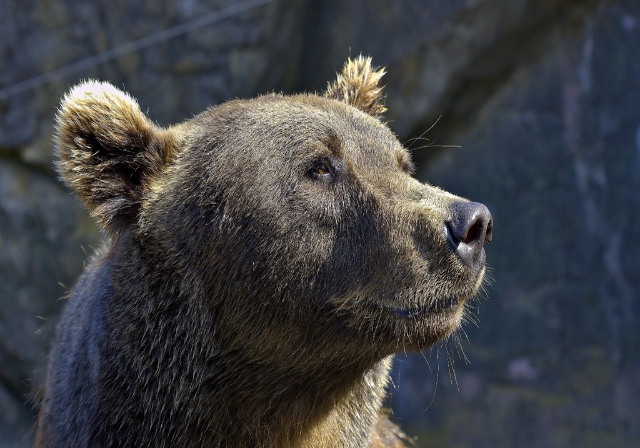}};
            \spy on \zoomfour in node [left] at \rebigone;
    	\end{tikzpicture}
    \end{subfigure}
    \begin{subfigure}{\depthWidth}
		\begin{tikzpicture}[spy using outlines={green,magnification=\ssmag,size=\ssizz},inner sep=0]
            \node [align=center, img] {\includegraphics[width=\textwidth]{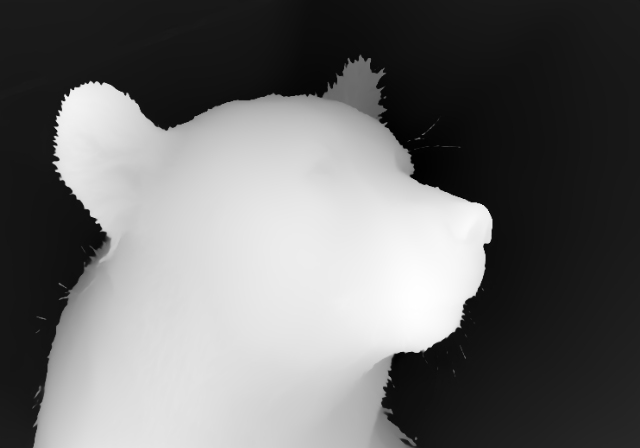}};
            \spy on \zoomfour in node [left] at \rebigone;
    	\end{tikzpicture}
    \end{subfigure}
    \begin{subfigure}{\depthWidth}
        \begin{tikzpicture}[spy using outlines={green,magnification=\ssmag,size=\ssizz},inner sep=0]
            \node [align=center, img] {\includegraphics[width=\textwidth]{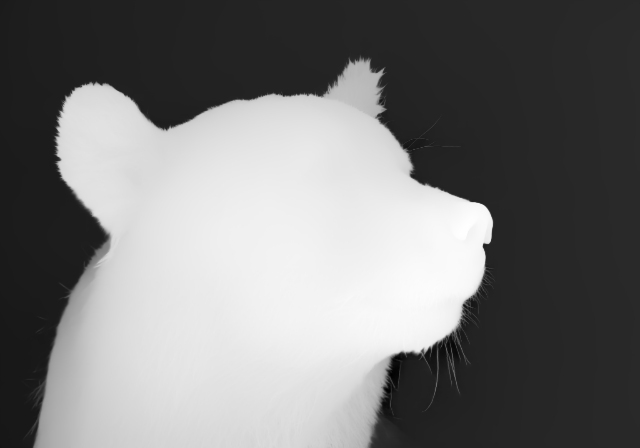}};
            \spy on \zoomfour in node [left] at \rebigone;
    	\end{tikzpicture}
      \end{subfigure}
    \begin{subfigure}{\depthWidth}
        \begin{tikzpicture}[spy using outlines={green,magnification=\ssmag,size=\ssizz},inner sep=0]
            \node [align=center, img] {\includegraphics[width=\textwidth]{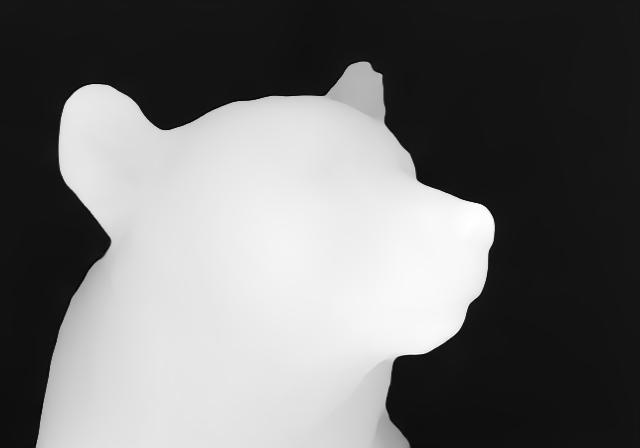}};
            \spy on \zoomfour in node [left] at \rebigone;
    	\end{tikzpicture}
      \end{subfigure}
    \begin{subfigure}{\depthWidth}
        \begin{tikzpicture}[spy using outlines={green,magnification=\ssmag,size=\ssizz},inner sep=0]
            \node [align=center, img] {\includegraphics[width=\textwidth]{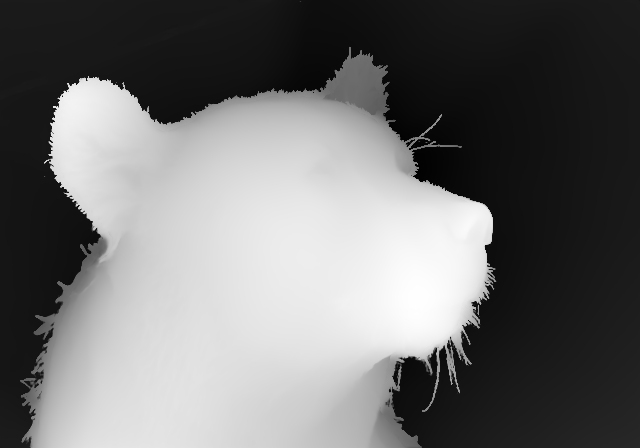}};
            \spy on \zoomfour in node [left] at \rebigone;
    	\end{tikzpicture}
      \end{subfigure}
    \end{subfigure}
    \begin{subfigure}{\linewidth}
    \centering
    \begin{subfigure}{\depthWidth}
        \begin{tikzpicture}[spy using outlines={green,magnification=\ssmag,size=\ssizz},inner sep=0]
            \node [align=center, img] {\includegraphics[width=\textwidth]{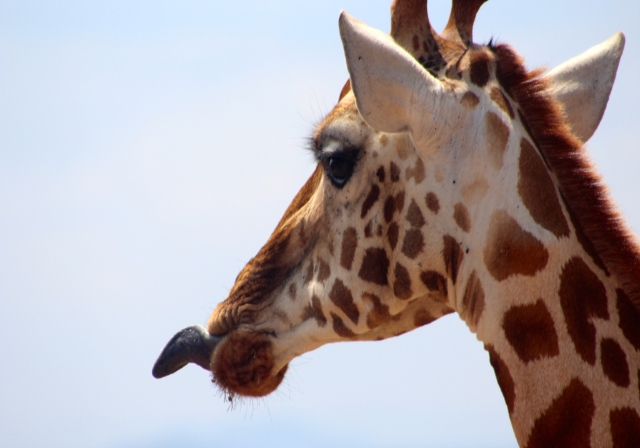}};
            \spy on \zoomfive in node [left] at \rebigtwo;
    	\end{tikzpicture}
    \end{subfigure}
    \begin{subfigure}{\depthWidth}
		\begin{tikzpicture}[spy using outlines={green,magnification=\ssmag,size=\ssizz},inner sep=0]
            \node [align=center, img] {\includegraphics[width=\textwidth]{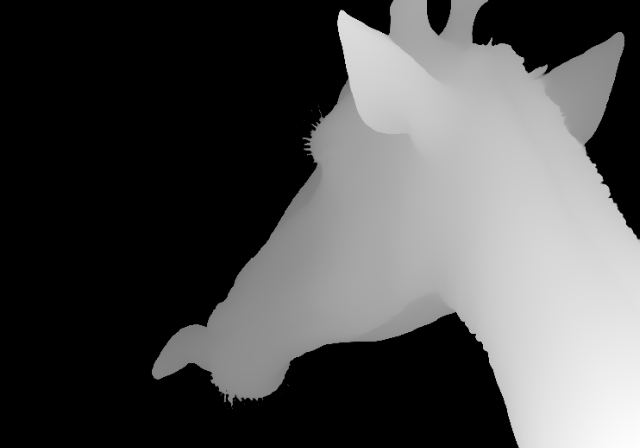}};
            \spy on \zoomfive in node [left] at \rebigtwo;
    	\end{tikzpicture}
    \end{subfigure}
    \begin{subfigure}{\depthWidth}
        \begin{tikzpicture}[spy using outlines={green,magnification=\ssmag,size=\ssizz},inner sep=0]
            \node [align=center, img] {\includegraphics[width=\textwidth]{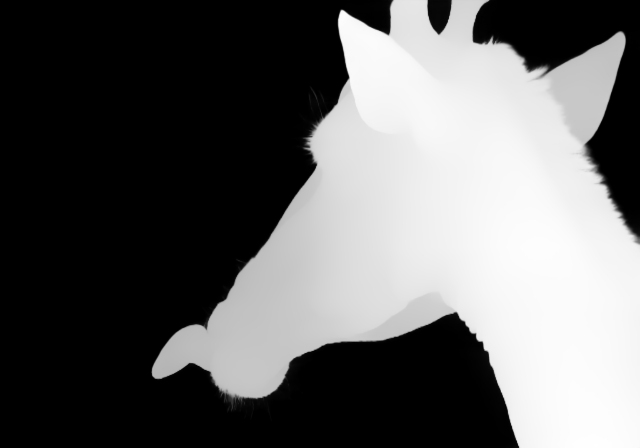}};
            \spy on \zoomfive in node [left] at \rebigtwo;
    	\end{tikzpicture}
      \end{subfigure}
    \begin{subfigure}{\depthWidth}
        \begin{tikzpicture}[spy using outlines={green,magnification=\ssmag,size=\ssizz},inner sep=0]
            \node [align=center, img] {\includegraphics[width=\textwidth]{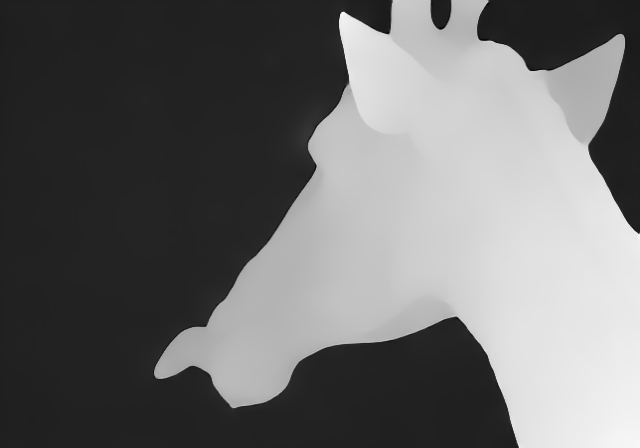}};
            \spy on \zoomfive in node [left] at \rebigtwo;
    	\end{tikzpicture}
      \end{subfigure}
    \begin{subfigure}{\depthWidth}
        \begin{tikzpicture}[spy using outlines={green,magnification=\ssmag,size=\ssizz},inner sep=0]
            \node [align=center, img] {\includegraphics[width=\textwidth]{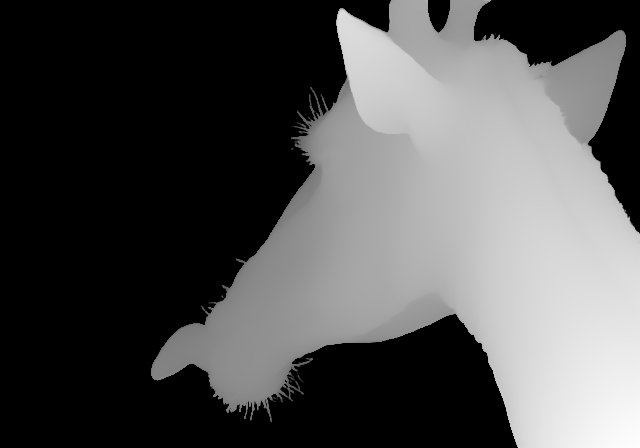}};
            \spy on \zoomfive in node [left] at \rebigtwo;
    	\end{tikzpicture}
      \end{subfigure}
    \end{subfigure}
    \begin{subfigure}{\linewidth}
    \centering
    \begin{subfigure}{\depthWidth}
        \begin{tikzpicture}[spy using outlines={green,magnification=\ssmag,size=\ssizz},inner sep=0]
            \node [align=center, img] {\includegraphics[width=\textwidth]{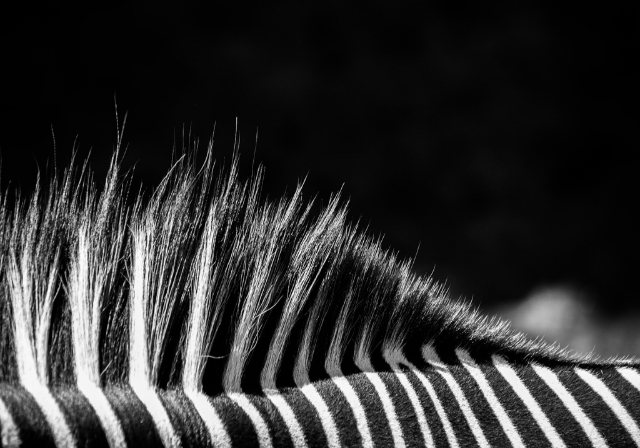}};
            \spy on \zoomsix in node [left] at \rebigtwo;
    	\end{tikzpicture}
    \end{subfigure}
    \begin{subfigure}{\depthWidth}
		\begin{tikzpicture}[spy using outlines={green,magnification=\ssmag,size=\ssizz},inner sep=0]
            \node [align=center, img] {\includegraphics[width=\textwidth]{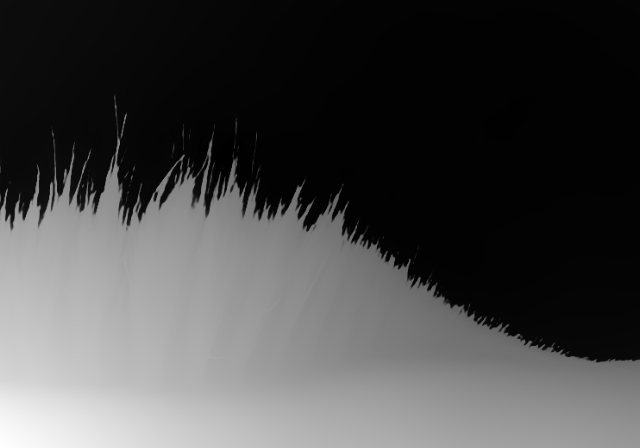}};
            \spy on \zoomsix in node [left] at \rebigtwo;
    	\end{tikzpicture}
    \end{subfigure}
    \begin{subfigure}{\depthWidth}
        \begin{tikzpicture}[spy using outlines={green,magnification=\ssmag,size=\ssizz},inner sep=0]
            \node [align=center, img] {\includegraphics[width=\textwidth]{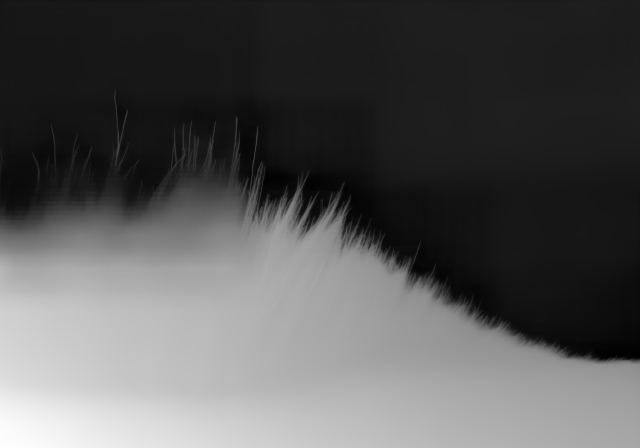}};
            \spy on \zoomsix in node [left] at \rebigtwo;
    	\end{tikzpicture}
      \end{subfigure}
    \begin{subfigure}{\depthWidth}
        \begin{tikzpicture}[spy using outlines={green,magnification=\ssmag,size=\ssizz},inner sep=0]
            \node [align=center, img] {\includegraphics[width=\textwidth]{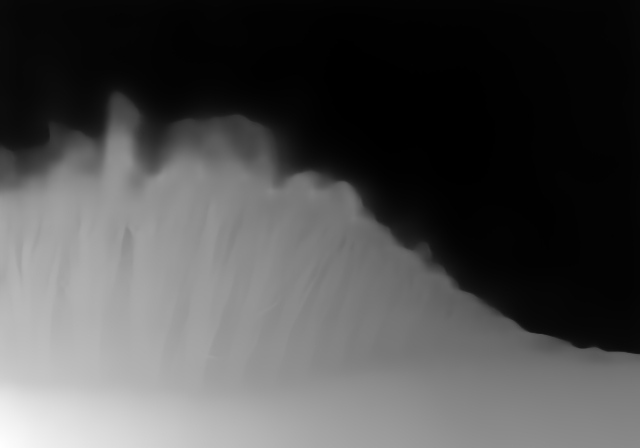}};
            \spy on \zoomsix in node [left] at \rebigtwo;
    	\end{tikzpicture}
      \end{subfigure}
    \begin{subfigure}{\depthWidth}
        \begin{tikzpicture}[spy using outlines={green,magnification=\ssmag,size=\ssizz},inner sep=0]
            \node [align=center, img] {\includegraphics[width=\textwidth]{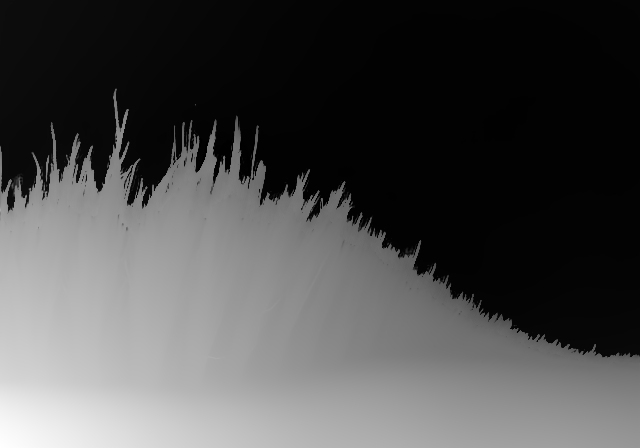}};
            \spy on \zoomsix in node [left] at \rebigtwo;
    	\end{tikzpicture}
      \end{subfigure}
    \end{subfigure}
    \begin{subfigure}{\linewidth}
    \centering
    \begin{subfigure}{\depthWidth}
        \begin{tikzpicture}[spy using outlines={green,magnification=\ssmag,size=\ssizz},inner sep=0]
            \node [align=center, img] {\includegraphics[width=\textwidth]{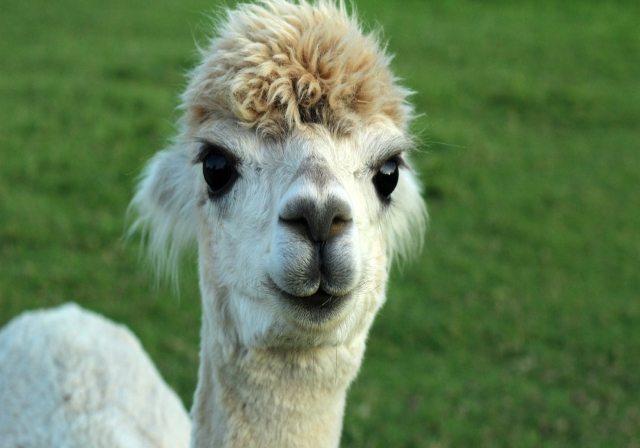}};
            \spy on \zoomseven in node [left] at \rebigtwo;
    	\end{tikzpicture}
    \end{subfigure}
    \begin{subfigure}{\depthWidth}
		\begin{tikzpicture}[spy using outlines={green,magnification=\ssmag,size=\ssizz},inner sep=0]
            \node [align=center, img] {\includegraphics[width=\textwidth]{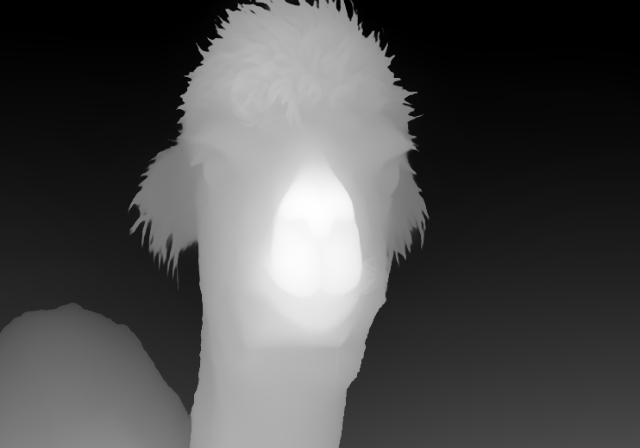}};
            \spy on \zoomseven in node [left] at \rebigtwo;
    	\end{tikzpicture}
    \end{subfigure}
    \begin{subfigure}{\depthWidth}
        \begin{tikzpicture}[spy using outlines={green,magnification=\ssmag,size=\ssizz},inner sep=0]
            \node [align=center, img] {\includegraphics[width=\textwidth]{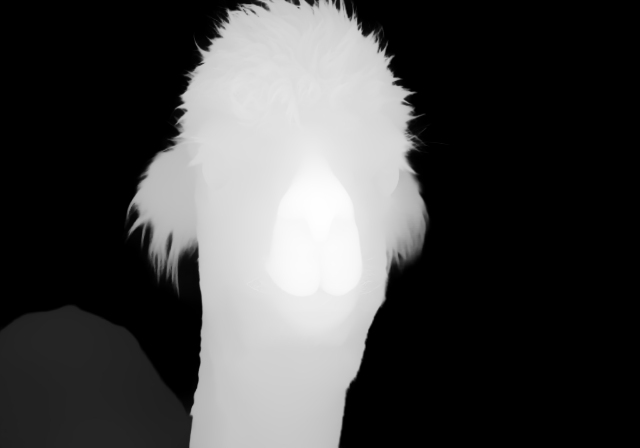}};
            \spy on \zoomseven in node [left] at \rebigtwo;
    	\end{tikzpicture}
      \end{subfigure}
    \begin{subfigure}{\depthWidth}
        \begin{tikzpicture}[spy using outlines={green,magnification=\ssmag,size=\ssizz},inner sep=0]
            \node [align=center, img] {\includegraphics[width=\textwidth]{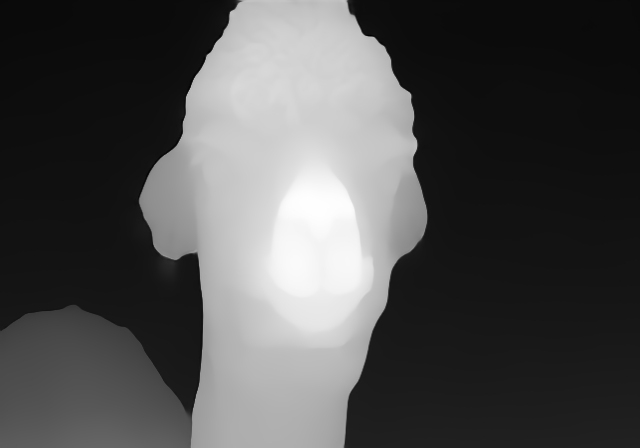}};
            \spy on \zoomseven in node [left] at \rebigtwo;
    	\end{tikzpicture}
      \end{subfigure}
    \begin{subfigure}{\depthWidth}
        \begin{tikzpicture}[spy using outlines={green,magnification=\ssmag,size=\ssizz},inner sep=0]
            \node [align=center, img] {\includegraphics[width=\textwidth]{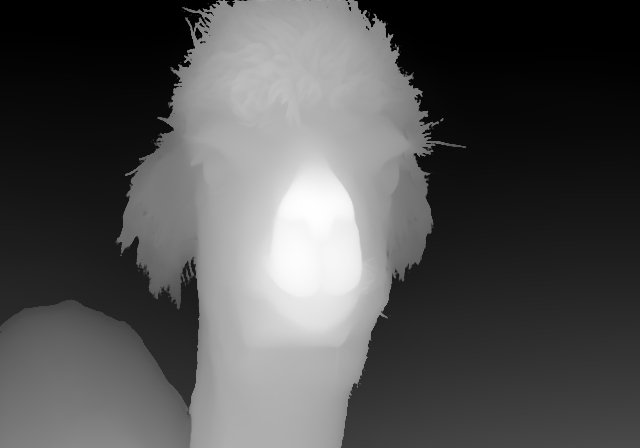}};
            \spy on \zoomseven in node [left] at \rebigtwo;
    	\end{tikzpicture}
      \end{subfigure}
    \end{subfigure}
    \begin{subfigure}{\linewidth}
    \centering
    \begin{subfigure}{\depthWidth}
        \begin{tikzpicture}[spy using outlines={green,magnification=\ssmag,size=\ssizz},inner sep=0]
            \node [align=center, img] {\includegraphics[width=\textwidth]{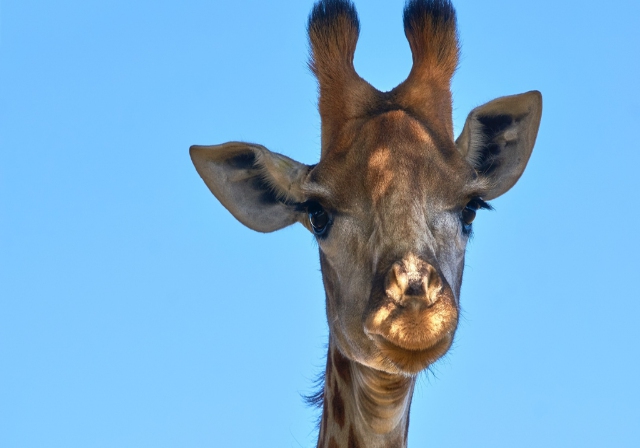}};
            \spy on \zoomeight in node [left] at \rebigone;
    	\end{tikzpicture}
    \end{subfigure}
    \begin{subfigure}{\depthWidth}
		\begin{tikzpicture}[spy using outlines={green,magnification=\ssmag,size=\ssizz},inner sep=0]
            \node [align=center, img] {\includegraphics[width=\textwidth]{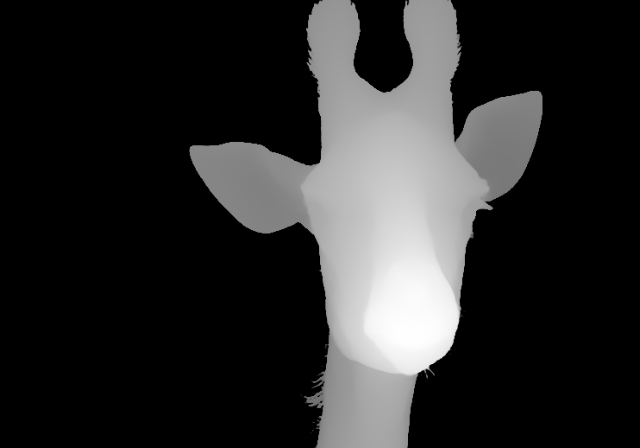}};
            \spy on \zoomeight in node [left] at \rebigone;
    	\end{tikzpicture}
    \end{subfigure}
    \begin{subfigure}{\depthWidth}
        \begin{tikzpicture}[spy using outlines={green,magnification=\ssmag,size=\ssizz},inner sep=0]
            \node [align=center, img] {\includegraphics[width=\textwidth]{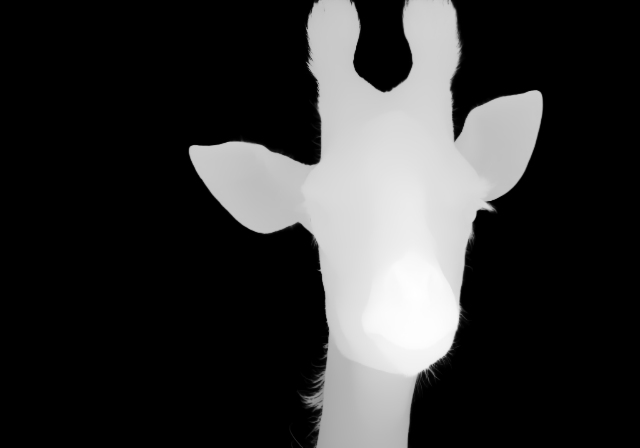}};
            \spy on \zoomeight in node [left] at \rebigone;
    	\end{tikzpicture}
      \end{subfigure}
    \begin{subfigure}{\depthWidth}
        \begin{tikzpicture}[spy using outlines={green,magnification=\ssmag,size=\ssizz},inner sep=0]
            \node [align=center, img] {\includegraphics[width=\textwidth]{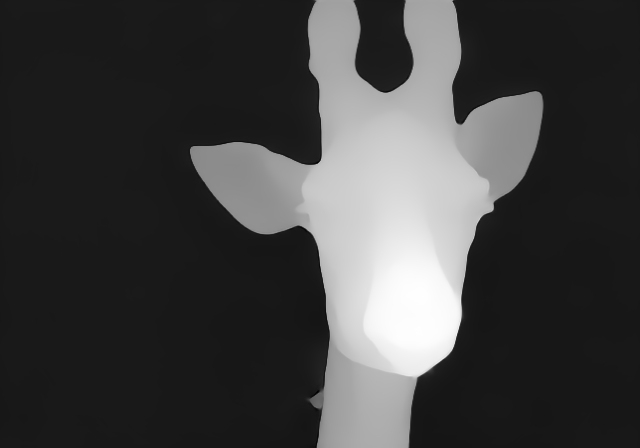}};
            \spy on \zoomeight in node [left] at \rebigone;
    	\end{tikzpicture}
      \end{subfigure}
    \begin{subfigure}{\depthWidth}
        \begin{tikzpicture}[spy using outlines={green,magnification=\ssmag,size=\ssizz},inner sep=0]
            \node [align=center, img] {\includegraphics[width=\textwidth]{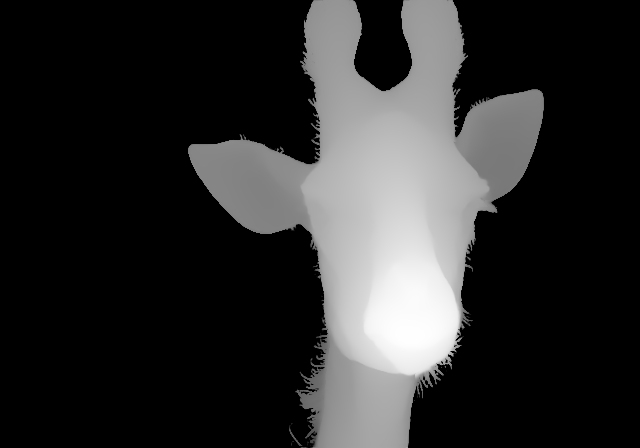}};
            \spy on \zoomeight in node [left] at \rebigone;
    	\end{tikzpicture}
      \end{subfigure}
    \end{subfigure}
    \begin{subfigure}{\linewidth}
    \centering
    \begin{subfigure}{\depthWidth}
        \begin{tikzpicture}[spy using outlines={green,magnification=\ssmag,size=\ssizz},inner sep=0]
            \node [align=center, img] {\includegraphics[width=\textwidth]{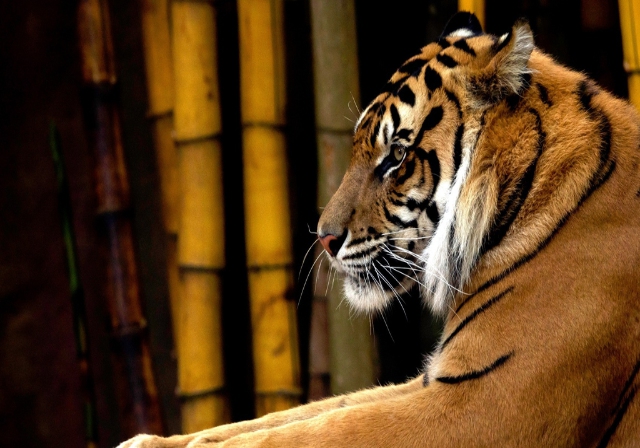}};
            \spy on \zoomnine in node [left] at \rebigone;
    	\end{tikzpicture}
        \caption*{Input Image}
    \end{subfigure}
    \begin{subfigure}{\depthWidth}
		\begin{tikzpicture}[spy using outlines={green,magnification=\ssmag,size=\ssizz},inner sep=0]
            \node [align=center, img] {\includegraphics[width=\textwidth]{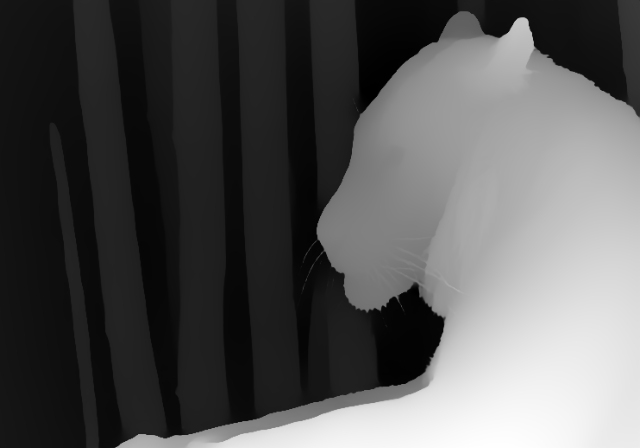}};
            \spy on \zoomnine in node [left] at \rebigone;
    	\end{tikzpicture}
        \caption*{Depth Anything V2}
    \end{subfigure}
    \begin{subfigure}{\depthWidth}
        \begin{tikzpicture}[spy using outlines={green,magnification=\ssmag,size=\ssizz},inner sep=0]
            \node [align=center, img] {\includegraphics[width=\textwidth]{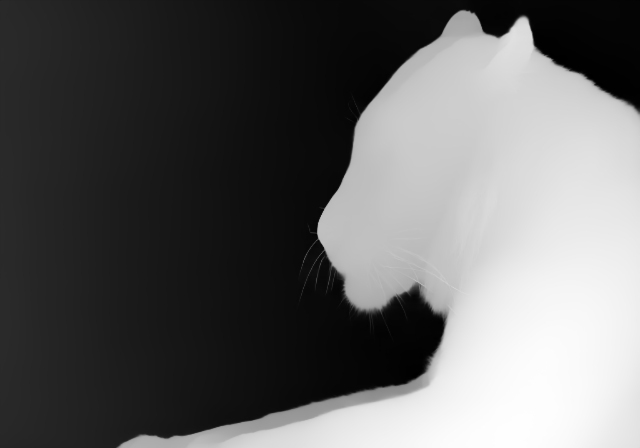}};
            \spy on \zoomnine in node [left] at \rebigone;
    	\end{tikzpicture}
        \caption*{Depth Pro}
      \end{subfigure}
    \begin{subfigure}{\depthWidth}
        \begin{tikzpicture}[spy using outlines={green,magnification=\ssmag,size=\ssizz},inner sep=0]
            \node [align=center, img] {\includegraphics[width=\textwidth]{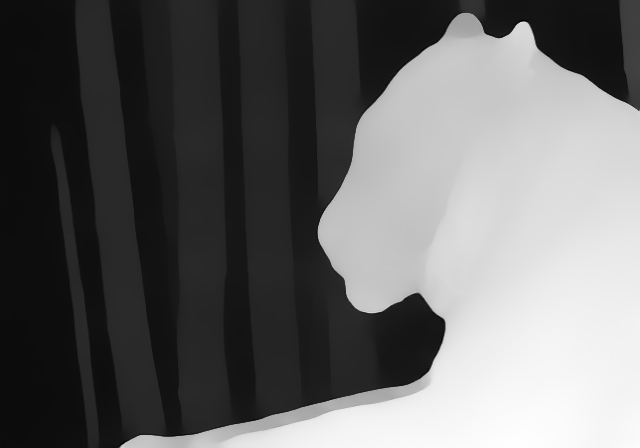}};
            \spy on \zoomnine in node [left] at \rebigone;
    	\end{tikzpicture}
        \caption*{UniDepthV2}
      \end{subfigure}
    \begin{subfigure}{\depthWidth}
        \begin{tikzpicture}[spy using outlines={green,magnification=\ssmag,size=\ssizz},inner sep=0]
            \node [align=center, img] {\includegraphics[width=\textwidth]{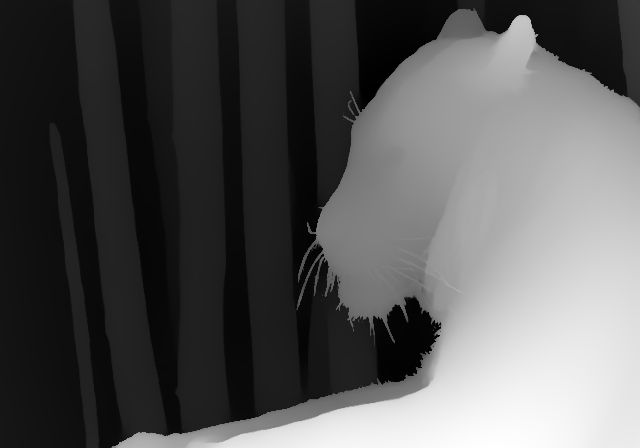}};
            \spy on \zoomnine in node [left] at \rebigone;
    	\end{tikzpicture}
        \caption*{Ours}
      \end{subfigure}
    \end{subfigure}
    \caption{\textbf{Qualitative comparison of depth estimation} on the AIM-500 dataset. }
    \label{fig:supp-depth-visual-aim}
\end{figure*}

%% file: figs/supp/fig-depth_est_visuals_p3m.tex
\def\imgWidth{0.32\linewidth} %
\def\depthWidth{0.19\linewidth} %
\def\pointWidth{0.24\linewidth} %
\def\scc{(-1.9,-1.4)}

\def\rebigone{(-0.5, -0.55)} %
\def\rebigtwo{(1.6, -0.55)} %

\def\zoomone{(1,0.15)} %
\def\zoomtwo{(0.7,0.5)} %
\def\zoomthree{(-1,0.9)} %

\def\zoomfour{(-0.8,0)} %
\def\zoomfive{(0.15,-0.15)} %

\def\zoomsix{(-0.6,0.5)} %
\def\zoomseven{(-0.12,0.95)} %
\def\zoomeight{(1.2,-0.8)} %
\def\zoomnine{(0.3,0.2)} %

\def\ssizz{1.1cm} %
\def\ssmag{3}

\begin{figure*}[t]
\centering
\tikzstyle{img} = [rectangle, minimum width=\imgWidth]
    \centering
    \begin{subfigure}{\linewidth}
    \centering
    \begin{subfigure}{\depthWidth}
        \begin{tikzpicture}[spy using outlines={green,magnification=\ssmag,size=\ssizz},inner sep=0]
            \node [align=center, img] {\includegraphics[width=\textwidth]{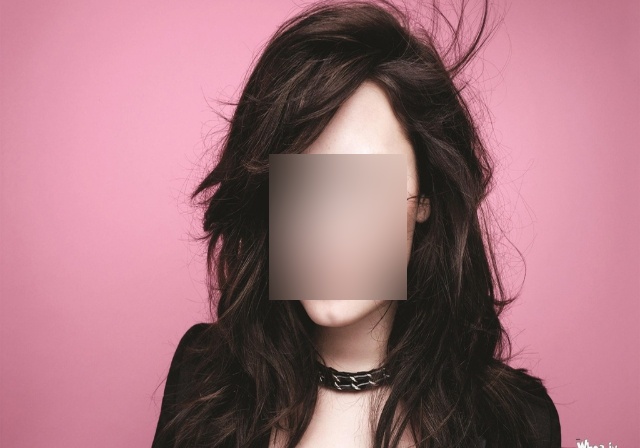}};
            \spy on \zoomone in node [left] at \rebigone;
    	\end{tikzpicture}
    \end{subfigure}
    \begin{subfigure}{\depthWidth}
		\begin{tikzpicture}[spy using outlines={green,magnification=\ssmag,size=\ssizz},inner sep=0]
            \node [align=center, img] {\includegraphics[width=\textwidth]{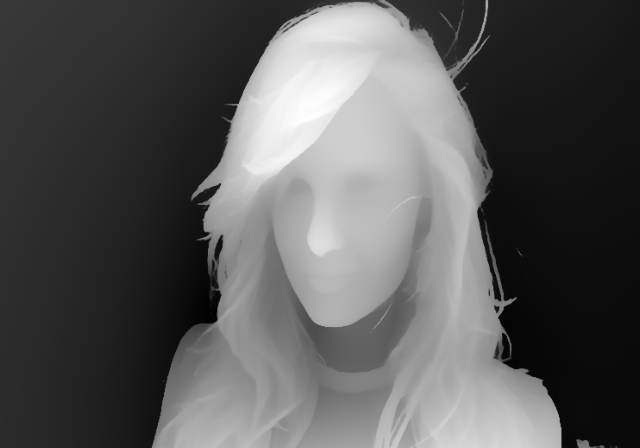}};
            \spy on \zoomone in node [left] at \rebigone;
    	\end{tikzpicture}
    \end{subfigure}
    \begin{subfigure}{\depthWidth}
        \begin{tikzpicture}[spy using outlines={green,magnification=\ssmag,size=\ssizz},inner sep=0]
            \node [align=center, img] {\includegraphics[width=\textwidth]{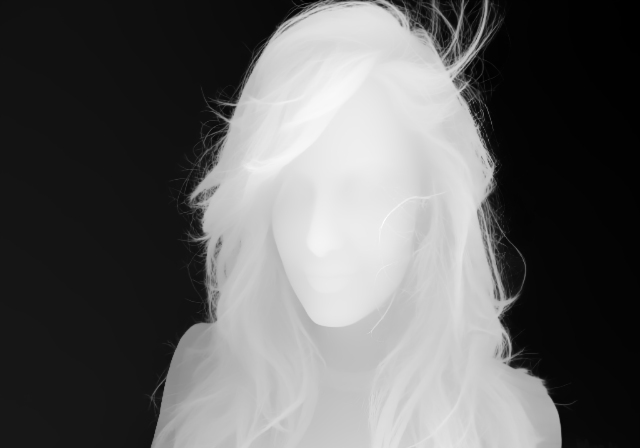}};
            \spy on \zoomone in node [left] at \rebigone;
    	\end{tikzpicture}
      \end{subfigure}
    \begin{subfigure}{\depthWidth}
        \begin{tikzpicture}[spy using outlines={green,magnification=\ssmag,size=\ssizz},inner sep=0]
            \node [align=center, img] {\includegraphics[width=\textwidth]{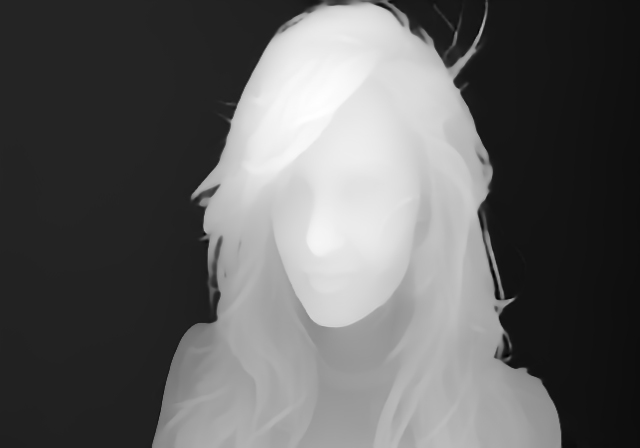}};
            \spy on \zoomone in node [left] at \rebigone;
    	\end{tikzpicture}
      \end{subfigure}
    \begin{subfigure}{\depthWidth}
        \begin{tikzpicture}[spy using outlines={green,magnification=\ssmag,size=\ssizz},inner sep=0]
            \node [align=center, img] {\includegraphics[width=\textwidth]{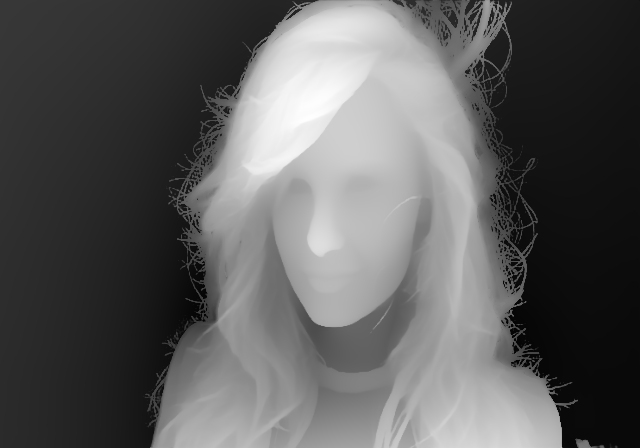}};
            \spy on \zoomone in node [left] at \rebigone;
    	\end{tikzpicture}
      \end{subfigure}
    \end{subfigure}
    \begin{subfigure}{\linewidth}
    \centering
    \begin{subfigure}{\depthWidth}
        \begin{tikzpicture}[spy using outlines={green,magnification=\ssmag,size=\ssizz},inner sep=0]
            \node [align=center, img] {\includegraphics[width=\textwidth]{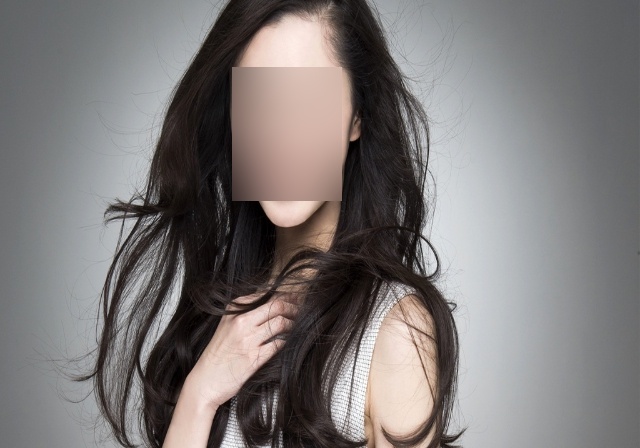}};
            \spy on \zoomtwo in node [left] at \rebigone;
    	\end{tikzpicture}
    \end{subfigure}
    \begin{subfigure}{\depthWidth}
		\begin{tikzpicture}[spy using outlines={green,magnification=\ssmag,size=\ssizz},inner sep=0]
            \node [align=center, img] {\includegraphics[width=\textwidth]{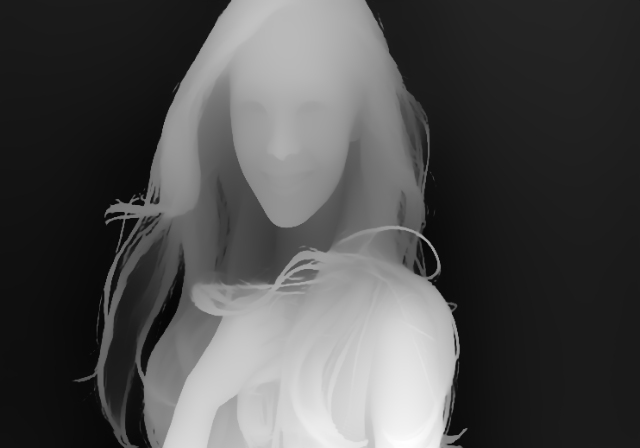}};
            \spy on \zoomtwo in node [left] at \rebigone;
    	\end{tikzpicture}
    \end{subfigure}
    \begin{subfigure}{\depthWidth}
        \begin{tikzpicture}[spy using outlines={green,magnification=\ssmag,size=\ssizz},inner sep=0]
            \node [align=center, img] {\includegraphics[width=\textwidth]{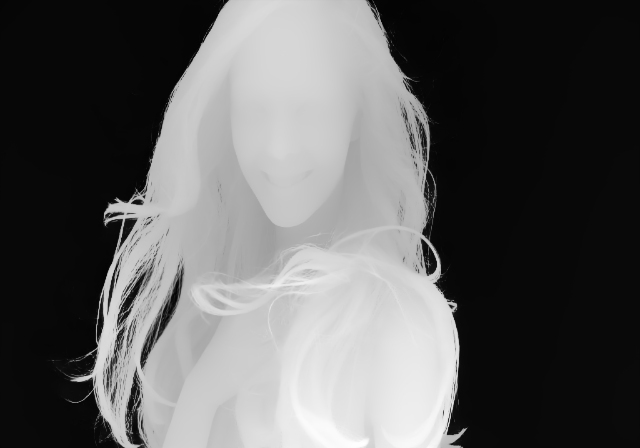}};
            \spy on \zoomtwo in node [left] at \rebigone;
    	\end{tikzpicture}
      \end{subfigure}
    \begin{subfigure}{\depthWidth}
        \begin{tikzpicture}[spy using outlines={green,magnification=\ssmag,size=\ssizz},inner sep=0]
            \node [align=center, img] {\includegraphics[width=\textwidth]{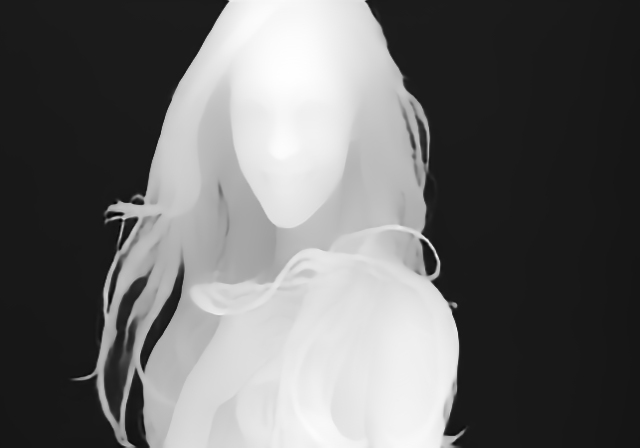}};
            \spy on \zoomtwo in node [left] at \rebigone;
    	\end{tikzpicture}
      \end{subfigure}
    \begin{subfigure}{\depthWidth}
        \begin{tikzpicture}[spy using outlines={green,magnification=\ssmag,size=\ssizz},inner sep=0]
            \node [align=center, img] {\includegraphics[width=\textwidth]{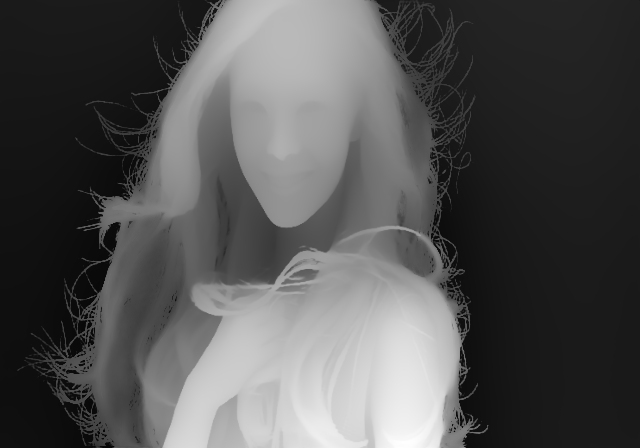}};
            \spy on \zoomtwo in node [left] at \rebigone;
    	\end{tikzpicture}
      \end{subfigure}
    \end{subfigure}
    \begin{subfigure}{\linewidth}
    \centering
    \begin{subfigure}{\depthWidth}
        \begin{tikzpicture}[spy using outlines={green,magnification=\ssmag,size=\ssizz},inner sep=0]
            \node [align=center, img] {\includegraphics[width=\textwidth]{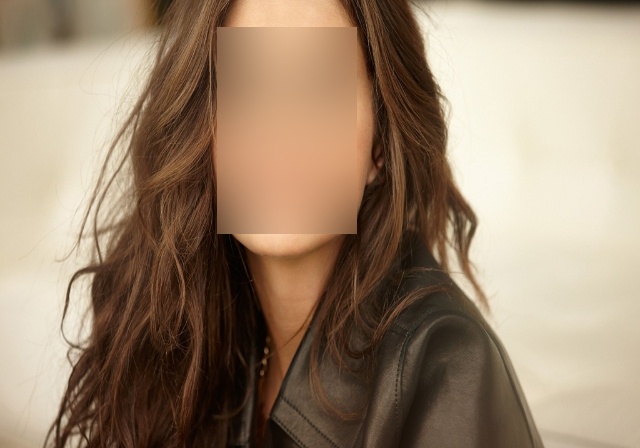}};
            \spy on \zoomthree in node [left] at \rebigone;
    	\end{tikzpicture}
    \end{subfigure}
    \begin{subfigure}{\depthWidth}
		\begin{tikzpicture}[spy using outlines={green,magnification=\ssmag,size=\ssizz},inner sep=0]
            \node [align=center, img] {\includegraphics[width=\textwidth]{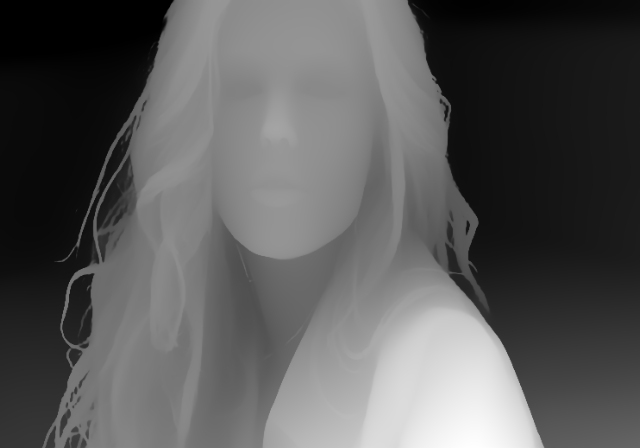}};
            \spy on \zoomthree in node [left] at \rebigone;
    	\end{tikzpicture}
    \end{subfigure}
    \begin{subfigure}{\depthWidth}
        \begin{tikzpicture}[spy using outlines={green,magnification=\ssmag,size=\ssizz},inner sep=0]
            \node [align=center, img] {\includegraphics[width=\textwidth]{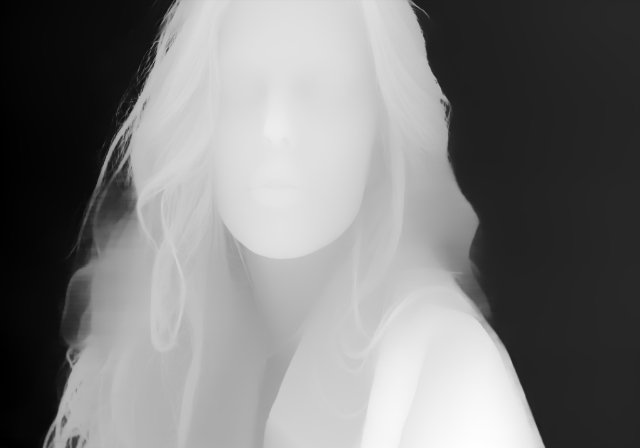}};
            \spy on \zoomthree in node [left] at \rebigone;
    	\end{tikzpicture}
      \end{subfigure}
    \begin{subfigure}{\depthWidth}
        \begin{tikzpicture}[spy using outlines={green,magnification=\ssmag,size=\ssizz},inner sep=0]
            \node [align=center, img] {\includegraphics[width=\textwidth]{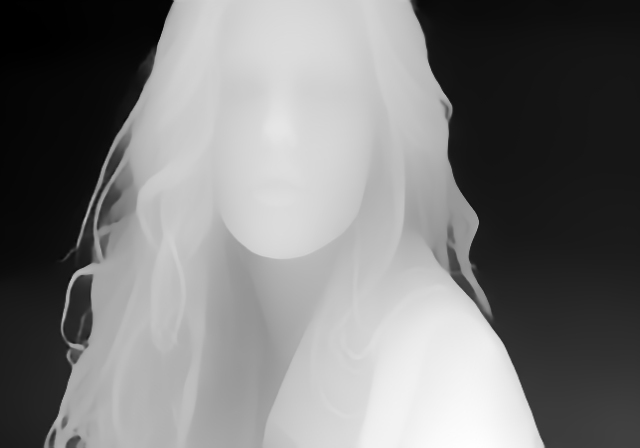}};
            \spy on \zoomthree in node [left] at \rebigone;
    	\end{tikzpicture}
      \end{subfigure}
    \begin{subfigure}{\depthWidth}
        \begin{tikzpicture}[spy using outlines={green,magnification=\ssmag,size=\ssizz},inner sep=0]
            \node [align=center, img] {\includegraphics[width=\textwidth]{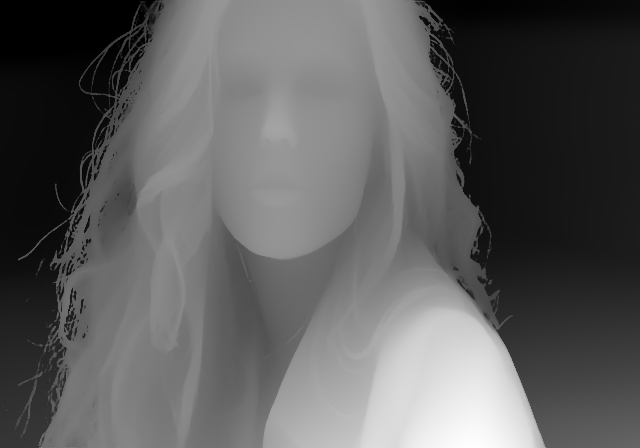}};
            \spy on \zoomthree in node [left] at \rebigone;
    	\end{tikzpicture}
      \end{subfigure}
    \end{subfigure}
    \begin{subfigure}{\linewidth}
    \centering
    \begin{subfigure}{\depthWidth}
        \begin{tikzpicture}[spy using outlines={green,magnification=\ssmag,size=\ssizz},inner sep=0]
            \node [align=center, img] {\includegraphics[width=\textwidth]{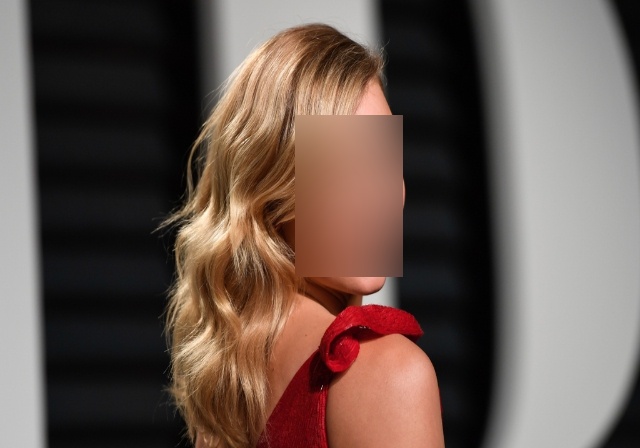}};
            \spy on \zoomfour in node [left] at \rebigtwo;
    	\end{tikzpicture}
    \end{subfigure}
    \begin{subfigure}{\depthWidth}
		\begin{tikzpicture}[spy using outlines={green,magnification=\ssmag,size=\ssizz},inner sep=0]
            \node [align=center, img] {\includegraphics[width=\textwidth]{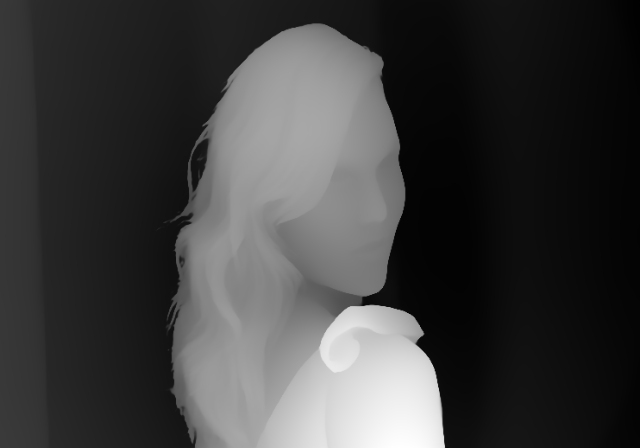}};
            \spy on \zoomfour in node [left] at \rebigtwo;
    	\end{tikzpicture}
    \end{subfigure}
    \begin{subfigure}{\depthWidth}
        \begin{tikzpicture}[spy using outlines={green,magnification=\ssmag,size=\ssizz},inner sep=0]
            \node [align=center, img] {\includegraphics[width=\textwidth]{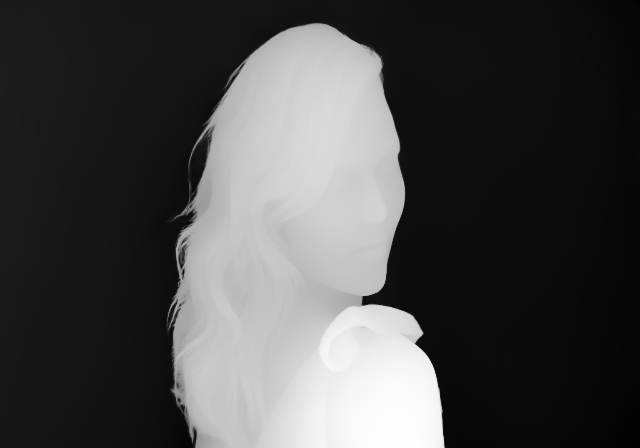}};
            \spy on \zoomfour in node [left] at \rebigtwo;
    	\end{tikzpicture}
      \end{subfigure}
    \begin{subfigure}{\depthWidth}
        \begin{tikzpicture}[spy using outlines={green,magnification=\ssmag,size=\ssizz},inner sep=0]
            \node [align=center, img] {\includegraphics[width=\textwidth]{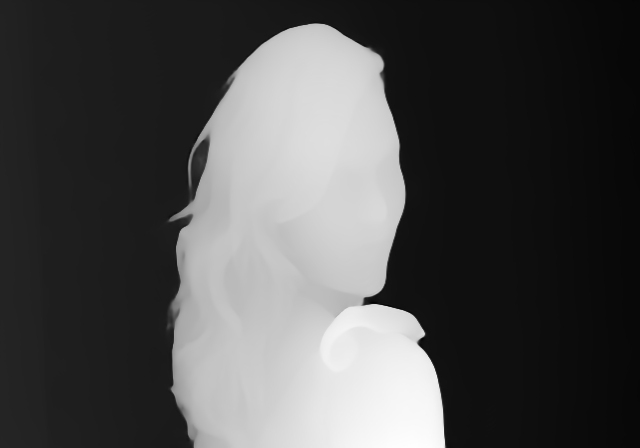}};
            \spy on \zoomfour in node [left] at \rebigtwo;
    	\end{tikzpicture}
      \end{subfigure}
    \begin{subfigure}{\depthWidth}
        \begin{tikzpicture}[spy using outlines={green,magnification=\ssmag,size=\ssizz},inner sep=0]
            \node [align=center, img] {\includegraphics[width=\textwidth]{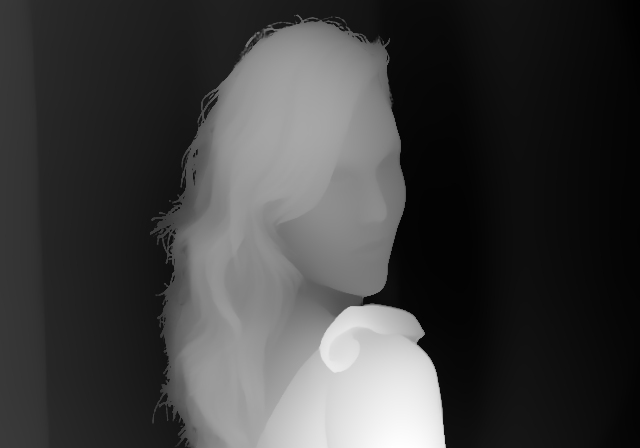}};
            \spy on \zoomfour in node [left] at \rebigtwo;
    	\end{tikzpicture}
      \end{subfigure}
    \end{subfigure}
    \begin{subfigure}{\linewidth}
    \centering
    \begin{subfigure}{\depthWidth}
        \begin{tikzpicture}[spy using outlines={green,magnification=\ssmag,size=\ssizz},inner sep=0]
            \node [align=center, img] {\includegraphics[width=\textwidth]{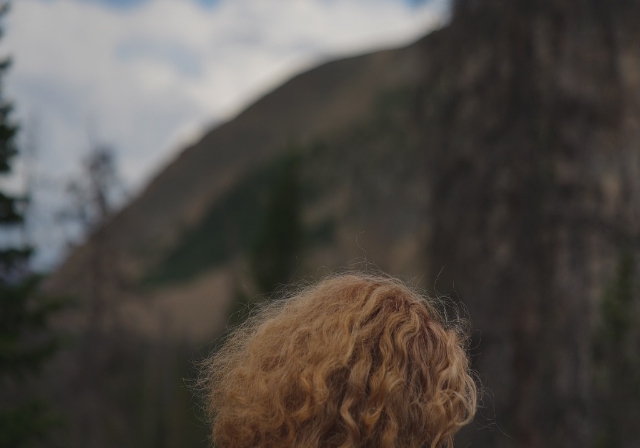}};
            \spy on \zoomfive in node [left] at \rebigone;
    	\end{tikzpicture}
    \end{subfigure}
    \begin{subfigure}{\depthWidth}
		\begin{tikzpicture}[spy using outlines={green,magnification=\ssmag,size=\ssizz},inner sep=0]
            \node [align=center, img] {\includegraphics[width=\textwidth]{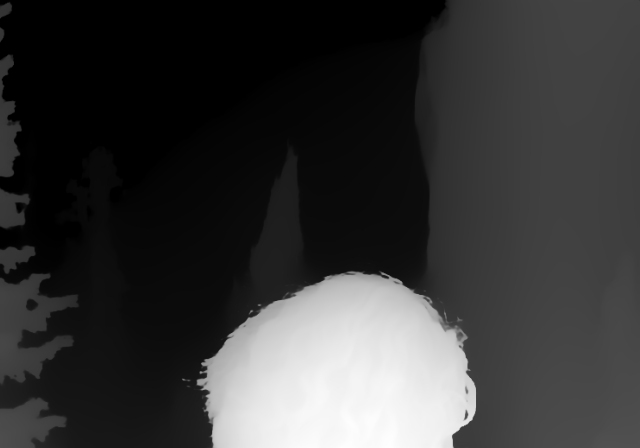}};
            \spy on \zoomfive in node [left] at \rebigone;
    	\end{tikzpicture}
    \end{subfigure}
    \begin{subfigure}{\depthWidth}
        \begin{tikzpicture}[spy using outlines={green,magnification=\ssmag,size=\ssizz},inner sep=0]
            \node [align=center, img] {\includegraphics[width=\textwidth]{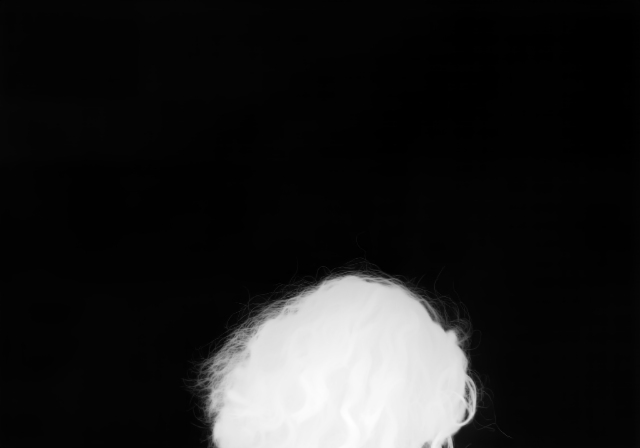}};
            \spy on \zoomfive in node [left] at \rebigone;
    	\end{tikzpicture}
      \end{subfigure}
    \begin{subfigure}{\depthWidth}
        \begin{tikzpicture}[spy using outlines={green,magnification=\ssmag,size=\ssizz},inner sep=0]
            \node [align=center, img] {\includegraphics[width=\textwidth]{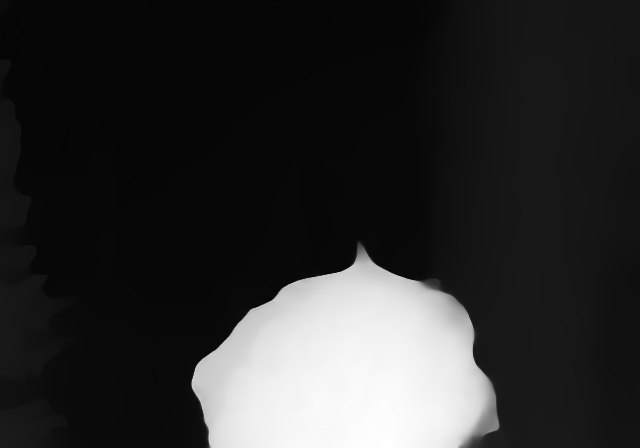}};
            \spy on \zoomfive in node [left] at \rebigone;
    	\end{tikzpicture}
      \end{subfigure}
    \begin{subfigure}{\depthWidth}
        \begin{tikzpicture}[spy using outlines={green,magnification=\ssmag,size=\ssizz},inner sep=0]
            \node [align=center, img] {\includegraphics[width=\textwidth]{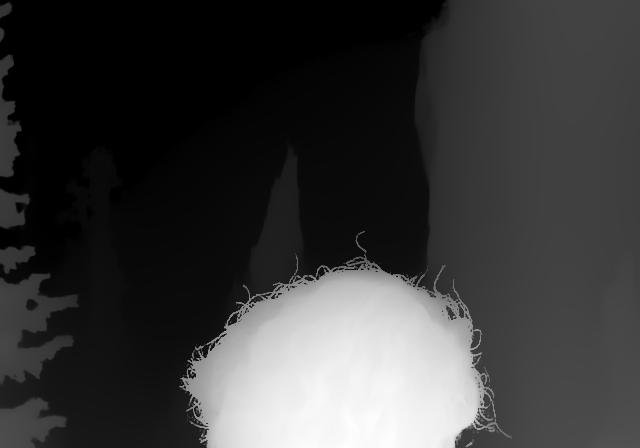}};
            \spy on \zoomfive in node [left] at \rebigone;
    	\end{tikzpicture}
      \end{subfigure}
    \end{subfigure}
    \begin{subfigure}{\linewidth}
    \centering
    \begin{subfigure}{\depthWidth}
        \begin{tikzpicture}[spy using outlines={green,magnification=\ssmag,size=\ssizz},inner sep=0]
            \node [align=center, img] {\includegraphics[width=\textwidth]{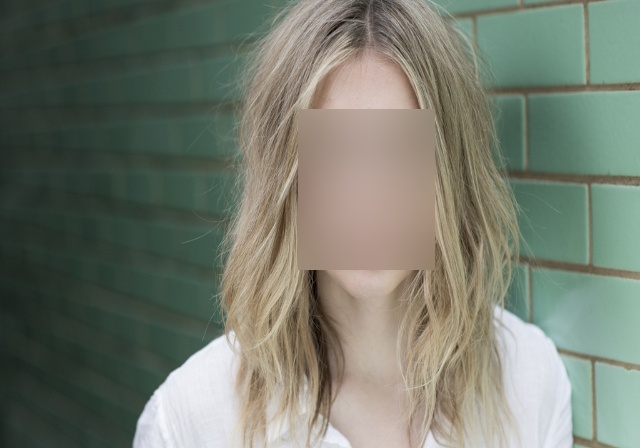}};
            \spy on \zoomsix in node [left] at \rebigone;
    	\end{tikzpicture}
    \end{subfigure}
    \begin{subfigure}{\depthWidth}
		\begin{tikzpicture}[spy using outlines={green,magnification=\ssmag,size=\ssizz},inner sep=0]
            \node [align=center, img] {\includegraphics[width=\textwidth]{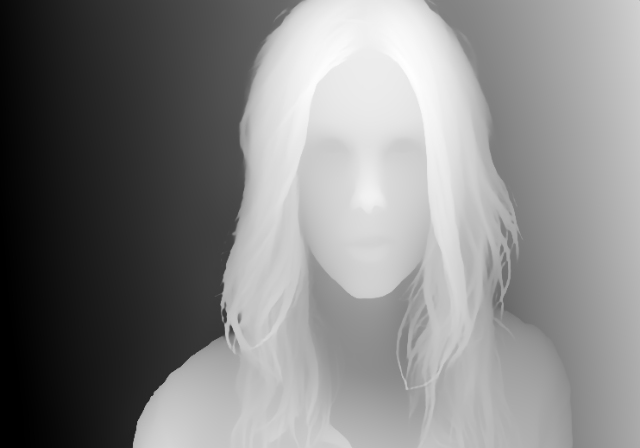}};
            \spy on \zoomsix in node [left] at \rebigone;
    	\end{tikzpicture}
    \end{subfigure}
    \begin{subfigure}{\depthWidth}
        \begin{tikzpicture}[spy using outlines={green,magnification=\ssmag,size=\ssizz},inner sep=0]
            \node [align=center, img] {\includegraphics[width=\textwidth]{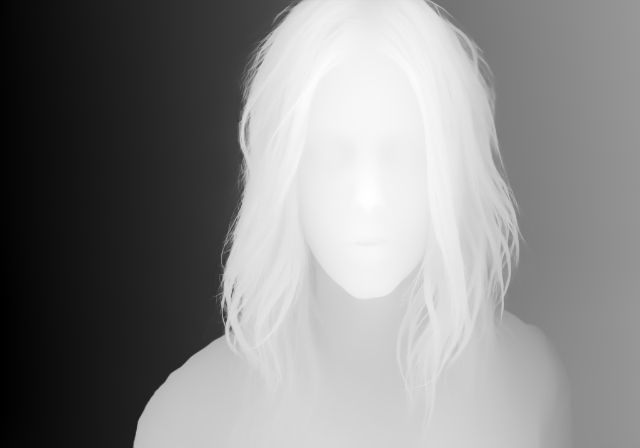}};
            \spy on \zoomsix in node [left] at \rebigone;
    	\end{tikzpicture}
      \end{subfigure}
    \begin{subfigure}{\depthWidth}
        \begin{tikzpicture}[spy using outlines={green,magnification=\ssmag,size=\ssizz},inner sep=0]
            \node [align=center, img] {\includegraphics[width=\textwidth]{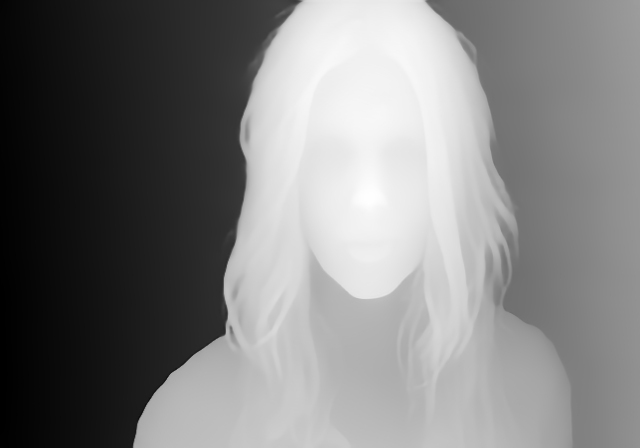}};
            \spy on \zoomsix in node [left] at \rebigone;
    	\end{tikzpicture}
      \end{subfigure}
    \begin{subfigure}{\depthWidth}
        \begin{tikzpicture}[spy using outlines={green,magnification=\ssmag,size=\ssizz},inner sep=0]
            \node [align=center, img] {\includegraphics[width=\textwidth]{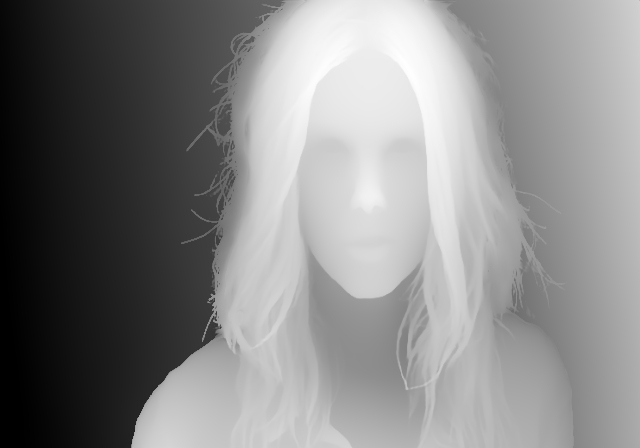}};
            \spy on \zoomsix in node [left] at \rebigone;
    	\end{tikzpicture}
      \end{subfigure}
    \end{subfigure}
    \begin{subfigure}{\linewidth}
    \centering
    \begin{subfigure}{\depthWidth}
        \begin{tikzpicture}[spy using outlines={green,magnification=\ssmag,size=\ssizz},inner sep=0]
            \node [align=center, img] {\includegraphics[width=\textwidth]{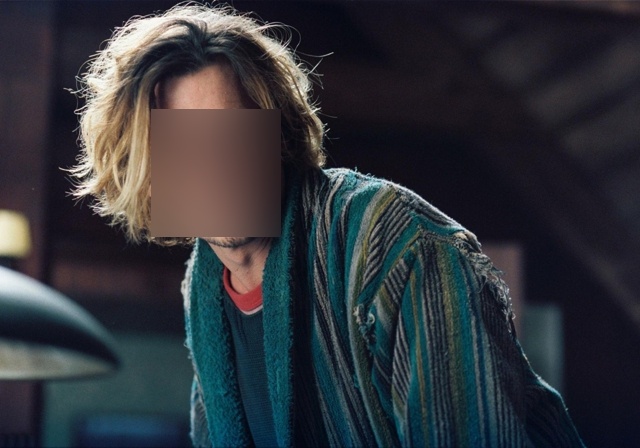}};
            \spy on \zoomseven in node [left] at \rebigone;
    	\end{tikzpicture}
    \end{subfigure}
    \begin{subfigure}{\depthWidth}
		\begin{tikzpicture}[spy using outlines={green,magnification=\ssmag,size=\ssizz},inner sep=0]
            \node [align=center, img] {\includegraphics[width=\textwidth]{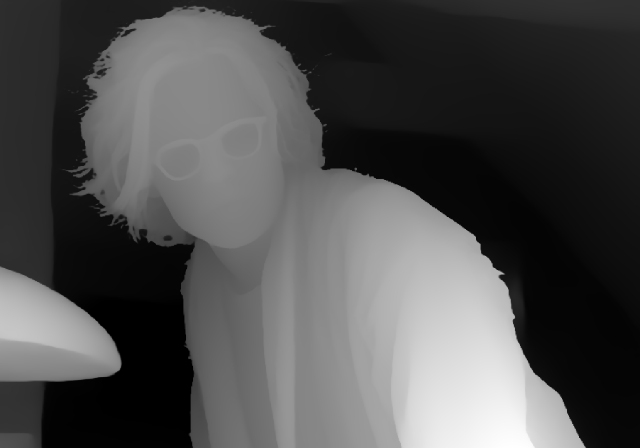}};
            \spy on \zoomseven in node [left] at \rebigone;
    	\end{tikzpicture}
    \end{subfigure}
    \begin{subfigure}{\depthWidth}
        \begin{tikzpicture}[spy using outlines={green,magnification=\ssmag,size=\ssizz},inner sep=0]
            \node [align=center, img] {\includegraphics[width=\textwidth]{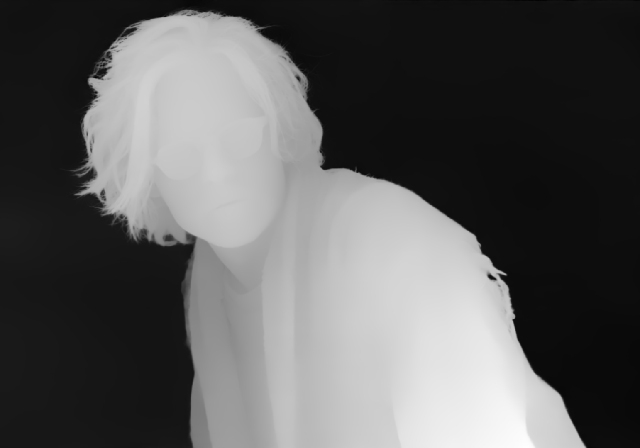}};
            \spy on \zoomseven in node [left] at \rebigone;
    	\end{tikzpicture}
      \end{subfigure}
    \begin{subfigure}{\depthWidth}
        \begin{tikzpicture}[spy using outlines={green,magnification=\ssmag,size=\ssizz},inner sep=0]
            \node [align=center, img] {\includegraphics[width=\textwidth]{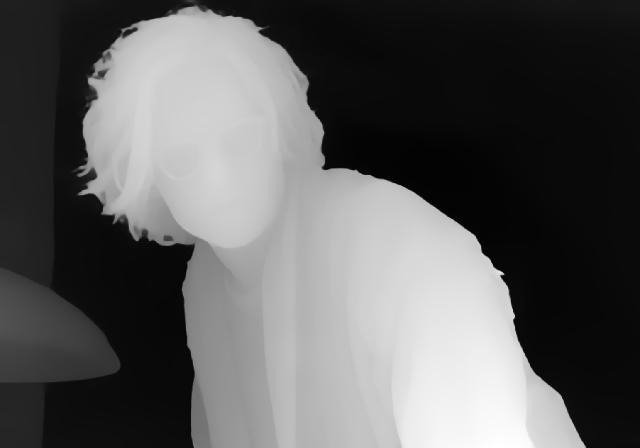}};
            \spy on \zoomseven in node [left] at \rebigone;
    	\end{tikzpicture}
      \end{subfigure}
    \begin{subfigure}{\depthWidth}
        \begin{tikzpicture}[spy using outlines={green,magnification=\ssmag,size=\ssizz},inner sep=0]
            \node [align=center, img] {\includegraphics[width=\textwidth]{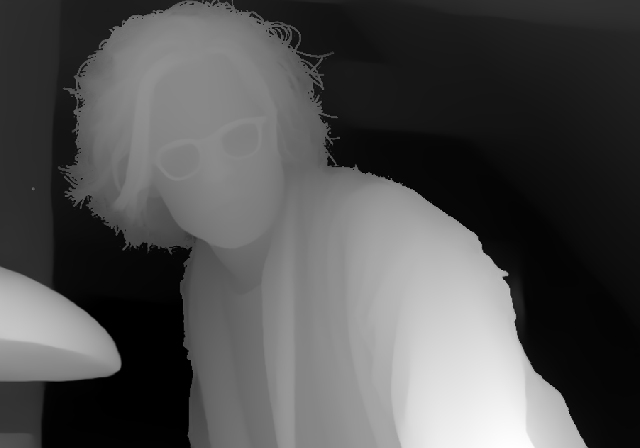}};
            \spy on \zoomseven in node [left] at \rebigone;
    	\end{tikzpicture}
      \end{subfigure}
    \end{subfigure}
    \begin{subfigure}{\linewidth}
    \centering
    \begin{subfigure}{\depthWidth}
        \begin{tikzpicture}[spy using outlines={green,magnification=\ssmag,size=\ssizz},inner sep=0]
            \node [align=center, img] {\includegraphics[width=\textwidth]{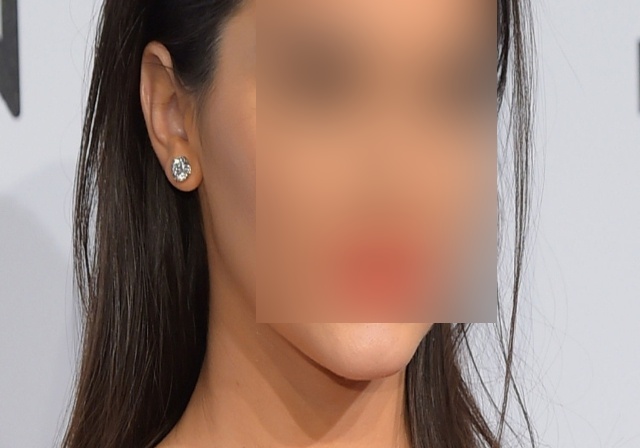}};
            \spy on \zoomeight in node [left] at \rebigone;
    	\end{tikzpicture}
    \end{subfigure}
    \begin{subfigure}{\depthWidth}
		\begin{tikzpicture}[spy using outlines={green,magnification=\ssmag,size=\ssizz},inner sep=0]
            \node [align=center, img] {\includegraphics[width=\textwidth]{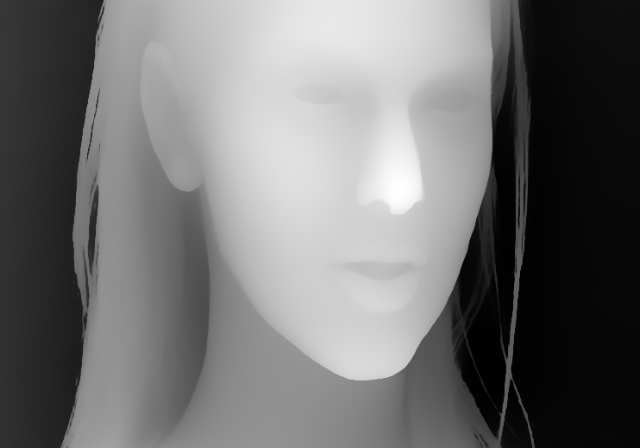}};
            \spy on \zoomeight in node [left] at \rebigone;
    	\end{tikzpicture}
    \end{subfigure}
    \begin{subfigure}{\depthWidth}
        \begin{tikzpicture}[spy using outlines={green,magnification=\ssmag,size=\ssizz},inner sep=0]
            \node [align=center, img] {\includegraphics[width=\textwidth]{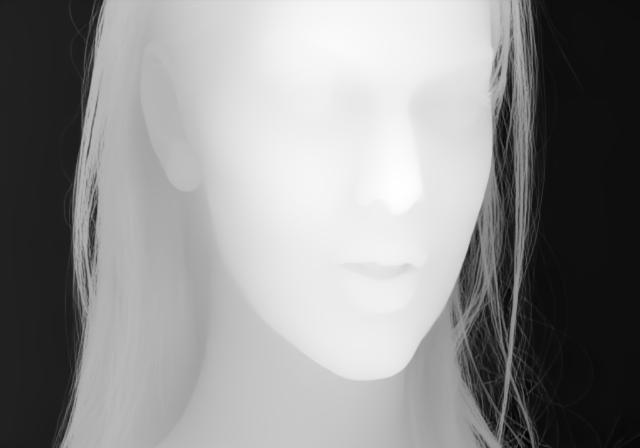}};
            \spy on \zoomeight in node [left] at \rebigone;
    	\end{tikzpicture}
      \end{subfigure}
    \begin{subfigure}{\depthWidth}
        \begin{tikzpicture}[spy using outlines={green,magnification=\ssmag,size=\ssizz},inner sep=0]
            \node [align=center, img] {\includegraphics[width=\textwidth]{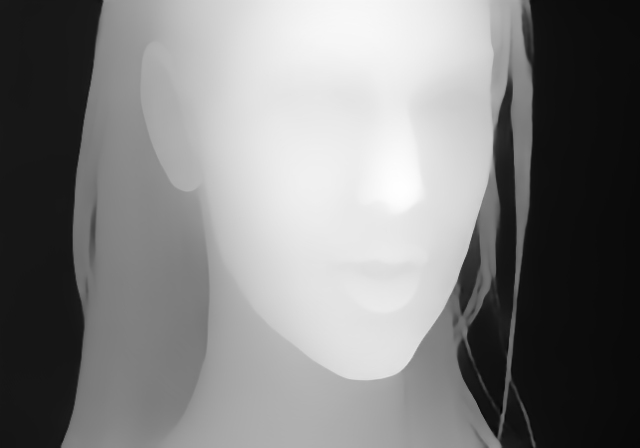}};
            \spy on \zoomeight in node [left] at \rebigone;
    	\end{tikzpicture}
      \end{subfigure}
    \begin{subfigure}{\depthWidth}
        \begin{tikzpicture}[spy using outlines={green,magnification=\ssmag,size=\ssizz},inner sep=0]
            \node [align=center, img] {\includegraphics[width=\textwidth]{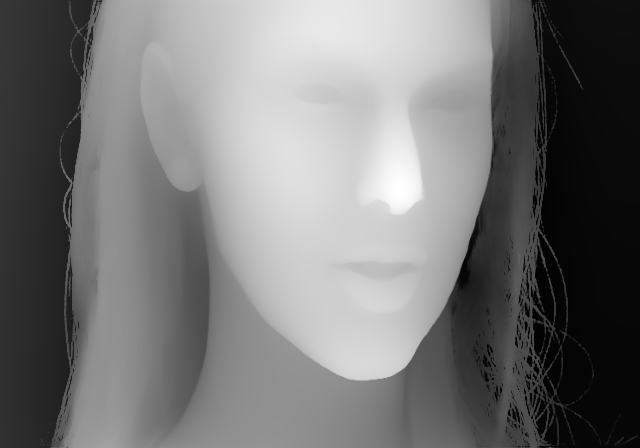}};
            \spy on \zoomeight in node [left] at \rebigone;
    	\end{tikzpicture}
      \end{subfigure}
    \end{subfigure}
    \begin{subfigure}{\linewidth}
    \centering
    \begin{subfigure}{\depthWidth}
        \begin{tikzpicture}[spy using outlines={green,magnification=\ssmag,size=\ssizz},inner sep=0]
            \node [align=center, img] {\includegraphics[width=\textwidth]{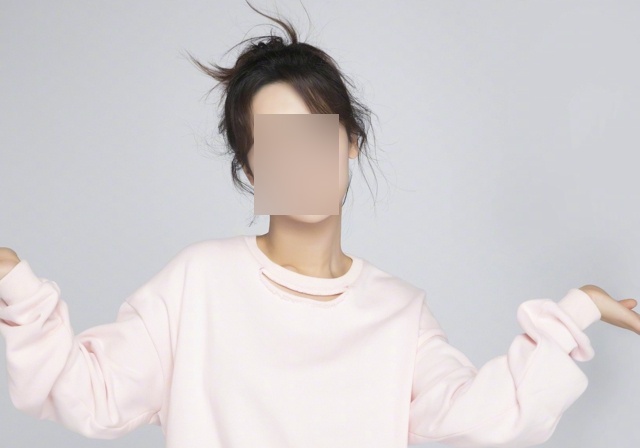}};
            \spy on \zoomnine in node [left] at \rebigone;
    	\end{tikzpicture}
        \caption*{Input Image}
    \end{subfigure}
    \begin{subfigure}{\depthWidth}
		\begin{tikzpicture}[spy using outlines={green,magnification=\ssmag,size=\ssizz},inner sep=0]
            \node [align=center, img] {\includegraphics[width=\textwidth]{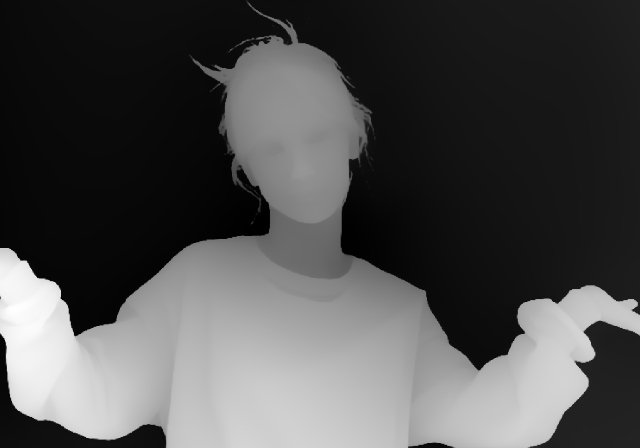}};
            \spy on \zoomnine in node [left] at \rebigone;
    	\end{tikzpicture}
        \caption*{Depth Anything V2}
    \end{subfigure}
    \begin{subfigure}{\depthWidth}
        \begin{tikzpicture}[spy using outlines={green,magnification=\ssmag,size=\ssizz},inner sep=0]
            \node [align=center, img] {\includegraphics[width=\textwidth]{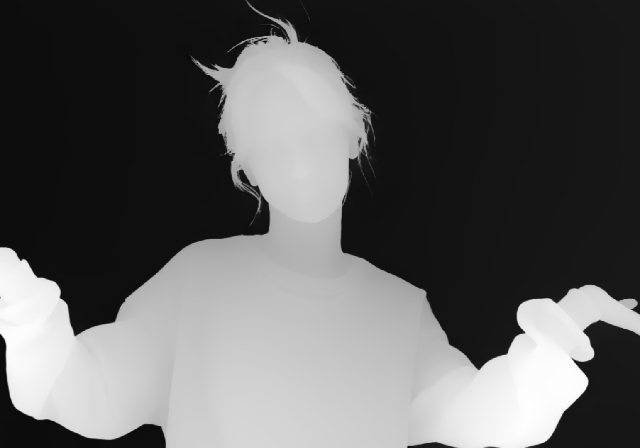}};
            \spy on \zoomnine in node [left] at \rebigone;
    	\end{tikzpicture}
        \caption*{Depth Pro}
      \end{subfigure}
    \begin{subfigure}{\depthWidth}
        \begin{tikzpicture}[spy using outlines={green,magnification=\ssmag,size=\ssizz},inner sep=0]
            \node [align=center, img] {\includegraphics[width=\textwidth]{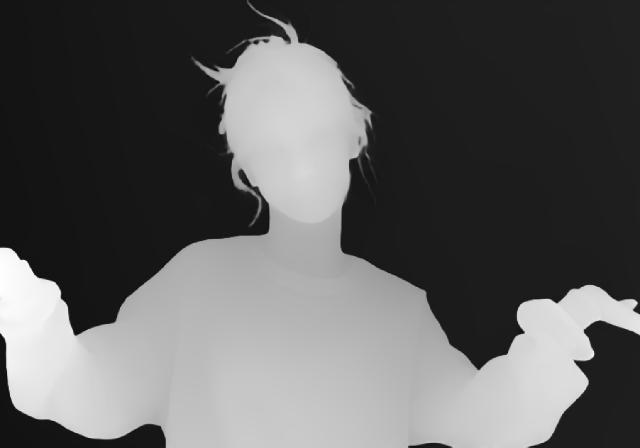}};
            \spy on \zoomnine in node [left] at \rebigone;
    	\end{tikzpicture}
        \caption*{UniDepthV2}
      \end{subfigure}
    \begin{subfigure}{\depthWidth}
        \begin{tikzpicture}[spy using outlines={green,magnification=\ssmag,size=\ssizz},inner sep=0]
            \node [align=center, img] {\includegraphics[width=\textwidth]{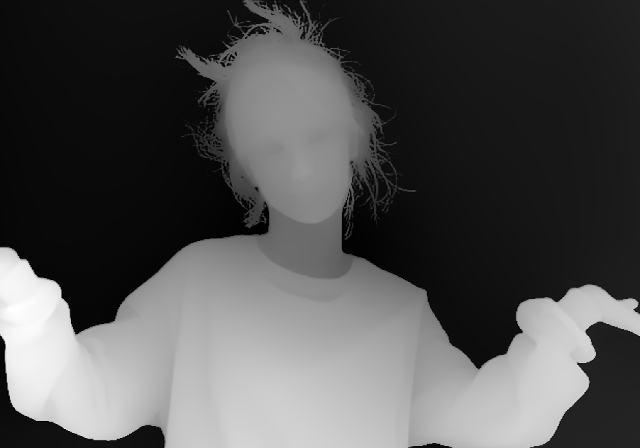}};
            \spy on \zoomnine in node [left] at \rebigone;
    	\end{tikzpicture}
        \caption*{Ours}
      \end{subfigure}
    \end{subfigure}
    \caption{\textbf{Qualitative comparison of depth estimation} on the P3M-10K dataset. Human faces are manually blurred to protect privacy.}
    \label{fig:supp-depth-visual-p3m}
\end{figure*}

%% file: figs/supp/fig-stereo_visuals_marvel.tex
\def\imgWidth{0.32\linewidth} %
\def\depthWidth{0.19\linewidth} %
\def\pointWidth{0.24\linewidth} %
\def\scc{(-1.9,-1.4)}

\def\rebigone{(-0.6, -0.44)} %
\def\rebigtwo{(1.6, -0.44)} %

\def\zoomone{(-1.48,0.6)} %
\def\zoomtwo{(-0.5,0.2)} %
\def\zoomthree{(-0.3,0.65)} %
\def\zoomfour{(-1.2,-0.5)} %
\def\zoomfive{(-0.3,-0.35)} %

\def\zoomsix{(-0.45,0.35)} %
\def\zoomsixsec{(-0.52,0.35)} %

\def\zoomseven{(-0.5,-0.8)} %
\def\zoomeight{(0.15,0.15)} %
\def\zoomnine{(-0.88,-0.05)} %
\def\zoomten{(-1,0.35)} %

\def\ssizz{1cm} %
\def\ssmag{3}

\begin{figure*}[t]
\centering
\tikzstyle{img} = [rectangle, minimum width=\imgWidth]
    \centering
    \begin{subfigure}{\linewidth}
    \centering
    \begin{subfigure}{\depthWidth}
        \begin{tikzpicture}[spy using outlines={green,magnification=\ssmag,size=\ssizz},inner sep=0]
            \node [align=center, img] {\includegraphics[width=\textwidth]{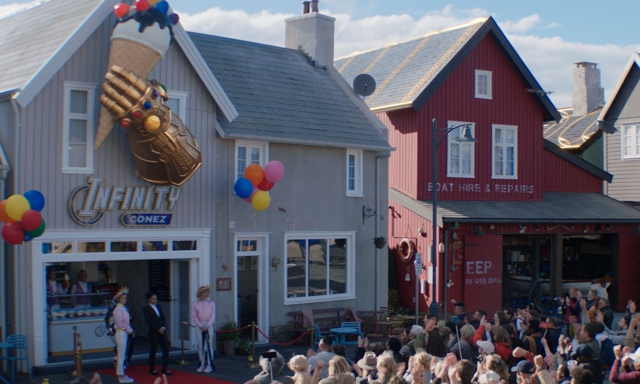}};
            \spy on \zoomnine in node [left] at \rebigtwo;
    	\end{tikzpicture}
    \end{subfigure}
    \begin{subfigure}{\depthWidth}
		\begin{tikzpicture}[spy using outlines={green,magnification=\ssmag,size=\ssizz},inner sep=0]
            \node [align=center, img] {\includegraphics[width=\textwidth]{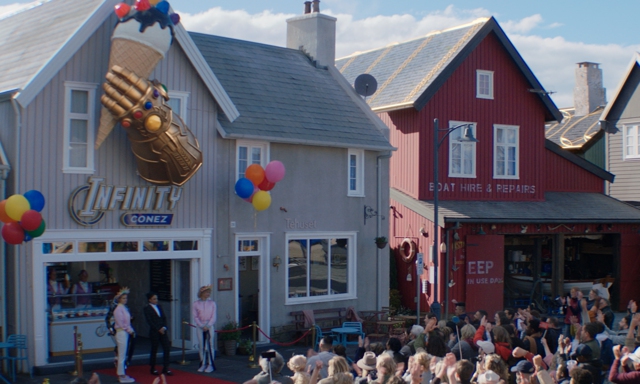}};
            \spy on \zoomnine in node [left] at \rebigtwo;
    	\end{tikzpicture}
    \end{subfigure}
    \begin{subfigure}{\depthWidth}
        \begin{tikzpicture}[spy using outlines={green,magnification=\ssmag,size=\ssizz},inner sep=0]
            \node [align=center, img] {\includegraphics[width=\textwidth]{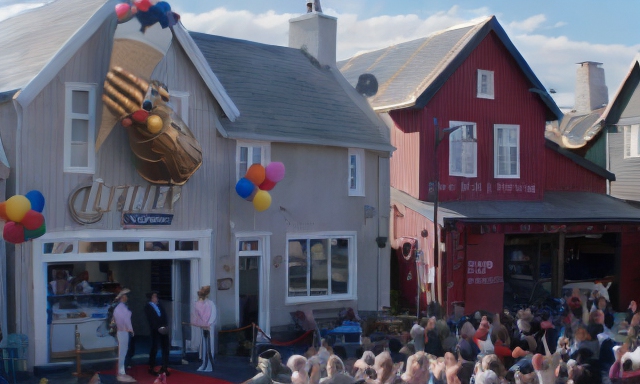}};
            \spy on \zoomnine in node [left] at \rebigtwo;
    	\end{tikzpicture}
      \end{subfigure}
    \begin{subfigure}{\depthWidth}
        \begin{tikzpicture}[spy using outlines={green,magnification=\ssmag,size=\ssizz},inner sep=0]
            \node [align=center, img] {\includegraphics[width=\textwidth]{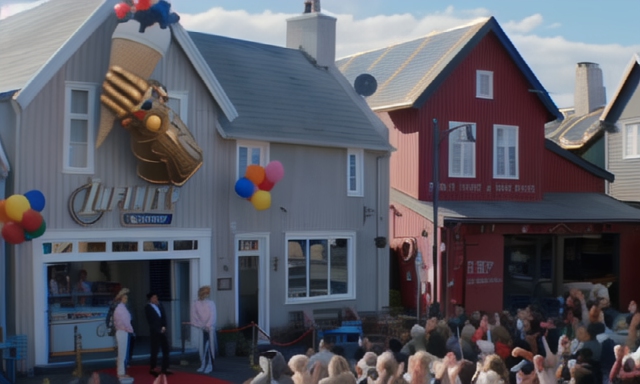}};
            \spy on \zoomnine in node [left] at \rebigtwo;
    	\end{tikzpicture}
      \end{subfigure}
    \begin{subfigure}{\depthWidth}
        \begin{tikzpicture}[spy using outlines={green,magnification=\ssmag,size=\ssizz},inner sep=0]
            \node [align=center, img] {\includegraphics[width=\textwidth]{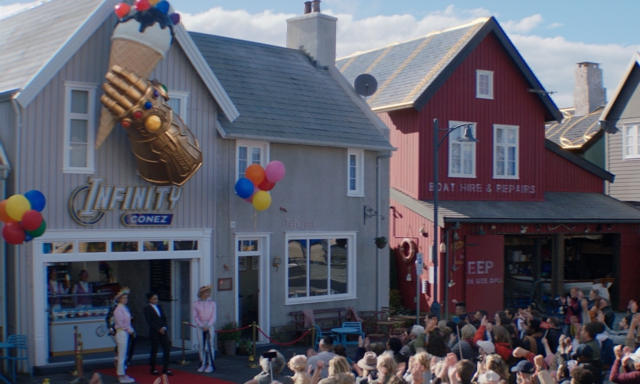}};
            \spy on \zoomnine in node [left] at \rebigtwo;
    	\end{tikzpicture}
      \end{subfigure}
    \end{subfigure}
    \begin{subfigure}{\linewidth}
    \centering
    \begin{subfigure}{\depthWidth}
        \begin{tikzpicture}[spy using outlines={green,magnification=\ssmag,size=\ssizz},inner sep=0]
            \node [align=center, img] {\includegraphics[width=\textwidth]{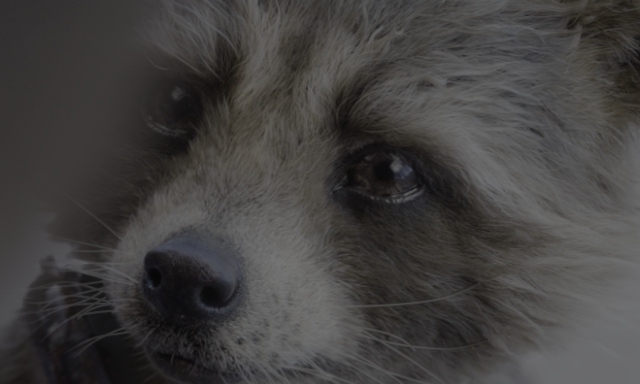}};
            \spy on \zoomfour in node [left] at \rebigtwo;
    	\end{tikzpicture}
    \end{subfigure}
    \begin{subfigure}{\depthWidth}
		\begin{tikzpicture}[spy using outlines={green,magnification=\ssmag,size=\ssizz},inner sep=0]
            \node [align=center, img] {\includegraphics[width=\textwidth]{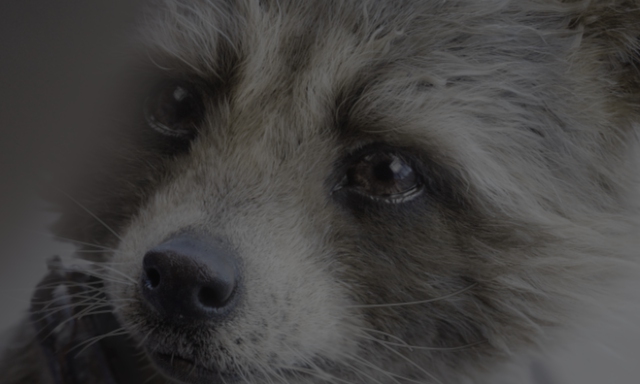}};
            \spy on \zoomfour in node [left] at \rebigtwo;
    	\end{tikzpicture}
    \end{subfigure}
    \begin{subfigure}{\depthWidth}
        \begin{tikzpicture}[spy using outlines={green,magnification=\ssmag,size=\ssizz},inner sep=0]
            \node [align=center, img] {\includegraphics[width=\textwidth]{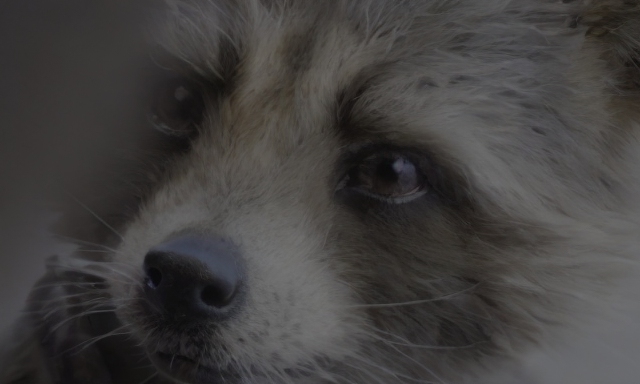}};
            \spy on \zoomfour in node [left] at \rebigtwo;
    	\end{tikzpicture}
      \end{subfigure}
    \begin{subfigure}{\depthWidth}
        \begin{tikzpicture}[spy using outlines={green,magnification=\ssmag,size=\ssizz},inner sep=0]
            \node [align=center, img] {\includegraphics[width=\textwidth]{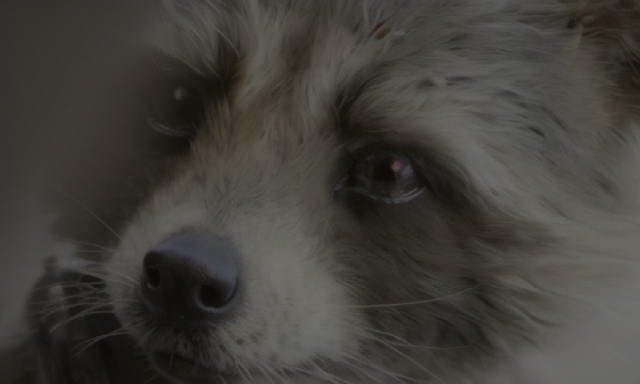}};
            \spy on \zoomfour in node [left] at \rebigtwo;
    	\end{tikzpicture}
      \end{subfigure}
    \begin{subfigure}{\depthWidth}
        \begin{tikzpicture}[spy using outlines={green,magnification=\ssmag,size=\ssizz},inner sep=0]
            \node [align=center, img] {\includegraphics[width=\textwidth]{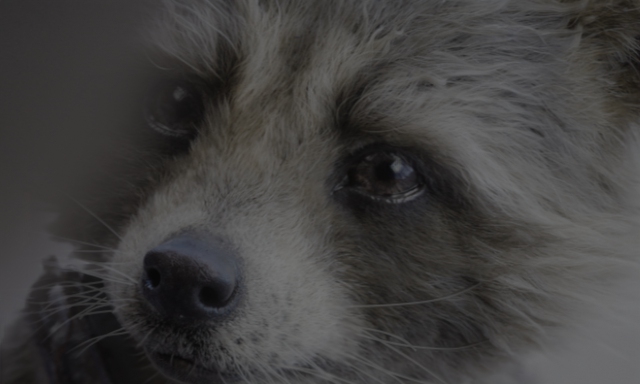}};
            \spy on \zoomfour in node [left] at \rebigtwo;
    	\end{tikzpicture}
      \end{subfigure}
    \end{subfigure}
    
    \begin{subfigure}{\linewidth}
    \centering
    \begin{subfigure}{\depthWidth}
        \begin{tikzpicture}[spy using outlines={green,magnification=\ssmag,size=\ssizz},inner sep=0]
            \node [align=center, img] {\includegraphics[width=\textwidth]{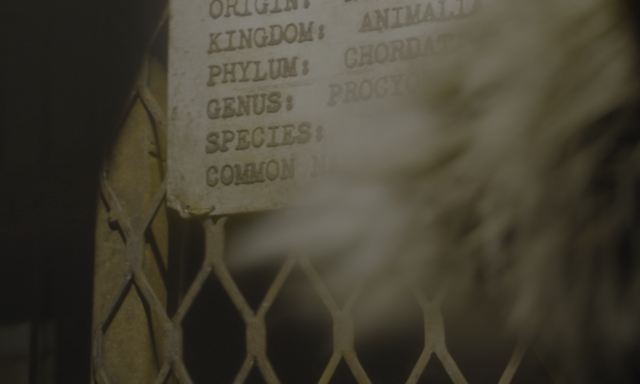}};
            \spy on \zoomsix in node [left] at \rebigone;
    	\end{tikzpicture}
    \end{subfigure}
    \begin{subfigure}{\depthWidth}
		\begin{tikzpicture}[spy using outlines={green,magnification=\ssmag,size=\ssizz},inner sep=0]
            \node [align=center, img] {\includegraphics[width=\textwidth]{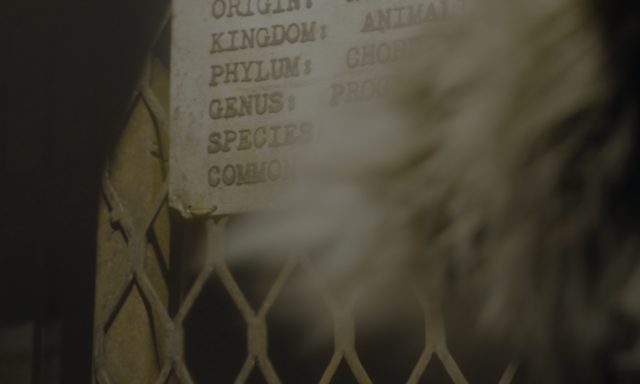}};
            \spy on \zoomsix in node [left] at \rebigone;
    	\end{tikzpicture}
    \end{subfigure}
    \begin{subfigure}{\depthWidth}
        \begin{tikzpicture}[spy using outlines={green,magnification=\ssmag,size=\ssizz},inner sep=0]
            \node [align=center, img] {\includegraphics[width=\textwidth]{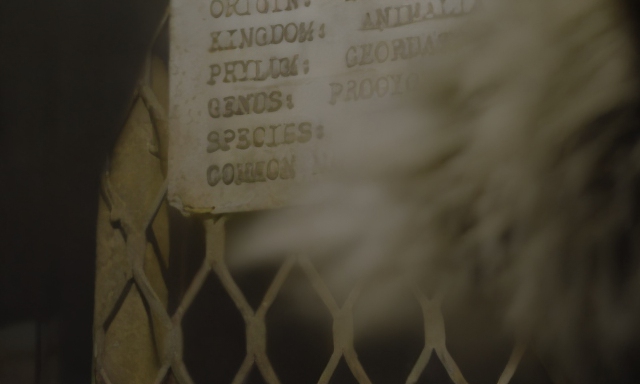}};
            \spy on \zoomsix in node [left] at \rebigone;
    	\end{tikzpicture}
      \end{subfigure}
    \begin{subfigure}{\depthWidth}
        \begin{tikzpicture}[spy using outlines={green,magnification=\ssmag,size=\ssizz},inner sep=0]
            \node [align=center, img] {\includegraphics[width=\textwidth]{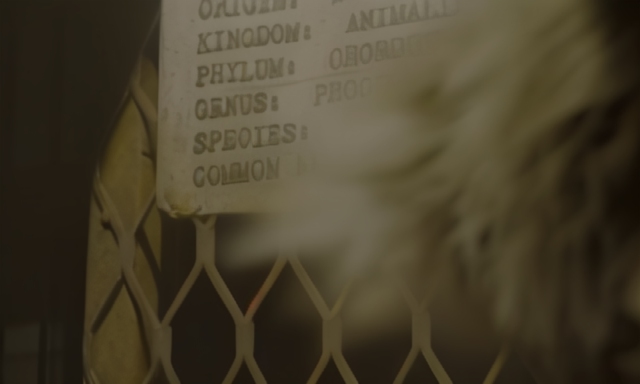}};
            \spy on \zoomsixsec in node [left] at \rebigone;
    	\end{tikzpicture}
      \end{subfigure}
    \begin{subfigure}{\depthWidth}
        \begin{tikzpicture}[spy using outlines={green,magnification=\ssmag,size=\ssizz},inner sep=0]
            \node [align=center, img] {\includegraphics[width=\textwidth]{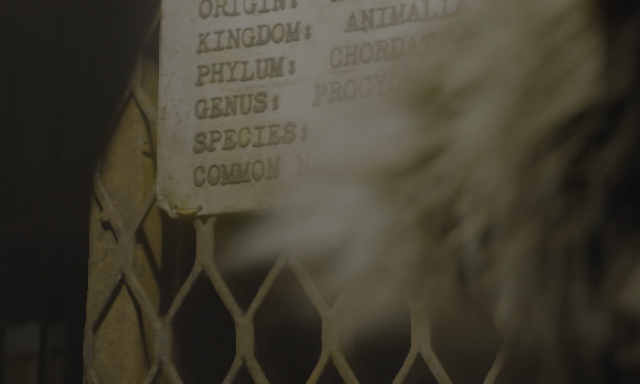}};
            \spy on \zoomsixsec in node [left] at \rebigone;
    	\end{tikzpicture}
      \end{subfigure}
    \end{subfigure}
    \begin{subfigure}{\linewidth}
    \centering
    \begin{subfigure}{\depthWidth}
        \begin{tikzpicture}[spy using outlines={green,magnification=\ssmag,size=\ssizz},inner sep=0]
            \node [align=center, img] {\includegraphics[width=\textwidth]{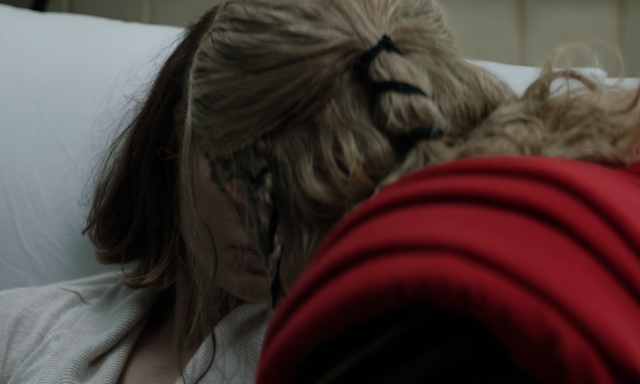}};
            \spy on \zoomseven in node [left] at \rebigtwo;
    	\end{tikzpicture}
    \end{subfigure}
    \begin{subfigure}{\depthWidth}
		\begin{tikzpicture}[spy using outlines={green,magnification=\ssmag,size=\ssizz},inner sep=0]
            \node [align=center, img] {\includegraphics[width=\textwidth]{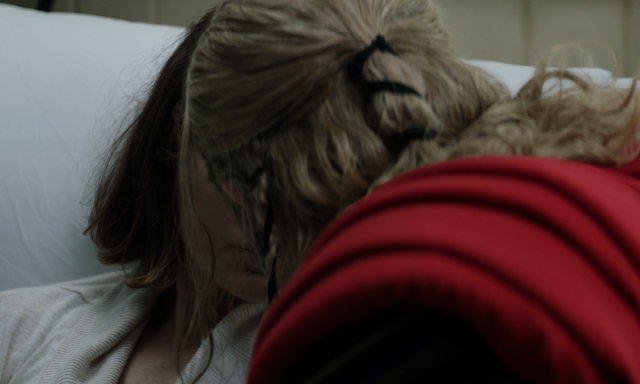}};
            \spy on \zoomseven in node [left] at \rebigtwo;
    	\end{tikzpicture}
    \end{subfigure}
    \begin{subfigure}{\depthWidth}
        \begin{tikzpicture}[spy using outlines={green,magnification=\ssmag,size=\ssizz},inner sep=0]
            \node [align=center, img] {\includegraphics[width=\textwidth]{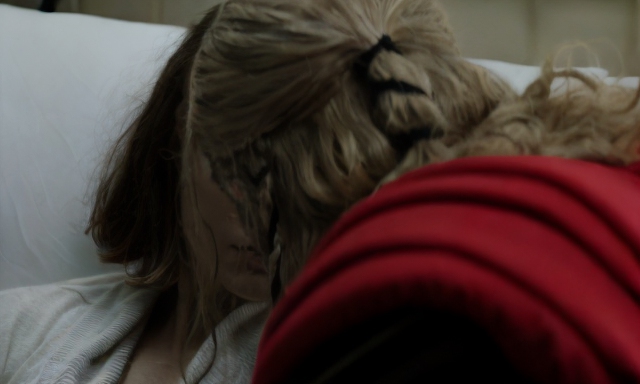}};
            \spy on \zoomseven in node [left] at \rebigtwo;
    	\end{tikzpicture}
      \end{subfigure}
    \begin{subfigure}{\depthWidth}
        \begin{tikzpicture}[spy using outlines={green,magnification=\ssmag,size=\ssizz},inner sep=0]
            \node [align=center, img] {\includegraphics[width=\textwidth]{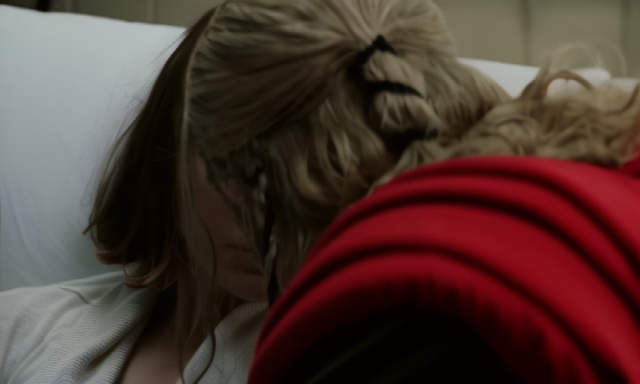}};
            \spy on \zoomseven in node [left] at \rebigtwo;
    	\end{tikzpicture}
      \end{subfigure}
    \begin{subfigure}{\depthWidth}
        \begin{tikzpicture}[spy using outlines={green,magnification=\ssmag,size=\ssizz},inner sep=0]
            \node [align=center, img] {\includegraphics[width=\textwidth]{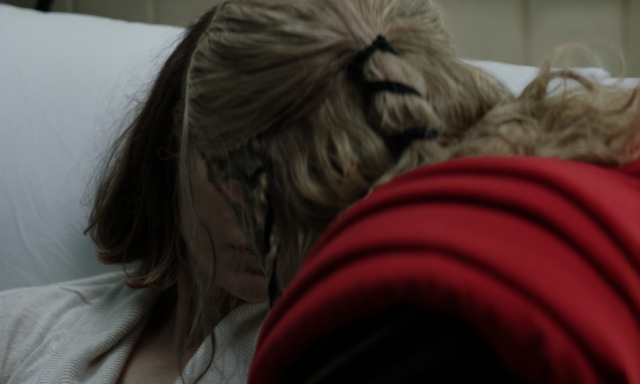}};
            \spy on \zoomseven in node [left] at \rebigtwo;
    	\end{tikzpicture}
      \end{subfigure}
    \end{subfigure}
    \begin{subfigure}{\linewidth}
    \centering
    \begin{subfigure}{\depthWidth}
        \begin{tikzpicture}[spy using outlines={green,magnification=\ssmag,size=\ssizz},inner sep=0]
            \node [align=center, img] {\includegraphics[width=\textwidth]{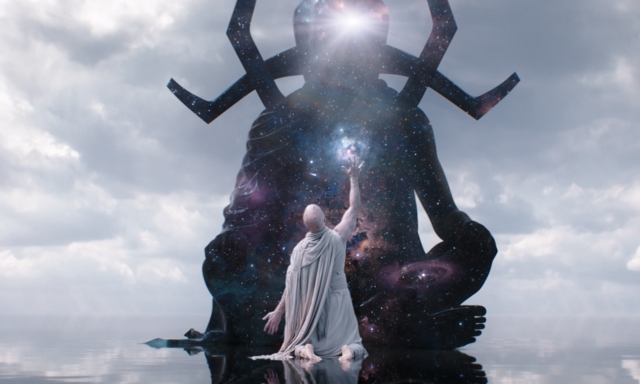}};
            \spy on \zoomeight in node [left] at \rebigone;
    	\end{tikzpicture}
    \end{subfigure}
    \begin{subfigure}{\depthWidth}
		\begin{tikzpicture}[spy using outlines={green,magnification=\ssmag,size=\ssizz},inner sep=0]
            \node [align=center, img] {\includegraphics[width=\textwidth]{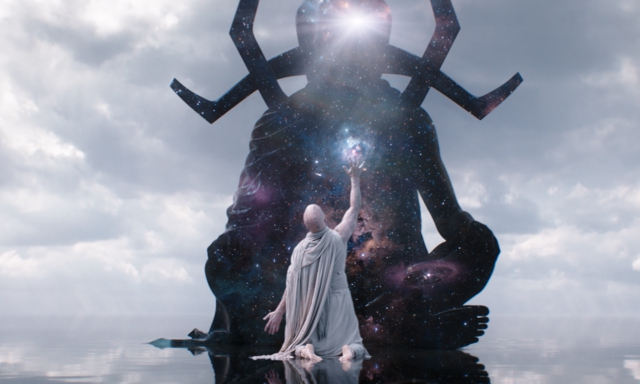}};
            \spy on \zoomeight in node [left] at \rebigone;
    	\end{tikzpicture}
    \end{subfigure}
    \begin{subfigure}{\depthWidth}
        \begin{tikzpicture}[spy using outlines={green,magnification=\ssmag,size=\ssizz},inner sep=0]
            \node [align=center, img] {\includegraphics[width=\textwidth]{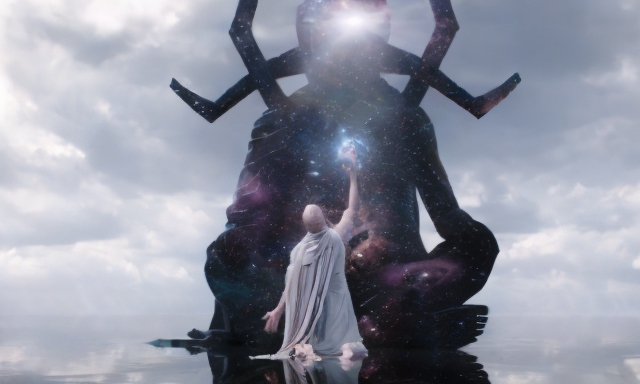}};
            \spy on \zoomeight in node [left] at \rebigone;
    	\end{tikzpicture}
      \end{subfigure}
    \begin{subfigure}{\depthWidth}
        \begin{tikzpicture}[spy using outlines={green,magnification=\ssmag,size=\ssizz},inner sep=0]
            \node [align=center, img] {\includegraphics[width=\textwidth]{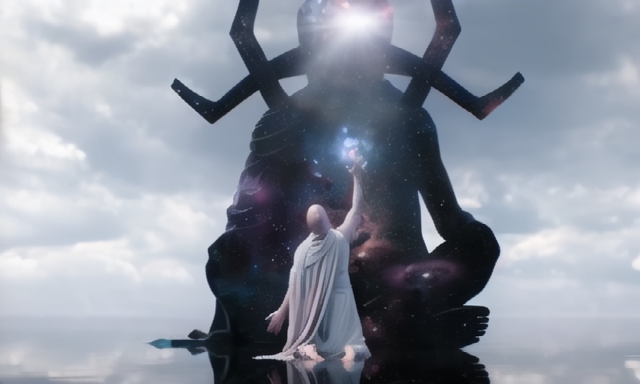}};
            \spy on \zoomeight in node [left] at \rebigone;
    	\end{tikzpicture}
      \end{subfigure}
    \begin{subfigure}{\depthWidth}
        \begin{tikzpicture}[spy using outlines={green,magnification=\ssmag,size=\ssizz},inner sep=0]
            \node [align=center, img] {\includegraphics[width=\textwidth]{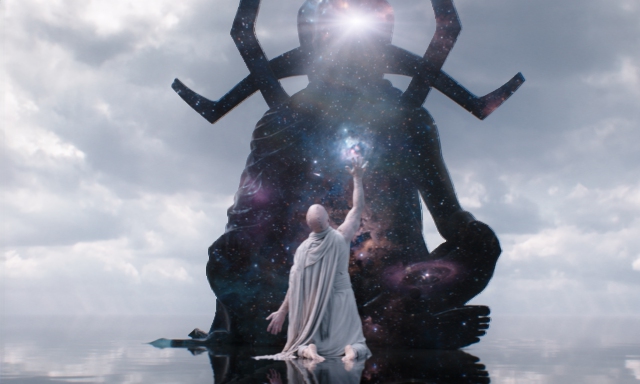}};
            \spy on \zoomeight in node [left] at \rebigone;
    	\end{tikzpicture}
      \end{subfigure}
    \end{subfigure}
    \begin{subfigure}{\linewidth}
    \centering
    \begin{subfigure}{\depthWidth}
        \begin{tikzpicture}[spy using outlines={green,magnification=\ssmag,size=\ssizz},inner sep=0]
            \node [align=center, img] {\includegraphics[width=\textwidth]{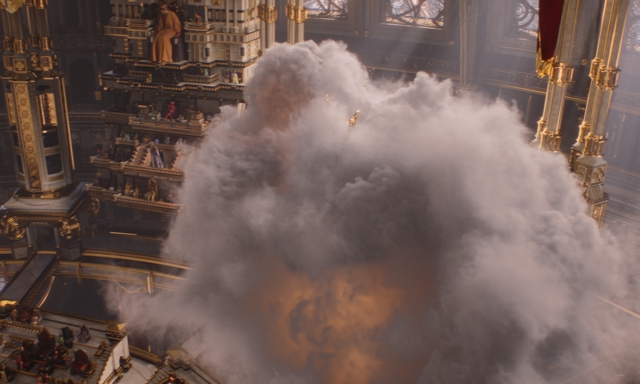}};
            \spy on \zoomone in node [left] at \rebigone;
    	\end{tikzpicture}
    \end{subfigure}
    \begin{subfigure}{\depthWidth}
		\begin{tikzpicture}[spy using outlines={green,magnification=\ssmag,size=\ssizz},inner sep=0]
            \node [align=center, img] {\includegraphics[width=\textwidth]{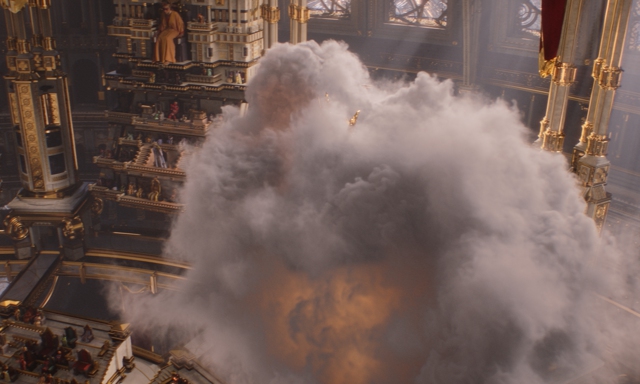}};
            \spy on \zoomone in node [left] at \rebigone;
    	\end{tikzpicture}
    \end{subfigure}
    \begin{subfigure}{\depthWidth}
        \begin{tikzpicture}[spy using outlines={green,magnification=\ssmag,size=\ssizz},inner sep=0]
            \node [align=center, img] {\includegraphics[width=\textwidth]{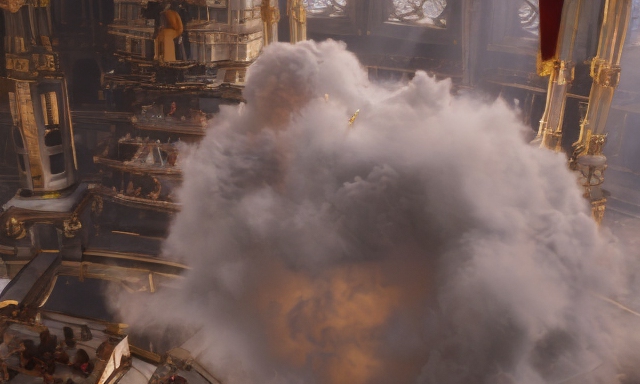}};
            \spy on \zoomone in node [left] at \rebigone;
    	\end{tikzpicture}
      \end{subfigure}
    \begin{subfigure}{\depthWidth}
        \begin{tikzpicture}[spy using outlines={green,magnification=\ssmag,size=\ssizz},inner sep=0]
            \node [align=center, img] {\includegraphics[width=\textwidth]{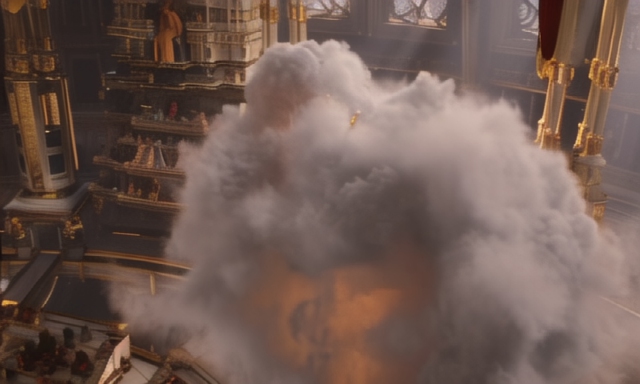}};
            \spy on \zoomone in node [left] at \rebigone;
    	\end{tikzpicture}
      \end{subfigure}
    \begin{subfigure}{\depthWidth}
        \begin{tikzpicture}[spy using outlines={green,magnification=\ssmag,size=\ssizz},inner sep=0]
            \node [align=center, img] {\includegraphics[width=\textwidth]{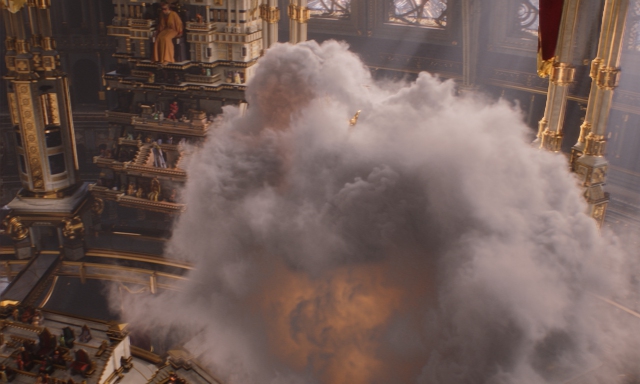}};
            \spy on \zoomone in node [left] at \rebigone;
    	\end{tikzpicture}
      \end{subfigure}
    \end{subfigure}
    \begin{subfigure}{\linewidth}
    \centering
    \begin{subfigure}{\depthWidth}
        \begin{tikzpicture}[spy using outlines={green,magnification=\ssmag,size=\ssizz},inner sep=0]
            \node [align=center, img] {\includegraphics[width=\textwidth]{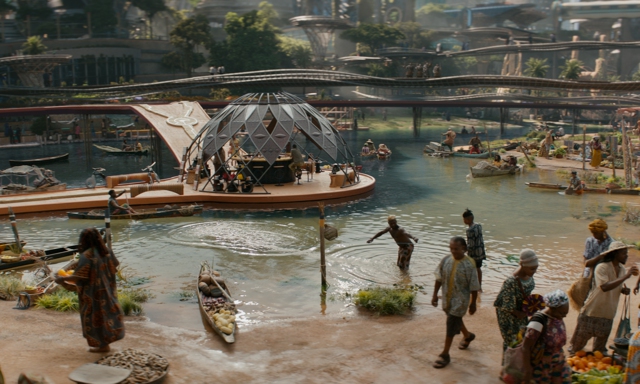}};
            \spy on \zoomtwo in node [left] at \rebigone;
    	\end{tikzpicture}
    \end{subfigure}
    \begin{subfigure}{\depthWidth}
		\begin{tikzpicture}[spy using outlines={green,magnification=\ssmag,size=\ssizz},inner sep=0]
            \node [align=center, img] {\includegraphics[width=\textwidth]{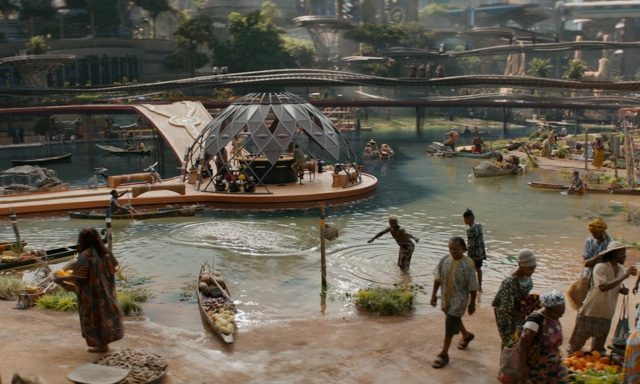}};
            \spy on \zoomtwo in node [left] at \rebigone;
    	\end{tikzpicture}
    \end{subfigure}
    \begin{subfigure}{\depthWidth}
        \begin{tikzpicture}[spy using outlines={green,magnification=\ssmag,size=\ssizz},inner sep=0]
            \node [align=center, img] {\includegraphics[width=\textwidth]{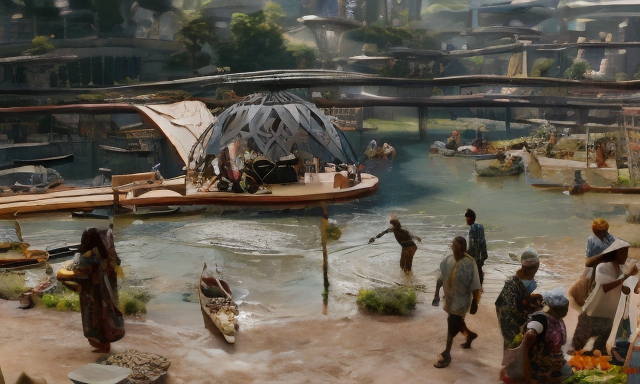}};
            \spy on \zoomtwo in node [left] at \rebigone;
    	\end{tikzpicture}
      \end{subfigure}
    \begin{subfigure}{\depthWidth}
        \begin{tikzpicture}[spy using outlines={green,magnification=\ssmag,size=\ssizz},inner sep=0]
            \node [align=center, img] {\includegraphics[width=\textwidth]{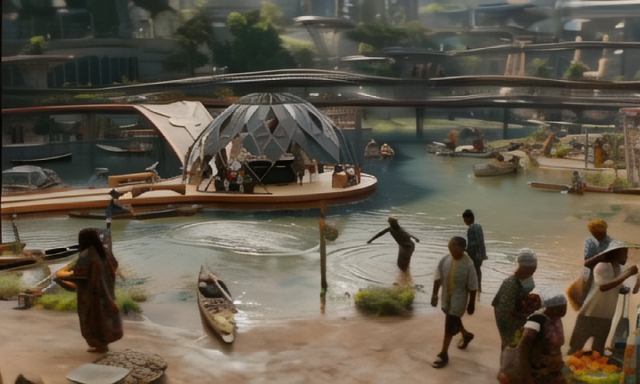}};
            \spy on \zoomtwo in node [left] at \rebigone;
    	\end{tikzpicture}
      \end{subfigure}
    \begin{subfigure}{\depthWidth}
        \begin{tikzpicture}[spy using outlines={green,magnification=\ssmag,size=\ssizz},inner sep=0]
            \node [align=center, img] {\includegraphics[width=\textwidth]{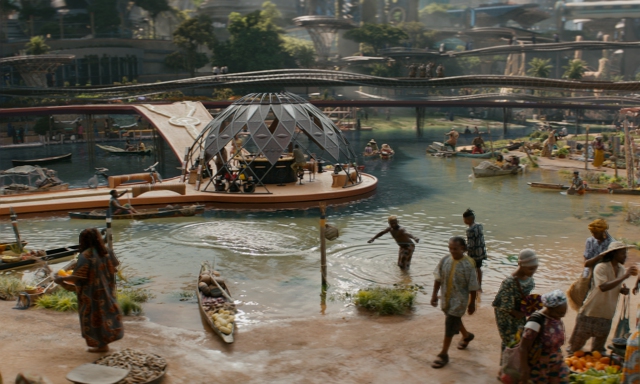}};
            \spy on \zoomtwo in node [left] at \rebigone;
    	\end{tikzpicture}
      \end{subfigure}
    \end{subfigure}
    \begin{subfigure}{\linewidth}
    \centering
    \begin{subfigure}{\depthWidth}
        \begin{tikzpicture}[spy using outlines={green,magnification=\ssmag,size=\ssizz},inner sep=0]
            \node [align=center, img] {\includegraphics[width=\textwidth]{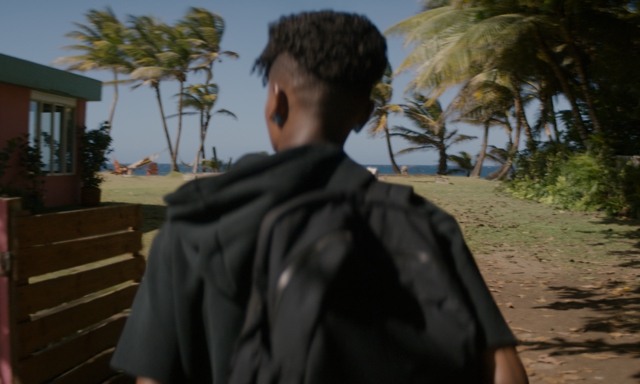}};
            \spy on \zoomthree in node [left] at \rebigone;
    	\end{tikzpicture}
    \end{subfigure}
    \begin{subfigure}{\depthWidth}
		\begin{tikzpicture}[spy using outlines={green,magnification=\ssmag,size=\ssizz},inner sep=0]
            \node [align=center, img] {\includegraphics[width=\textwidth]{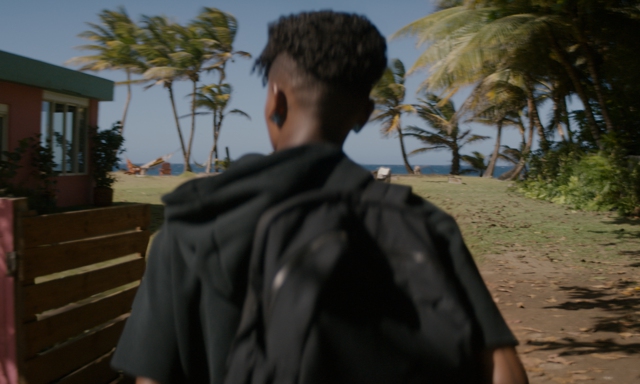}};
            \spy on \zoomthree in node [left] at \rebigone;
    	\end{tikzpicture}
    \end{subfigure}
    \begin{subfigure}{\depthWidth}
        \begin{tikzpicture}[spy using outlines={green,magnification=\ssmag,size=\ssizz},inner sep=0]
            \node [align=center, img] {\includegraphics[width=\textwidth]{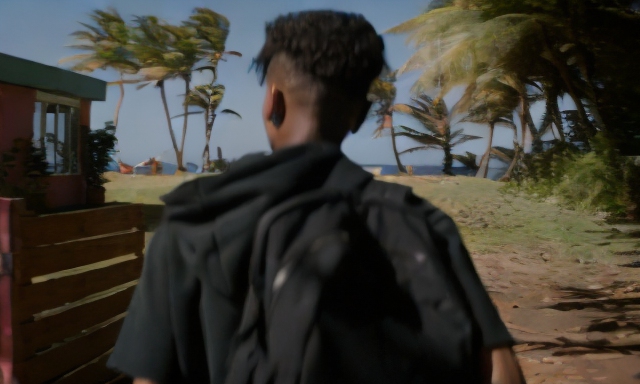}};
            \spy on \zoomthree in node [left] at \rebigone;
    	\end{tikzpicture}
      \end{subfigure}
    \begin{subfigure}{\depthWidth}
        \begin{tikzpicture}[spy using outlines={green,magnification=\ssmag,size=\ssizz},inner sep=0]
            \node [align=center, img] {\includegraphics[width=\textwidth]{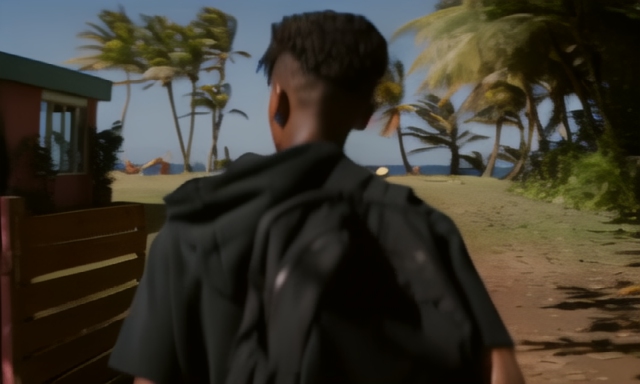}};
            \spy on \zoomthree in node [left] at \rebigone;
    	\end{tikzpicture}
      \end{subfigure}
    \begin{subfigure}{\depthWidth}
        \begin{tikzpicture}[spy using outlines={green,magnification=\ssmag,size=\ssizz},inner sep=0]
            \node [align=center, img] {\includegraphics[width=\textwidth]{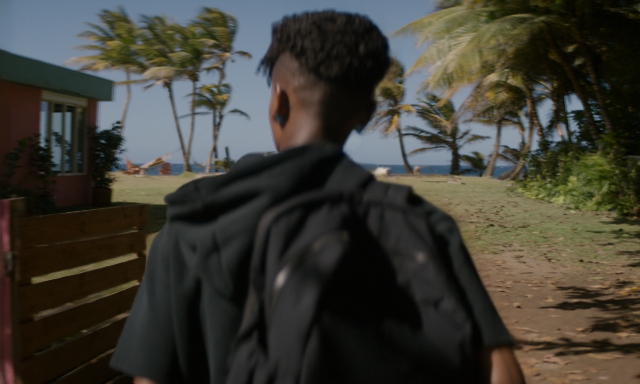}};
            \spy on \zoomthree in node [left] at \rebigone;
    	\end{tikzpicture}
      \end{subfigure}
    \end{subfigure}
    \begin{subfigure}{\linewidth}
    \centering
    \begin{subfigure}{\depthWidth}
        \begin{tikzpicture}[spy using outlines={green,magnification=\ssmag,size=\ssizz},inner sep=0]
            \node [align=center, img] {\includegraphics[width=\textwidth]{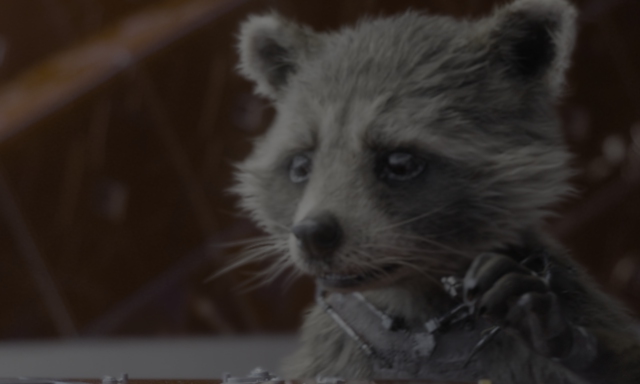}};
            \spy on \zoomfive in node [left] at \rebigone;
    	\end{tikzpicture}
    \end{subfigure}
    \begin{subfigure}{\depthWidth}
		\begin{tikzpicture}[spy using outlines={green,magnification=\ssmag,size=\ssizz},inner sep=0]
            \node [align=center, img] {\includegraphics[width=\textwidth]{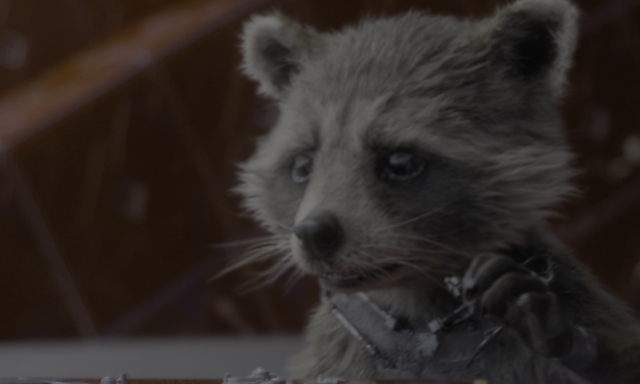}};
            \spy on \zoomfive in node [left] at \rebigone;
    	\end{tikzpicture}
    \end{subfigure}
    \begin{subfigure}{\depthWidth}
        \begin{tikzpicture}[spy using outlines={green,magnification=\ssmag,size=\ssizz},inner sep=0]
            \node [align=center, img] {\includegraphics[width=\textwidth]{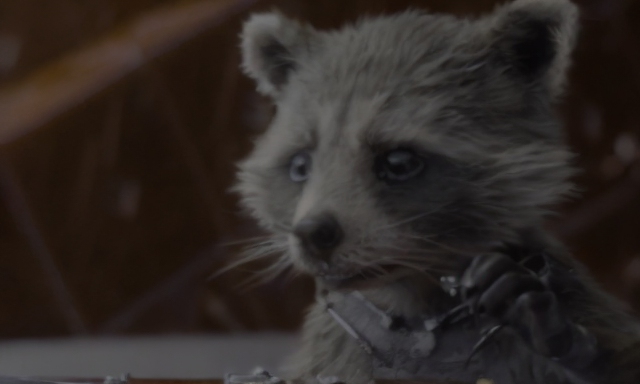}};
            \spy on \zoomfive in node [left] at \rebigone;
    	\end{tikzpicture}
      \end{subfigure}
    \begin{subfigure}{\depthWidth}
        \begin{tikzpicture}[spy using outlines={green,magnification=\ssmag,size=\ssizz},inner sep=0]
            \node [align=center, img] {\includegraphics[width=\textwidth]{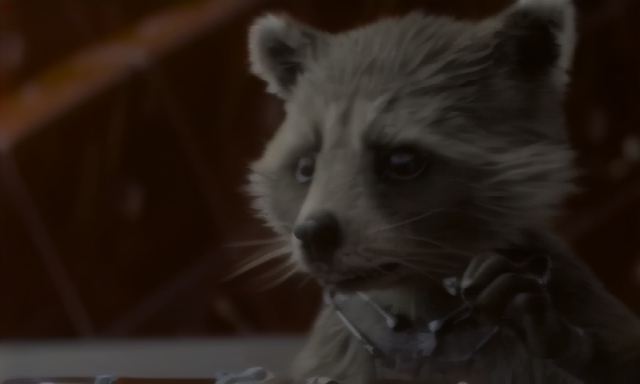}};
            \spy on \zoomfive in node [left] at \rebigone;
    	\end{tikzpicture}
      \end{subfigure}
    \begin{subfigure}{\depthWidth}
        \begin{tikzpicture}[spy using outlines={green,magnification=\ssmag,size=\ssizz},inner sep=0]
            \node [align=center, img] {\includegraphics[width=\textwidth]{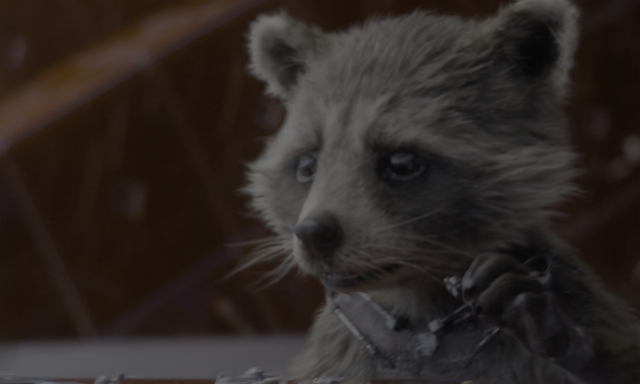}};
            \spy on \zoomfive in node [left] at \rebigone;
    	\end{tikzpicture}
      \end{subfigure}
    \end{subfigure}
    \begin{subfigure}{\linewidth}
    \centering
    \begin{subfigure}{\depthWidth}
        \begin{tikzpicture}[spy using outlines={green,magnification=\ssmag,size=\ssizz},inner sep=0]
            \node [align=center, img] {\includegraphics[width=\textwidth]{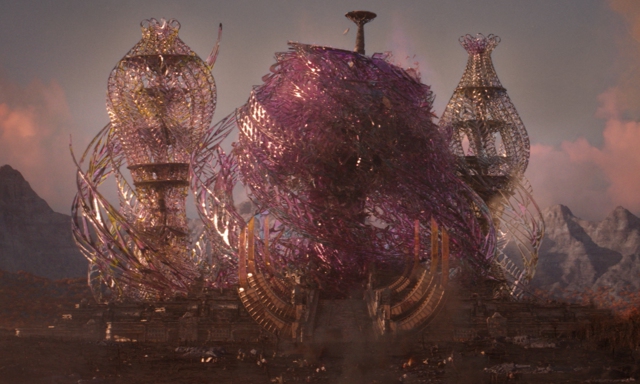}};
            \spy on \zoomten in node [left] at \rebigone;
    	\end{tikzpicture}
        \caption*{Left View}
    \end{subfigure}
    \begin{subfigure}{\depthWidth}
		\begin{tikzpicture}[spy using outlines={green,magnification=\ssmag,size=\ssizz},inner sep=0]
            \node [align=center, img] {\includegraphics[width=\textwidth]{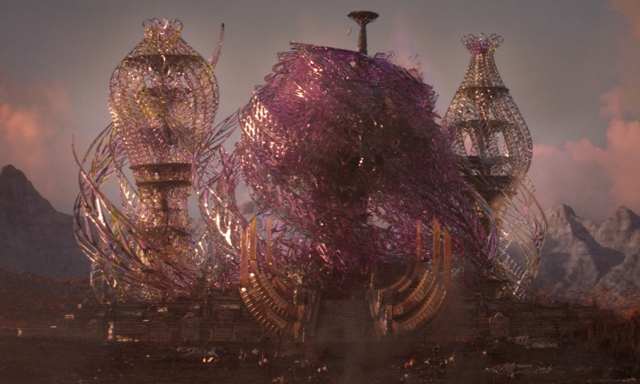}};
            \spy on \zoomten in node [left] at \rebigone;
    	\end{tikzpicture}
        \caption*{Right View}
    \end{subfigure}
    \begin{subfigure}{\depthWidth}
        \begin{tikzpicture}[spy using outlines={green,magnification=\ssmag,size=\ssizz},inner sep=0]
            \node [align=center, img] {\includegraphics[width=\textwidth]{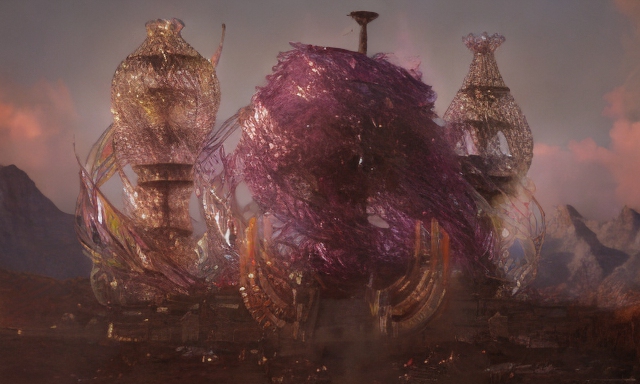}};
            \spy on \zoomten in node [left] at \rebigone;
    	\end{tikzpicture}
        \caption*{Mono2Stereo}
      \end{subfigure}
    \begin{subfigure}{\depthWidth}
        \begin{tikzpicture}[spy using outlines={green,magnification=\ssmag,size=\ssizz},inner sep=0]
            \node [align=center, img] {\includegraphics[width=\textwidth]{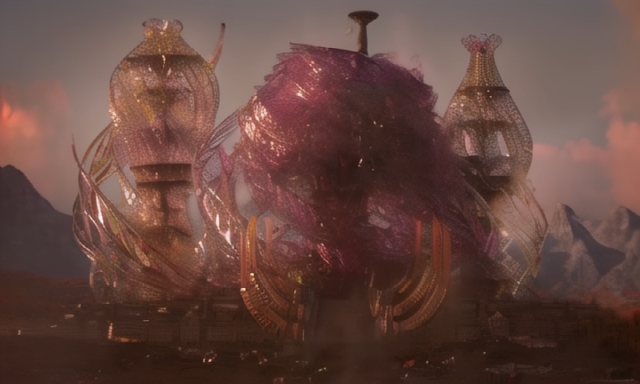}};
            \spy on \zoomten in node [left] at \rebigone;
    	\end{tikzpicture}
        \caption*{StereoCrafter}
      \end{subfigure}
    \begin{subfigure}{\depthWidth}
        \begin{tikzpicture}[spy using outlines={green,magnification=\ssmag,size=\ssizz},inner sep=0]
            \node [align=center, img] {\includegraphics[width=\textwidth]{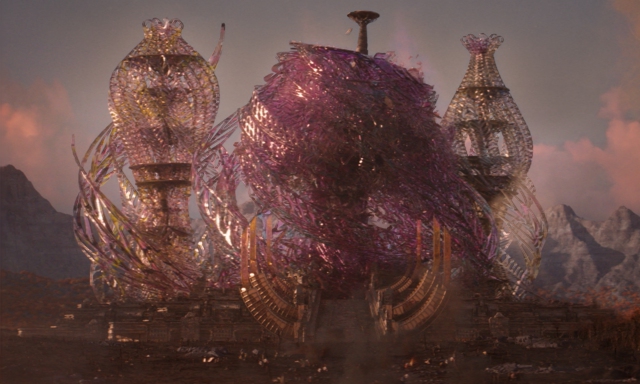}};
            \spy on \zoomten in node [left] at \rebigone;
    	\end{tikzpicture}
        \caption*{Ours}
      \end{subfigure}
    \end{subfigure}
    \caption{\textbf{Qualitative comparison of stereo conversion} on the Marvel-10K dataset.}
    \label{fig:supp-stereo-visuals-marvel}
\end{figure*}

%% file: figs/supp/fig-nvs_visuals_aim.tex
\def\imgWidth{0.32\linewidth} %
\def\depthWidth{0.19\linewidth} %
\def\pointWidth{0.24\linewidth} %
\def\scc{(-1.9,-1.4)}

\def\rebigone{(-0.5, -0.55)} %
\def\rebigtwo{(1.6, -0.55)} %

\def\zoomone{(-0.3,0.75)} %
\def\zoomoneori{(-0.4,0.75)} %
\def\zoomonerecam{(-0.1,0.75)} %

\def\zoomtwo{(-0.8,-0.3)} %
\def\zoomtwoori{(-0.6,-0.25)} %

\def\zoomthree{(0,-0.9)} %
\def\zoomthreeori{(0,-0.8)} %

\def\zoomfour{(-0.65,0.75)} %
\def\zoomfourori{(-0.5,0.7)} %

\def\zoomfive{(-1.3,-0.95)} %
\def\zoomfiveori{(-1.2,-0.85)} %

\def\zoomsix{(0.15,0.3)} %
\def\zoomsixori{(0,0.25)} %

\def\zoomseven{(-0.7,-0.9)} %
\def\zoomsevenori{(-0.65,-0.8)} %

\def\zoomeight{(-0.4,0.2)} %
\def\zoomeightori{(-0.3,0.2)} %

\def\zoomnine{(-0.45,0.3)} %
\def\zoomnineori{(-0.3,0.25)} %

\def\ssizz{1.1cm} %
\def\ssmag{3}

\begin{figure*}[t]
\centering
\tikzstyle{img} = [rectangle, minimum width=\imgWidth]
    \centering
    
    \begin{subfigure}{\linewidth}
    \centering
    \begin{subfigure}{\depthWidth}
        \begin{tikzpicture}[spy using outlines={green,magnification=\ssmag,size=\ssizz},inner sep=0]
            \node [align=center, img] {\includegraphics[width=\textwidth]{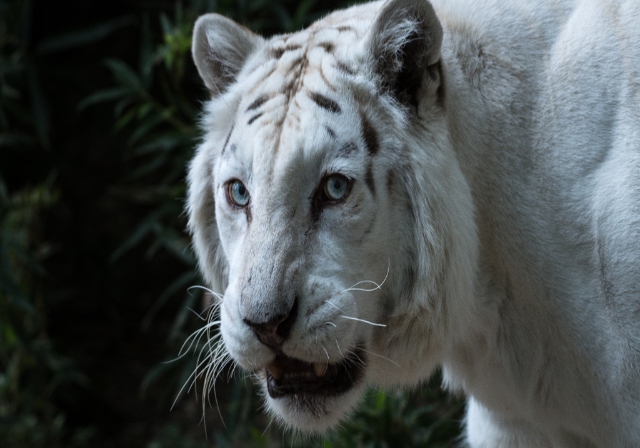}};
            \spy on \zoomsevenori in node [left] at \rebigtwo;
    	\end{tikzpicture}
    \end{subfigure}
    \begin{subfigure}{\depthWidth}
		\begin{tikzpicture}[spy using outlines={green,magnification=\ssmag,size=\ssizz},inner sep=0]
            \node [align=center, img] {\includegraphics[width=\textwidth]{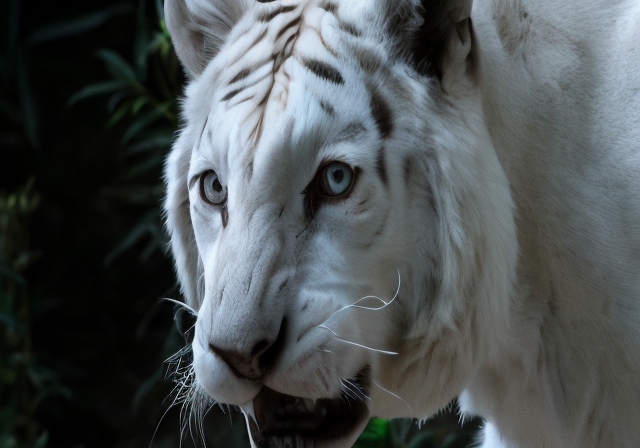}};
            \spy on \zoomseven in node [left] at \rebigtwo;
    	\end{tikzpicture}
    \end{subfigure}
    \begin{subfigure}{\depthWidth}
        \begin{tikzpicture}[spy using outlines={green,magnification=\ssmag,size=\ssizz},inner sep=0]
            \node [align=center, img] {\includegraphics[width=\textwidth]{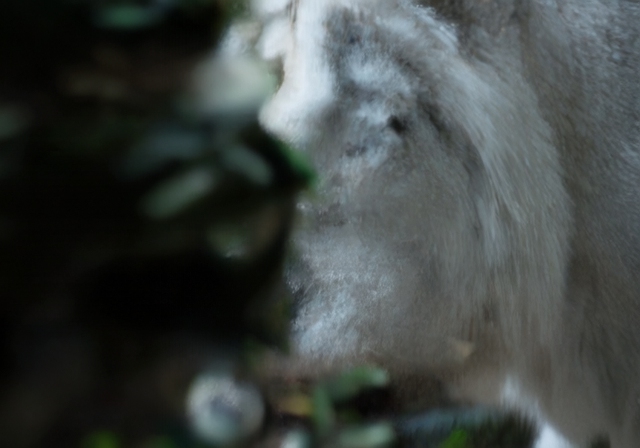}};
            \spy on \zoomseven in node [left] at \rebigtwo;
    	\end{tikzpicture}
      \end{subfigure}
    \begin{subfigure}{\depthWidth}
        \begin{tikzpicture}[spy using outlines={green,magnification=\ssmag,size=\ssizz},inner sep=0]
            \node [align=center, img] {\includegraphics[width=\textwidth]{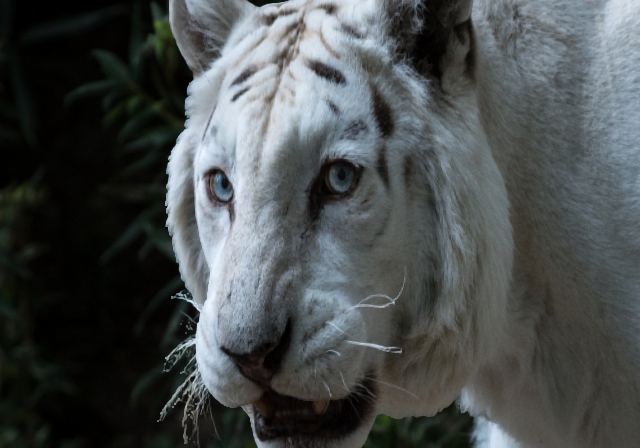}};
            \spy on \zoomseven in node [left] at \rebigtwo;
    	\end{tikzpicture}
      \end{subfigure}
    \begin{subfigure}{\depthWidth}
        \begin{tikzpicture}[spy using outlines={green,magnification=\ssmag,size=\ssizz},inner sep=0]
            \node [align=center, img] {\includegraphics[width=\textwidth]{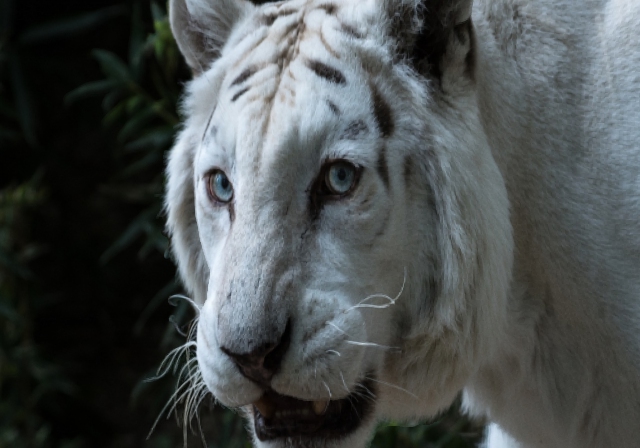}};
            \spy on \zoomseven in node [left] at \rebigtwo;
    	\end{tikzpicture}
      \end{subfigure}
    \end{subfigure}
    \begin{subfigure}{\linewidth}
    \centering
    \begin{subfigure}{\depthWidth}
        \begin{tikzpicture}[spy using outlines={green,magnification=\ssmag,size=\ssizz},inner sep=0]
            \node [align=center, img] {\includegraphics[width=\textwidth]{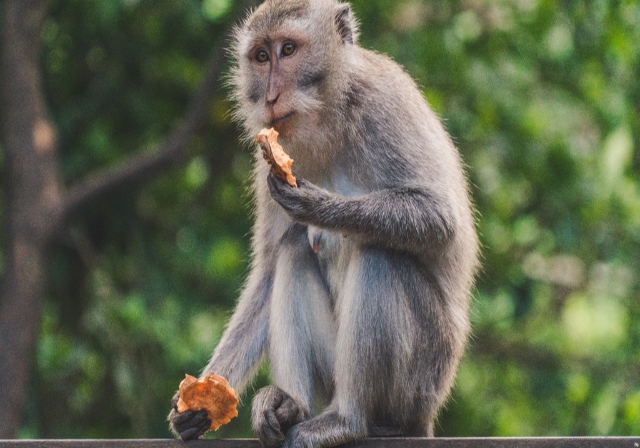}};
            \spy on \zoomoneori in node [left] at \rebigone;
    	\end{tikzpicture}
    \end{subfigure}
    \begin{subfigure}{\depthWidth}
		\begin{tikzpicture}[spy using outlines={green,magnification=\ssmag,size=\ssizz},inner sep=0]
            \node [align=center, img] {\includegraphics[width=\textwidth]{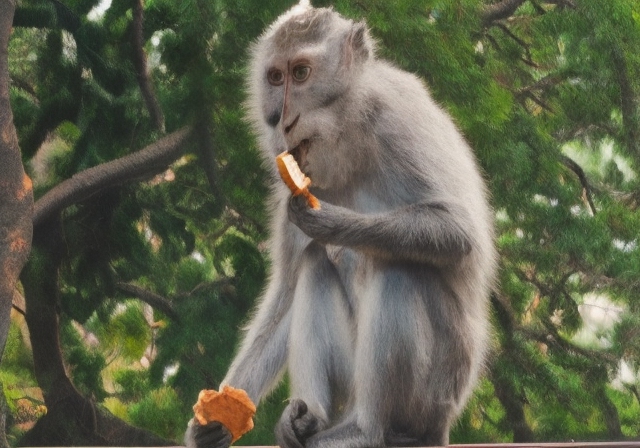}};
            \spy on \zoomone in node [left] at \rebigone;
    	\end{tikzpicture}
    \end{subfigure}
    \begin{subfigure}{\depthWidth}
        \begin{tikzpicture}[spy using outlines={green,magnification=\ssmag,size=\ssizz},inner sep=0]
            \node [align=center, img] {\includegraphics[width=\textwidth]{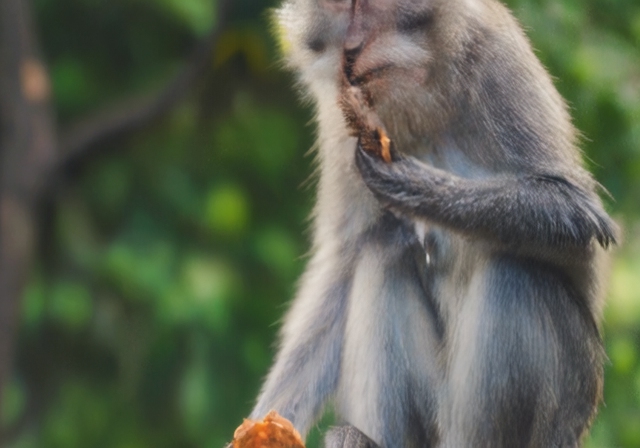}};
            \spy on \zoomonerecam in node [left] at \rebigone;
    	\end{tikzpicture}
      \end{subfigure}
    \begin{subfigure}{\depthWidth}
        \begin{tikzpicture}[spy using outlines={green,magnification=\ssmag,size=\ssizz},inner sep=0]
            \node [align=center, img] {\includegraphics[width=\textwidth]{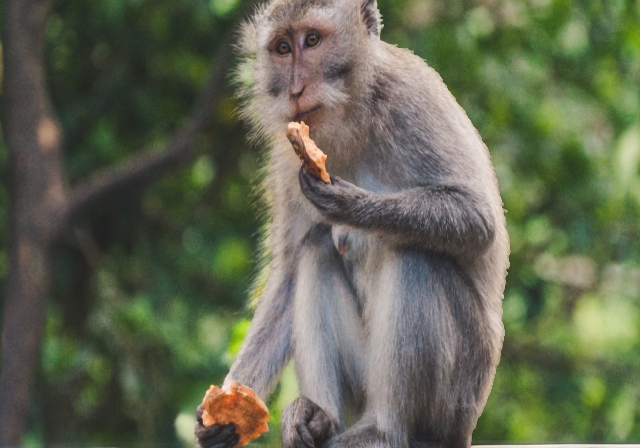}};
            \spy on \zoomone in node [left] at \rebigone;
    	\end{tikzpicture}
      \end{subfigure}
    \begin{subfigure}{\depthWidth}
        \begin{tikzpicture}[spy using outlines={green,magnification=\ssmag,size=\ssizz},inner sep=0]
            \node [align=center, img] {\includegraphics[width=\textwidth]{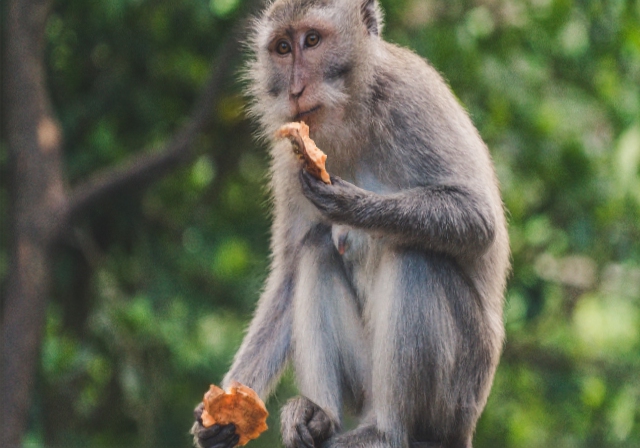}};
            \spy on \zoomone in node [left] at \rebigone;
    	\end{tikzpicture}
      \end{subfigure}
    \end{subfigure}
    \begin{subfigure}{\linewidth}
    \centering
    \begin{subfigure}{\depthWidth}
        \begin{tikzpicture}[spy using outlines={green,magnification=\ssmag,size=\ssizz},inner sep=0]
            \node [align=center, img] {\includegraphics[width=\textwidth]{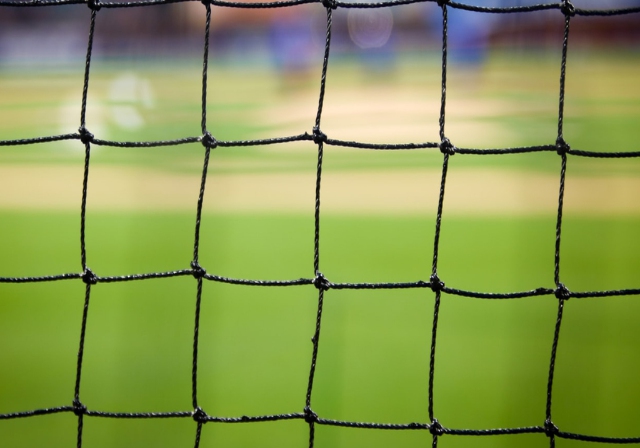}};
            \spy on \zoomtwoori in node [left] at \rebigtwo;
    	\end{tikzpicture}
    \end{subfigure}
    \begin{subfigure}{\depthWidth}
		\begin{tikzpicture}[spy using outlines={green,magnification=\ssmag,size=\ssizz},inner sep=0]
            \node [align=center, img] {\includegraphics[width=\textwidth]{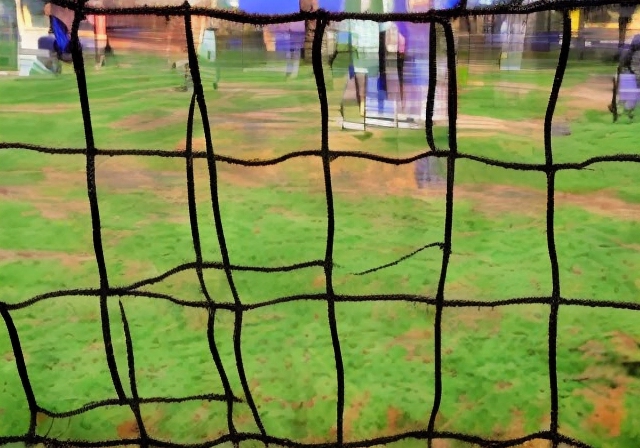}};
            \spy on \zoomtwo in node [left] at \rebigtwo;
    	\end{tikzpicture}
    \end{subfigure}
    \begin{subfigure}{\depthWidth}
        \begin{tikzpicture}[spy using outlines={green,magnification=\ssmag,size=\ssizz},inner sep=0]
            \node [align=center, img] {\includegraphics[width=\textwidth]{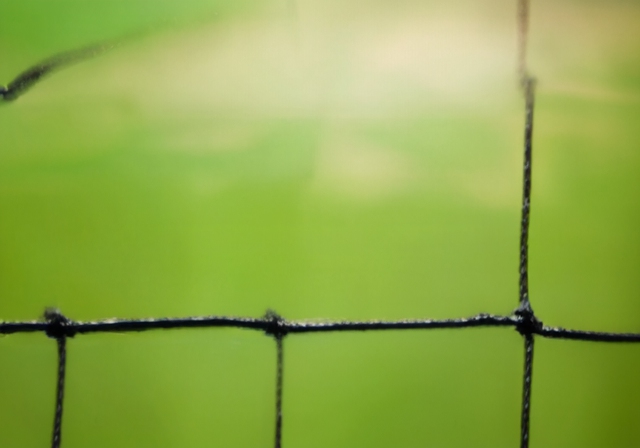}};
            \spy on \zoomtwo in node [left] at \rebigtwo;
    	\end{tikzpicture}
      \end{subfigure}
    \begin{subfigure}{\depthWidth}
        \begin{tikzpicture}[spy using outlines={green,magnification=\ssmag,size=\ssizz},inner sep=0]
            \node [align=center, img] {\includegraphics[width=\textwidth]{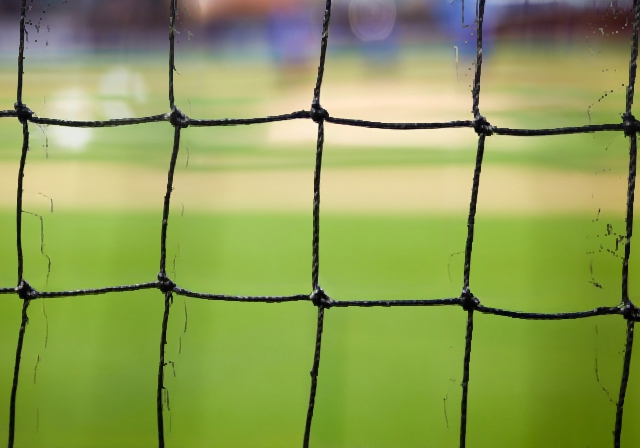}};
            \spy on \zoomtwo in node [left] at \rebigtwo;
    	\end{tikzpicture}
      \end{subfigure}
    \begin{subfigure}{\depthWidth}
        \begin{tikzpicture}[spy using outlines={green,magnification=\ssmag,size=\ssizz},inner sep=0]
            \node [align=center, img] {\includegraphics[width=\textwidth]{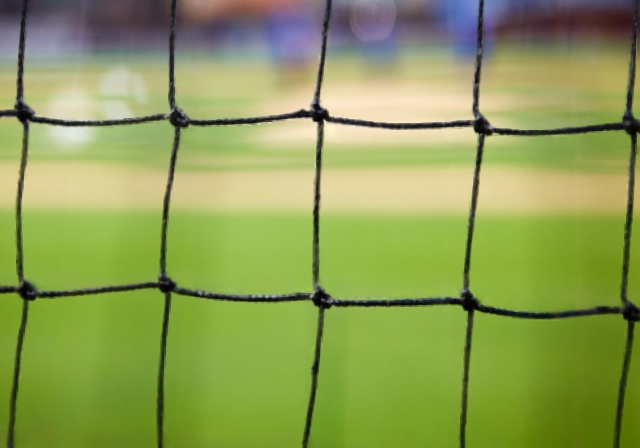}};
            \spy on \zoomtwo in node [left] at \rebigtwo;
    	\end{tikzpicture}
      \end{subfigure}
    \end{subfigure}
    \begin{subfigure}{\linewidth}
    \centering
    \begin{subfigure}{\depthWidth}
        \begin{tikzpicture}[spy using outlines={green,magnification=\ssmag,size=\ssizz},inner sep=0]
            \node [align=center, img] {\includegraphics[width=\textwidth]{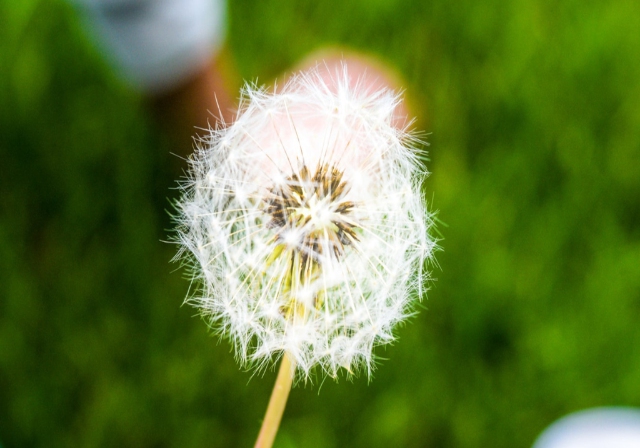}};
            \spy on \zoomthreeori in node [left] at \rebigone;
    	\end{tikzpicture}
    \end{subfigure}
    \begin{subfigure}{\depthWidth}
		\begin{tikzpicture}[spy using outlines={green,magnification=\ssmag,size=\ssizz},inner sep=0]
            \node [align=center, img] {\includegraphics[width=\textwidth]{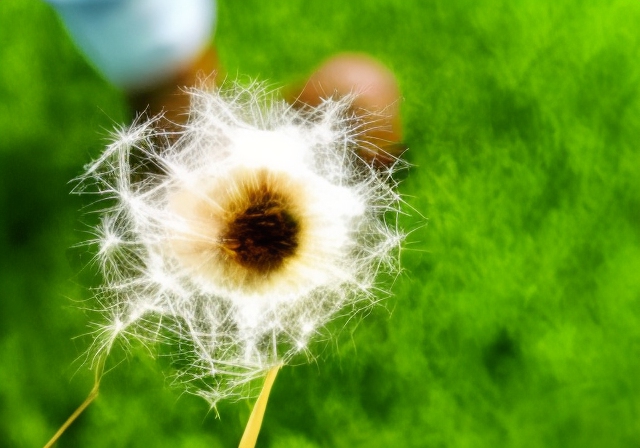}};
            \spy on \zoomthree in node [left] at \rebigone;
    	\end{tikzpicture}
    \end{subfigure}
    \begin{subfigure}{\depthWidth}
        \begin{tikzpicture}[spy using outlines={green,magnification=\ssmag,size=\ssizz},inner sep=0]
            \node [align=center, img] {\includegraphics[width=\textwidth]{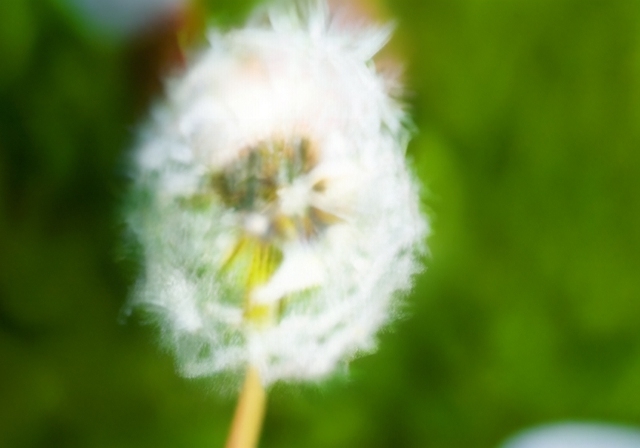}};
            \spy on \zoomthree in node [left] at \rebigone;
    	\end{tikzpicture}
      \end{subfigure}
    \begin{subfigure}{\depthWidth}
        \begin{tikzpicture}[spy using outlines={green,magnification=\ssmag,size=\ssizz},inner sep=0]
            \node [align=center, img] {\includegraphics[width=\textwidth]{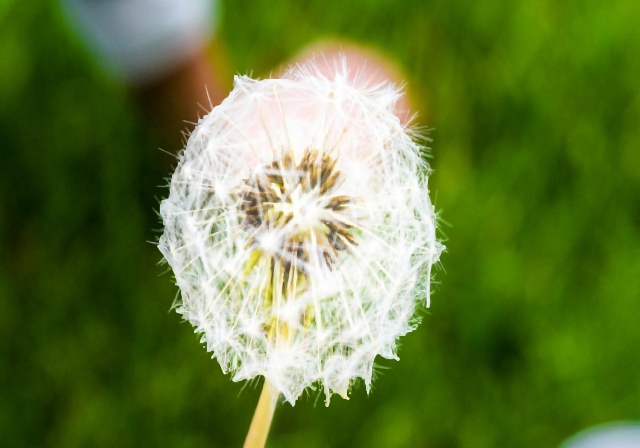}};
            \spy on \zoomthree in node [left] at \rebigone;
    	\end{tikzpicture}
      \end{subfigure}
    \begin{subfigure}{\depthWidth}
        \begin{tikzpicture}[spy using outlines={green,magnification=\ssmag,size=\ssizz},inner sep=0]
            \node [align=center, img] {\includegraphics[width=\textwidth]{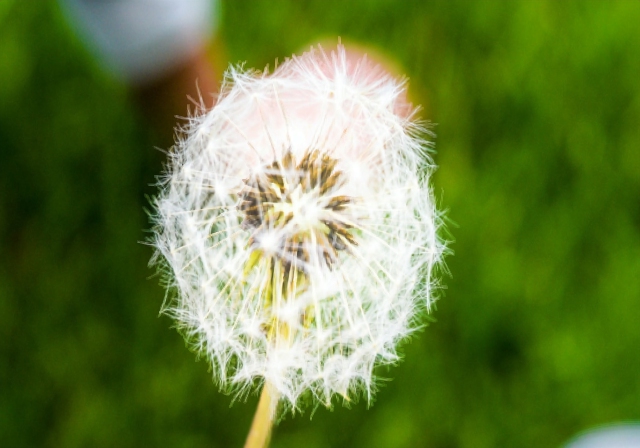}};
            \spy on \zoomthree in node [left] at \rebigone;
    	\end{tikzpicture}
      \end{subfigure}
    \end{subfigure}
    \begin{subfigure}{\linewidth}
    \centering
    \begin{subfigure}{\depthWidth}
        \begin{tikzpicture}[spy using outlines={green,magnification=\ssmag,size=\ssizz},inner sep=0]
            \node [align=center, img] {\includegraphics[width=\textwidth]{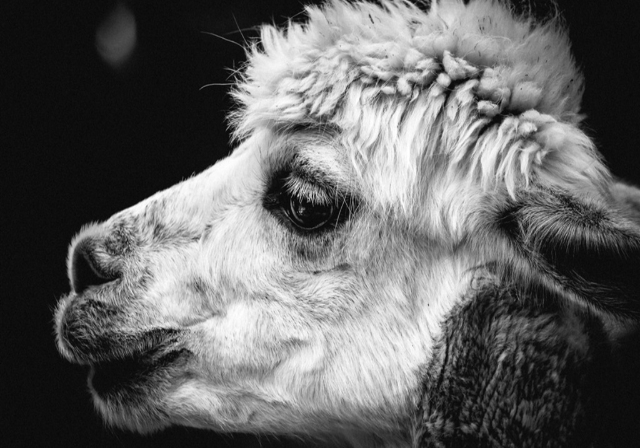}};
            \spy on \zoomfourori in node [left] at \rebigtwo;
    	\end{tikzpicture}
    \end{subfigure}
    \begin{subfigure}{\depthWidth}
		\begin{tikzpicture}[spy using outlines={green,magnification=\ssmag,size=\ssizz},inner sep=0]
            \node [align=center, img] {\includegraphics[width=\textwidth]{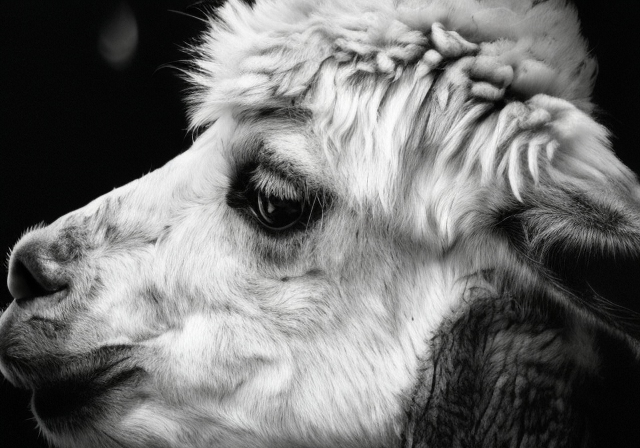}};
            \spy on \zoomfour in node [left] at \rebigtwo;
    	\end{tikzpicture}
    \end{subfigure}
    \begin{subfigure}{\depthWidth}
        \begin{tikzpicture}[spy using outlines={green,magnification=\ssmag,size=\ssizz},inner sep=0]
            \node [align=center, img] {\includegraphics[width=\textwidth]{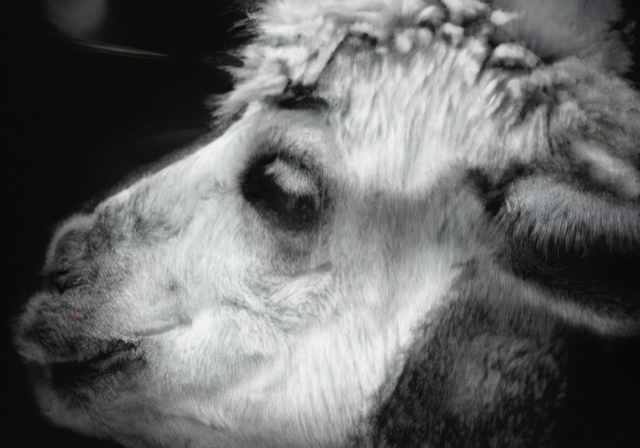}};
            \spy on \zoomfour in node [left] at \rebigtwo;
    	\end{tikzpicture}
      \end{subfigure}
    \begin{subfigure}{\depthWidth}
        \begin{tikzpicture}[spy using outlines={green,magnification=\ssmag,size=\ssizz},inner sep=0]
            \node [align=center, img] {\includegraphics[width=\textwidth]{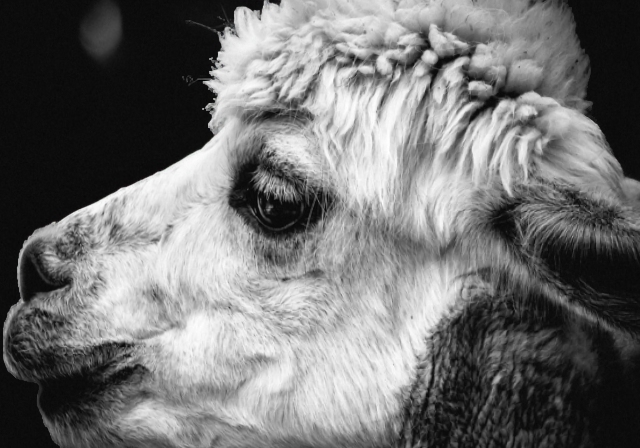}};
            \spy on \zoomfour in node [left] at \rebigtwo;
    	\end{tikzpicture}
      \end{subfigure}
    \begin{subfigure}{\depthWidth}
        \begin{tikzpicture}[spy using outlines={green,magnification=\ssmag,size=\ssizz},inner sep=0]
            \node [align=center, img] {\includegraphics[width=\textwidth]{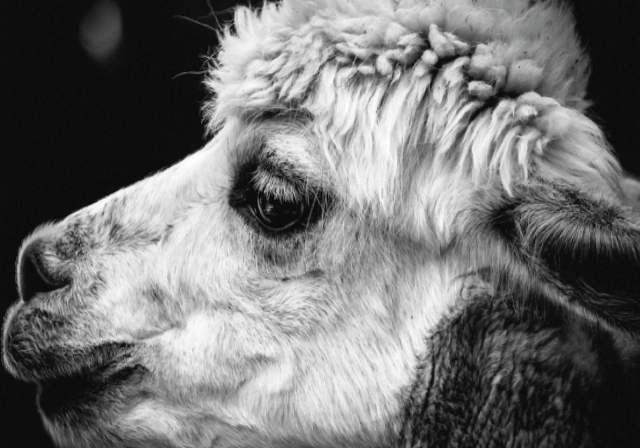}};
            \spy on \zoomfour in node [left] at \rebigtwo;
    	\end{tikzpicture}
      \end{subfigure}
    \end{subfigure}
    \begin{subfigure}{\linewidth}
    \centering
    \begin{subfigure}{\depthWidth}
        \begin{tikzpicture}[spy using outlines={green,magnification=\ssmag,size=\ssizz},inner sep=0]
            \node [align=center, img] {\includegraphics[width=\textwidth]{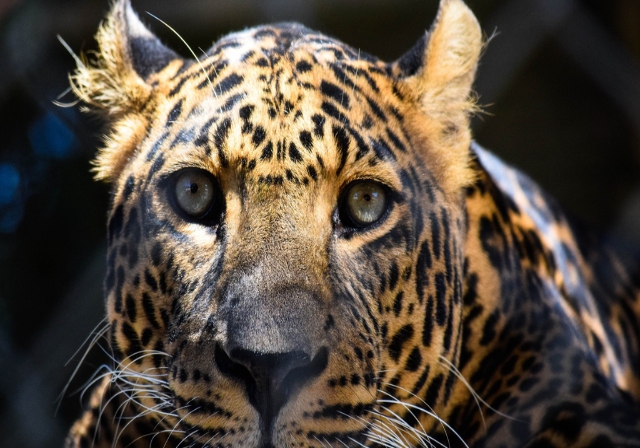}};
            \spy on \zoomfiveori in node [left] at \rebigtwo;
    	\end{tikzpicture}
    \end{subfigure}
    \begin{subfigure}{\depthWidth}
		\begin{tikzpicture}[spy using outlines={green,magnification=\ssmag,size=\ssizz},inner sep=0]
            \node [align=center, img] {\includegraphics[width=\textwidth]{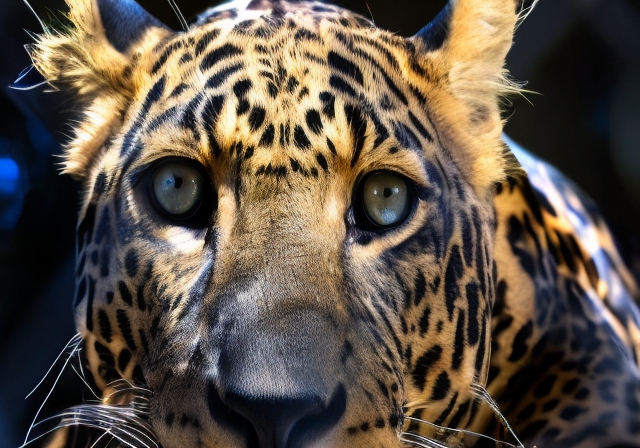}};
            \spy on \zoomfive in node [left] at \rebigtwo;
    	\end{tikzpicture}
    \end{subfigure}
    \begin{subfigure}{\depthWidth}
        \begin{tikzpicture}[spy using outlines={green,magnification=\ssmag,size=\ssizz},inner sep=0]
            \node [align=center, img] {\includegraphics[width=\textwidth]{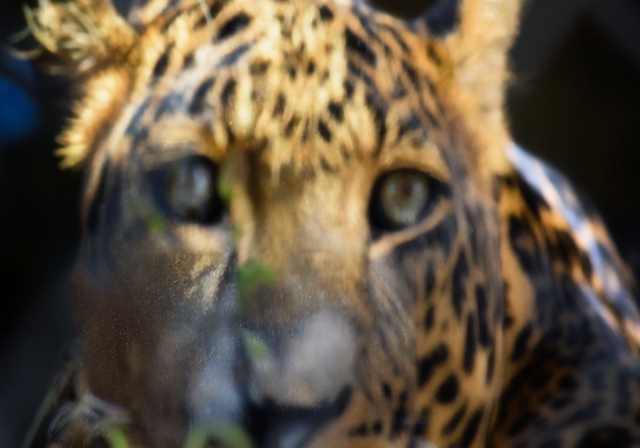}};
            \spy on \zoomfive in node [left] at \rebigtwo;
    	\end{tikzpicture}
      \end{subfigure}
    \begin{subfigure}{\depthWidth}
        \begin{tikzpicture}[spy using outlines={green,magnification=\ssmag,size=\ssizz},inner sep=0]
            \node [align=center, img] {\includegraphics[width=\textwidth]{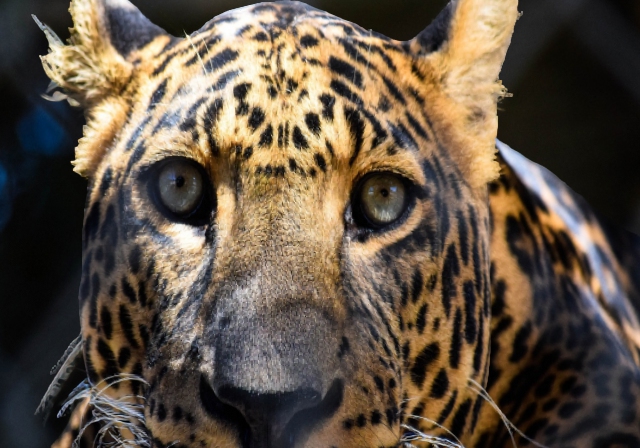}};
            \spy on \zoomfive in node [left] at \rebigtwo;
    	\end{tikzpicture}
      \end{subfigure}
    \begin{subfigure}{\depthWidth}
        \begin{tikzpicture}[spy using outlines={green,magnification=\ssmag,size=\ssizz},inner sep=0]
            \node [align=center, img] {\includegraphics[width=\textwidth]{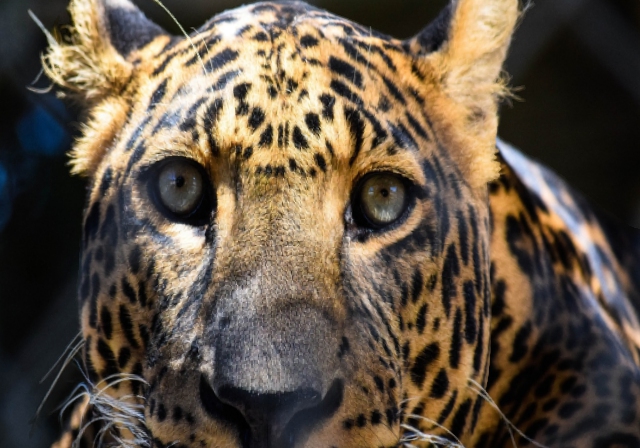}};
            \spy on \zoomfive in node [left] at \rebigtwo;
    	\end{tikzpicture}
      \end{subfigure}
    \end{subfigure}
    \begin{subfigure}{\linewidth}
    \centering
    \begin{subfigure}{\depthWidth}
        \begin{tikzpicture}[spy using outlines={green,magnification=\ssmag,size=\ssizz},inner sep=0]
            \node [align=center, img] {\includegraphics[width=\textwidth]{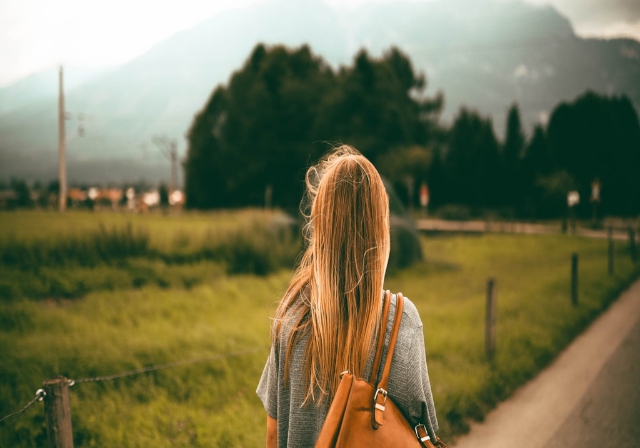}};
            \spy on \zoomsixori in node [left] at \rebigone;
    	\end{tikzpicture}
    \end{subfigure}
    \begin{subfigure}{\depthWidth}
		\begin{tikzpicture}[spy using outlines={green,magnification=\ssmag,size=\ssizz},inner sep=0]
            \node [align=center, img] {\includegraphics[width=\textwidth]{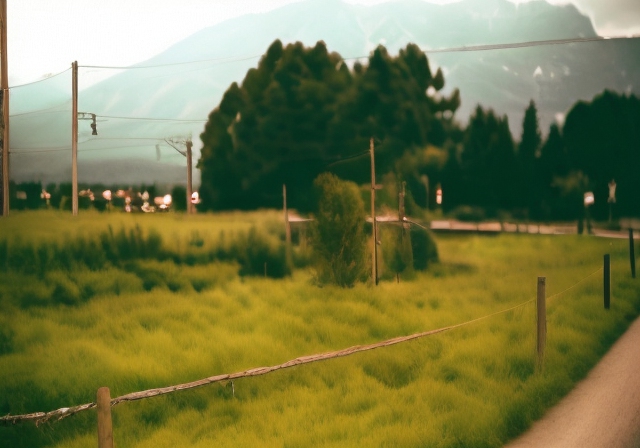}};
            \spy on \zoomsix in node [left] at \rebigone;
    	\end{tikzpicture}
    \end{subfigure}
    \begin{subfigure}{\depthWidth}
        \begin{tikzpicture}[spy using outlines={green,magnification=\ssmag,size=\ssizz},inner sep=0]
            \node [align=center, img] {\includegraphics[width=\textwidth]{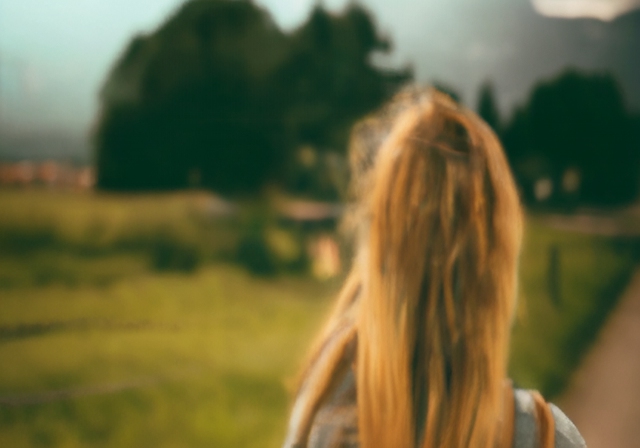}};
            \spy on \zoomsix in node [left] at \rebigone;
    	\end{tikzpicture}
      \end{subfigure}
    \begin{subfigure}{\depthWidth}
        \begin{tikzpicture}[spy using outlines={green,magnification=\ssmag,size=\ssizz},inner sep=0]
            \node [align=center, img] {\includegraphics[width=\textwidth]{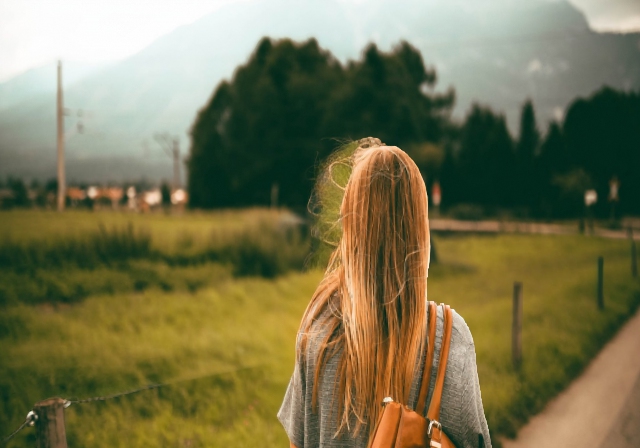}};
            \spy on \zoomsix in node [left] at \rebigone;
    	\end{tikzpicture}
      \end{subfigure}
    \begin{subfigure}{\depthWidth}
        \begin{tikzpicture}[spy using outlines={green,magnification=\ssmag,size=\ssizz},inner sep=0]
            \node [align=center, img] {\includegraphics[width=\textwidth]{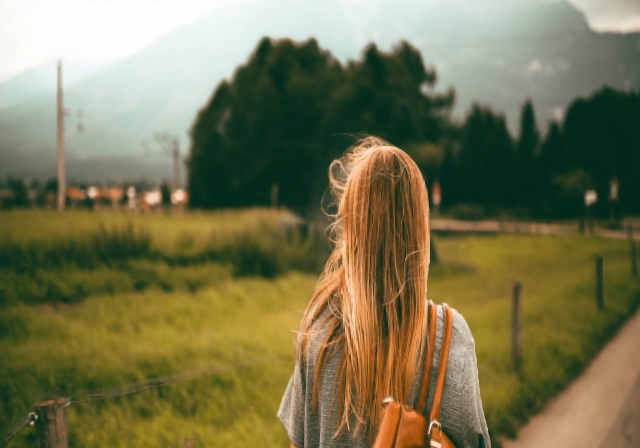}};
            \spy on \zoomsix in node [left] at \rebigone;
    	\end{tikzpicture}
      \end{subfigure}
    \end{subfigure}
    \begin{subfigure}{\linewidth}
    \centering
    \begin{subfigure}{\depthWidth}
        \begin{tikzpicture}[spy using outlines={green,magnification=\ssmag,size=\ssizz},inner sep=0]
            \node [align=center, img] {\includegraphics[width=\textwidth]{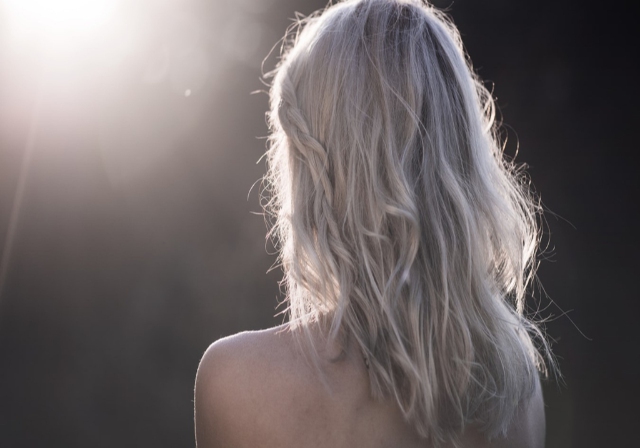}};
            \spy on \zoomeightori in node [left] at \rebigone;
    	\end{tikzpicture}
    \end{subfigure}
    \begin{subfigure}{\depthWidth}
		\begin{tikzpicture}[spy using outlines={green,magnification=\ssmag,size=\ssizz},inner sep=0]
            \node [align=center, img] {\includegraphics[width=\textwidth]{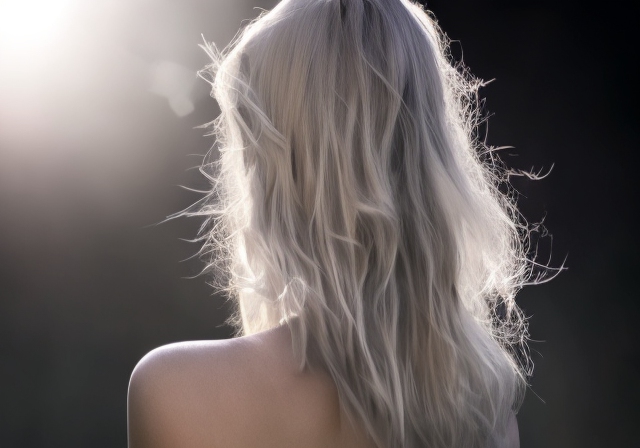}};
            \spy on \zoomeight in node [left] at \rebigone;
    	\end{tikzpicture}
    \end{subfigure}
    \begin{subfigure}{\depthWidth}
        \begin{tikzpicture}[spy using outlines={green,magnification=\ssmag,size=\ssizz},inner sep=0]
            \node [align=center, img] {\includegraphics[width=\textwidth]{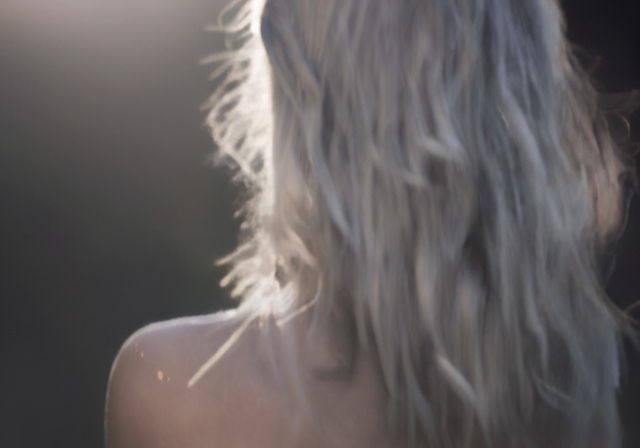}};
            \spy on \zoomeight in node [left] at \rebigone;
    	\end{tikzpicture}
      \end{subfigure}
    \begin{subfigure}{\depthWidth}
        \begin{tikzpicture}[spy using outlines={green,magnification=\ssmag,size=\ssizz},inner sep=0]
            \node [align=center, img] {\includegraphics[width=\textwidth]{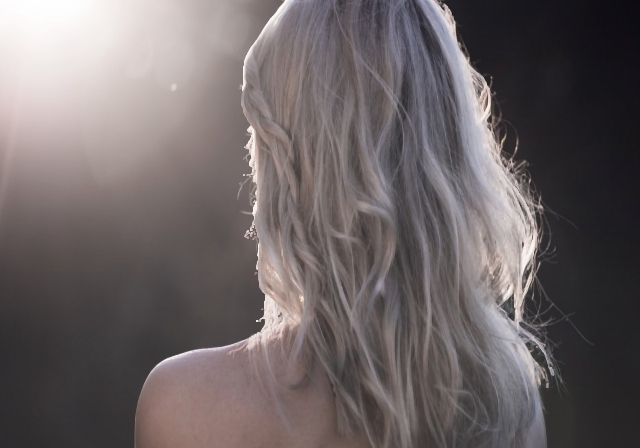}};
            \spy on \zoomeight in node [left] at \rebigone;
    	\end{tikzpicture}
      \end{subfigure}
    \begin{subfigure}{\depthWidth}
        \begin{tikzpicture}[spy using outlines={green,magnification=\ssmag,size=\ssizz},inner sep=0]
            \node [align=center, img] {\includegraphics[width=\textwidth]{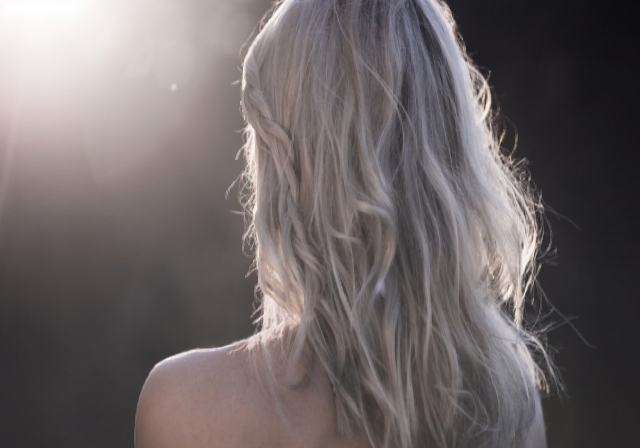}};
            \spy on \zoomeight in node [left] at \rebigone;
    	\end{tikzpicture}
      \end{subfigure}
    \end{subfigure}
    \begin{subfigure}{\linewidth}
    \centering
    \begin{subfigure}{\depthWidth}
        \begin{tikzpicture}[spy using outlines={green,magnification=\ssmag,size=\ssizz},inner sep=0]
            \node [align=center, img] {\includegraphics[width=\textwidth]{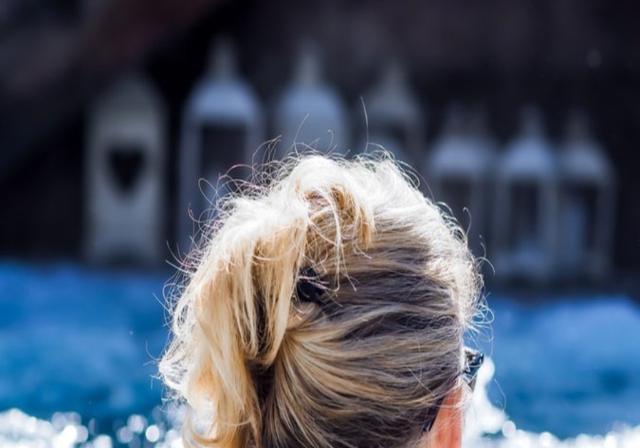}};
            \spy on \zoomnineori in node [left] at \rebigone;
    	\end{tikzpicture}
        \caption*{Input Image}
    \end{subfigure}
    \begin{subfigure}{\depthWidth}
		\begin{tikzpicture}[spy using outlines={green,magnification=\ssmag,size=\ssizz},inner sep=0]
            \node [align=center, img] {\includegraphics[width=\textwidth]{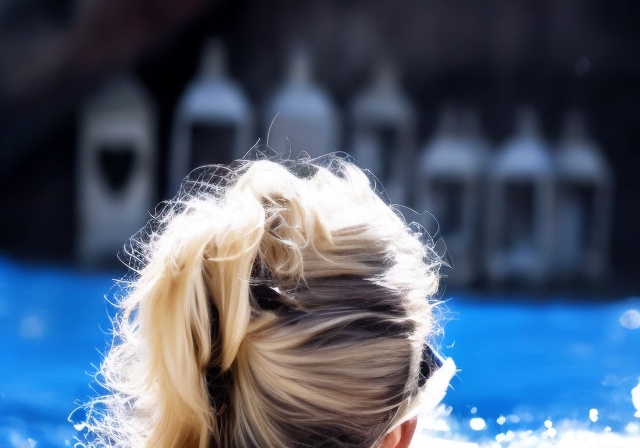}};
            \spy on \zoomnine in node [left] at \rebigone;
    	\end{tikzpicture}
        \caption*{ViewCrafter}
    \end{subfigure}
    \begin{subfigure}{\depthWidth}
        \begin{tikzpicture}[spy using outlines={green,magnification=\ssmag,size=\ssizz},inner sep=0]
            \node [align=center, img] {\includegraphics[width=\textwidth]{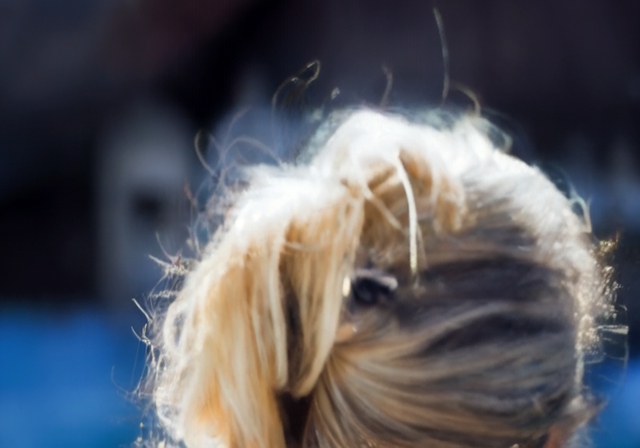}};
            \spy on \zoomnine in node [left] at \rebigone;
    	\end{tikzpicture}
        \caption*{ReCamMaster}
      \end{subfigure}
    \begin{subfigure}{\depthWidth}
        \begin{tikzpicture}[spy using outlines={green,magnification=\ssmag,size=\ssizz},inner sep=0]
            \node [align=center, img] {\includegraphics[width=\textwidth]{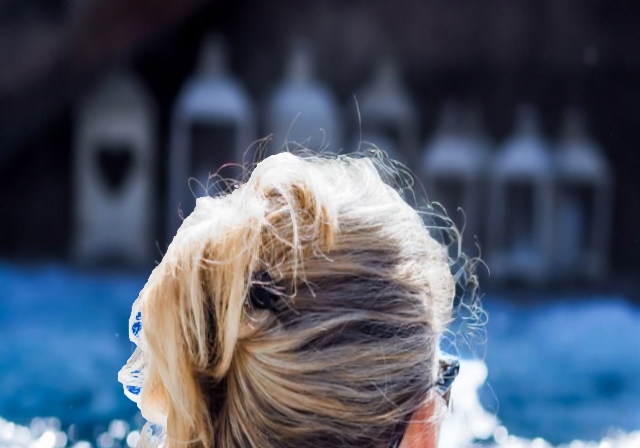}};
            \spy on \zoomnine in node [left] at \rebigone;
    	\end{tikzpicture}
        \caption*{SplatDiff}
      \end{subfigure}
    \begin{subfigure}{\depthWidth}
        \begin{tikzpicture}[spy using outlines={green,magnification=\ssmag,size=\ssizz},inner sep=0]
            \node [align=center, img] {\includegraphics[width=\textwidth]{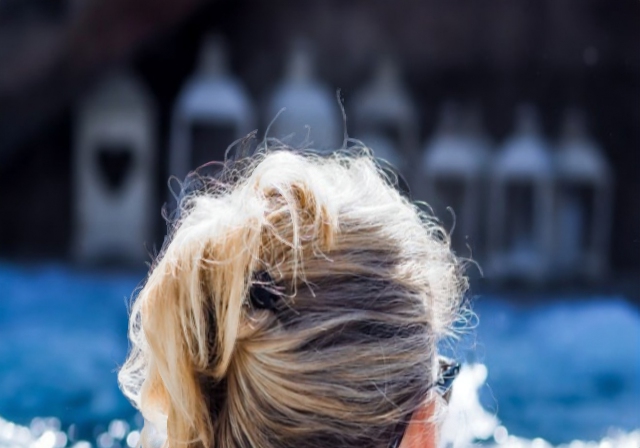}};
            \spy on \zoomnine in node [left] at \rebigone;
    	\end{tikzpicture}
        \caption*{Ours}
      \end{subfigure}
    \end{subfigure}
    \caption{\textbf{Qualitative comparison of novel view synthesis} on the AIM-500 dataset.}
    \label{fig:supp-nvs-visual-aim}
\end{figure*}

%% file: figs/supp/fig-nvs_visuals_p3m.tex
\def\imgWidth{0.32\linewidth} %
\def\depthWidth{0.19\linewidth} %
\def\pointWidth{0.24\linewidth} %
\def\scc{(-1.9,-1.4)}

\def\rebigone{(-0.5, -0.55)} %
\def\rebigtwo{(1.6, -0.55)} %

\def\zoomone{(1.4,-0.05)} %
\def\zoomoneori{(1.02,-0.05)} %

\def\zoomtwo{(1.2,0.)} %
\def\zoomtwoori{(0.95,0.)} %

\def\zoomthree{(1,-0.6)} %
\def\zoomthreeori{(0.95,-0.6)} %
\def\zoomthreeview{(1.4,-0.6)} %
\def\zoomthreerecam{(1.3,-0.6)} %

\def\zoomfour{(1.3,0.8)} %
\def\zoomfourori{(1.05,0.75)} %

\def\zoomfive{(0.8,0.7)} %
\def\zoomfiveori{(0.6,0.6)} %

\def\zoomsix{(0.8,0.95)} %
\def\zoomsixori{(0.75,0.8)} %

\def\zoomseven{(0.5,0.5)} %
\def\zoomsevenori{(0.6,0.45)} %

\def\zoomeight{(-0.9,0.6)} %
\def\zoomeightori{(-0.7,0.55)} %

\def\zoomnine{(0.25,0.55)} %
\def\zoomnineori{(0.4,0.5)} %
\def\zoomnineview{(0.1,0.45)} %

\def\ssizz{1.1cm} %
\def\ssmag{3}
\def\ssizz{1.1cm} %
\def\ssmag{3}

\begin{figure*}[t]
\centering
\tikzstyle{img} = [rectangle, minimum width=\imgWidth]
    \centering
    \begin{subfigure}{\linewidth}
    \centering
    \begin{subfigure}{\depthWidth}
        \begin{tikzpicture}[spy using outlines={green,magnification=\ssmag,size=\ssizz},inner sep=0]
            \node [align=center, img] {\includegraphics[width=\textwidth]{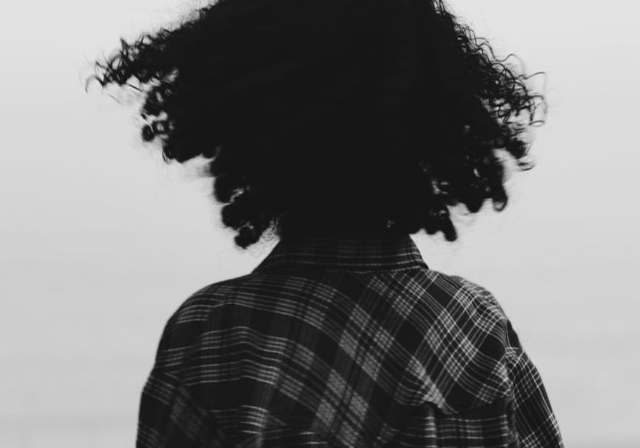}};
            \spy on \zoomfourori in node [left] at \rebigone;
    	\end{tikzpicture}
    \end{subfigure}
    \begin{subfigure}{\depthWidth}
		\begin{tikzpicture}[spy using outlines={green,magnification=\ssmag,size=\ssizz},inner sep=0]
            \node [align=center, img] {\includegraphics[width=\textwidth]{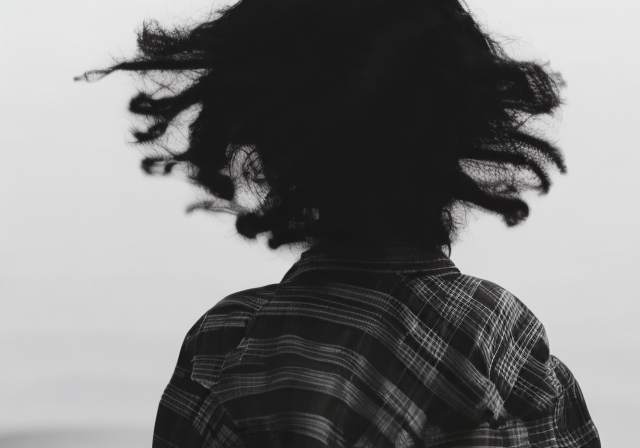}};
            \spy on \zoomfour in node [left] at \rebigone;
    	\end{tikzpicture}
    \end{subfigure}
    \begin{subfigure}{\depthWidth}
        \begin{tikzpicture}[spy using outlines={green,magnification=\ssmag,size=\ssizz},inner sep=0]
            \node [align=center, img] {\includegraphics[width=\textwidth]{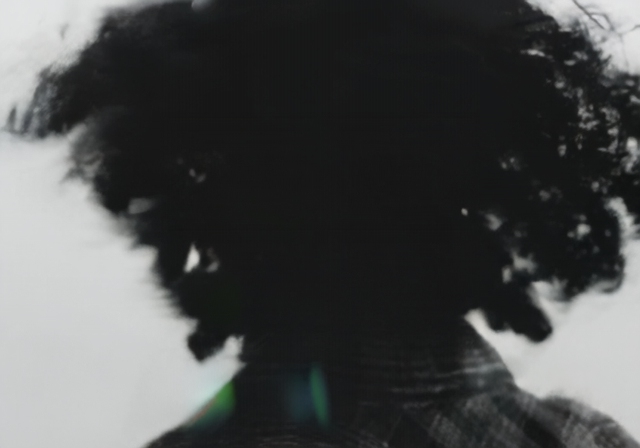}};
            \spy on \zoomfour in node [left] at \rebigone;
    	\end{tikzpicture}
      \end{subfigure}
    \begin{subfigure}{\depthWidth}
        \begin{tikzpicture}[spy using outlines={green,magnification=\ssmag,size=\ssizz},inner sep=0]
            \node [align=center, img] {\includegraphics[width=\textwidth]{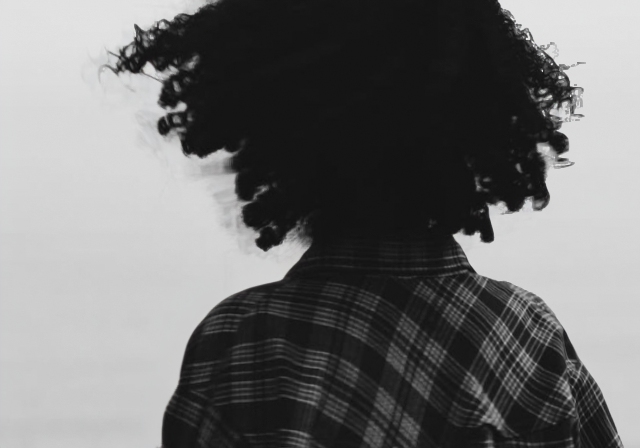}};
            \spy on \zoomfour in node [left] at \rebigone;
    	\end{tikzpicture}
      \end{subfigure}
    \begin{subfigure}{\depthWidth}
        \begin{tikzpicture}[spy using outlines={green,magnification=\ssmag,size=\ssizz},inner sep=0]
            \node [align=center, img] {\includegraphics[width=\textwidth]{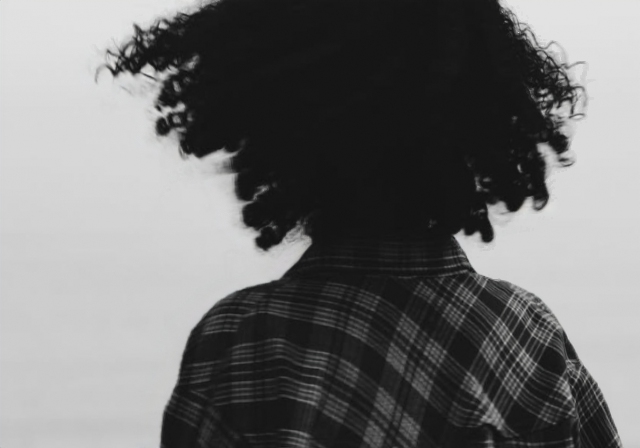}};
            \spy on \zoomfour in node [left] at \rebigone;
    	\end{tikzpicture}
      \end{subfigure}
    \end{subfigure}
    \begin{subfigure}{\linewidth}
    \centering
    \begin{subfigure}{\depthWidth}
        \begin{tikzpicture}[spy using outlines={green,magnification=\ssmag,size=\ssizz},inner sep=0]
            \node [align=center, img] {\includegraphics[width=\textwidth]{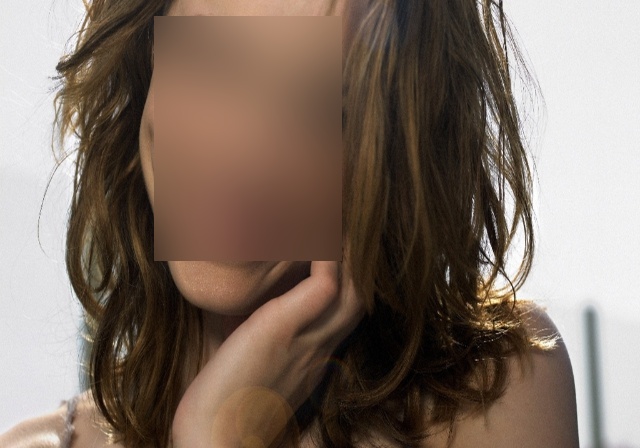}};
            \spy on \zoomtwoori in node [left] at \rebigone;
    	\end{tikzpicture}
    \end{subfigure}
    \begin{subfigure}{\depthWidth}
		\begin{tikzpicture}[spy using outlines={green,magnification=\ssmag,size=\ssizz},inner sep=0]
            \node [align=center, img] {\includegraphics[width=\textwidth]{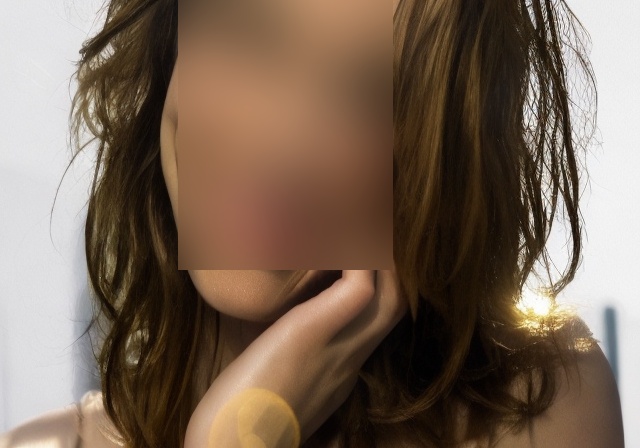}};
            \spy on \zoomtwo in node [left] at \rebigone;
    	\end{tikzpicture}
    \end{subfigure}
    \begin{subfigure}{\depthWidth}
        \begin{tikzpicture}[spy using outlines={green,magnification=\ssmag,size=\ssizz},inner sep=0]
            \node [align=center, img] {\includegraphics[width=\textwidth]{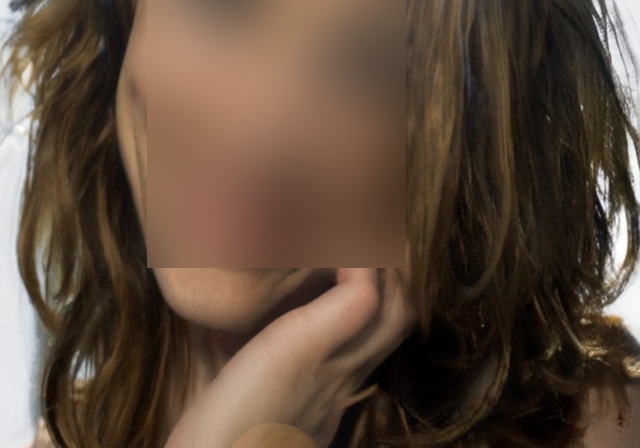}};
            \spy on \zoomtwo in node [left] at \rebigone;
    	\end{tikzpicture}
      \end{subfigure}
    \begin{subfigure}{\depthWidth}
        \begin{tikzpicture}[spy using outlines={green,magnification=\ssmag,size=\ssizz},inner sep=0]
            \node [align=center, img] {\includegraphics[width=\textwidth]{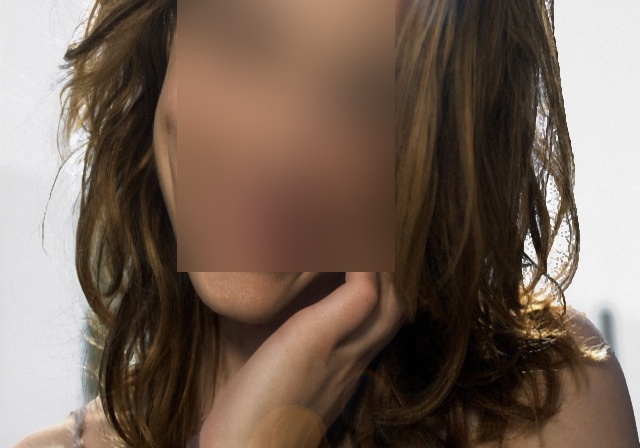}};
            \spy on \zoomtwo in node [left] at \rebigone;
    	\end{tikzpicture}
      \end{subfigure}
    \begin{subfigure}{\depthWidth}
        \begin{tikzpicture}[spy using outlines={green,magnification=\ssmag,size=\ssizz},inner sep=0]
            \node [align=center, img] {\includegraphics[width=\textwidth]{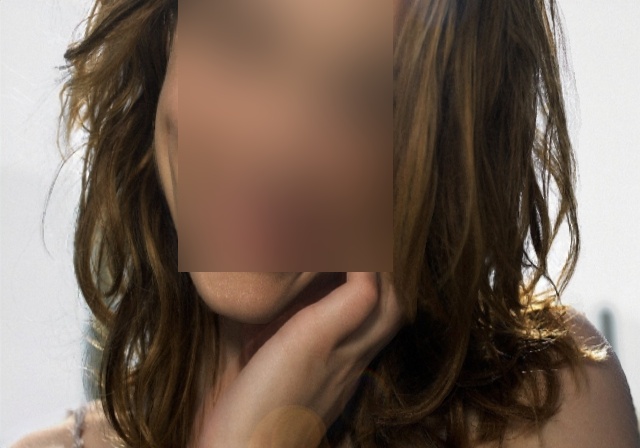}};
            \spy on \zoomtwo in node [left] at \rebigone;
    	\end{tikzpicture}
      \end{subfigure}
    \end{subfigure}
    \begin{subfigure}{\linewidth}
    \centering
    \begin{subfigure}{\depthWidth}
        \begin{tikzpicture}[spy using outlines={green,magnification=\ssmag,size=\ssizz},inner sep=0]
            \node [align=center, img] {\includegraphics[width=\textwidth]{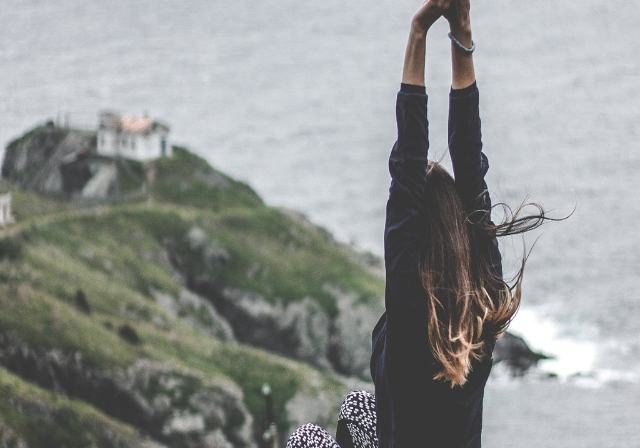}};
            \spy on \zoomoneori in node [left] at \rebigone;
    	\end{tikzpicture}
    \end{subfigure}
    \begin{subfigure}{\depthWidth}
		\begin{tikzpicture}[spy using outlines={green,magnification=\ssmag,size=\ssizz},inner sep=0]
            \node [align=center, img] {\includegraphics[width=\textwidth]{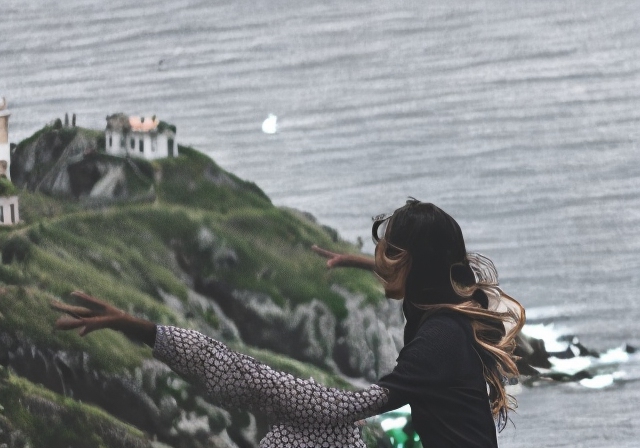}};
            \spy on \zoomone in node [left] at \rebigone;
    	\end{tikzpicture}
    \end{subfigure}
    \begin{subfigure}{\depthWidth}
        \begin{tikzpicture}[spy using outlines={green,magnification=\ssmag,size=\ssizz},inner sep=0]
            \node [align=center, img] {\includegraphics[width=\textwidth]{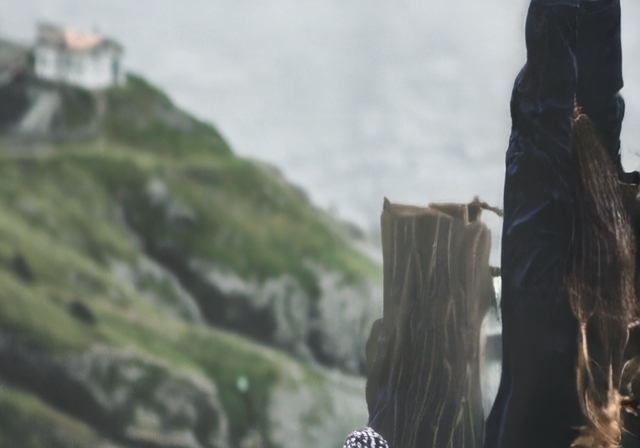}};
            \spy on \zoomone in node [left] at \rebigone;
    	\end{tikzpicture}
      \end{subfigure}
    \begin{subfigure}{\depthWidth}
        \begin{tikzpicture}[spy using outlines={green,magnification=\ssmag,size=\ssizz},inner sep=0]
            \node [align=center, img] {\includegraphics[width=\textwidth]{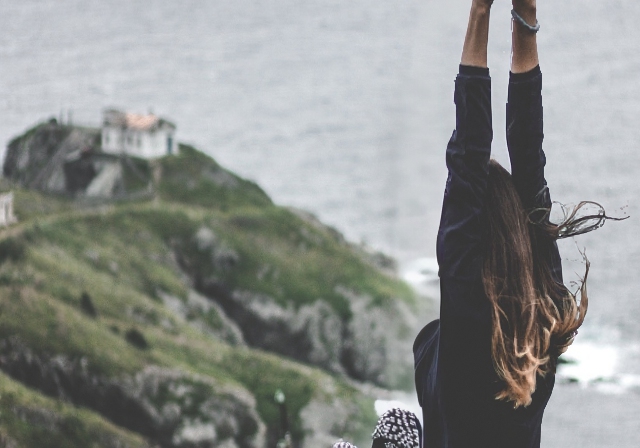}};
            \spy on \zoomone in node [left] at \rebigone;
    	\end{tikzpicture}
      \end{subfigure}
    \begin{subfigure}{\depthWidth}
        \begin{tikzpicture}[spy using outlines={green,magnification=\ssmag,size=\ssizz},inner sep=0]
            \node [align=center, img] {\includegraphics[width=\textwidth]{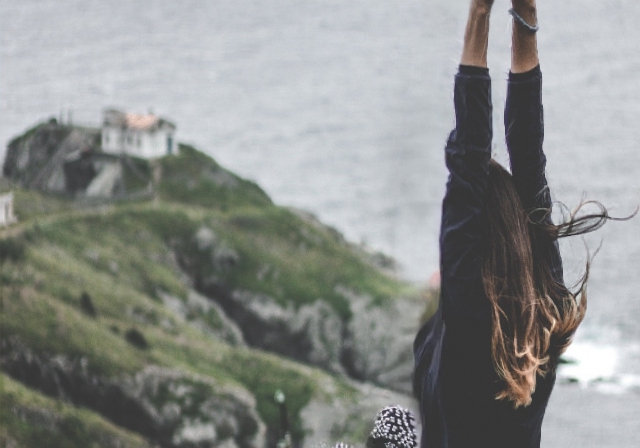}};
            \spy on \zoomone in node [left] at \rebigone;
    	\end{tikzpicture}
      \end{subfigure}
    \end{subfigure}
    \begin{subfigure}{\linewidth}
    \centering
    \begin{subfigure}{\depthWidth}
        \begin{tikzpicture}[spy using outlines={green,magnification=\ssmag,size=\ssizz},inner sep=0]
            \node [align=center, img] {\includegraphics[width=\textwidth]{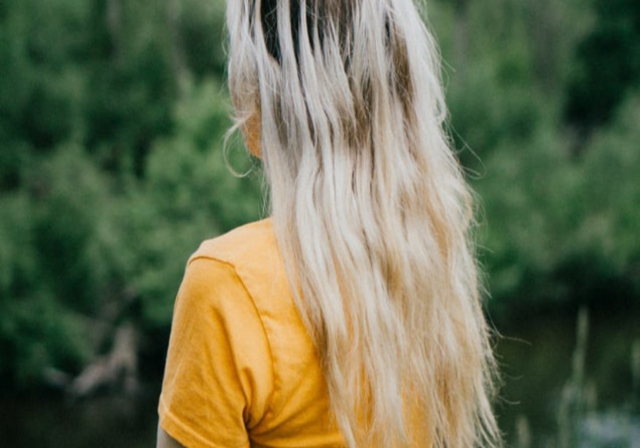}};
            \spy on \zoomthreeori in node [left] at \rebigone;
    	\end{tikzpicture}
    \end{subfigure}
    \begin{subfigure}{\depthWidth}
		\begin{tikzpicture}[spy using outlines={green,magnification=\ssmag,size=\ssizz},inner sep=0]
            \node [align=center, img] {\includegraphics[width=\textwidth]{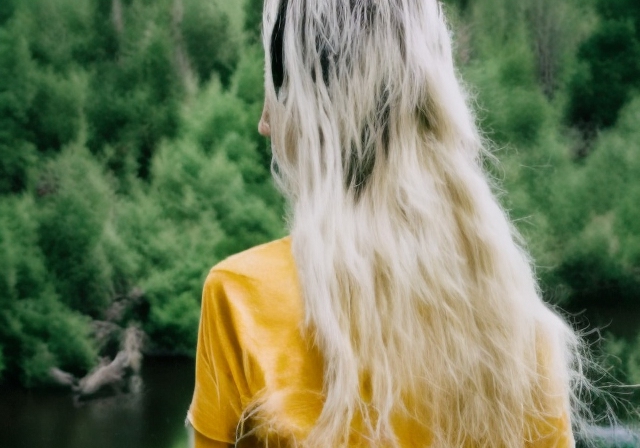}};
            \spy on \zoomthreeview in node [left] at \rebigone;
    	\end{tikzpicture}
    \end{subfigure}
    \begin{subfigure}{\depthWidth}
        \begin{tikzpicture}[spy using outlines={green,magnification=\ssmag,size=\ssizz},inner sep=0]
            \node [align=center, img] {\includegraphics[width=\textwidth]{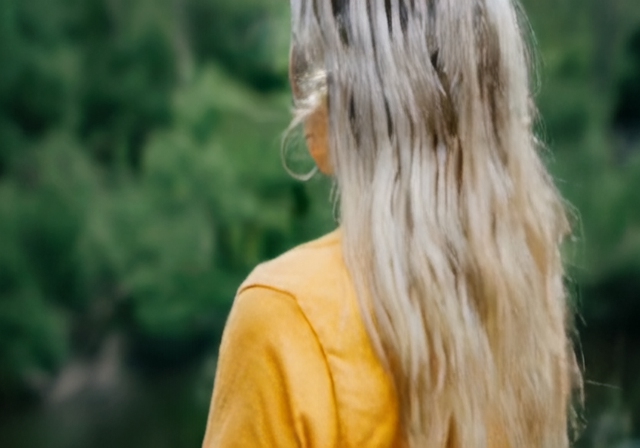}};
            \spy on \zoomthreerecam in node [left] at \rebigone;
    	\end{tikzpicture}
      \end{subfigure}
    \begin{subfigure}{\depthWidth}
        \begin{tikzpicture}[spy using outlines={green,magnification=\ssmag,size=\ssizz},inner sep=0]
            \node [align=center, img] {\includegraphics[width=\textwidth]{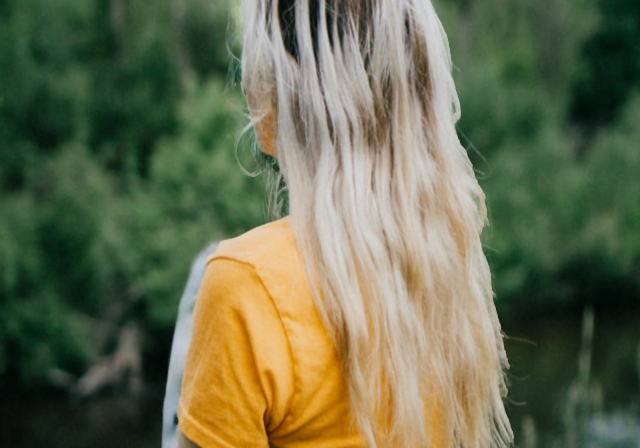}};
            \spy on \zoomthree in node [left] at \rebigone;
    	\end{tikzpicture}
      \end{subfigure}
    \begin{subfigure}{\depthWidth}
        \begin{tikzpicture}[spy using outlines={green,magnification=\ssmag,size=\ssizz},inner sep=0]
            \node [align=center, img] {\includegraphics[width=\textwidth]{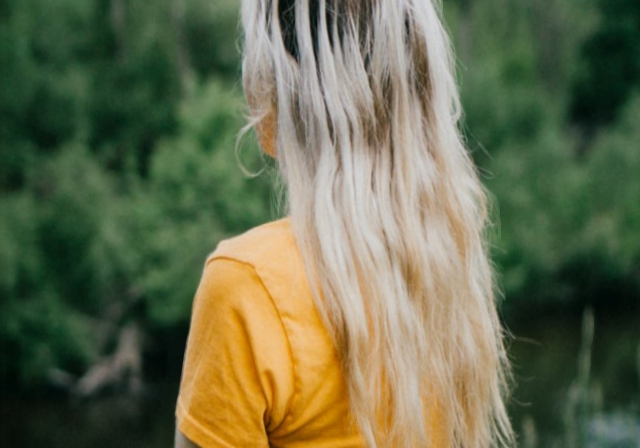}};
            \spy on \zoomthree in node [left] at \rebigone;
    	\end{tikzpicture}
      \end{subfigure}
    \end{subfigure}
    \begin{subfigure}{\linewidth}
    \centering
    \begin{subfigure}{\depthWidth}
        \begin{tikzpicture}[spy using outlines={green,magnification=\ssmag,size=\ssizz},inner sep=0]
            \node [align=center, img] {\includegraphics[width=\textwidth]{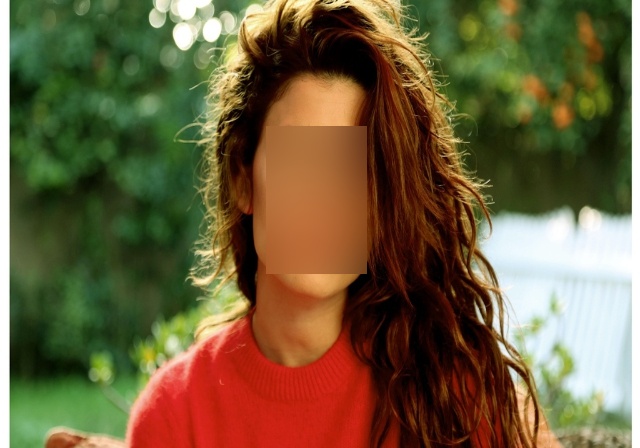}};
            \spy on \zoomfiveori in node [left] at \rebigone;
    	\end{tikzpicture}
    \end{subfigure}
    \begin{subfigure}{\depthWidth}
		\begin{tikzpicture}[spy using outlines={green,magnification=\ssmag,size=\ssizz},inner sep=0]
            \node [align=center, img] {\includegraphics[width=\textwidth]{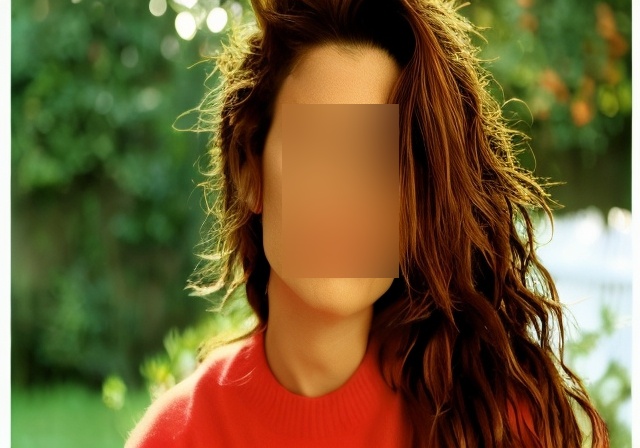}};
            \spy on \zoomfive in node [left] at \rebigone;
    	\end{tikzpicture}
    \end{subfigure}
    \begin{subfigure}{\depthWidth}
        \begin{tikzpicture}[spy using outlines={green,magnification=\ssmag,size=\ssizz},inner sep=0]
            \node [align=center, img] {\includegraphics[width=\textwidth]{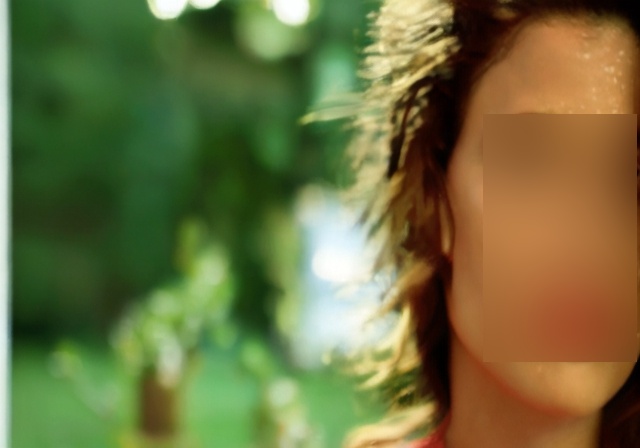}};
            \spy on \zoomfive in node [left] at \rebigone;
    	\end{tikzpicture}
      \end{subfigure}
    \begin{subfigure}{\depthWidth}
        \begin{tikzpicture}[spy using outlines={green,magnification=\ssmag,size=\ssizz},inner sep=0]
            \node [align=center, img] {\includegraphics[width=\textwidth]{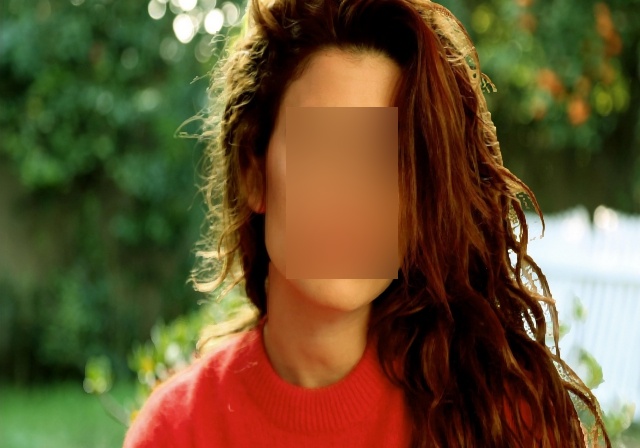}};
            \spy on \zoomfive in node [left] at \rebigone;
    	\end{tikzpicture}
      \end{subfigure}
    \begin{subfigure}{\depthWidth}
        \begin{tikzpicture}[spy using outlines={green,magnification=\ssmag,size=\ssizz},inner sep=0]
            \node [align=center, img] {\includegraphics[width=\textwidth]{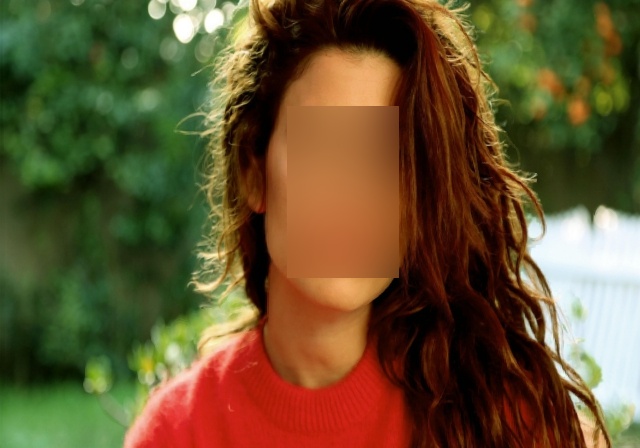}};
            \spy on \zoomfive in node [left] at \rebigone;
    	\end{tikzpicture}
      \end{subfigure}
    \end{subfigure}
    \begin{subfigure}{\linewidth}
    \centering
    \begin{subfigure}{\depthWidth}
        \begin{tikzpicture}[spy using outlines={green,magnification=\ssmag,size=\ssizz},inner sep=0]
            \node [align=center, img] {\includegraphics[width=\textwidth]{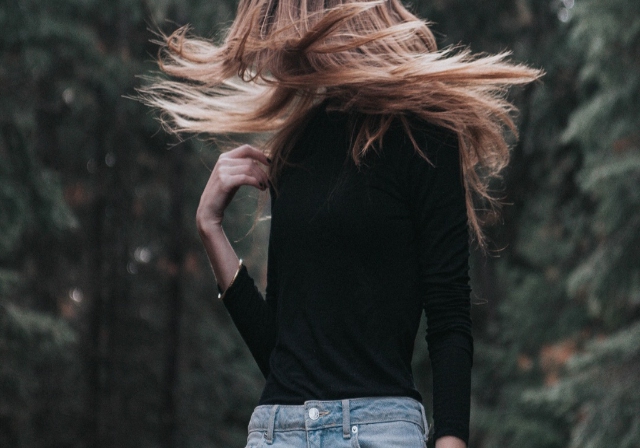}};
            \spy on \zoomsixori in node [left] at \rebigone;
    	\end{tikzpicture}
    \end{subfigure}
    \begin{subfigure}{\depthWidth}
		\begin{tikzpicture}[spy using outlines={green,magnification=\ssmag,size=\ssizz},inner sep=0]
            \node [align=center, img] {\includegraphics[width=\textwidth]{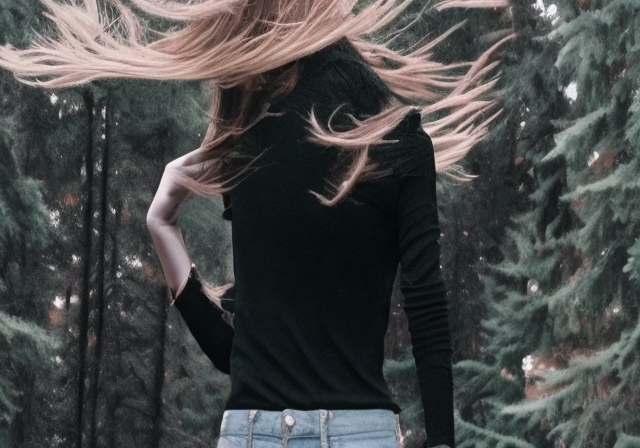}};
            \spy on \zoomsix in node [left] at \rebigone;
    	\end{tikzpicture}
    \end{subfigure}
    \begin{subfigure}{\depthWidth}
        \begin{tikzpicture}[spy using outlines={green,magnification=\ssmag,size=\ssizz},inner sep=0]
            \node [align=center, img] {\includegraphics[width=\textwidth]{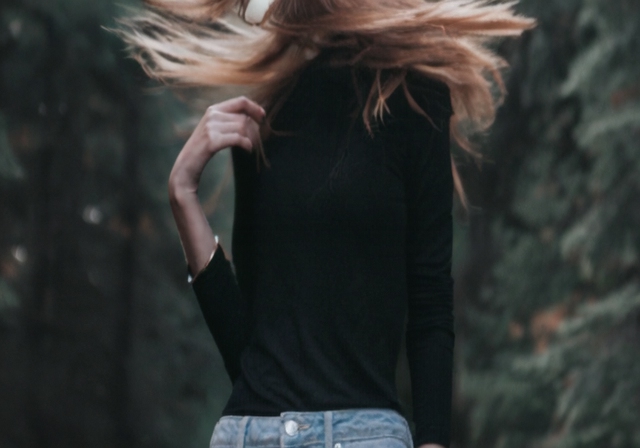}};
            \spy on \zoomsix in node [left] at \rebigone;
    	\end{tikzpicture}
      \end{subfigure}
    \begin{subfigure}{\depthWidth}
        \begin{tikzpicture}[spy using outlines={green,magnification=\ssmag,size=\ssizz},inner sep=0]
            \node [align=center, img] {\includegraphics[width=\textwidth]{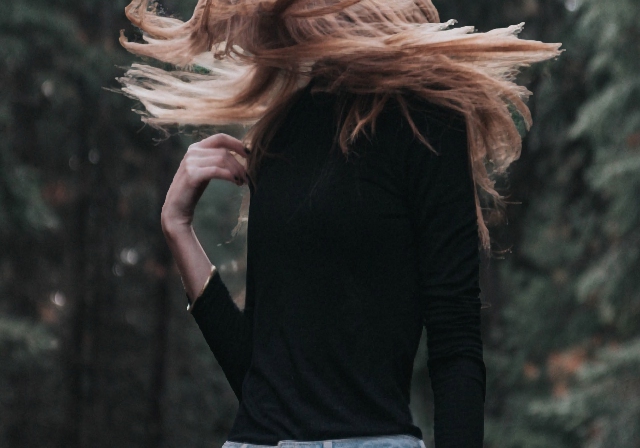}};
            \spy on \zoomsix in node [left] at \rebigone;
    	\end{tikzpicture}
      \end{subfigure}
    \begin{subfigure}{\depthWidth}
        \begin{tikzpicture}[spy using outlines={green,magnification=\ssmag,size=\ssizz},inner sep=0]
            \node [align=center, img] {\includegraphics[width=\textwidth]{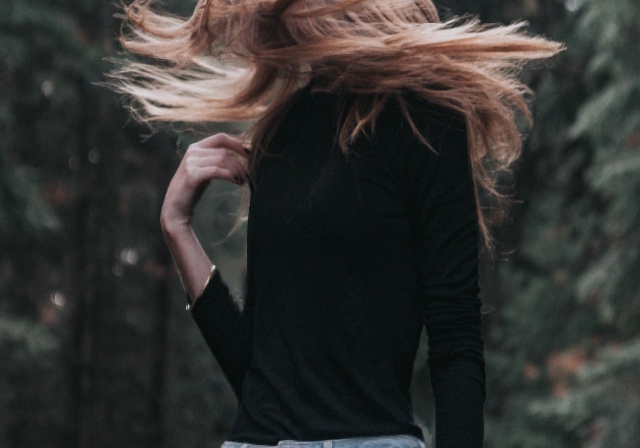}};
            \spy on \zoomsix in node [left] at \rebigone;
    	\end{tikzpicture}
      \end{subfigure}
    \end{subfigure}
    \begin{subfigure}{\linewidth}
    \centering
    \begin{subfigure}{\depthWidth}
        \begin{tikzpicture}[spy using outlines={green,magnification=\ssmag,size=\ssizz},inner sep=0]
            \node [align=center, img] {\includegraphics[width=\textwidth]{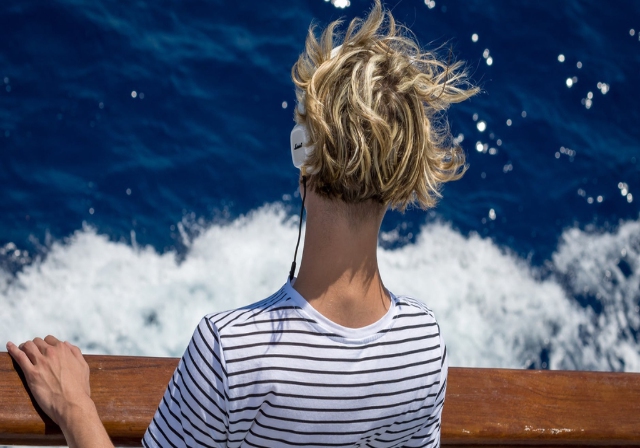}};
            \spy on \zoomsevenori in node [left] at \rebigone;
    	\end{tikzpicture}
    \end{subfigure}
    \begin{subfigure}{\depthWidth}
		\begin{tikzpicture}[spy using outlines={green,magnification=\ssmag,size=\ssizz},inner sep=0]
            \node [align=center, img] {\includegraphics[width=\textwidth]{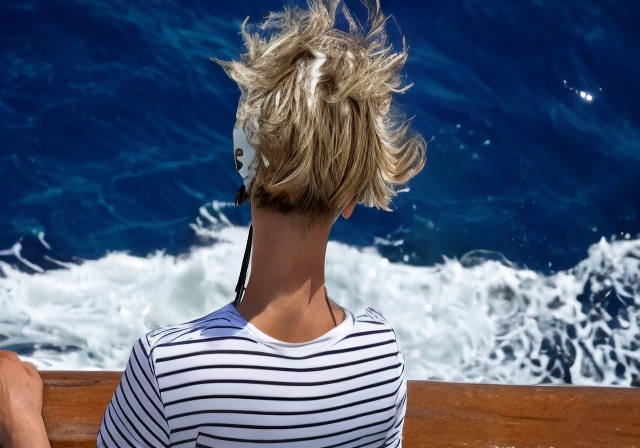}};
            \spy on \zoomseven in node [left] at \rebigone;
    	\end{tikzpicture}
    \end{subfigure}
    \begin{subfigure}{\depthWidth}
        \begin{tikzpicture}[spy using outlines={green,magnification=\ssmag,size=\ssizz},inner sep=0]
            \node [align=center, img] {\includegraphics[width=\textwidth]{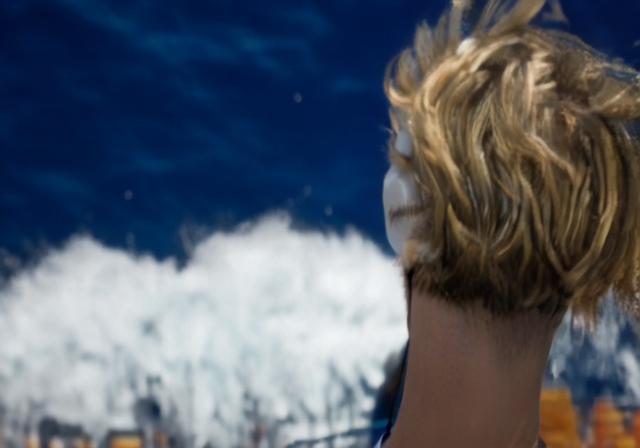}};
            \spy on \zoomseven in node [left] at \rebigone;
    	\end{tikzpicture}
      \end{subfigure}
    \begin{subfigure}{\depthWidth}
        \begin{tikzpicture}[spy using outlines={green,magnification=\ssmag,size=\ssizz},inner sep=0]
            \node [align=center, img] {\includegraphics[width=\textwidth]{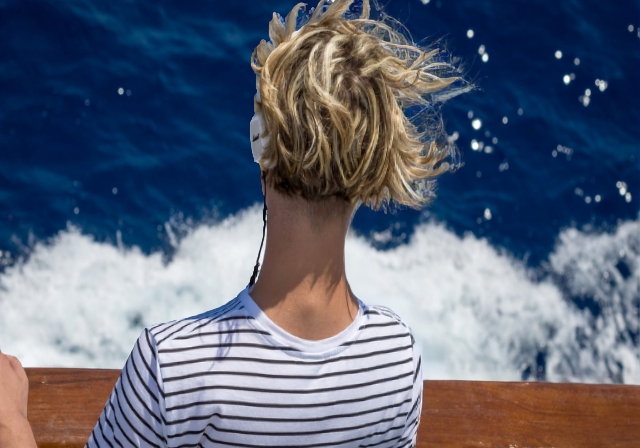}};
            \spy on \zoomseven in node [left] at \rebigone;
    	\end{tikzpicture}
      \end{subfigure}
    \begin{subfigure}{\depthWidth}
        \begin{tikzpicture}[spy using outlines={green,magnification=\ssmag,size=\ssizz},inner sep=0]
            \node [align=center, img] {\includegraphics[width=\textwidth]{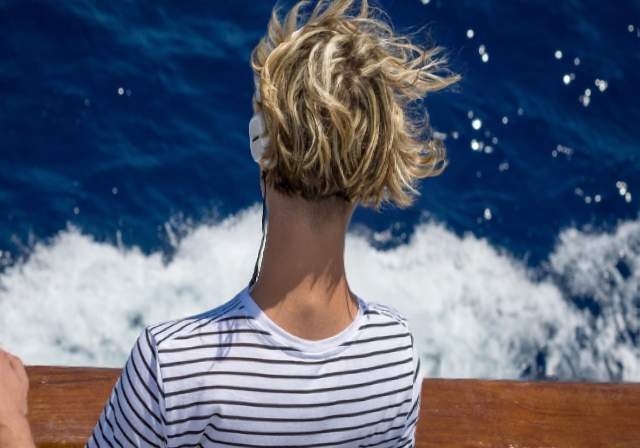}};
            \spy on \zoomseven in node [left] at \rebigone;
    	\end{tikzpicture}
      \end{subfigure}
    \end{subfigure}
    \begin{subfigure}{\linewidth}
    \centering
    \begin{subfigure}{\depthWidth}
        \begin{tikzpicture}[spy using outlines={green,magnification=\ssmag,size=\ssizz},inner sep=0]
            \node [align=center, img] {\includegraphics[width=\textwidth]{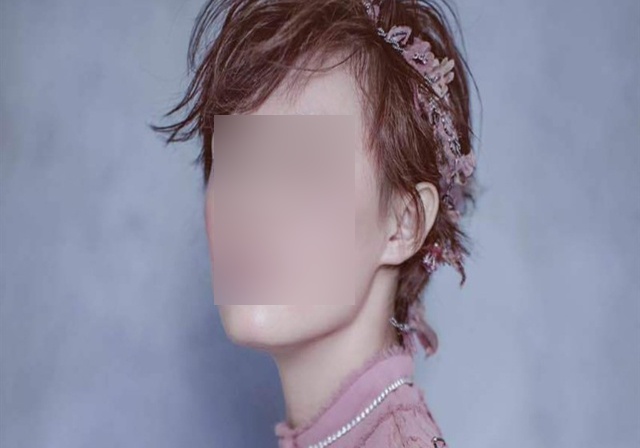}};
            \spy on \zoomeightori in node [left] at \rebigone;
    	\end{tikzpicture}
    \end{subfigure}
    \begin{subfigure}{\depthWidth}
		\begin{tikzpicture}[spy using outlines={green,magnification=\ssmag,size=\ssizz},inner sep=0]
            \node [align=center, img] {\includegraphics[width=\textwidth]{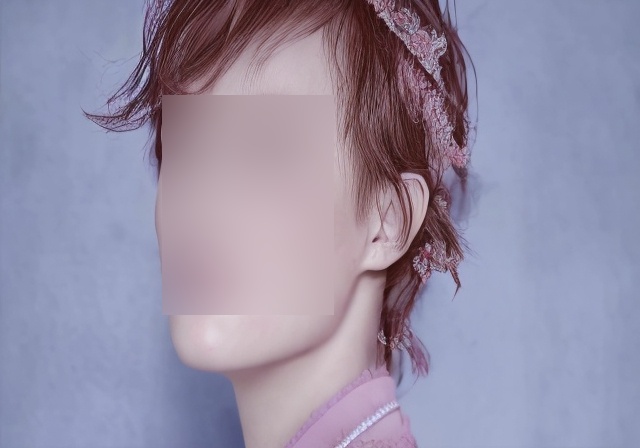}};
            \spy on \zoomeight in node [left] at \rebigone;
    	\end{tikzpicture}
    \end{subfigure}
    \begin{subfigure}{\depthWidth}
        \begin{tikzpicture}[spy using outlines={green,magnification=\ssmag,size=\ssizz},inner sep=0]
            \node [align=center, img] {\includegraphics[width=\textwidth]{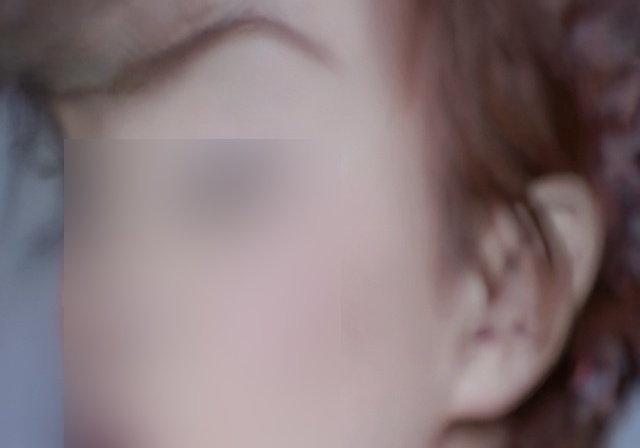}};
            \spy on \zoomeight in node [left] at \rebigone;
    	\end{tikzpicture}
      \end{subfigure}
    \begin{subfigure}{\depthWidth}
        \begin{tikzpicture}[spy using outlines={green,magnification=\ssmag,size=\ssizz},inner sep=0]
            \node [align=center, img] {\includegraphics[width=\textwidth]{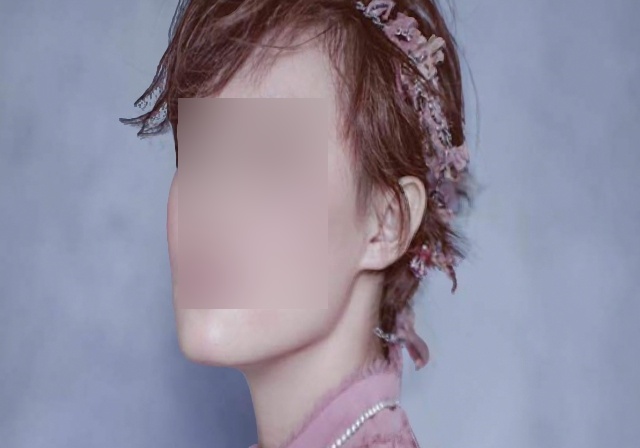}};
            \spy on \zoomeight in node [left] at \rebigone;
    	\end{tikzpicture}
      \end{subfigure}
    \begin{subfigure}{\depthWidth}
        \begin{tikzpicture}[spy using outlines={green,magnification=\ssmag,size=\ssizz},inner sep=0]
            \node [align=center, img] {\includegraphics[width=\textwidth]{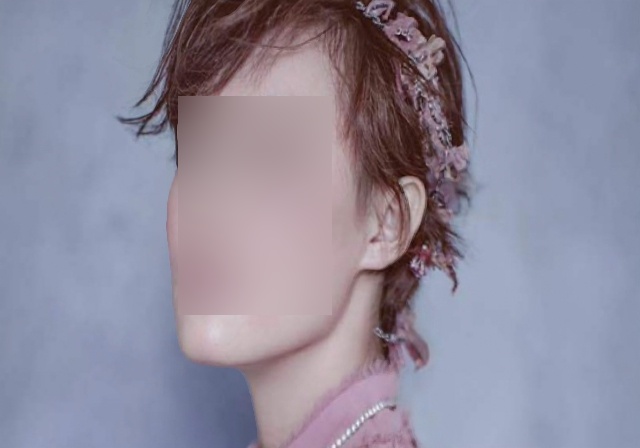}};
            \spy on \zoomeight in node [left] at \rebigone;
    	\end{tikzpicture}
      \end{subfigure}
    \end{subfigure}
    \begin{subfigure}{\linewidth}
    \centering
    \begin{subfigure}{\depthWidth}
        \begin{tikzpicture}[spy using outlines={green,magnification=\ssmag,size=\ssizz},inner sep=0]
            \node [align=center, img] {\includegraphics[width=\textwidth]{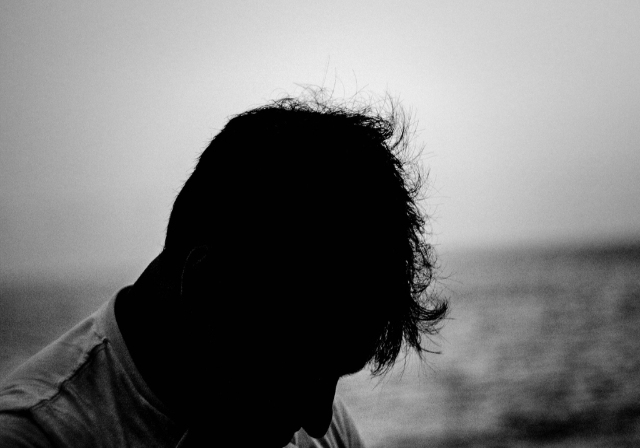}};
            \spy on \zoomnineori in node [left] at \rebigone;
    	\end{tikzpicture}
        \caption*{Input Image}
    \end{subfigure}
    \begin{subfigure}{\depthWidth}
		\begin{tikzpicture}[spy using outlines={green,magnification=\ssmag,size=\ssizz},inner sep=0]
            \node [align=center, img] {\includegraphics[width=\textwidth]{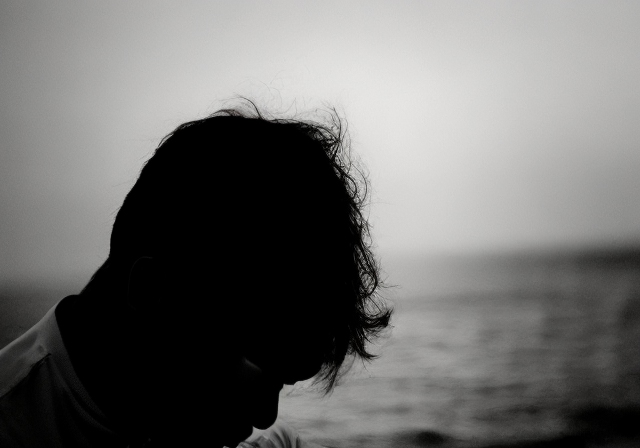}};
            \spy on \zoomnineview in node [left] at \rebigone;
    	\end{tikzpicture}
        \caption*{ViewCrafter}
    \end{subfigure}
    \begin{subfigure}{\depthWidth}
        \begin{tikzpicture}[spy using outlines={green,magnification=\ssmag,size=\ssizz},inner sep=0]
            \node [align=center, img] {\includegraphics[width=\textwidth]{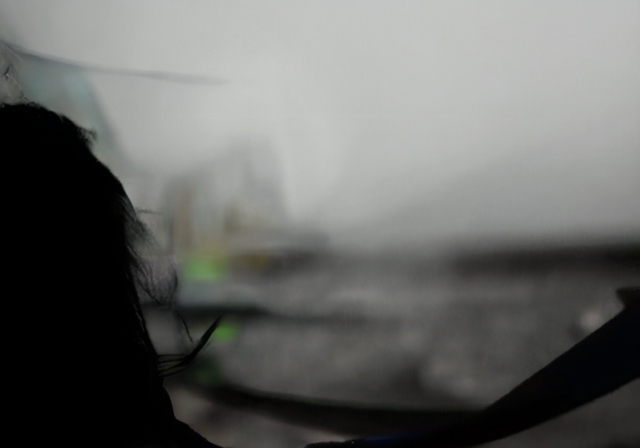}};
            \spy on \zoomnine in node [left] at \rebigone;
    	\end{tikzpicture}
        \caption*{ReCamMaster}
      \end{subfigure}
    \begin{subfigure}{\depthWidth}
        \begin{tikzpicture}[spy using outlines={green,magnification=\ssmag,size=\ssizz},inner sep=0]
            \node [align=center, img] {\includegraphics[width=\textwidth]{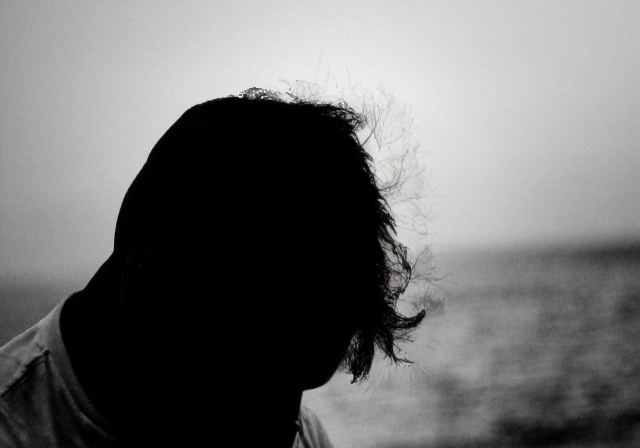}};
            \spy on \zoomnine in node [left] at \rebigone;
    	\end{tikzpicture}
        \caption*{SplatDiff}
      \end{subfigure}
    \begin{subfigure}{\depthWidth}
        \begin{tikzpicture}[spy using outlines={green,magnification=\ssmag,size=\ssizz},inner sep=0]
            \node [align=center, img] {\includegraphics[width=\textwidth]{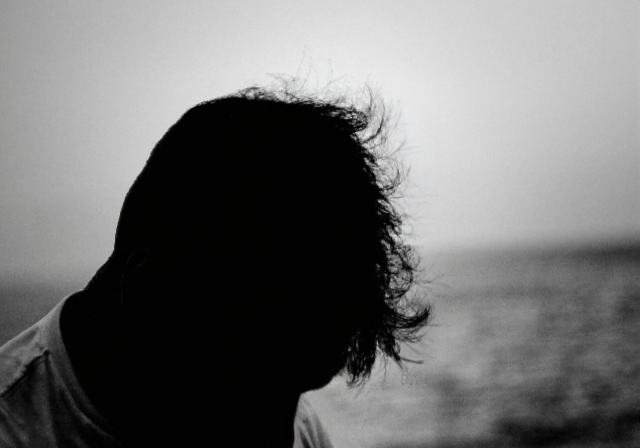}};
            \spy on \zoomnine in node [left] at \rebigone;
    	\end{tikzpicture}
        \caption*{Ours}
      \end{subfigure}
    \end{subfigure}
    \caption{\textbf{Qualitative comparison of novel view synthesis} on the P3M-10K dataset. Human faces are manually blurred to protect privacy.}
    \label{fig:supp-nvs-visual-p3m}
\end{figure*}